\documentclass[lettersize,journal]{IEEEtran}
   \usepackage{setspace}
\usepackage{lineno}
\usepackage{wrapfig}

\usepackage{placeins}   

\usepackage{graphicx}
\usepackage{floatrow}
\usepackage{sidecap}
\usepackage{caption}
\usepackage{calc} 

\usepackage{algorithm}

\usepackage{algorithm}
\usepackage{algorithmicx}
\usepackage{algpseudocode}
\usepackage{amsmath}
\usepackage{amssymb}
\usepackage{bm} 

\usepackage{enumitem}
\usepackage{changepage}

   \usepackage[colorlinks=true,
               linkcolor=red,       
               citecolor=green,      
               urlcolor=blue,         
               filecolor=magenta     
               ]{hyperref}

\usepackage{natbib}
\usepackage{xcolor}         
\usepackage[utf8]{inputenc} 
\usepackage[T1]{fontenc}    
\usepackage{url}            
\usepackage{booktabs}       
\usepackage{tablefootnote}
\usepackage{amsfonts}       
\usepackage{nicefrac}       
\usepackage{microtype}      

\definecolor{myblue}{rgb}{0.255, 0.412, 0.882} 

\definecolor{green1}{RGB}{10,158,10}
\definecolor{blue1}{RGB}{17,85,204}
\definecolor{red1}{RGB}{204,0,0}
\definecolor{mygray}{gray}{0.85}

\definecolor{myblue2}{RGB}{187, 213, 232} 

\usepackage{url}            
\usepackage{booktabs}       
\usepackage{amsfonts}       
\usepackage{nicefrac}       
\usepackage{microtype}      
\usepackage{lipsum}
\usepackage{wrapfig}
\usepackage{siunitx}
\usepackage{sidecap}

\usepackage{amsmath,amsfonts,amssymb,amsthm}
\usepackage{mathtools}
\usepackage{multirow}

\usepackage{bbm}
\usepackage{colortbl}
\usepackage{booktabs}
\usepackage{multirow}
\usepackage{siunitx}
\usepackage{array}

\usepackage{booktabs}
\usepackage{diagbox}

\usepackage{subcaption}

\renewcommand{\mathbf}[1]{\boldsymbol{\mathit{#1}}}

   \usepackage[colorlinks=true,
               linkcolor=red,       
               citecolor=green,      
               urlcolor=blue,         
               filecolor=magenta     
               ]{hyperref}

\usepackage{enumitem}

\newcolumntype{P}[1]{>{\hspace{1ex}}p{#1}<{\hspace{1ex}}}

\newtheorem{theorem}{Theorem}
\newtheorem{corollary}{Corollary}[theorem]
\newtheorem{lemma}[theorem]{Lemma}
\newtheorem{prop}[theorem]{Proposition}

\newtheorem{property}[theorem]{Property}

\title{Contrastive Entropy Bounds for Density and Conditional Density Decomposition}

\author{%
  Bo Hu, \; Jos{\'e}~C.~Pr{\'\i}ncipe\\
  Department of Electrical and Computer Engineering\\
  University of Florida\\
  \texttt{hubo@ufl.edu\; principe@cnel.ufl.edu }\\}

\begin{document}


\maketitle

\begin{abstract}

This paper studies the interpretability of neural network features through a Bayesian, Gaussianity viewpoint, where optimizing a cost is to reach a probabilistic bound; learning a model approximates a probability density that makes the bound tight and the cost optimal, often with a Gaussian mixture density. The two examples are Mixture Density Networks (MDNs) using the bound for the marginal and autoencoders using the conditional bound. It is a known result that minimizing the error between the inputs and outputs maximizes the dependence between inputs and the middle, not only for autoencoders.



This paper uses Hilbert space and the idea of decomposition to address cases in which a multiple-output neural network produces multiple centers to define a Gaussian mixture. Our first finding is that an autoencoder’s objective is equivalent to maximizing the trace of a Gaussian operator, the sum of eigenvalues under bases orthonormal w.r.t. the data and model distributions. This observation inspires us to test that, when a one-to-one correspondence as needed in autoencoders is unnecessary, we can maximize the nuclear norm of this operator, the sum of singular values, to maximize the overall rank rather than the trace. So, the trace of a Gaussian operator can be used to train autoencoders, and the nuclear norm of it can be used as a divergence measure to train MDNs, under bases orthonormal w.r.t. the data and model distributions.

Our second test is to use the inner product and norms in a Hilbert space to define bounds and costs. Such bounds often have an extra norm term compared to KL-based bounds, which has been shown to increase sample diversity and prevent the trivial solution when a multiple-output network produces the same constant. But the trade-off is that it requires a batch of samples to estimate and optimize. Motivated by this, we propose an encoder–mixture–decoder architecture whose decoder is multiple-output producing multiple centers per sample, potentially tightening the bound. Assuming the data as small-variance Gaussian mixtures, this upper bound can be estimated, tracked during training, and analyzed quantitatively.

\end{abstract}

\section{Introduction}

Whether neural network features are interpretable has long been investigated. Our previous series of work proposed the hypothesis of the orthonormal decomposition of a density ratio, claiming that multivariate features, particularly in contrastive learning or self-supervised learning, can be interpreted as the orthonormal basis functions of a density ratio, coming from statistical dependence. We proposed that such multivariate features must satisfy an orthonormal condition to be diverse, and an equilibrium condition to be meaningful. In this paper, we tackle this from a different angle, coming from Gaussianity and Bayes's analysis.

Throughout paper, we use $p(X)$ for the data density, $q(X)$ for the model density, $p(Y)$ for the prior or feature density, $p(Y|X)$ for the encoder, and $q(X|Y)$ for the decoder if there are any. The general objective in Bayes's analysis can be described as approximating a data density $p(X)$ with a model density $q(X)$, preferably a Gaussian mixture $q(X) = \int q(Y)\mathcal{N}(X - m(Y)) dY$, through defining a bound, a cost, empirical estimation and optimization, using the inequality of Kullback–Leibler divergence. Standard methods often use the following two bounds, the bound for $q(X)$ and the bound for $q(X|Y)$:
\begin{equation}
\resizebox{.77\linewidth}{!}{
$\begin{gathered}
\int p(X)\log q(X) dX \leq \int p(X) \log p(X) dX,
\end{gathered}$}
\label{cross_entropy_3}
\end{equation}
\begin{equation}
\resizebox{1\linewidth}{!}{
$\begin{aligned}
\iint p(X,Y) \log q(X|Y) dXdY &\leq \iint p(X,Y)\log p(X|Y) dXdY.
\end{aligned}$}
\label{conditional_entropy_bound_3}
\end{equation}
Eq.~\eqref{cross_entropy_3} says that the negative value of the cross entropy between $p(X)$ and $q(X)$ is upper bounded by the negative value of the data entropy itself, and the bound is tight only when $q(X)= p(X)$ a.e. $p$. Directly using the left side of~\eqref{cross_entropy_3} as a cost for gradient ascent is called the Mixture Density Network (MDN). 

However, since $q(X)$ is an integral over the prior $q(Y)$, taking the $\log$ of $q(X)$ makes the optimization more nonconvex, i.e., it needs to maximize the $\log$ of a sum. Observe that $q(X|Y) = \mathcal{N}(X-m(Y))$ is a single Gaussian, taking the $\log$ of $q(X|Y)$ becomes an $L_2$ distance. Thus a common practice is to use the bound for $q(X|Y)$ as in Eq.~\eqref{conditional_entropy_bound_3}. The left side of~\eqref{conditional_entropy_bound_3} is the MSE. The right side of~\eqref{conditional_entropy_bound_3} is the negative value of the conditional entropy of $X$ given $Y$ for $p$. 

For an autoencoder, the two conditional densities are Gaussians: $p(Y|X) = \mathcal{N}(Y-\textbf{E}(X))$ and $q(X|Y) = \mathcal{N}(X-\textbf{D}(Y))$, where $\textbf{E}$ and $\textbf{D}$ are two deterministic neural networks. The cost $\iint p(X)p(Y|X)\log q(X|Y) dXdY$ describes the data $p(X)$ being fed through an encoder $p(Y|X)$ and a decoder $q(X|Y)$, reduced to the mean-squared error since $q(X|Y) = \mathcal{N}(X-\textbf{D}(Y))$ is parameterized by a Gaussian.\vspace{3pt}

\begin{enumerate}[leftmargin=*]
\item The bound is tight only when $q(X|Y) = p(X|Y)$;\vspace{5pt}
\item Since $q(Y) = p(Y)$ is the density of the features, which is shared by two cascaded networks, when the bound is tight and $q(X|Y) = p(X|Y)$, we have $q(X,Y) = p(X,Y)$ and thus $q(X) = p(X)$;\vspace{5pt}
\item When the bound is tight and $q(X|Y) = p(X|Y)$, maximizing the cost further maximizes its upper bound, the negative value of the conditional entropy, which is equivalent to mutual information because the data entropy is fixed. This leads to the established result that in an architecture of two cascading neural networks, minimizing the error between the input and the output also maximizes the dependence between the input and the middle or features.\vspace{5pt}
\item The evidence lower bound (ELBO) is a special case that does not have this maximal mutual information property, which we explain further in the Appendix.\vspace{5pt}
\end{enumerate}
\noindent A further clarification for these probability densities is\vspace{3pt}
\begin{enumerate}[leftmargin=*]
\item $p(Y|X) = \mathcal{N}(Y-\textbf{E}(X))$ is a Gaussian over the encoder;\vspace{3pt}
\item $q(X|Y) = \mathcal{N}(X-\textbf{D}(Y))$ is a Gaussian over the decoder;\vspace{3pt}
\item $p(X|Y)$ is applying Bayes's rule to $p(Y|X)$;\vspace{3pt}
\item Ideally $q(X|Y)$ and $p(X|Y)$ should match.\vspace{3pt}
\item When they match, maximizing the cost further maximizes the upper bound, which gives the maximal mutual information solution.\vspace{5pt}
\end{enumerate}
\noindent In this paper, our discussion primarily revolves around the divergence bound for $q(X)$ in MDNs and the divergence bound for $q(X|Y)$ in autoencoders. A brief discussion about the use of ELBO is provided in the Appendix. Our method can also derive a simplified version of ELBO to achieve the same effect. However, we acknowledge that ELBO requires the additivity of $\log$ and cannot simply be substituted. 

The most obvious advantage of taking the $\log$ of $q(X|Y)$ is that it transforms the cost into the MSE, which simplifies optimization. A more subtle advantage is that neither the bound Eq.~\eqref{cross_entropy_3} nor~\eqref{conditional_entropy_bound_3} involves the entropy from $q$, because the Kullback-Leibler divergence as a distance measure does not necessitate any entropy from $q$. The divergence is over $p(X)$, and the optimal condition is that the error is minimized almost everywhere relative to $p(X)$, thereby justifying the empirical optimization of the cost. However, this also shows the disadvantage: since the entropy of $q$ is not involved, the shape of $q$ is not regularized. For example, in autoencoders, a model may find a trivial solution that makes all samples go to the global mean if parameters are not properly chosen. Or when the feature dimension is insufficient, the samples often appear blurry, seemingly mixing two samples close in the sample space. In generative models, a model may miss certain parts of the distributions or simply produce a delta function. Since no regularizations of $q$ are in the cost of Eq.~\eqref{cross_entropy_3} and~\eqref{conditional_entropy_bound_3}, the model may be unable to avoid this trivial delta-function solution or emphasize sample diversity. The implicit constraint is that the $\log$ of a probability density must be greater than $0$ a.e. $p$, which can be a loose constraint since we often add a small constant in the $\log$ to ensure numerical stability in implementations. However, the tradeoff is that if we add an extra term for regularizing $q$, then we need a batch of samples to estimate this term, so the optimization will no longer be purely empirical over the sample distribution $p$.\vspace{5pt}

For this consideration, our approach addresses two issues:\vspace{3pt}
\begin{enumerate}[leftmargin=*]
\item Whether we can consider the regularization or the shape of $q$ to avoid the trivial delta-function solution and consider sample diversity;\vspace{3pt}
\item In autoencoders, the bound is tight only when $q(X|Y) = p(X|Y)$; whether our new formulation will increase the tightness of this bound.\vspace{5pt}
\end{enumerate}

\noindent The concepts of divergence and entropy in a Hilbert space are constructed using the inner product and the norms. The bounds equivalent to Eq.~\eqref{cross_entropy_3} and Eq.~\eqref{conditional_entropy_bound_3} are based on the Cauchy-Schwarz inequality:
\begin{equation}
\begin{aligned}
\frac{\langle p, q\rangle^2}{\langle q, q\rangle} \leq {\langle p, p\rangle}.
\end{aligned}
\label{cs_inequality_2}
\end{equation}
\begin{equation}
\begin{aligned}
\frac{\langle p(X|Y), q(X|Y)  \rangle_{p(Y)} ^2}{||q(X|Y)||_{p(Y)}^2} \leq ||p(X|Y)||_{p(Y)}^2.
\end{aligned}
\label{conditional_cs_inequality_2}
\end{equation}
Eq.~\eqref{cs_inequality_2} is the bound for $q(X)$, equivalent to Eq.~\eqref{cross_entropy_3}; Eq.~\eqref{conditional_cs_inequality_2} is the bound for $q(X|Y)$, equivalent to Eq.~\eqref{conditional_entropy_bound_3}. The inner product and norms in Eq.~\eqref{cs_inequality_2} can be assumed to be taken over the Lebesgue measure $\mu$ for simplicity, whereas in Eq.~\eqref{conditional_cs_inequality_2} they are over the feature distribution of $p(Y)$. The inner product $\langle p(X|Y), q(X|Y) \rangle_{p(Y)}$ is simply $\iint p(X|Y)q(X|Y) p(Y) dXdY$, with the norms calculated similarly. The left sides of the two bounds are the cost. The right sides are the upper bound, the norm of $p$ or $p(X|Y)$.

The costs in these two bounds now depend on the norm of $q$ or $q(X|Y)$, which is missing in the bounds with Kullback-Leibler divergence. A bound like this in $L_2$ with a norm in the denominator is often used to prevent trivial delta-function-like solutions for neural network outputs. In contrastive and self-supervised learning, normalizing multidimensional features avoids the trivial constant solution, which we have demonstrated to be related to the orthonormal decomposition of the density ratio. Similar scenarios include batch normalization to prevent gradients from vanishing, and the penalties for the entropy to prevent "mode collapse" in generative models. The entropy penalties in particular are often not well theoretically justified, while our approach differs. 

Our proposal is consistent with the advantages of the $L_2$ bound. For approximating $q(X)$ in MDNs, we demonstrate that when approximating a data density with a neural network, using the inner product and norm form of the bound can increase sample diversity and ensure that the model does not produce a trivial delta-function-like solution. For approximating $q(X|Y)$ in autoencoders, we design a new structure in which the decoder is no longer a single Gaussian over the decoder, but a mixture. We argue that this encoder-mixture-decoder structure potentially makes the bound tighter compared to traditional autoencoders. We explain our paper as follows:\vspace{3pt}
\begin{enumerate}[leftmargin=*]
\item The inner product of two Gaussian mixtures and the norm of a Gaussian mixture both have closed forms, determined only by their means and variances. Using this property, when $q(X)$ or $q(X|Y)$ are parameterized as a Gaussian mixture, the terms in the cost of the bounds can be computed empirically. With the extra normalization term in the denominator, we aim to show that it increases the sample diversity and ensures numerical stability;\vspace{3pt}
\item For approximating $q(X)$ in MDNs, we compare the standard $\log$ bound (Eq.~\eqref{cross_entropy_3}), the Hilbert space normalized inner product bound (Eq.~\eqref{cs_inequality_2}), and the nuclear norm of the Gaussian linear operator. We find that the nuclear norm performs the best, followed by the Hilbert space bound, and lastly the $\log$ bound. Our findings decisively show that the two decomposition-involved costs, the normalized inner product and the nuclear norm form, outperform the regular $\log$ bound;\vspace{3pt}
\item For approximating $q(X|Y)$ in autoencoders, we have discussed that the optimal condition requires $q(X|Y)$ of the decoder to match $p(X|Y)$ of the encoder such that the bound is tight, and further maximizing the cost maximizes its upper bound, until finding the minimal conditional entropy of $X$ given $Y$ for $p$. And the question is whether $q(X|Y)$ and $p(X|Y)$ can match. That is, when $p(Y|X)$ is a Gaussian over the encoder, whether $p(X|Y)$ after applying the Bayes's rule can be approximated by the decoder $q(X|Y)$, a Gaussian over the decoder. A simple illustration is given in Fig.~\ref{figure_illustration_3}.\vspace{5pt}

Since the decoder network $\textbf{D}$ is deterministic, $\textbf{D}(Y)$ will lie on a manifold in the sample space bounded by the dimensionality of $Y$. Then the density $q(X|Y) = \mathcal{N}(X-\textbf{D}(Y))$ is only this manifold with an additive Gaussian noise. Suppose the feature dimension is significantly insufficient, for example, if the data distribution is highly complex and the feature is only in 1D, then the reconstruction lies on a 1D manifold. This means that $q(X|Y)$ and $p(X|Y)$ can be far apart, so the bound is not tight. The deterministic neural network is many-to-one or, at most, one-to-one, which means that the entropy or the cardinality must decrease through the two networks.\vspace{5pt}

Our proposed fix is to change $q(X|Y)$ from a single Gaussian over the decoder to a mixture $q(X|Y) = \int p(c)\mathcal{N}(X-\textbf{D}(Y, c)) dc$. Now, the decoder $\textbf{D}$ receives not only features $Y$, but also a prior noise variable $\mathbf{c}$. By sampling multiple $\mathbf{c}$, the decoder can produce multiple outputs from one $Y$, essentially creating a one-to-many mapping. It is evident that this mixture form is more universal as a function approximator than a single Gaussian, so $q(X|Y)$ and $p(X|Y)$ can potentially be closer, making the bound tighter.\vspace{5pt}

After this change, the regular $\log$ bound Eq.~\eqref{conditional_entropy_bound_3} may no longer be useful since $q(Y|X)$ is now a mixture, so the cost cannot be reduced to the mean-squared error. But the normalized inner product bound Eq.~\eqref{conditional_cs_inequality_2} may become more fitting, as we can still use the property that the inner product and norms of Gaussian mixtures has a closed form. The nuclear norm form may no longer apply since it requires performing singular value decomposition for each image sample.\vspace{5pt}

Our results indeed show that this encoder-mixture-decoder structure can generate diverse samples from one sample when the model bears high uncertainty. We use simple datasets and a subset of the image dataset to verify that the bound becomes tighter, although it is still difficult to verify on the full image dataset.\vspace{5pt}

\begin{figure}[h]
  \centering
\includegraphics[width=1\linewidth]{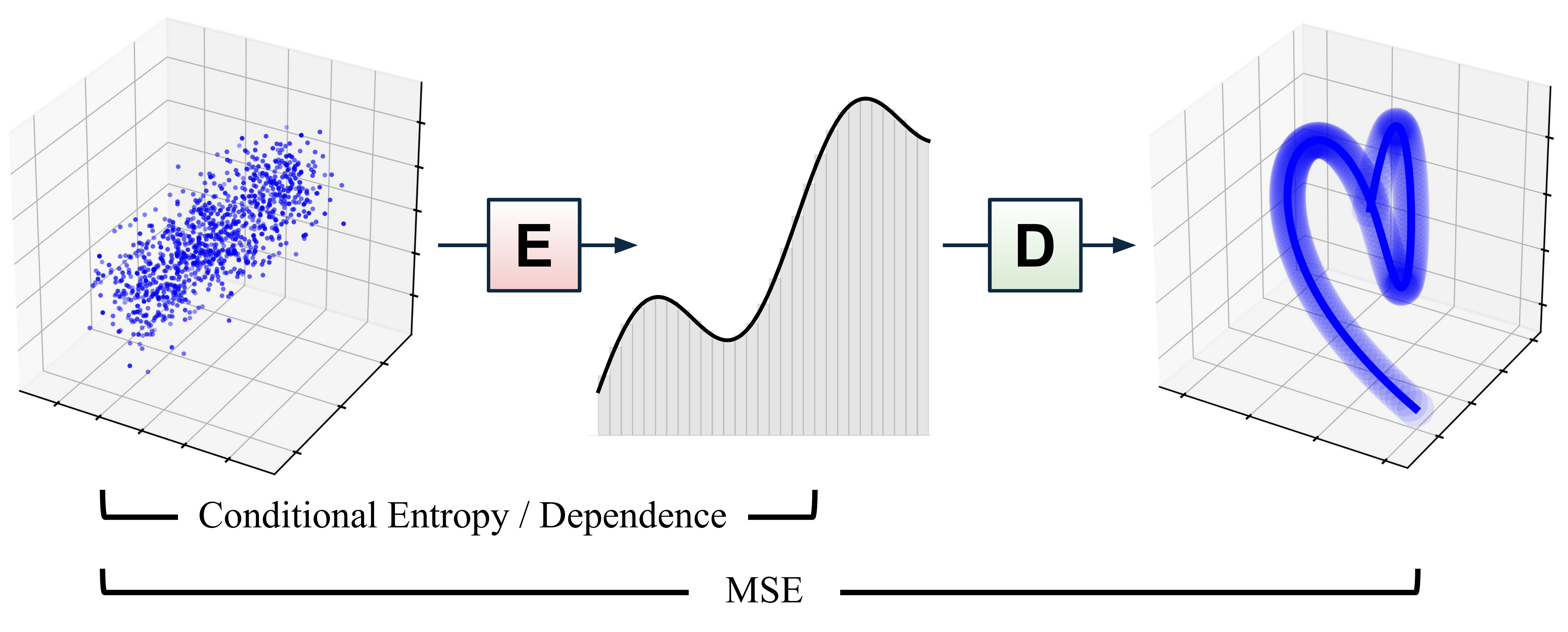}\vspace{-3pt}
\caption{The reconstruction lies on a manifold in the sample space with an additive Gaussian noise, bounded by the dimensionality of the features. If the feature dimension is insufficient, $q(X|Y)$ and $p(X|Y)$ might not match. Changing $q(X|Y)$ from a single Gaussian to a mixture can potentially make the bound tighter for insufficient feature dimensions.}
\label{figure_illustration_3}
\end{figure}

\item In most of our experiments, we find it possible to verify whether the upper bound is tight and the optimal upper bound value (the entropy of the data or the minimal conditional entropy the encoder can reach). We illustrated
the methods with extensive quantitative analysis.\vspace{5pt}
\end{enumerate}

\noindent Since both the approximations of $q(X)$ and $q(X|Y)$ use the advantage of a normalized inner product bound in $L_2$, making a neural network with multidimensional outputs produce diverse samples and avoid the trivial delta-function-like solution, which in this case is a network producing Gaussian centers for a mixture for the marginal or conditional density, we name the two bounds the contrastive entropy bound and the contrastive conditional entropy bound.

We will start with the properties of Gaussian mixtures in $L_2$ that their inner product and norm have closed forms. Then we will move to the comparisons of approximating $q(X)$ with different bounds and training an encoder-mixture-decoder architecture.

\newpage

\section{Inspirations from Autoencoders}\vspace{5pt}
\label{inspiration_from_autoencoders}

We begin our discussion on the standard autoencoder and draw inspiration from it. The discrete and continuous cases of an autoencoder can be summarized as the following two optimization objectives:

\begin{equation}
\begin{gathered}
\max_{\mathbf{P},\mathbf{Q}} \;\; {Trace}\left({diag}(\mathbf{P}_X) \mathbf{P}\log \mathbf{Q}\right).\vspace{-5pt}
\end{gathered}
\label{discrete_case}
\end{equation}
\begin{equation}
\begin{gathered}
\max_{p(Y|X), p(X|Y)}\; \iint p(X) \cdot p(Y|X)\log q(X|Y) dX dY.\vspace{6pt}
\end{gathered}
\label{continuous_case}
\end{equation}

\noindent The discrete case~\eqref{discrete_case} can be described as starting with the data density $\mathbf{P}_X$, supposedly a vector diagonalized into a matrix, fed through an ``encoder matrix'' and a ``decoder matrix''. Two matrices are Markovian and have to satisfy the Markov property, with columns summing to $1$ for each row. 

\begin{figure}[h]
\centering
\includegraphics[width=1\linewidth]{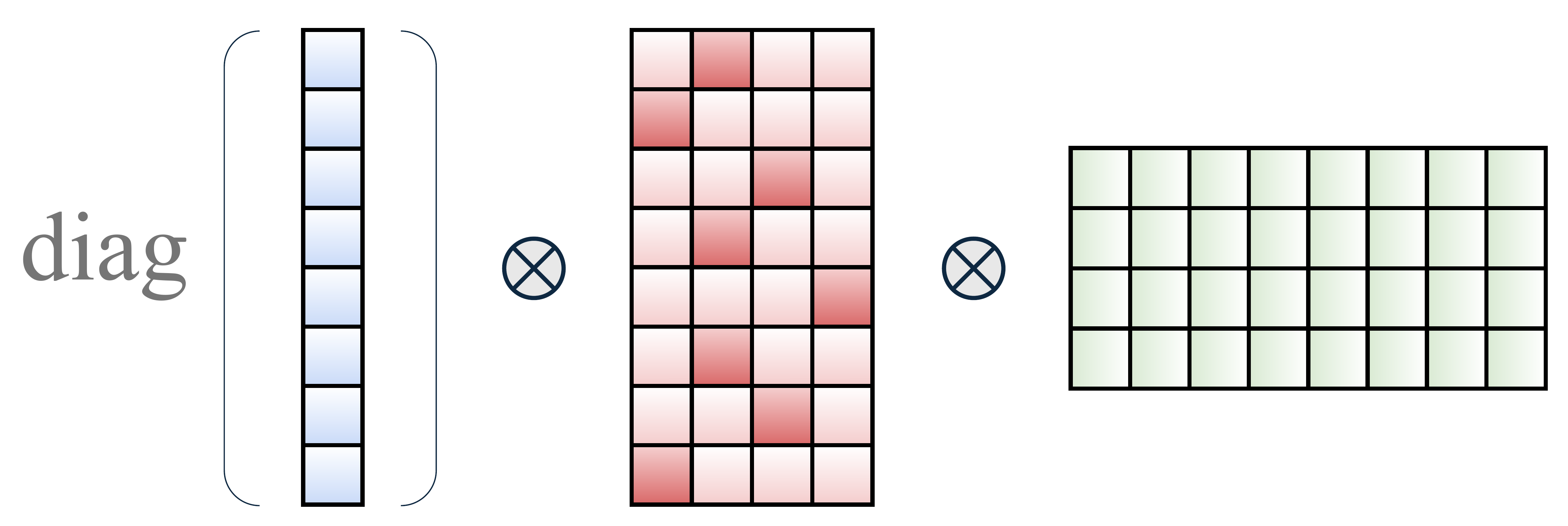}
\caption{A discrete equivalent of an autoencoder is to feed a diagonalized density vector of $p(X)$ through an encoder and a decoder matrix, both Markovian. The mean-squared error is the average of the errors between a sample's reconstruction and the sample itself, so it is the trace of the final matrix after the matrix product. As neural networks are deterministic, the Markov matrices are very sparse, for example with just one positive entry in each row.}
\label{fig1}
\end{figure}

Without the trace, the $i$-th row and $j$-th column of the matrix inside the trace is a score, which is the negative value of the mean-squared error between the sample $X=i$ and a possible reconstruction $X'=j$. In autoencoders, we only care about the diagonal entries, i.e., the error between a sample's reconstruction and the sample itself, essentially the matrix trace. A simple diagram is in Fig.~\ref{fig1}. 

\noindent We give the following remarks from this simple example:\vspace{3pt}
\begin{enumerate}[leftmargin=*]
\item \textit{\textbf{Transition matrices are sparse.}} Neural networks are often deterministic, meaning that their mapping is either many-to-one, or at most one-to-one with additive Gaussian noise. Correspondingly the two Markov matrices in the discrete case will be sparse, for example with each row having only one positive entry. That is, in practice they are commonly rank-one parameterized by a vector.

\vspace{7pt}
\item \textit{\textbf{If no constraint, the optimal $\mathbf{Q}$ is from $\mathbf{P}$ thus not needed.}} If the Markovian matrices are allowed to be arbitrary, not restricted to be sparse and many-to-one, then the optimal $\mathbf{Q}$ would be by applying the Bayes's rule to $\mathbf{P}$ and get $\mathbf{Q}^*(X=i|Y=j) = \mathbf{P}(X=i|Y=j)$ given by the inequality of Kullback-Leibler divergence. So one can directly set $\mathbf{Q}^*$ to be this conditional from $\mathbf{P}$, not requiring any optimization. 

\vspace{7pt}
\item \textit{\textbf{If no constraint, the upper bound is the negative value of the conditional entropy.}} After setting $\mathbf{Q}^*$ to be the conditional from $\mathbf{P}$, the cost is no longer relevant to $\mathbf{Q}$ but only depends on the encoder $\mathbf{P}$, and becomes the negative value of the conditional entropy of $Y$ given $X$ for $\mathbf{P}$. Further maximizing the cost becomes maximizing this upper bound w.r.t. the encoder $\mathbf{P}$. Thus the optimal condition becomes finding features $Y$ with an encoder $\mathbf{P}$ with the maximal possible conditional entropy, or equivalently the maximal mutual information since the data entropy is fixed.

\vspace{7pt}
\item \textit{\textbf{The $\log$ can be omitted.}} This can be explained in two ways. First, minimizing mean squared error and minimizing Gaussian differences in an autoencoder often have similar results, which is the case without taking the $\log$ of $\mathbf{Q}$. A second way is looking at the Cauchy-Schwarz inequality (Eq.~\eqref{conditional_cs_inequality_2}), which introduces an extra norm term ${||q(X|Y)||_{p(Y)}^2} = \iint q^2(X|Y) q(Y) dX dY$ in the denominator compared to the Kullback-Leibler divergence. In an autoencoder when $q(X|Y)$ is parameterized as a single Gaussian, $q^2(X|Y)$ is also Gaussian, and thus $\int q^2(X|Y) dX$ becomes a constant. Our paper will provide an extensive discussion on this.\vspace{3pt}

\;\; The advantage is that if we create a matrix $\mathbf{P}_{X,X'} = {diag}(\mathbf{P}_X) \mathbf{P}\log \mathbf{Q}$, since both $\mathbf{P}$ and $\mathbf{Q}$ are Markovian, this matrix $\mathbf{P}_{X,X'}$ is essentially a joint pdf between samples $X$ and reconstructions $X'$. Then maximizing its trace is making this joint pdf diagonalized.\vspace{7pt}

\item \textit{\textbf{Maximizing the trace is maximizing the eigenvalues.}} The trace of a matrix is the sum of its eigenvalues. This inspires us to consider that in many applications, such as MDNs or generative models, where we want to match two distributions without a one-to-one correspondence from samples to reconstructions, whether we can maximize the overall rank, not the trace or the sum of eigenvalues as doing so would only enforce the diagonal elements to be dominant.\vspace{5pt}
\end{enumerate}

The continuous case Eq.~\eqref{continuous_case} is an extension of the discrete case. The difference is that the encoder $p(Y|X)$ and the decoder $q(X|Y)$ are restricted to \textbf{\textit{a single Gaussian over centers}}: $p(Y|X) = \mathcal{N}(Y-\textbf{E}(X))$ and $q(X|Y) = \mathcal{N}(X-\textbf{D}(Y))$, where $\textbf{E}$ and $\textbf{D}$ are two deterministic neural networks. So in the continuous case, the two conditionals are restricted to a very special class of functions and cannot be set arbitrarily, even though we often analyze the bound assuming the functions to be arbitrary.


\vspace{5pt}

Considering these, we have the following propositions:
\begin{enumerate}[leftmargin=*]
\item In applications like MDNs or generative models when we want to match a distribution with another distribution, without requiring a one-to-one correspondence from samples to reconstructions as in autoencoders, we should be able to maximize the overall rank of a Gaussian Gram matrix to achieve this, contrasting with maximizing the trace or the sum of eigenvalues.\vspace{3pt}

\;\; This inspires us to propose maximizing the nuclear norm in Sec.~\ref{nuclear_norm_section}. \vspace{5pt}

\item Taking out the $\log$ has the advantage of turning the cost into the diagonal entries of a joint pdf, which is permissible because of the Cauchy-Schwarz inequality. But if the Gaussian variances are trainable, or the functions are not parameterized by a single Gaussian (a mixture, for example), then this extra norm is not a constant and must be included in optimization.\vspace{3pt}

\;\; This is addressed in Sec.~\ref{inner_product_norm}, where we propose a similar bound but using inner products and norms in a Hilbert space. We find that the extra norm term also effectively increases sample diversity and helps avoid the trivial solution, but the trade-off is that it requires a batch of samples to estimate.\vspace{5pt}

\item The analysis on the conditional entropy and optimal conditions is based on the assumption that the two matrices $\mathbf{P}$ and $\mathbf{Q}$, or $p(Y|X)$ and $q(X|Y)$ in the continuous case can be arbitrary. But in practice they are restricted to a single Gaussian over centers.\vspace{3pt}

\;\; So we propose that if we change $p(Y|X)$ and $q(X|Y)$ into Gaussian mixtures, which undoubtedly have greater representation power than single Gaussians, the bounds might become tight. This is especially important for the decoder: since the encoder $p(Y|X)$ will likely be many-to-one, the $p(X|Y)$ after applying the Bayes's rule will likely be one-to-many. Thus for the decoder $q(X|Y)$ to approximate $p(X|Y)$ such that the bound is tight, it also has to be one-to-many. Then parameterizing $q(X|Y)$ as a mixture of Gaussians becomes reasonable.\vspace{3pt}

\;\; This is addressed in Sec.~\ref{encoder_mixture_decoder}, where we propose a new encoder-mixture-decoder architecture. The decoder is multiple-output, defining a mixture density, consistent with the inner product and norm form of the bound we defined in Sec.~\ref{inner_product_norm}, which is the only approach that achieves a mapping from one sample to multiple samples by a mixture decoder.\vspace{7pt}
\end{enumerate}

\noindent In this paper, we propose two costs: one with the nuclear norm and one with the inner product and norm in the Hilbert space, both of which explicitly exploit contrastivity. The former works for MDNs. The latter works for both MDNs and the proposed encoder-mixture-decoder architecture, which we refer as the contrastive entropy and conditional entropy bound, because the cost will be bounded by the norm of the data density, a form of entropy.

\newpage\;\newpage
\vspace{10pt}
\section{MDNs: Using the Nuclear Norm}\label{nuclear_norm_section}

\subsection{Trace for Autoencoders; Nuclear Norm for MDNs}

Following our proposition, we maximize the sum of eigenvalues to train an autoencoder and maximize the nuclear norm, the sum of singular values to train a generative model such as a Mixture Density Network (MDN).

We choose the Gaussian cross Gram matrix to decompose. Suppose $X$ is the data samples, paired with $X'$ as an autoencoder's output, their Gaussian difference is a conditional probability $p(X'|X) = \int p(Y|X)p(X'|Y) dY \approx \mathcal{N}(X' - \textbf{D}(\textbf{E}(X)))$, which describes the probability of the output landing on a reconstruction $X'$. The objective of an autoencoder, in the case when the $\log$ is omitted, is a sum over the Gaussian difference:\vspace{3pt}
\begin{equation}
\begin{aligned}
&\iint p(X'|X) \cdot \mathbf{1}\{X'=X\}\cdot p(X) dX'dX  \\ 
&= \iint p(X,X') \cdot \mathbf{1}\{X'=X\} dX'dX\\
&= \int \mathcal{N}(X-\textbf{D}(\textbf{E}(X))) \cdot p(X) dX \approx \frac{1}{N}\sum_{n=1}^N \mathcal{N}(X_n-X'_n).
\end{aligned}
\end{equation}
\noindent That is, the objective only looks at the diagonal entries of the joint $p(X, X')$ when $X' = X$. For a batch of samples $X_1, X_2, \cdots, X_N$ and their reconstructions $X_1', X_2', \cdots, X_N'$, the objective can be empirically approximated by the sum over their Gaussian differences, which is the trace of a Gaussian cross Gram matrix. Thus it is the matrix we should decompose.

Now suppose $X_1, X_2, \cdots, X_N$ and $X_1', X_2', \cdots, X'_N$ are arbitrary, not necessarily from an autoencoder, we construct the matrix of $L_2$ differences $\mathbf{M}$, and the exponential of the negative $\mathbf{M}$ as $\mathbf{K}$:
\begin{enumerate}[leftmargin=*]
\item To match each $X_n$ with $X_n'$ with an one-to-one correspondence, like in autoencoders, we maximize the sum of eigenvalues of $\mathbf{K}$ as in Eq.~\eqref{eigenvalues}, which is the trace of $\mathbf{K}$ so the actual eigendecomposition is unnecessary.\vspace{5pt}
\item To match the distribution of $X$ with the distribution of $X'$ without an one-to-one correspondence, like in MDNs, we maximize the sum of singular values, the nuclear norm of $\mathbf{K}$ in Eq.~\eqref{singular_values}.\vspace{5pt}
\item The SVD in Eq.~\eqref{singular_values} is necessary. We tested the Frobenius norm (the trace of $\mathbf{K}\mathbf{K}^\intercal$) and it was not effective. The SVD must also be done on the Gaussian Gram matrix. We tested decomposing the matrix of $L_2$ distances, which was also not effective.
\end{enumerate}

\begin{equation}
\resizebox{1\linewidth}{!}{
$\begin{gathered}
\mathbf{M} \coloneqq
\frac{1}{d_X}\begin{bmatrix}
||X_1-X_1'||_2^2 & \cdots & ||X_1-X_N'||_2^2 \\
\vdots      & \ddots & \vdots  \\
||X_N-X_1'||_2^2 & \cdots & ||X_N-X_N'||_2^2  
\end{bmatrix}, \;\mathbf{K} \coloneqq \exp(-\frac{1}{2v_X}\mathbf{M}).
\end{gathered}$}
\label{gram_matrices}
\end{equation}

\begin{equation}
\resizebox{1\linewidth}{!}{
$\begin{gathered}
\mathbf{K} = \mathbf{Q}\mathbf{\Lambda}\mathbf{Q}^{-1}, \; \mathbf{\Lambda} = \begin{bmatrix}
    \lambda_{1} &  \\
    & \ddots  \\
    & & \lambda_{N}
\end{bmatrix},\; \max\;\;\sum_{n=1}^N \lambda_n:= Trace(\mathbf{K}).
\end{gathered}$}
\label{eigenvalues}
\end{equation}

\begin{equation}
\begin{gathered}
\mathbf{K}  = \mathbf{U}\mathbf{S}\mathbf{V},\;
\mathbf{U}\mathbf{U}^\intercal = \mathbf{I}, \; \mathbf{V}\mathbf{V}^\intercal = \mathbf{I}, \; \mathbf{S} = \begin{bmatrix}
    \sigma_{1} &  \\
    & \ddots  \\
    & & \sigma_{N}
\end{bmatrix}, \vspace{-10pt} \\ 
\text{max} \;\; \sum_{n=1}^N \sigma_n.
\end{gathered}\vspace{10pt}
\label{singular_values}
\end{equation}

\noindent \textbf{How to Train MDNs.} First sample $u_1,u_2,\cdots,u_N$, a batch of noise from a prior distribution, which could be uniform, Gaussian, or hybrid noise created by concatenating continuous noises with one-hot vectors sampled from a discrete categorical distribution. Next, use a neural network to map the noises to generated samples $X_1',X_2',\cdots,X_N'$ in the sample space, then construct the Gram matrix $\mathbf{K}$ and maximize its singular values through gradient ascent. 

\vspace{10pt}

\noindent \textbf{Numerical Stability.} We need to divide the $L_2$ distances by the data dimension $d_X$, for example the image size or the number of pixels. This means calculating the mean over the data dimension in the Gaussian's $\exp$, not the sum. The Gaussian difference $v_X$ is also a hyperparameter that needs to be chosen. We ignore the scaling factor in front of the Gaussian pdf, as for high-dimensional data, that factor will be arbitrarily small. See Appendix~\ref{algorithm_appendix} for the full algorithm and implementation details.

\vspace{10pt}

\noindent \textbf{Analysis of the Optimal Solution.} Appendix~\ref{Section_2_nuclear_norm} provides an extensive analysis in terms of the theory. The conclusion we obtain can be summarized as follows: Decomposing $\mathbf{K}$ is equivalent to decomposing the continuous Gaussian functions $\mathcal{N}(X-X')$ with two sets of basis functions orthonormal w.r.t. $p(X)$ and $q(X)$, the probability measures associated with the data and model distributions. It is also equivalent to decomposing the function $\sqrt{p(X)}\mathcal{N}(X-X')\sqrt{q(X')}$, with basis functions orthonormal w.r.t. the Lebesgue measure $\mu$.

We find that the easiest way to analyze this is by assuming the discretized version of $p(X)$, $\mathcal{N}(X-X')$, and $q(X)$ using the Nyström method, denoted as $\mathbf{P}_X$, $\mathbf{N}_{XX'}$, and $\mathbf{Q}_{X}$. Then the decomposition becomes the SVD of the matrix ${diag}(\sqrt{\mathbf{P}_X}) \mathbf{N}_{XX'} diag(\sqrt{\mathbf{Q}_{X}})$. Since the matrix $\mathbf{N}_{XX'}$ is Hermitian, we can decompose it with $\mathbf{N}_{XX'} = \mathbf{Q}_{\mathbf{N}}\mathbf{\Lambda}_{\mathbf{N}}\mathbf{Q}_{\mathbf{N}}$. 

Define $\mathbf{A}:= diag(\sqrt{\mathbf{P}_X}) \mathbf{Q}_{\mathbf{N}} \mathbf{\Lambda}_{\mathbf{N}}^{\frac{1}{2}}$ and $\mathbf{B} :=  diag(\sqrt{\mathbf{Q}_{X}}) \mathbf{Q}_{\mathbf{N}} \mathbf{\Lambda}_{\mathbf{N}}^{\frac{1}{2}}$, using the inequality of the nuclear norm, we have\vspace{3pt}
\begin{equation}
\begin{aligned}
& \hspace{-6pt} ||\mathbf{A}\mathbf{B}^\intercal||_*  \leq \sqrt{||\mathbf{A}\mathbf{A}^\intercal||_*} \cdot \sqrt{||\mathbf{B}\mathbf{B}^\intercal||_*}, \\
||\mathbf{A}\mathbf{A}^\intercal||_* &=  ||{diag}(\sqrt{\mathbf{P}_X}) \mathbf{N}_{XX'} diag(\sqrt{\mathbf{P}_{X}})||_* \\
&= Trace({diag}(\sqrt{\mathbf{P}_X}) \mathbf{N}_{XX'} diag(\sqrt{\mathbf{P}_{X}})) \\
&= \mathcal{N}(0) = ||\mathbf{B}\mathbf{B}^\intercal||_*. \\
\end{aligned}\vspace{3pt}
\label{bound_AB}
\end{equation}

\noindent That is, the nuclear norm of the defined matrix ${diag}(\sqrt{\mathbf{P}_X}) \mathbf{N}_{XX'} diag(\sqrt{\mathbf{Q}_{X}})$ is upper bounded by the center of the Gaussian $\mathcal{N}(0)$, and the bound is tight when $\mathbf{A} = \mathbf{B}$, i.e., when $diag(\sqrt{\mathbf{P}_X}) \mathbf{Q}_{\mathbf{N}} =  diag(\sqrt{\mathbf{Q}_{X}}) \mathbf{Q}_{\mathbf{N}}$ for all positive eigenvalues. 

Eq.~\eqref{bound_AB} uses the fact that $\mathbf{A}\mathbf{A}^\intercal$ is Hermitian, so its nuclear norm is the sum of eigenvalues, which is the trace. The diagonal of $\mathbf{A}\mathbf{A}^\intercal$ is the constant $\mathcal{N}(0)$ multiplied by the density $\mathbf{P}_X$, so the trace, which is integrating over $\mathbf{P}_X$, is $\mathcal{N}(0)$. Considering the constant gap in discretization from the Nyström method and that our cost ignores the scaling factor in the Gaussian pdf, the singular values of the Gram matrix $\mathbf{K}$ in our cost will be bounded by $N$, the number of samples in a batch. 

The eigenvectors in the optimal conditional can also be analyzed. Suppose the Gaussian variance used in the cost for $\mathbf{K}$ is $v_X$. Initiating a Gaussian with a variance of $\frac{v_X}{2}$, denoted as $\mathcal{N}(X-X';\frac{v_X}{2})$, it can be shown that the left singular functions for $\sqrt{p(X)\cdot\mathcal{N}(X-X';\frac{v_X}{2})}$ are the optimal eigenfunctions for $\sqrt{p(X)}\mathcal{N}(X-X';v_X)\sqrt{p(X’)}$. Suppose the top singular values of the former are $\sqrt{\sigma_1}, \sqrt{\sigma_2},\cdots$, the top singular values for the latter are ${\sigma_1}, {\sigma_2},\cdots$, and we have shown that the sum of singular values is a constant $\mathcal{N}(0)$. Proving this uses the fact that $\mathcal{N}(X-X';v_X) = \int \mathcal{N}(X-X'';\frac{v_X}{2}) \mathcal{N}(X'-X'';\frac{v_X}{2}) dX''$. That is, the Gaussian difference $\mathcal{N}(X-X';v_X)$ can be written as the convolution of two Gaussian functions, each with a variance of $\frac{v_X}{2}$. So we can decompose $\mathcal{N}(X-X')$ or the discrete $\mathbf{N}_{XX'}$ in this way, not through eigendecomposition. The full details are provided in Appendix~\ref{Section_2_nuclear_norm}.

\subsection{An ELBO Variation}
\label{sectionb_elbo}

This framework of SVD-based MDNs can also be applied to the scenario of the evidence lower bound (ELBO), although the forms are not exactly the same.\vspace{10pt}

\begin{figure}[t]
\centering
\includegraphics[width=.5\linewidth]{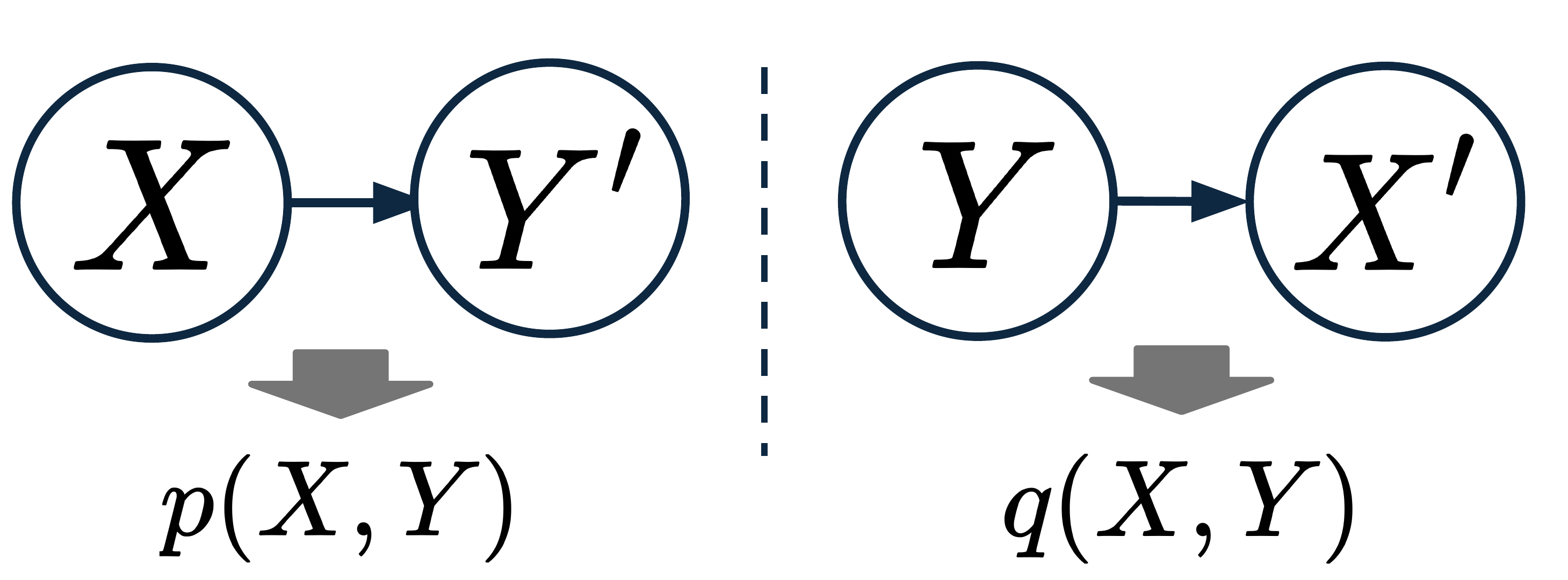}
\caption{We are only allowed to independently sample $X$ from the data distribution $p(X)$ and $Y$ from a chosen prior distribution $q(Y)$. The goal is to find deterministic mappings between them. The standard procedure is creating an encoder for $p(X,Y) = p(X)p(Y|X)$ and a decoder for $q(X,Y) = q(Y)q(X|Y)$, then minimize their divergence.}
\label{fig3_2}
\end{figure}

\noindent \textbf{The Standard ELBO.} The ELBO is applied in the following scenario, as illustrated in Fig.~\ref{fig3_2}. In addition to the data distribution $p(X)$ for $X$, we also need to pick a prior distribution $q(Y)$ for features $Y$. We can only sample them independently. And from the independent joint we want to find the deterministic mappings from $X$ to $Y$, from $Y$ to $X$, and create a dependent joint. 

Since we can only sample from $p(X)$ and $q(Y)$ independently, the standard procedure is to create an encoder first mapping $X$ to an auxiliary $Y'$ and a decoder mapping $Y$ to an auxiliary $X'$. In this way, the encoder defines $p(Y|X)$, creating a joint distribution $p(X,Y)=p(X)p(Y|X)$, representing the probabilistic system $p$. The decoder defines $q(X|Y)$, creating a joint distribution $q(X,Y)=q(Y)q(X|Y)$, representing the probabilistic system $q$.

It is not sufficient to minimize the divergence between $p(Y)$ and $q(Y)$ or between $p(X)$ and $q(X)$, as it does not enforce dependence. Instead, the cost of ELBO is minimizing the divergence between $p(X,Y)$ and $q(X,Y)$.\vspace{10pt}

\noindent \textbf{Our Approach.} We can use the decomposition framework to achieve this task of minimizing the divergence between $p(X,Y)$ and $q(X,Y)$. The idea is to sample $X,Y'$ from $p(X,Y)$ and $Y,X'$ from $q(X,Y)$ independently, view them as pairs, construct the Gaussian cross Gram matrix and decompose it. 


Suppose the encoder takes a batch of inputs $X_1,\cdots,X_N$ and produces outputs $Y_1',\cdots,Y_N'$. We then sample a batch of prior noises $Y_1,\cdots,Y_N$ as inputs for the decoder, which generates outputs $X_1',\cdots,X_N'$. We create two matrices of $L_2$ distances $\mathbf{M}_X$ and $\mathbf{M}_Y$, and construct the Gaussian Gram matrix of the joint $\mathbf{K}_{XY}$, illustrated in Eq.~\eqref{singular_value_elbo}. The objective is to maximize the nuclear norm, the sum of singular values of $\mathbf{K}_{XY}$.

\begin{equation}
\resizebox{1\linewidth}{!}{
$\begin{aligned}
&\;\;\;\mathbf{M}_X \coloneqq
\frac{1}{d_X}\begin{bmatrix}
\;\; ||X_1-X_1'||_2^2 & \cdots & ||X_1-X_N'||_2^2\;\;\; \\
\;\; \vdots      & \ddots & \vdots \;\;\;\; \\
\;\; ||X_N-X_1'||_2^2 & \cdots & ||X_N-X_N'||_2^2 \;\;\; 
\end{bmatrix},\\ \\
&\;\;\;\mathbf{M}_Y \coloneqq \frac{1}{d_Y}\begin{bmatrix}
\;\;\; ||Y_1-Y_1'||_2^2 & \cdots & \;\;||Y_1-Y_N'||_2^2\;\;\;\; \\
\;\;\; \vdots      & \ddots & \;\; \vdots \;\;\;\;\; \\
\;\;\; ||Y_N-Y_1'||_2^2 & \cdots & \;\;||Y_N-Y_N'||_2^2 \;\;\;\; 
\end{bmatrix},\\ \vspace{7pt}
&\mathbf{K}_{XY} = \exp(-\frac{1}{2v_X}\mathbf{M}_X + -\frac{1}{2v_Y}\mathbf{M}_Y),\; \text{max} \;\; \sum_{k=1}^K \sigma_k(\mathbf{K}_{XY}).\\
\end{aligned}$}
\label{singular_value_elbo}
\end{equation}

\noindent The implementation we found stable is scaling the two matrices of $L_2$ distances by their own data dimensions $d_X$ and $d_Y$, for example the image size and the feature dimension, and also scaling them by the chosen variances $v_X$ and $v_Y$, one for each of them, before taking the $\exp$. The variances $v_X$ and $v_Y$ do matter and may need to be different. The scaling factor in the Gaussian pdf is ignored since it is arbitrarily small in high dimensions. 

Our approach is not training a conventional autoencoder, strictly speaking. In a variational autoencoder trained with ELBO, a sample still needs to go through both the encoder and the decoder to generate reconstructions. There, minimizing the MSE with the reconstruction is still needed, which comes from the term $\iint p(X)p(Y|X)\log q(Y|X) dXdY$. In our approach, the inputs to the encoder are samples; the inputs to the decoder are noises sampled from the prior; the output of the encoder does not necessarily go through the decoder. How we sample from $p(X,Y)$ and $q(X,Y)$ are separate and independent. The dependence and interactions happen at calculating the $L_2$ distances and taking the exponential of the sum. There is no specific one-to-one correspondence between $X$ and $X'$ or $Y$ and $Y'$. 

The trade-off is that a batch of samples is required at each training step, and it might only find a suboptimal solution in terms of the full dataset. But we do observe that the generation quality and generalization capability improve as we train for more iterations.

\newpage 

\section{MDNs: Using Inner Products and Norms}\label{inner_product_norm}
 
\subsection{Inspirations}

The second approach we find worth mentioning is that in order to omit the $\log$ in the mean-squared error and optimizes with the Gaussians, one needs to apply the Cauchy-Schwarz inequality:
\begin{equation}
\begin{aligned}
\frac{\langle p, q\rangle^2}{\langle q, q\rangle} \leq {\langle p, p\rangle}.
\end{aligned}
\label{CS_inequality}
\end{equation}
Here $p$ is the density of the data and $q$ is the density of the model. So the square of the inner product, normalized by the norm of $q$, is upper-bounded by the norm of $p$. 

In the case where the denominator is a fixed constant, which occurs when $q$ is a single Gaussian as the norm of a Gaussian function is constant, optimizing just the numerator is enough. But if the denominator is not fixed, one needs to take the gradient of the entire ratio including the norm of $q$. This happens particularly when $q$ is a mixture of Gaussians:\vspace{3pt}
\begin{enumerate}[leftmargin=*]
\item In MDNs, the density $q$ is a mixture of Gaussians used to approximate $p$, and thus can use this cost;\vspace{5pt}
\item In autoencoders, the densities are conditional. Conventionally $q$ is a conditional $q(Y|X)$, which is a single Gaussian. In Sec.~\ref{encoder_mixture_decoder}, we propose an encoder-mixture-decoder architecture where the decoder becomes a Gaussian mixture as a one-to-many mapping, which can be trained with this cost;\vspace{5pt}
\item The density of a Gaussian mixture has a closed-form norm, determined only by the mean, variance, and weights, the parameters of the mixtures. The inner product of the densities between two Gaussian mixtures also has a closed form. This is why a Gaussian mixture is particularly suitable as $q$ in this cost.\vspace{5pt}
\end{enumerate}

\noindent The closed-form properties are as presented follows: Property~\ref{property_1} shows the inner product between two Gaussian functions; Property~\ref{property_2} shows closed forms for norms and inner products of Gaussian mixtures with discrete priors; Property~\ref{property_3} shows closed forms for norms and inner products of Gaussian mixtures with continuous priors.\vspace{5pt}

\begin{property}
Given two Gaussian density functions $f_1(X) = \mathcal{N}(X-m_1;v_1)$ and $f_2(X) = \mathcal{N}(X-m_2;v_2)$, their inner product has a closed form:
\begin{equation}
\begin{gathered}
\langle f_1, f_2 \rangle = \mathcal{N}(m_1-m_2; {v_1+v_2}).\vspace{5pt}
\end{gathered}
\end{equation}
\label{property_1}
\end{property}

\begin{property}
(Gaussian mixtures with discrete priors) Given a discrete Gaussian mixture $p(X) = \sum_{k=1}^K w_k \mathcal{N}(X - m_k;v_k)$, the $L_2$ norm of $p$ satisfies
\begin{equation}
\begin{gathered}
\langle p ,p\rangle = \sum_{i=1}^K \sum_{j=1}^K w_i w_j\mathcal{N}(m_i-m_j;v_i+v_j).
\end{gathered}
\end{equation}
Given another mixture $q(X) = \sum_{k=1}^{K'} w_k' \mathcal{N}(X - m_k';v_k')$, the inner product between $p$ and $q$ satisfies
\begin{equation}
\begin{gathered}
\langle p ,q\rangle = \sum_{i=1}^K \sum_{j=1}^{K'} w_i w_j'\mathcal{N}(m_i-m_j';v_i+v_j').
\end{gathered}
\end{equation}
\label{property_2}
\end{property}

\begin{property}
(Gaussian mixtures with any priors) Given a Gaussian mixture with a prior distribution $p(X) = \int p(c) w(c) \mathcal{N}(X - m(c);v(c)) dc$. The norm of $p(X)$ satisfies
\begin{equation}
\begin{aligned}
& \langle p, p\rangle = \iint p(c_1)p(c_2) w(c_1)w(c_2) \\ & \;\;\;\;\;\;\;\;\;\;\;\;\;\;\;\;\;\;\;\;\mathcal{N}(m(c_1)-m(c_2);v(c_1)+v(c_2)) dc_1dc_2.\vspace{15pt}
\end{aligned}
\end{equation}
\noindent Given another Gaussian mixture $q(X) = \int p'(c) w'(c)\mathcal{N}(X-m'(c);v'(c)) dc$, the inner product between them satisfies
\begin{equation}
\begin{aligned}
& \langle p, q\rangle = \iint p(c)p'(c') w(c)w'(c') \\ & \;\;\;\;\;\;\;\;\;\;\;\;\;\;\;\;\;\;\;\;\mathcal{N}(m(c)-m'(c');v(c)+v'(c')) dc dc'.\vspace{15pt}
\end{aligned}
\end{equation}
\label{property_3}
\end{property}

\begin{corollary}
The norm defined with the polynomial of any order of a Gaussian mixture, regardless of discrete or continuous prior, has a closed form.\vspace{5pt}
\end{corollary}

That is, the norm and inner products for Gaussian mixtures, whether their prior is discrete $p(X) = \sum_{k=1}^K w_k \mathcal{N}(X - m_k;v_k)$ or arbitrary $p(X) = \int p(c) w(c) \mathcal{N}(X - m(c);v(c)) dc$, have closed forms determined by the distances of the mean values, the sum of variances, and the product of weights. The final form is the double sum of Gaussian differences in the discrete case or the expectation in the continuous case, which is also equivalent to constructing the Gaussian cross Gram matrix first, and then averaging all entries.\vspace{9pt}

\noindent \textbf{Difference from the Kullback–Leibler bound.} The big difference between this bound and the Kullback–Leibler divergence bound $\int p(X) \log q(X) dX \leq \int p(X) \log p(X) dX$ is the extra norm term of $q$ in the denominator. When $q$ is a mixture of Gaussians, its norm is the average of all pairwise Gaussian differences. By maximizing the cost, we are minimizing this norm, thus enforcing contrastivity and maximizing the distances between all pairs of samples. So we name this bound and its variations, which use the inner product between two densities normalized by the norm of the model density, the contrastive entropy bounds.

\subsection{Contrastive Entropy Bound for MDNs}

In MDNs, the model density $q$ is by a Gaussian mixture with centers defined by the outputs of a neural network. The inputs to the network can be uniform or hybrid noises. Thus we can directly use this contrastive bound to train an MDN:\vspace{3pt}
\begin{enumerate}[leftmargin=*]
\item Eq.~\eqref{cs_inequality} is Cauchy-Schwarz inequality applied to probability density functions;\vspace{5pt}
\item From~\eqref{cs_inequality} we derive the cost Eq.~\eqref{maximization_cost} that is the inner product normalized by the norm of $q$. Maximizing the cost reaches its upper bound, the norm of $p$;\vspace{5pt}
\item The implementation we found to work the best needs to assume the data distribution also as a mixture. As shown in Eq.~\eqref{pair_wise_difference}, given a batch $X_1,X_2,\cdots,X_N$, we approximate $p(X)$ with a mixture defined by these samples as centers and a variance $v_p$. Then we create a batch of prior noises $u_1,u_2,\cdots,u_N$, feed them through a neural network, and obtain generated samples $X_1',X_2', \cdots, X_N'$, and construct the density $q(X)$ with a variance $v_q$. 

\;\; The inner product is the double sum of cross Gaussian differences. The norm of $q$ is the double sum of the Gaussian differences from the $q$ itself. Then we can construct the cost and maximize it, and ideally the cost should approach the norm of $p$.




\vspace{5pt}
\end{enumerate}

\begin{equation}
\begin{gathered}
\langle p, q \rangle^2 \leq \langle p, p \rangle \cdot \langle q, q \rangle. 
\end{gathered}
\label{cs_inequality}
\end{equation}

\begin{equation}
\begin{gathered}
r(q) = \frac{\langle p,q \rangle^2}{\langle q, q \rangle},\; r(q) \leq \langle p, p \rangle, \;\max_q \; r(q)\rightarrow  \langle p, p\rangle. 
\end{gathered}
\label{maximization_cost} 
\end{equation}

\begin{equation}
\resizebox{1\linewidth}{!}{
$\begin{gathered}
p(X) \approx \frac{1}{N}\sum_{n=1}^N \mathcal{N}(X-X_n;v_p), \;\;q(X) \approx \frac{1}{K}\sum_{k=1}^K \mathcal{N}(X-X_k';v_q),\\
\langle p, q\rangle = \frac{1}{NK}\sum_{n=1}^N \sum_{k=1}^K \mathcal{N}(X_n - X_k';v_p+v_q), \\
\langle q, q\rangle = \frac{1}{K^2} \sum_{i=1}^K \sum_{j=1}^K \mathcal{N}(X_i' - X_j';2v_q), \\
\langle p, p\rangle = \frac{1}{N^2} \sum_{i=1}^N \sum_{j=1}^N \mathcal{N}(X_i - X_j;2v_p).
\end{gathered}$}
\label{pair_wise_difference}
\end{equation}

\noindent \textbf{How to Choose $v_p$ and $v_q$.} For real images, we find that setting $v_p=v_q$ to be a constant is the most effective. For simple datasets, $v_q$ is trainable by simply substituting $v_q$ with $v(u_n)$. That is, we can assign each noise sample a variance of itself, parametrized by a neural network, which will produce the exact approximation for the density and the norm of $p$, but it could be redundant for real images.\vspace{7pt}

\noindent \textbf{Suboptimal Solution.} Like any contrastive learning method, the trade-off is that at each training step, it requires a batch of samples to estimate the inner product and the norm, with a specific batch size. So theoretically, we only find a suboptimal solution that works for the random batches, but we do observe that the generation quality improves evidently during training.

\newpage \;
\newpage

\section{Results for MDNs}
\label{experiments_mdns}

This section presents the results for using our proposed methods, the SVD of the Gaussian cross Gram matrix and the normalized inner product of pdfs in the Hilbert space, to learn Mixture Density Networks (MDNs), on image datasets MNIST and CelebA. 

Assuming that $p(X)$ is the data density and $q(X)$ is the model density parameterized by a Gaussian mixture, given by $q(X) = \int p(c) \mathcal{N}(X - m(c); v_q) dc$, the task is to approximate $p$ with $q$. The procedure is first sampling noise $u_1, u_2, \cdots, u_K$ from a prior, mapping them through a neural network to generate samples $X_1', X_2', \cdots, X_K'$, and then applying the costs between $X_1, X_2, \cdots, X_N$ and $X_1', X_2', \cdots, X_K'$. At each training step, it is applied to a different batch. 

\subsection{How to Train a Regular MDN}

First, we discuss how to train a regular MDN using the inequality of Kullback–Leibler divergence, the cost of which follows\vspace{-5pt}
\begin{equation}
\resizebox{.9\linewidth}{!}{
$\begin{aligned}
\int p(X)\log q(X) dX & \approx \int p(X) \log \frac{1}{K}\sum_{k=1}^K \mathcal{N}(X-X_k') dX\\
& \approx \frac{1}{N}\sum_{n=1}^N \log \frac{1}{K}\sum_{k=1}^K \mathcal{N}(X_n-X_k'),\vspace{-5pt}
\end{aligned}$}
\label{regular_shannon_cost}
\end{equation}
where $X_1',X_2',\cdots,X_K'$ is a series of generated samples, generated by passing sampled noises $u_1,u_2,\cdots,u_K$ through a neural network $m$. The regular MDN maximizes the cost in Eq.~\eqref{regular_shannon_cost}. For a batch of data samples $X_1,X_2,\cdots,X_N$ and a batch of generated samples $X_1',X_2',\cdots,X_K'$, we compute all pairwise Gaussian differences between them, then take the average over $X'$, apply the nonlinear $\log$ function, and finally take the average over $X$.\vspace{-3pt}

\subsection{Comparison of Generation Quality}

\begin{figure}[t]
\centering
\begin{subfigure}{.45\textwidth}\includegraphics[width=\linewidth]{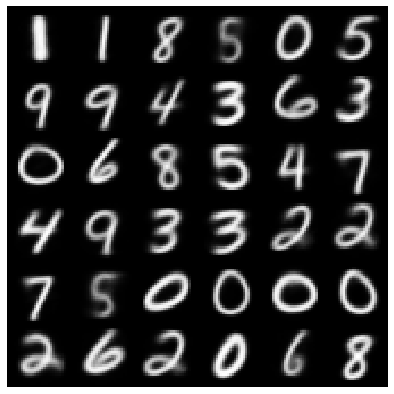}\vspace{-3pt}
\caption{{KL-Based Cost}}
\end{subfigure}
\begin{subfigure}{.45\textwidth}\includegraphics[width=\linewidth]{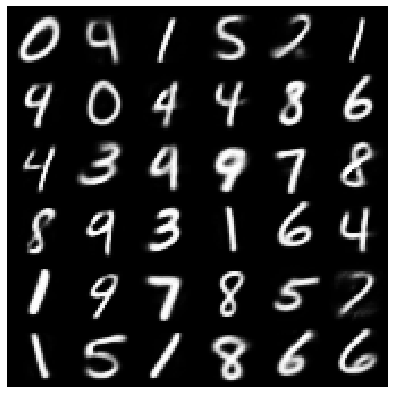}\vspace{-3pt}
\caption{{Normalized Inner Product}}
\end{subfigure}\vspace{3pt}
\begin{subfigure}{.45\textwidth}\includegraphics[width=\linewidth]{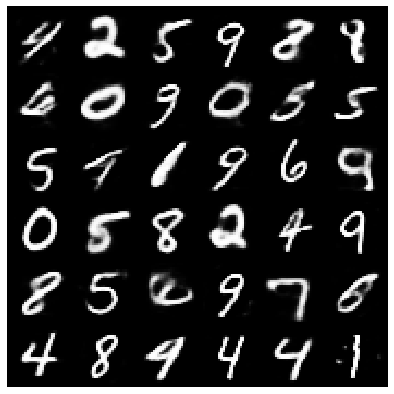}\vspace{-3pt}
\caption{{Nuclear Norm}}
\end{subfigure}
\begin{subfigure}{.45\textwidth}\includegraphics[width=\linewidth]{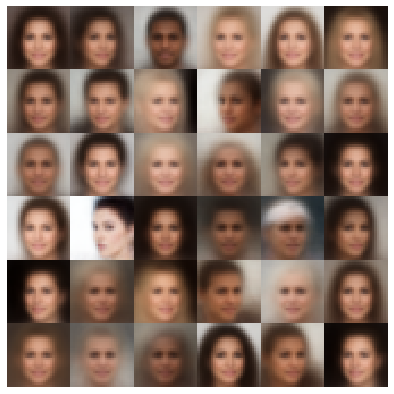}\vspace{-3pt}
\caption{{KL-Based Cost}}
\end{subfigure}\vspace{3pt}
\begin{subfigure}{.45\textwidth}\includegraphics[width=\linewidth]{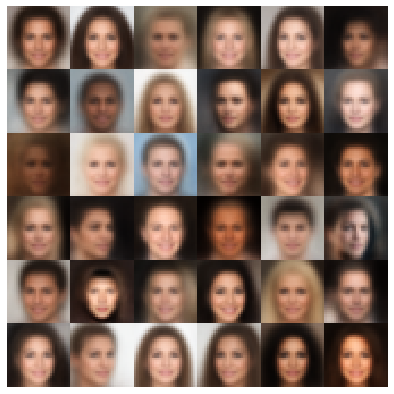}\vspace{-3pt}
\caption{{Normalized Inner Product}}
\end{subfigure}
\begin{subfigure}{.45\textwidth}\includegraphics[width=\linewidth]{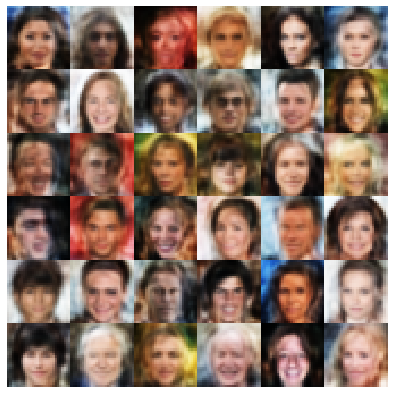}\vspace{-3pt}
\caption{{Nuclear Norm}}
\end{subfigure}\vspace{-5pt}
\caption{Side-by-side comparisons of generation qualities between KL-based cost, the normalized inner product between pdfs in the Hilbert space, and the nuclear norm (the sum of eigenvalues) of the Gaussian cross Gram matrix $\mathbf{K}_{XX'}$, for the datasets MNIST and CIFAR10. The nuclear norm works the best, followed by the normalized inner product, then the KL-based cost. We definitely observed that the normalization improves the sample diversity.}
\label{figure_4}
\end{figure}

Fig.~\ref{figure_4} is the side-by-side comparisons between the MDNs trained by the regular cost of the Kullback–Leibler divergence (Eq.~\eqref{regular_shannon_cost}), the normalized inner product cost (Eq.~\eqref{maximization_cost}~\eqref{pair_wise_difference}), and the nuclear norm of the Gaussian cross Gram matrix (Eq.~\eqref{singular_values}). First, we do find that the variances matter a lot for the generation quality. we have a bunch of observations in terms of the variance parameter in the Gaussian differences:\vspace{3pt}
\begin{enumerate}[leftmargin=*]
\item The regular Shannon's cost for an MDN does not require a normalization term, so we do not need to discuss whether $p$ is also a mixture distribution. That is, we only need to pick one variance parameter for the Gaussian in the $\log$ in Eq.~\eqref{regular_shannon_cost}. The same applies to the nuclear norm form, which only requires picking one variance parameter in the Gaussian cross Gram matrix $\mathbf{K}$. But it is not the same for the normalized inner product, because it has a numerator which has the Gaussian differences between data samples and the generated samples, and a denominator that has the Gaussian differences between generated samples and themselves. Thus there are two variances to choose.\vspace{5pt}

Our solution is to assume $p$ to also be a mixture with a variance $v_X$, and the generated samples have a variance $v_q$. So in the numerator $\langle p, q\rangle$, the Gaussian differences have a variance $v_X+v_q$, and in the denominator $\langle q, q \rangle$, the Gaussian differences have a variance $2 v_q$. We also find that choosing $v_X = v_q = v$ has the best performance. That is, the variance parameters in the numerator and the denominator are set to be the same.\vspace{3pt}
 
\item These methods have different trainable ranges. We find that the nuclear norm form is the most tolerant with the widest range. Then it is followed by the normalized inner product and the nuclear norm form.\vspace{3pt}

\item Generally, the smaller the variance, the more diverse the generated samples would be, and the higher the resolution would be. But specifically for the nuclear norm case, we find that the variance should be set in the middle. When we set the variance to be large, we find that the nuclear norm case becomes similar to decomposing a matrix of $L_2$ distances, which has a very poor performance.\vspace{5pt}

\item For the nuclear norm case that requires the SVD of a Gaussian cross Gram matrix, when we substitute the Gaussian cross Gram matrix with a matrix of $L_2$ distances, we find that the generated samples still will not go to the trivial solution of the global mean, but the generations are highly undistinguishable and does not seem to have semantic meanings. Only using the SVD of the Gaussian cross Gram matrix the generated samples will have semantic meanings and actually look like data samples, although a little more blurry. The reason of this requires more investigations.\vspace{3pt}

\;\; One should also be tempted to use the Frobenius's norm that has the form $Trace(\mathbf{K}\mathbf{K}^\intercal)$ so the SVD can be omitted. But still it does not work well. It could be that a matrix inverse is missing and a cost like $Trace(\mathbf{K}_{XY}\mathbf{K}_{XX}^{-1}\mathbf{K}_{XY}^\intercal)$ is needed. We leave the details in the Appendix.\vspace{5pt}

\item The choosing of these parameters is as follows. For the KL-based coast and the normalized inner product, we choose the variance to be as small as possible such that the model is trainable. For the nuclear norm, we found it simple to start with $v=1$, and find a value around $1$ that generates samples that have the best sample diversity and the clarity. We do find that for the nuclear norm case, if the variance is too small, the sample diversity will decrease (for example in the MNIST case all digits will look like ``1''). But if the variance is too large, the clarity will decrease, the samples will feel like the results from decomposing $L_2$ distances. But if the variance is set correctly, the nuclear norm SVD case is much easier to train and has a better performance than the other two.\vspace{3pt}
\end{enumerate}

In terms of the generation quality, we do find that the nuclear norm form beats the other two. The normalized inner product also beats the KL-based cost in terms of the sample diversity. We think the normalization definitely works. This is shown in Fig.~\ref{figure_4}. 

\begin{figure}[t]
\begin{subfigure}{.4\textwidth}\includegraphics[width=\linewidth]{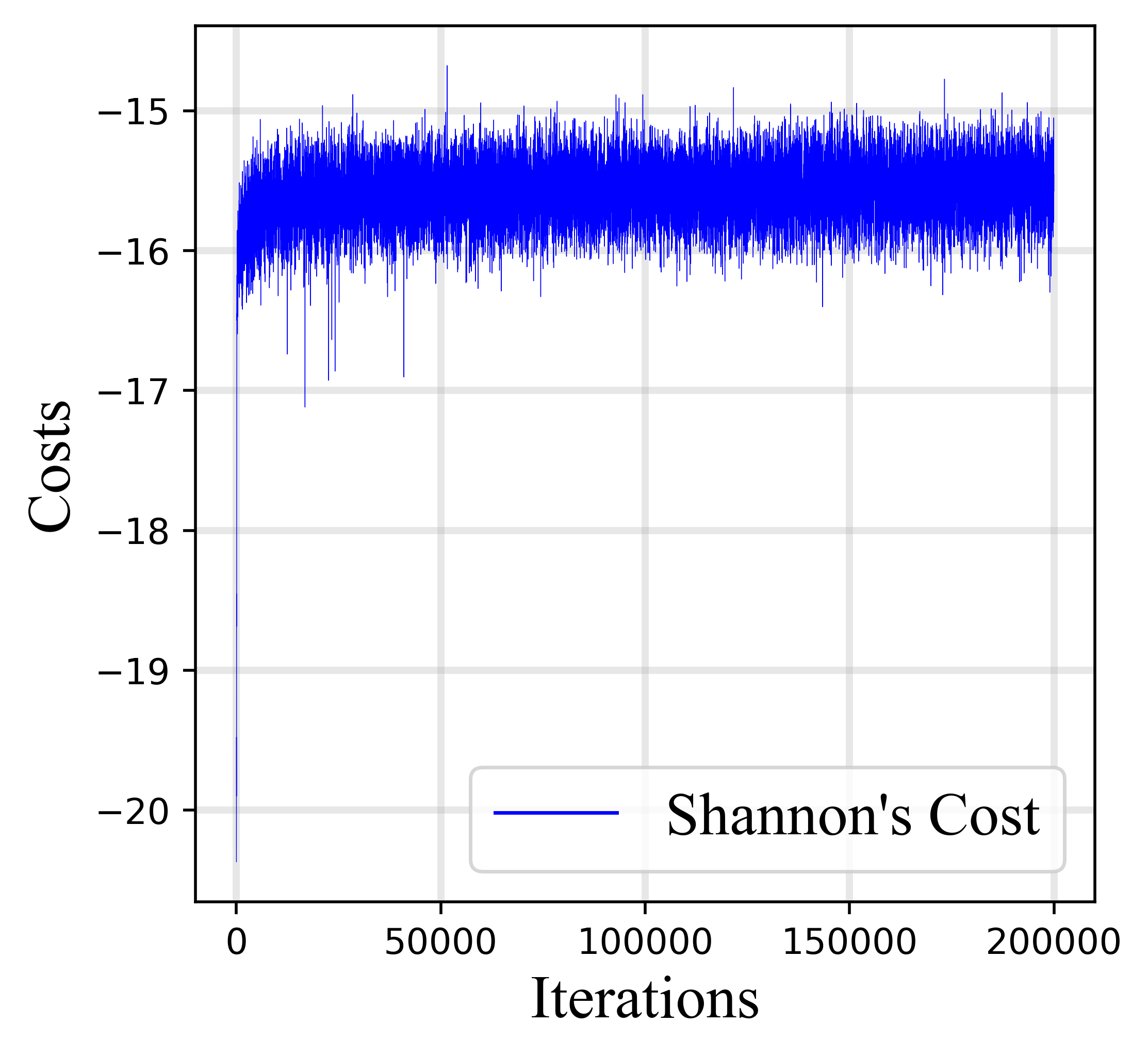}\vspace{-5pt}
\caption{KL: CelebA}
\end{subfigure}  
\begin{subfigure}{.4\textwidth}\includegraphics[width=\linewidth]{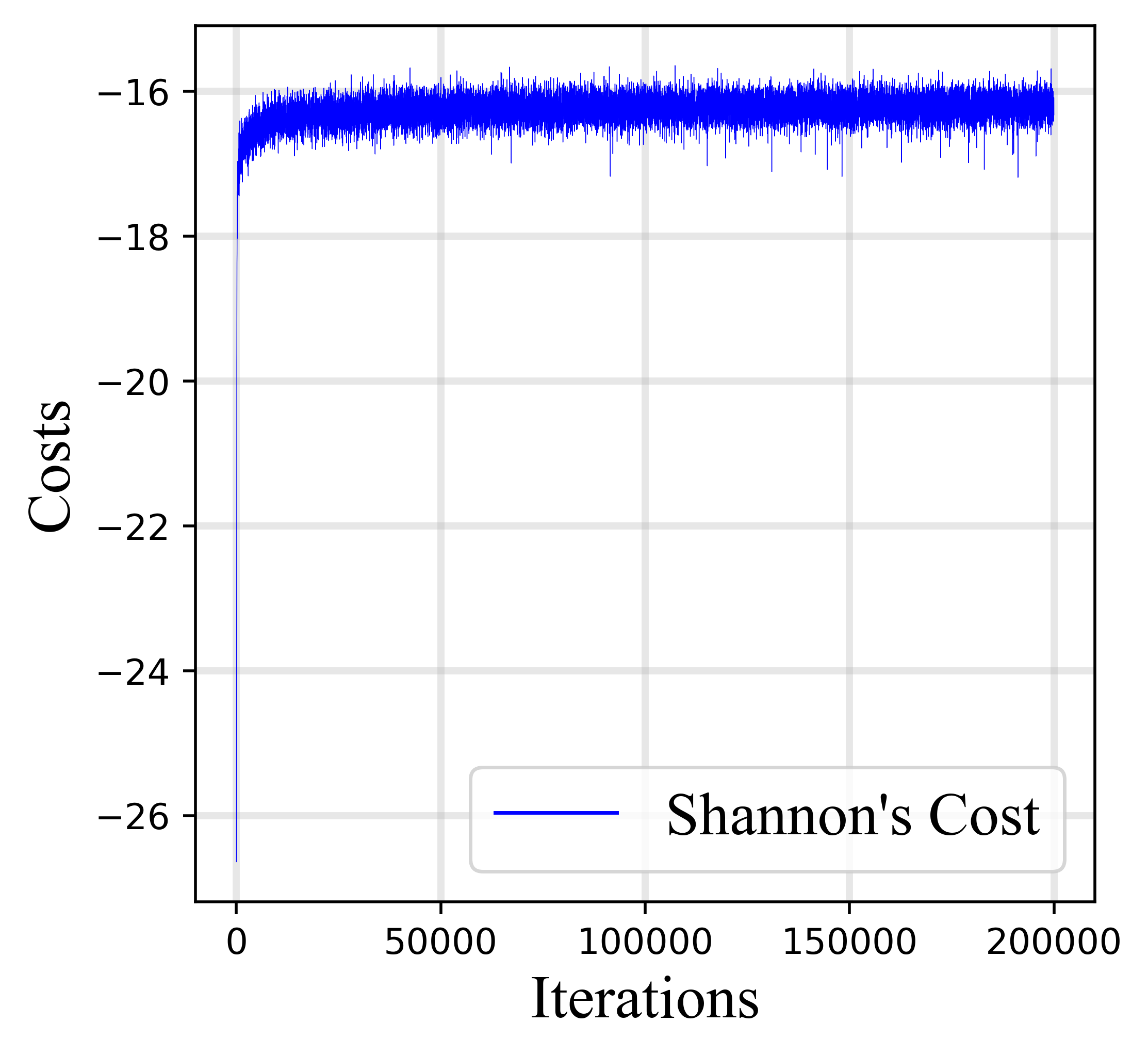}\vspace{-5pt}
\caption{KL: MNIST}
\end{subfigure}
\begin{subfigure}{.4\textwidth}\includegraphics[width=\linewidth]{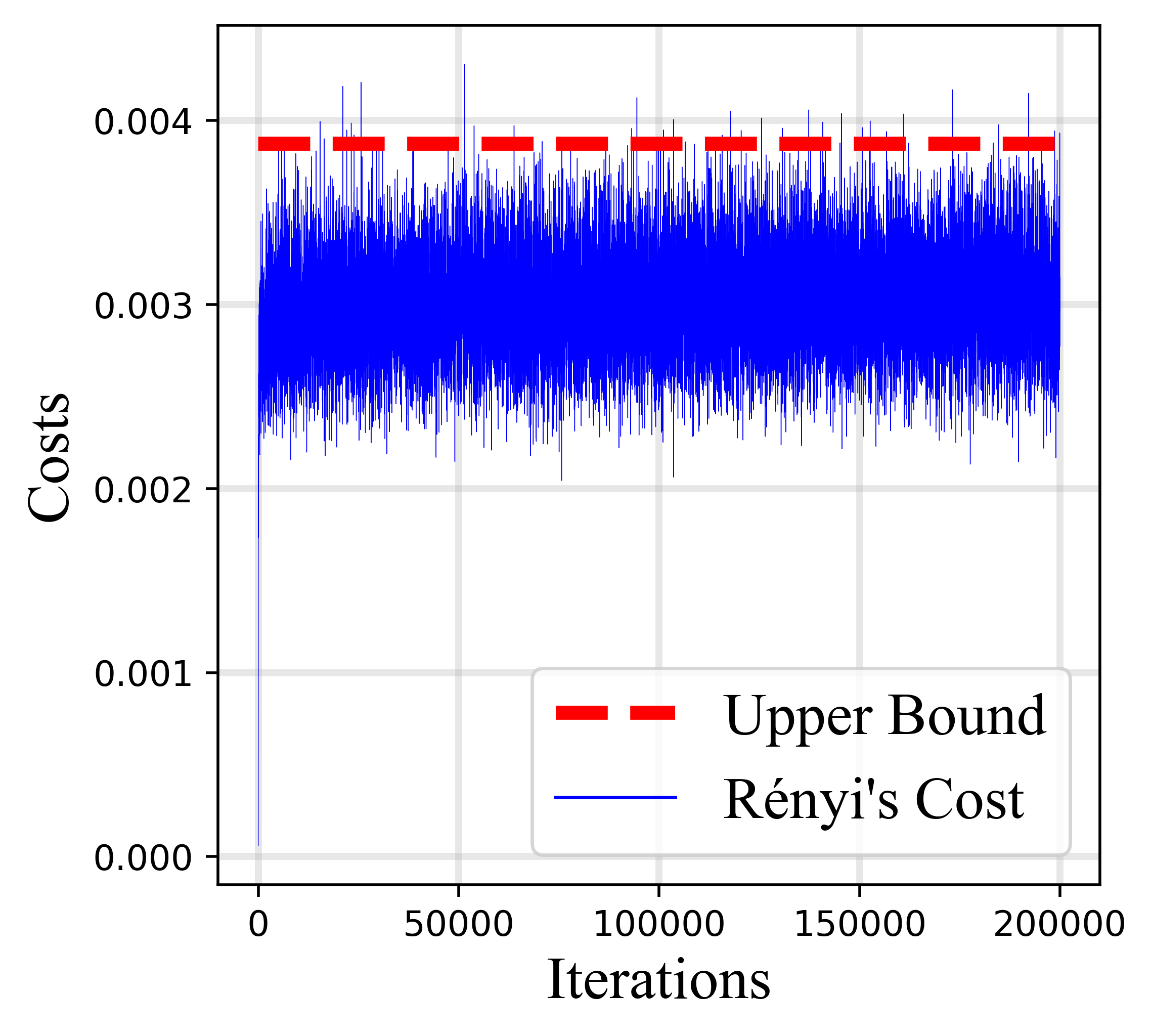}\vspace{-5pt}
\caption{NIP: CelebA}
\end{subfigure}
\begin{subfigure}{.4\textwidth}\includegraphics[width=\linewidth]{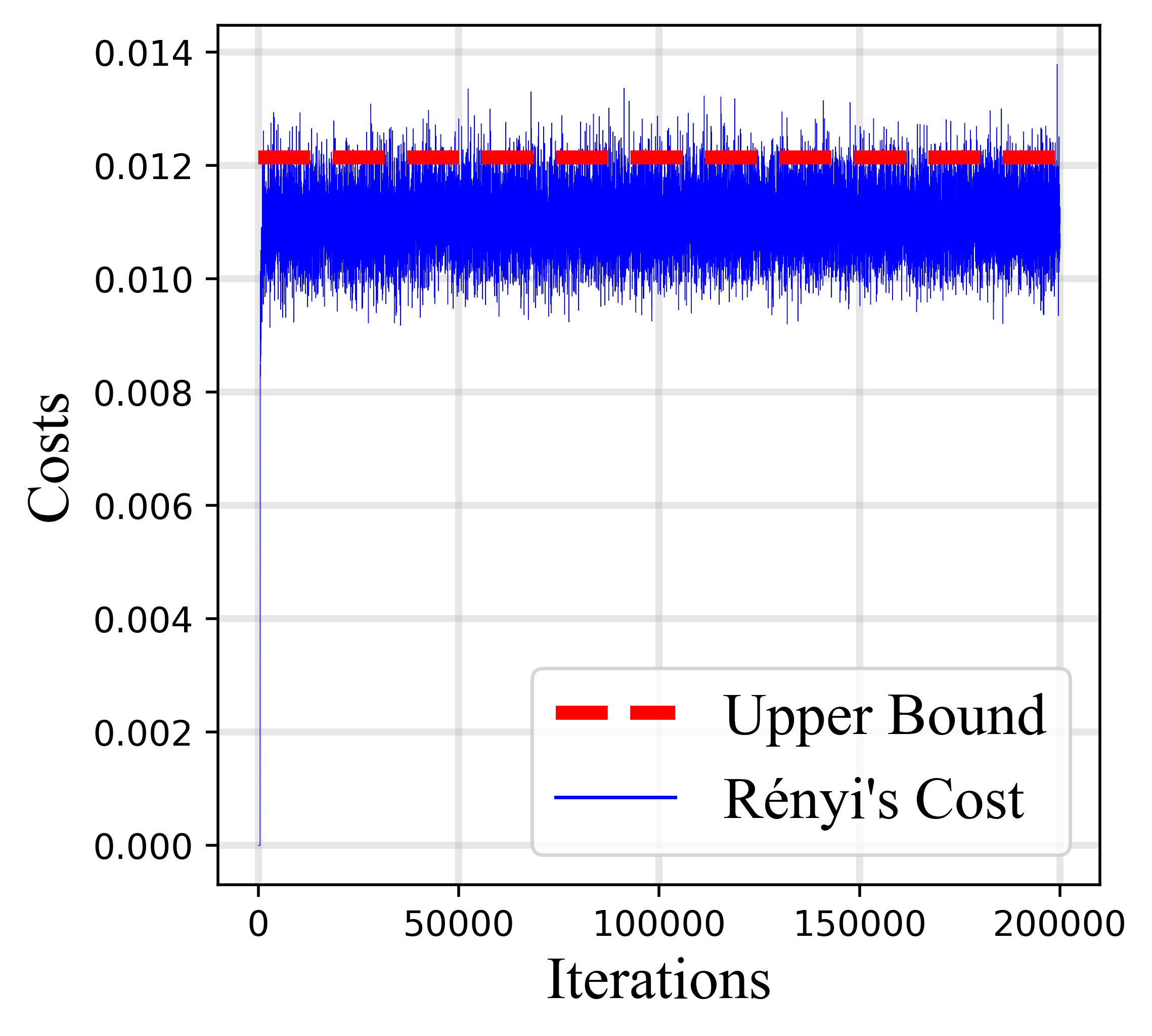}\vspace{-5pt}
\caption{NIP: MNIST}
\end{subfigure}
\begin{subfigure}{.4\textwidth}\includegraphics[width=\linewidth]{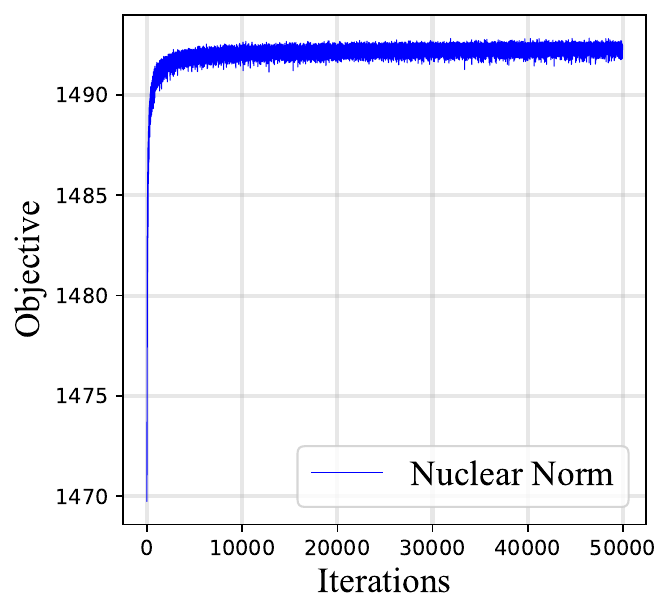}\vspace{-5pt}
\caption{NucNorm: CelebA}
\end{subfigure}
\begin{subfigure}{.4\textwidth}\includegraphics[width=\linewidth]{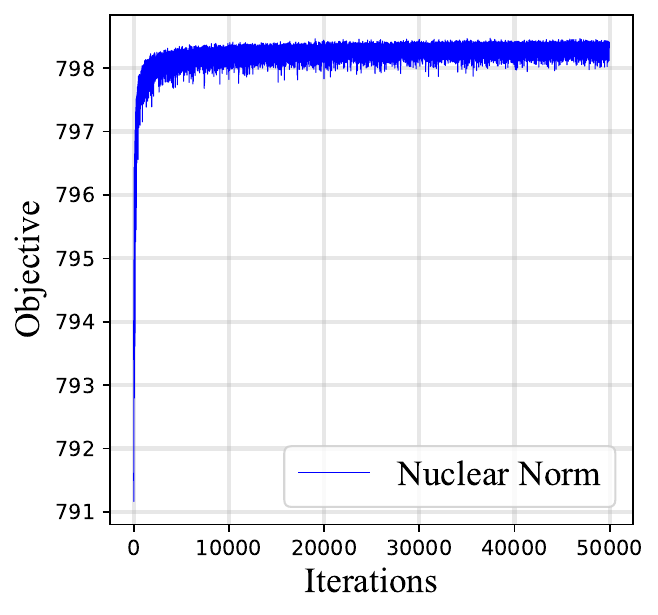}\vspace{-5pt}
\caption{NucNorm: MNIST}
\end{subfigure}\vspace{-7pt}
\caption{Learning curves of KL-based costs (KL), normalized inner products (NIP), and the nuclear norm (NucNorm) on datasets MNIST and CelebA.\vspace{-5pt}}
\label{figure_5}
\end{figure}

We also show the learning curves in Fig.~\ref{figure_5}. We find them all to be stable. Particularly for the normalized inner product, the upper bound is actually tractable, which is the norm of $p$, $\langle p , p\rangle$. Because we assume that $p$ is also a Gaussian mixture with a small variance, the norm of $p$ is just the pairwise distances, following Eq.~\eqref{pair_wise_difference}. So we also use red curves to indicate the upper bound, the norm of $p$, by computing the double sums of Gaussian differences over the training samples.

\begin{figure}[t]
\centering
\begin{subfigure}{.33\textwidth}\includegraphics[width=\linewidth]{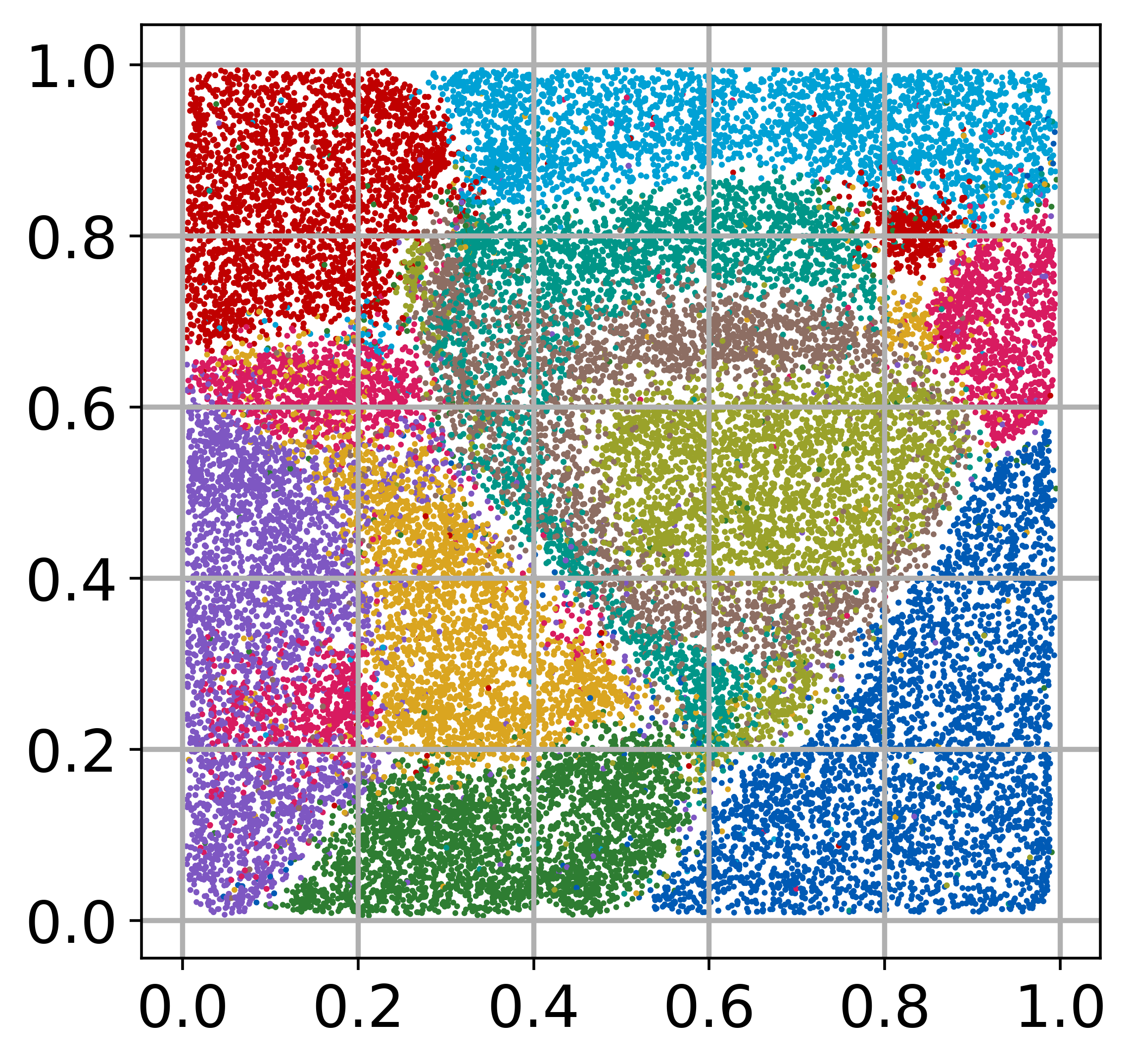}\vspace{-3pt}
\caption{}
\end{subfigure}
\begin{subfigure}{.33\textwidth}\includegraphics[width=\linewidth]{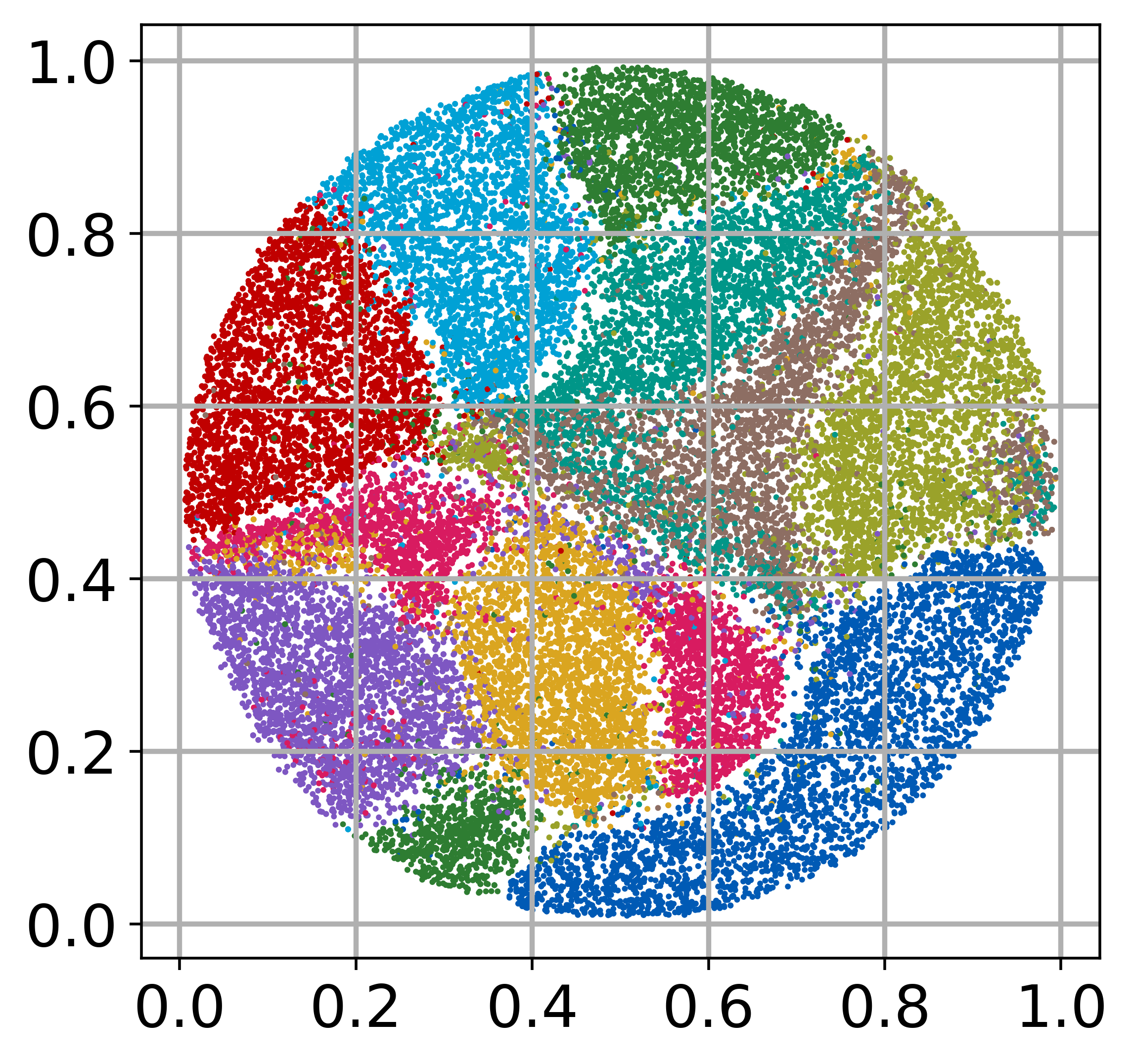}\vspace{-3pt}
\caption{}
\end{subfigure}
\begin{subfigure}{.33\textwidth}\includegraphics[width=\linewidth]{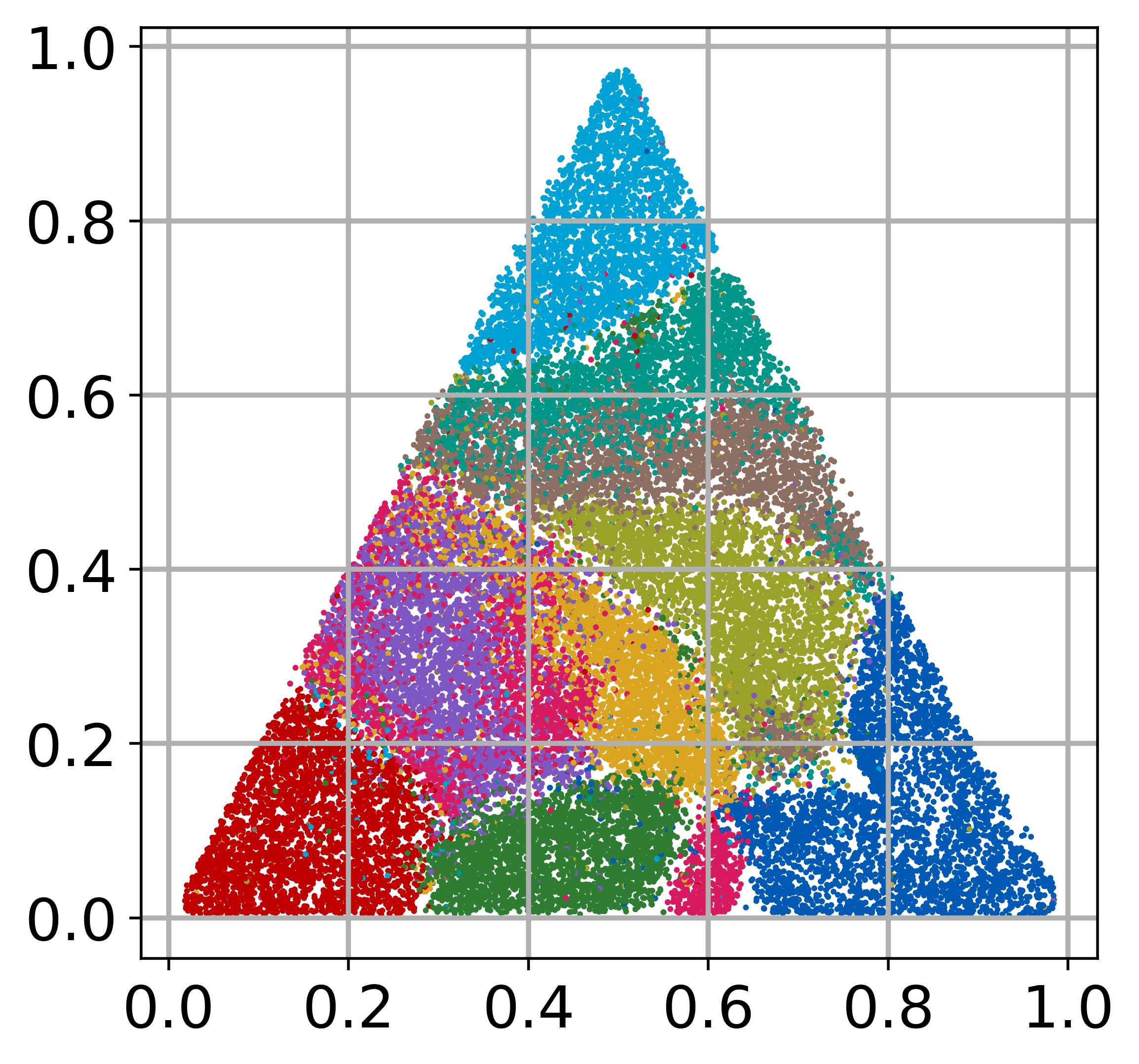}\vspace{-3pt}
\caption{}
\end{subfigure}
\begin{subfigure}{.33\textwidth}\includegraphics[width=\linewidth]{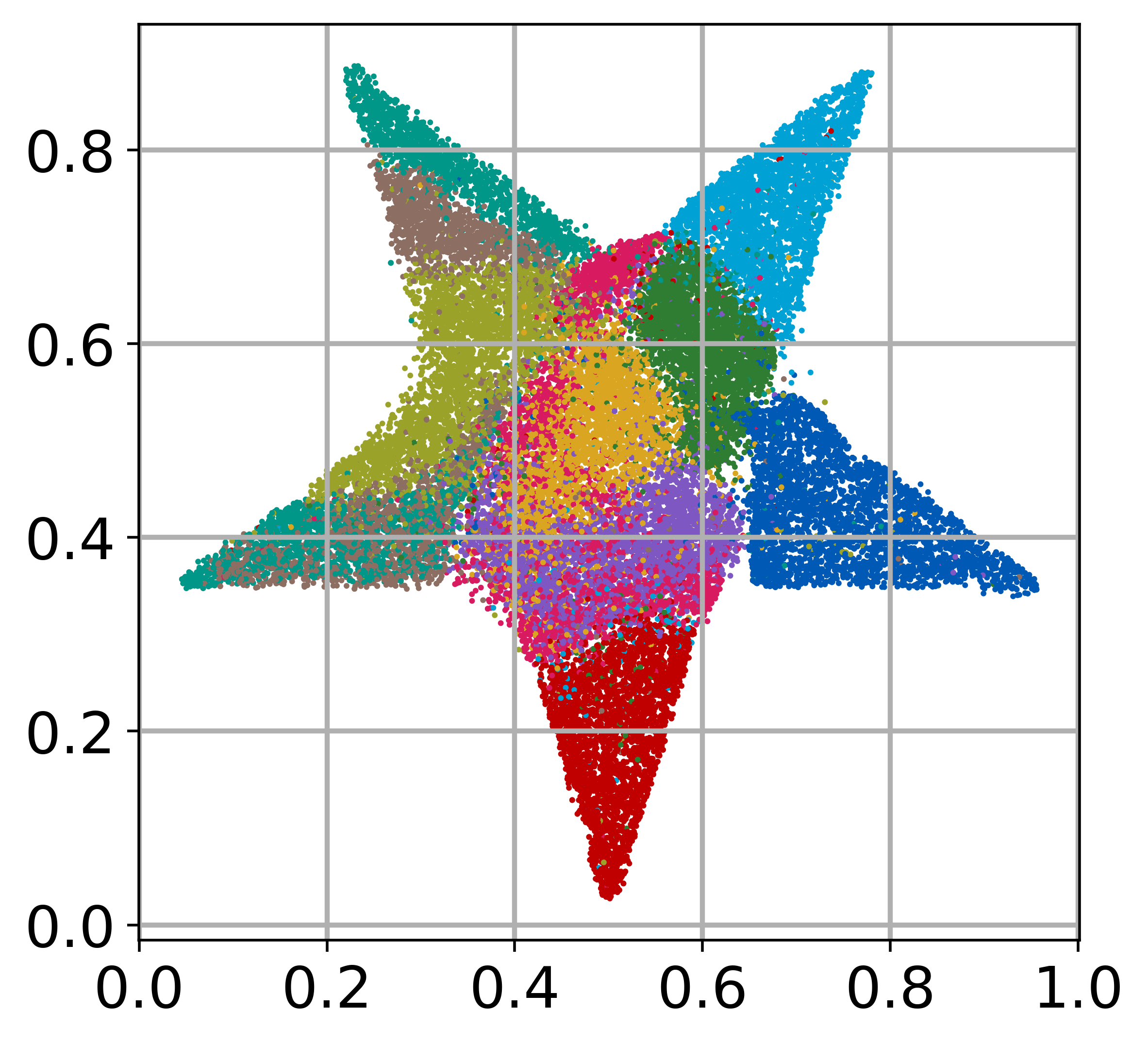}\vspace{-3pt}
\caption{}
\end{subfigure}\vspace{-9pt}
\caption{ELBO variation of the nuclear norm cost. We project MNIST into a $2D$ feature space while also minimizing the divergence between the encoder joint $p(X,Y) = p(X)p(Y|X)$ and the decoder joint $q(X,Y) = q(Y)q(X|Y)$. We tried four different priors and found that the features have good separations and semantically meaningful.\vspace{-12pt}}
\label{ELBO_variation}
\end{figure}

\subsection{The ELBO Variation}

In Sec.~\ref{sectionb_elbo}, we have shown that the nuclear norm cost can also achieve a task similar to what the evidence lower bound (ELBO) wants to achieve. As we have introduced, we can sample a batch of samples $X_1,X_2,\cdots,X_N$, and a batch of noises $Y_1, Y_2,\cdots,Y_N$. The samples are fed through the encoder to generate $Y_1', Y_2',\cdots,Y_N'$ in the dimension of the noise; the noises sampled from the prior are fed through the decoder to generate generated samples $X_1',X_2',\cdots,X_N'$. Then we apply the SVD to the Gaussian cross Gram matrix constructed by both $X$, $X'$, $Y$ and $Y'$, following Eq.~\eqref{singular_value_elbo}. 

Fig.~\ref{ELBO_variation} is the result when we pick four different shapes of priors. The noise samples are uniformly distributed in each of the region. We do find that the variances for the data and the noises matter significantly for the generalization capability. Suppose the variance of the data is $v_X$ and the variance for the noise is $v_Y$. They do need to be picked with different values. For this experiment it performs the best when $v_X = 0.1$ and $v_Y = 0.01$. Also notice that they are not strictly the variances of the Gaussian since the matrix of $L_2$ distances ($\mathbf{M}_X$ and $\mathbf{M}_Y$ in Eq,~\eqref{singular_value_elbo}) have to be divided by $d_X$ and $d_Y$, the data dimension and the noise dimension, such that after the exponential it still ensures numerical stability. 

\subsection{A Trick for Saving Computing Memories}
\label{saving_computing_memory}

In all these methods, we need to compute the $L_2$ distances and Gaussian differences between pairwise samples. For example given two samples $X_n$ and $X'_n$ we will have to compute $\exp(-\frac{1}{d_X}||X_n-X'_n||_2^2)$. Suppose $X_n(i)$ to be the $i$-th dimension of $X$, then the $L_2$ distance becomes $\frac{1}{d_X}\sum_{i=1}^{d_X}\left (X_n(i) -X'_n(i)\right )^2$. This needs to be done for each sample $X_n$ for $n=1,2,\cdots,N$. Taking MNIST for an example, it requires constructing a matrix of size $N\times N \times 786$ that store all pairwise distances between all samples, and then taking the average in the last dimension of the matrix. But this third dimension of $786$ can make the matrix very large. If we have further the color dimension, the batch size has to be chosen to be very small. But the creation of this matrix is actually not necessary. Suppose the data matrix is a matrix $\mathbf{X}$ with a size of $N\times 786$, and we want to construct the matrix of $L_2$ distances $\mathbf{M}$. We can construct \vspace{-3pt}
\begin{equation}
\begin{gathered}
\mathbf{M}_A =  \frac{1}{d_X}\mathbf{X} {\mathbf{X}'}^\intercal, \; \mathbf{M}_B = \frac{1}{d_X}\sum_{i=1}^{d_X}(\mathbf{X}\odot \mathbf{X})_{\cdot,i},\vspace{-7pt}\\
\mathbf{M}_C = \frac{1}{d_X}\sum_{i=1}^{d_X}(\mathbf{X}'\odot \mathbf{X}')_{\cdot,i}, \; \mathbf{M} = \mathbf{M}_B + \mathbf{M}_C^\intercal - 2\cdot \mathbf{M}_A.
\end{gathered}
\end{equation}
Here $\mathbf{M}_A$ is simply the matrix product of two data matrix in the data dimension. The matrix $\mathbf{M}_B$ is simply the data multiplying itself, taking the square of each element of this batch of data, then average over the data dimension. The matrix $\mathbf{M}_C$ is when we do this to the generated samples $\mathbf{X}'$, taking the square of each element in this batch of generated samples and square their each element, then average it in the data dimension. This makes $\mathbf{M}_A$ a matrix of size $N\times N$, and $\mathbf{M}_B$ and $\mathbf{M}_C$ two vectors, both of size $N$. Then we can construct the matrix of $L_2$ distances $\mathbf{M}$ by $\mathbf{M}_B + \mathbf{M}_C^\intercal - 2\cdot \mathbf{M}_A$, meaning that we are summing a vector of size $N\times 1$, a transposed vector of size $1\times N$, and subtracting twice the matrix $\mathbf{M}_A$ of size $N \times N$, this makes each element of $\mathbf{M}$ just $X_m^2 + X_n^2 - 2\cdot X_m \cdot X_n$. So indeed this matrix is the matrix of $L_2$ distances. And this way we can avoid constructing a matrix of size $N \times N \times d_X$ but only with matrix of size $N \times N$, which saves tons of memory. The way we write $\mathbf{M}_B + \mathbf{M}_C^\intercal$ might be a little less standard, but it simply means that we sum a vector and a transposed vector to produce a $2D$ matrix, with the $i$-th row and the $j$-th column of it to be $\mathbf{M}_B(i) + \mathbf{M}_C(j)$. This not only applies to the nuclear norm cost but also the normalized inner product where the extra step is to take the mean of the matrix $\mathbf{K}_{XX}$ and $\mathbf{K}_{XY}$. 



\newpage
\section{Encoder-mixture-decoder Architecture}\label{encoder_mixture_decoder}

Following our discussion on Section~\ref{inspiration_from_autoencoders} and Section~\ref{inner_product_norm}, we now introduce the encoder-mixture-decoder architecture based on the normalized inner product. The reasoning has been emphasized in Section~\ref{inspiration_from_autoencoders} and we reiterate here:\vspace{3pt}
\begin{enumerate}[leftmargin=*]
\item A conventional neural network is a many-to-one mapping or at most a one-to-one mapping, and cannot be a one-to-may mapping;\vspace{3pt}
\item Using the autoencoder as an example. Its objective is to minimize the term $\iint p(X)p(Y|X)\log q(X|Y) dXdY$ with $p(Y|X)$ as the encoder and $q(X|Y)$ as the decoder. There are two optimality conditions for maximizing this objective. First the $q(X|Y)$ has to be as close as to $p(X|Y)$, after applying the Bayes's rule to $p(Y|X)$, such that the bound of Kullback–Leibler divergence is tight. Second, after the bound is tight, we need to further maximize the objective $\iint p(X)p(Y|X) \log p(X|Y)$ which is now irrelevant to $q$ but only relevant to $p$. Maximizing this term is finding the optimal $Y$ such that the $p$ it comes with has the minimal conditional entropy, i.e., equivalent to the maximal mutual information for Shannon's bound because of the additivity of the $\log$.\vspace{3pt}
\item If $p(Y|X)$ and $q(X|Y)$ are single neural networks, then it means that they are many-to-one or one-to-one functions, and these two densities as functions of $X$ and $Y$ will be very sparse.\vspace{3pt}
\item The assumption is that if we change these two functions from sparse many-to-one mappings to one-to-many mappings, specifically for the decoder since it is the mapping from compressed features back to the sample space, whether the bound can be more tight and we can find a better solution. This inspires us to propose the architecture of an encoder-mixture-decoder.\vspace{3pt}
\item Now the decoder is multiple-output and defines a mixture density, using the $\log$ will produce a term that is the $\log$ of the sum, so the Shannon's cost of KL divergence is no longer suitable. For the same reason we train an MDN, using the Hilbert space definition of the normalized inner product is more suitable.\vspace{3pt}
\item Unfortunately the nuclear norm form is difficult to apply here as it will require an eigendecomposition for every sample or some sort. 
\end{enumerate}

\subsection{Contrastive Conditional Entropy Bound}

Here we introduce our proposed architecture, which comes with a bound on the conditional entropy, a cost based on this bound, and a network topology, the encoder-mixture-decoder architecture that implemented optimizing this cost. 

The bound is as follows. We have introduced a bound with the inner product and norms of densities in the Hilbert space with a Cauchy-Schwarz inequality (Eq.~\eqref{cs_inequality}), which has the form $\langle p, q\rangle^2 \leq \langle p, p\rangle \cdot \langle q, q\rangle$ that the inner product between $p$ and $q$ is smaller than the product of two norms. We need to change this inequality into an inequality on the conditional densities and the conditional entropy. 

The trick is to look a $\iint p(X,Y)q(X|Y) dX dY$, the objective of an autoencoder without the $\log$. Instead of writing $p(X,Y)$ as $p(X)p(Y|X)$, the data density multiplying the conditional of the encoder, we write it as $p(X,Y) = p(Y)p(X|Y)$, the feature density multiplying $p(X|Y)$ obtained by applying Bayes's rule to $p(Y|X)$. Then this inner product $\iint p(Y)p(X|Y) q(X|Y) dXdY$ with this new form can be seen as the inner product between $p(X|Y)$ and $q(X|Y)$ over the probability measure $p(Y)$. Similar to $\langle p, q\rangle$, we write the inner product of the conditional as $\langle p(X|Y), q(X|Y)\rangle_{p(Y)}$, meaning that we compute the inner product between $p(X|Y)$ and $q(X|Y)$ under the probability measure $p(Y)$. The bound follows Eq.~\eqref{bound_conditional_entropy}, which says that this inner product is bounded by the product of two norms of the conditional densities, also under the probability measure of $p(Y)$.

After defining the bound, we define the cost as before. We move the norm of the decoder $q(X|Y)$ under $p(Y)$ to the left-hand-side, so the right-hand-side only has $p(X|Y)$ and $p(Y)$. The cost is a ratio bounded by the norm of $p(X|Y)$, maximizing the cost will reach its upper bound, the norm of $p(X|Y)$, and further maximizing the bound is finding the $Y$ such that the norm of $p(X|Y)$ under $p(Y)$ is the maximal, similar to before (Eq.~\eqref{cs_inequality}~\eqref{maximization_cost}) but the only difference is that now this is on the conditionals.

\begin{equation}
\begin{aligned}
& \big ( \iint p(X,Y)q(X|Y) dX dY  \big )^2 \\ & \leq  \big(\iint p^2(X|Y) p(Y) dX dY  \big) \cdot  \big(\iint q^2(X|Y) p(Y) dX dY  \big).
\end{aligned}
\label{bound_conditional_entropy}
\end{equation}

\begin{equation}
\begin{aligned}
\langle p(X|Y), q(X|Y)\rangle_{p(Y)} &:=  \iint p(X,Y)q(X|Y) dX dY,\\
||p(X|Y)||_{p(Y)}^2 &:= \iint p^2(X|Y) p(Y) dX dY, \\
||q(X|Y)||_{p(Y)}^2 &:= \iint q^2(X|Y) p(Y) dX dY. 
\end{aligned}
\end{equation}

\begin{equation}
\begin{aligned}
r_c(p, q) = \frac{\langle p(X|Y), q(X|Y)\rangle_{p(Y)}^2}{||q(X|Y)||_{p(Y)}^2}, \; r_c(p, q) \leq ||p(X|Y)||_{p(Y)}^2. 
\end{aligned}
\label{conditional_cost1}
\end{equation}

\begin{equation}
\resizebox{1\linewidth}{!}{
$\begin{aligned}
\max_{p, q} \; r_c(p,q) = \frac{\langle p(X|Y), q(X|Y)\rangle_{p(Y)}^2}{||q(X|Y)||_{p(Y)}^2} \rightarrow \max_p ||p(X|Y)||_{p(Y)}^2. 
\end{aligned}$}
\label{conditional_cost2}
\end{equation}
\vspace{-20pt}

\subsection{Algorithm}
\label{SecVIB}

Based on the cost, the inner product $\langle p(X|Y), q(X|Y)\rangle_{p(Y)}^2$ divided by the norm $||q(X|Y)||_{p(Y)}^2$, we propose the encoder-mixture-decoder architecture. Basically, we change $q(X|Y)$ from a single Gaussian over the decoder to a mixture of Gaussian, following Eq.~\eqref{change_mixture_decoder}. Given one input $Y$, before, the decoder is just one Gaussian $q(X|Y) = \mathcal{N}(X-\textbf{D}(Y))$. Afterwards, we add an extra variable $c$ such that the decoder $q(X|Y)$ is an integral over $c$, such that the conditional density is a mixture density. In practice, we sample $c$ from a uniform distribution, concatenate it with $Y$ as the inputs to the decoder. Sampling different $c$, concatenating them with the same $Y$, going through the decoder $\textbf{D}(Y,c)$, will generate multiple outputs $\textbf{D}(Y,c)$. 

\begin{figure}[h]
  \centering
\includegraphics[width=.7\linewidth]{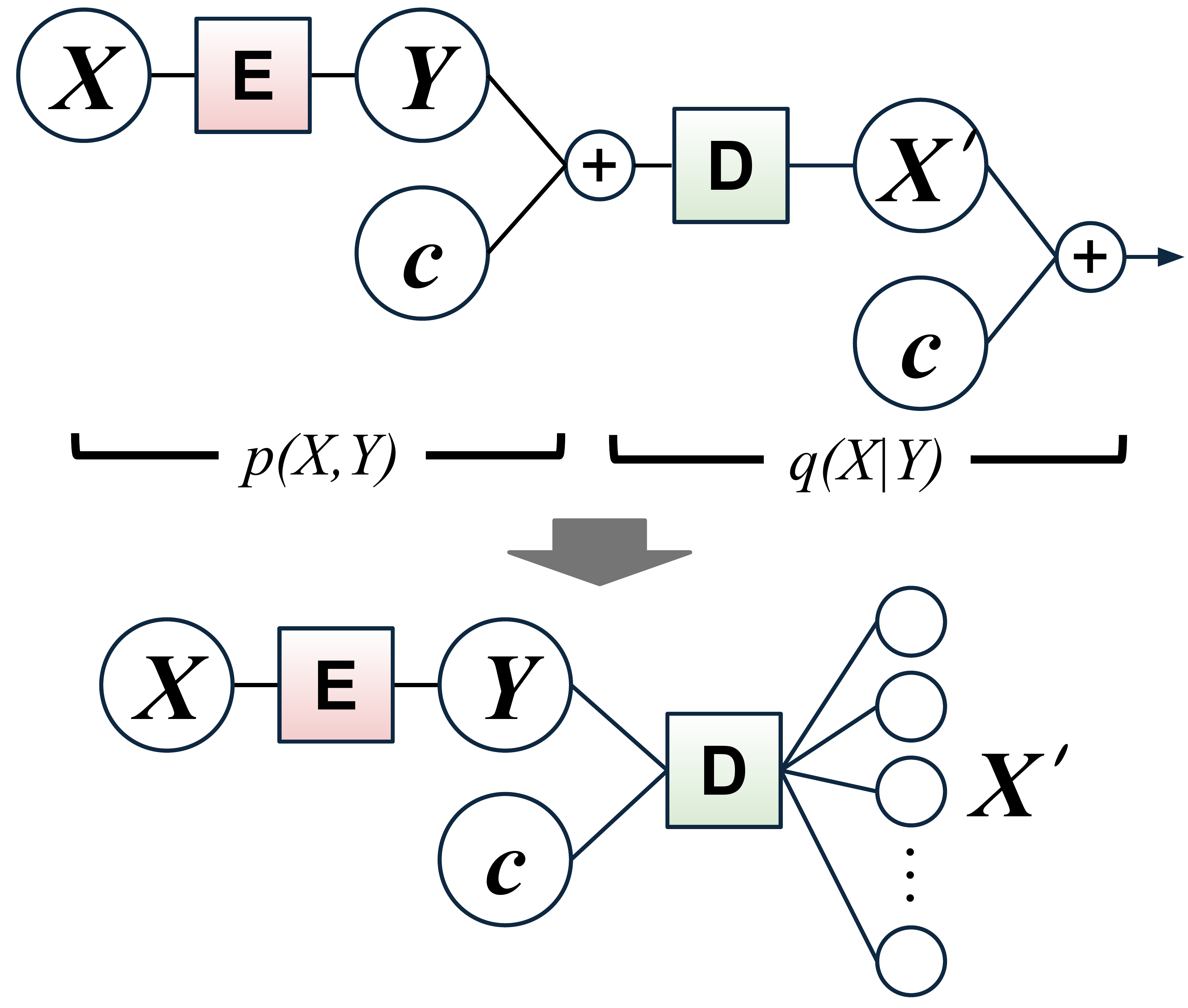}\vspace{-5pt}
\caption{A visualization of the changes we made in Eq.~\eqref{change_mixture_decoder}. (1) The noise $\mathbf{c}$ is not added to features $Y$ but concatenated with $Y$ as inputs to the decoder; (2) Sampling multiple $\mathbf{c}$ will generate multiple $X'$, making the decoder multiple-output and one-to-many; (3) The conditional density is taking the average over Gaussians defined by the multiple centers of $X'$; (4) The norm of the conditional has a closed form, so we can use the normalized inner product in the Hilbert space to optimize this model.}
\label{illustrative_figure}
\end{figure}

We add an illustrative figure in Fig.~\ref{illustrative_figure}. Before, an input $X$ in the decoder $\textbf{E}$ will generate one output $Y$, with a Gaussian noise variable $\mathbf{c}$ that is additive, added to the features $\mathbf{Y}$. The addition of them (so the probability is $p(Y|X)$ is a Gaussian) goes through the decoder $\textbf{D}$, produce $X'$, and added with another additive noise $\mathbf{c}$ such that the conditional density $q(X|Y)$ is also a Gaussian. Our newly proposed framework changes the decoder into one-to-many and defines a mixture. The encoder stays unchanged. But instead of the additive noise, we concatenate the noise and samples as inputs to the decoder. One $Y$ and multiple $\mathbf{c}$ can generate multiple reconstructed samples $X'$, making the decoder a one-to-many mapping. The conditional density of $q(X|Y)$ is given by $q(X|Y) = \int p(c) \mathcal{N}(X-\textbf{D}(Y,c)) dc$, an average over Gaussians with centers defined on outputs $X'$. With $q(X|Y)$ as a mixture density, it is potentially more powerful as a function approximator than a single Gaussian, and we want to claim that this improvement may make the bound more tight. 

By the property that the norm of a Gaussian mixture density has a closed form, the norm $||q(X|Y)||_{p(Y)}^2$ for each $Y$ will have a closed form. So we can use the bound and the cost defined in Eq.~\eqref{conditional_cost1} and~\eqref{conditional_cost2} to train this encoder-mixture-decoder. 

\begin{equation}
\begin{gathered}
q(X|Y) = \mathcal{N}(X-\textbf{D}(Y)) \\
\Downarrow \vspace{-3pt} \\
q(X|Y) = \int p(c) \mathcal{N}(X-\textbf{D}(Y, c)) \, dc. \vspace{-7pt}
\end{gathered}
\label{change_mixture_decoder}
\end{equation}

\noindent Now we include samples in the equations how to empirically construct the cost and optimize it. Given a batch of samples $X_1,X_2,\cdots,X_N$. Suppose they go through the encoder and produce $Y_1,Y_2,\cdots, Y_N$. For each such $Y_n$, we sample the noise from the prior $K$ times to have $c_n(1), c_n(2),\cdots,c_n(K)$. Then we concatenate one $Y_n$ with one $c_n(k)$, go through the decoder, and generate multiple $X_n'(1), X_n'(2),\cdots,X_n'(K)$ for each $n$. 

The numerator, the inner product, is computed as follows. We fix each $X_n$, gather the multiple reconstructions of it $X_n(1),X_n(2),\cdots,X_n(K)$, we construct the Gaussian differences between $X_n$ and each $X_n'(k)$, and compute the double sum between them. One $X_n$ will correspond to multiple $X_n'(k)$. We take the average, that is the double sum of Gaussian differences, with sum of two variances, $v_p$ for the data and $v_q$ for the model. 

The denominator, the norm of $q(X|Y)$ under $p(Y)$, is computed as follows. We fix each $X_n$, instead of computing all pair-wise Gaussian distances this sample $X_n$ with all its samples $X_n'$, we compute all the Gaussian pairwise differences among the reconstructions of $X_n$, that is the pairwise distances between all $X_n'(i)$ and $X_n'(j)$ from one $X_n$, and compute the double sum and take the average over all samples. Notice that the denominator only involves the pairwise distances between $X_n'(i)$ and $X_n'(j)$, reconstructions of one sample $X_n$. We do not need to compute Gaussian differences between $X_m'(i)$ and $X_n'(j)$, two different reconstructions from two different samples. It might be possible to construct a cost with them, but our purpose here is to match $q(X|Y)$ with $p(X|Y)$ which requires only the norm of $q(X|Y)$, $||q(X|Y)||_{p(Y)}^2$, not requiring any other terms like the norm of $q(X)$. 

\begin{equation}
\resizebox{1\linewidth}{!}{
$\begin{gathered}
X_n'(k) = \textbf{D}(\textbf{E}(X_n), c_n(k)), \;\; k=1,2,\cdots,K,\\
\langle p(X|Y), q(X|Y)\rangle_{p(Y)} = \frac{1}{NK}\sum_{n=1}^N\sum_{k=1}^K\mathcal{N}(X_n - X_n'(k);v_p + v_q), \\
||q(X|Y)||_{p(Y)}^2 = \frac{1}{NK^2}\sum_{n=1}^N \sum_{i=1}^K \sum_{j=1}^K \mathcal{N}(X_n'(i) - X_n'(j);2v_q). 
\end{gathered}$}
\label{sample_estimate}
\end{equation}

\noindent There are two ways of combining the prior $\mathbf{c}$ and the features $\mathbf{Y}$. If the prior $\mathbf{c}$ is discrete, then the first way is to use just $\mathbf{Y}$ as inputs, and make the neural network produce multiple outputs, and each output corresponds to a $\mathbf{c}$. In this way the $\mathbf{c}$ is on the output end of the decoder. The second way is to concatenate features $\mathbf{Y}$ with one-hot vectors. Each one-hot vector represents a different $\mathbf{c}$ and concatenating with different $\mathbf{c}$ will generate different reconstructions $\mathbf{X}'$. In this way the $\mathbf{c}$ is on the input end of the decoder. We find that putting $\mathbf{c}$ in the input of the decoder is slightly more stable than putting it in the output. But if $\mathbf{c}$ is continuous or a prior of combining discrete and continuous, then the only way is using the input priors. We need to concatenate $Y$ and $c$, and sample multiple $c$ to generate multiple reconstructions $X_n'(i)$. The model learned from this has better interpolations because the prior is continuous. Using outputs of the decoder is difficult to apply to a continuous prior, as the number of outputs correspond directly to the dimension of a discrete prior, but parameterizing a continuous prior in this way is much more difficult. 

\newpage 

\section{Results for Autoencoders}

\subsection{Random Walk Example}

\begin{figure}[t]
\centering
\begin{subfigure}{.5\textwidth}\includegraphics[width=\linewidth]{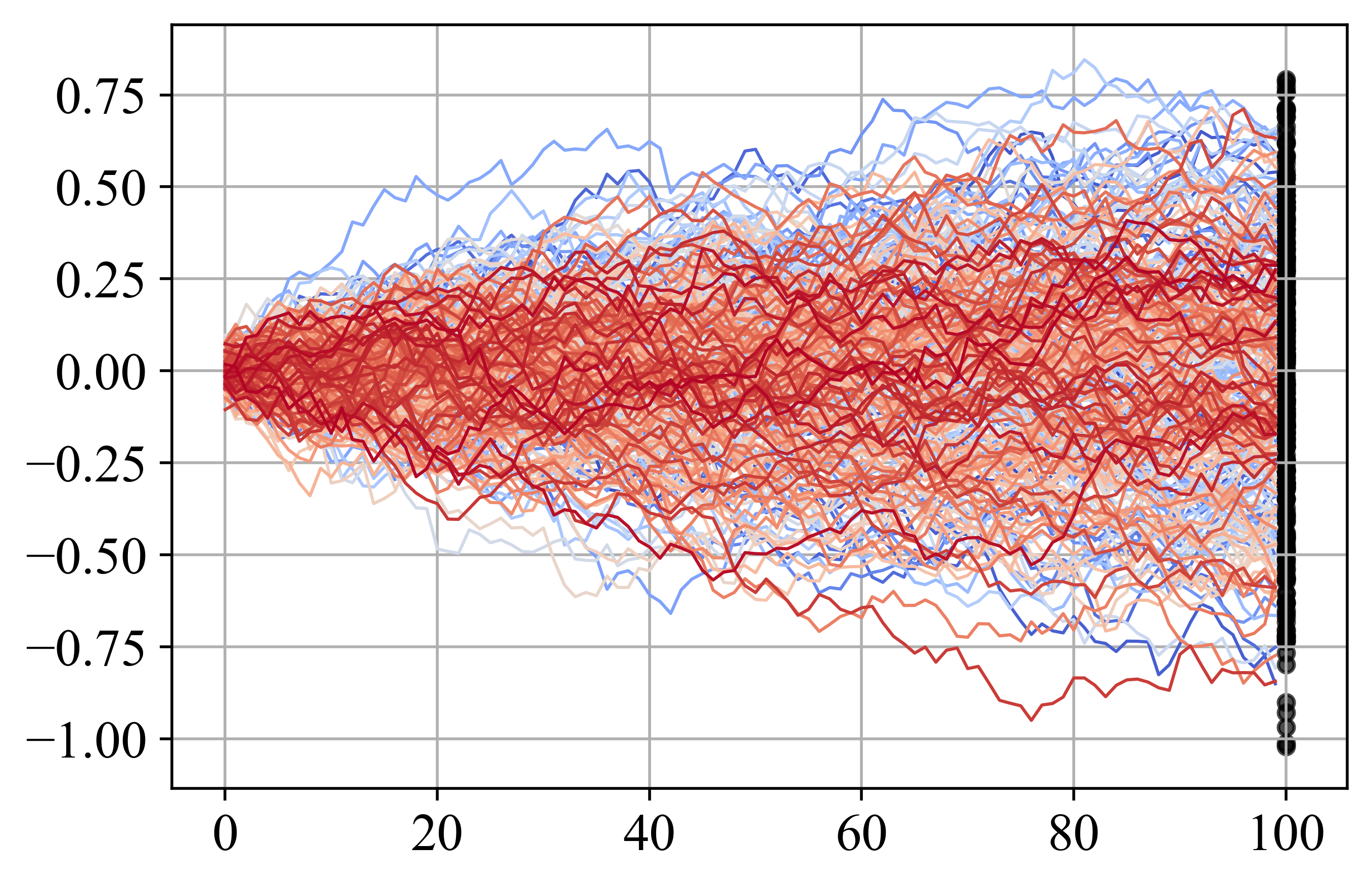}\vspace{-3pt}
\caption{\footnotesize Random Walk Simulations}
\label{Figure_8a}
\end{subfigure}\hspace{-2pt}%
\begin{subfigure}{.5\textwidth}\includegraphics[width=\linewidth]{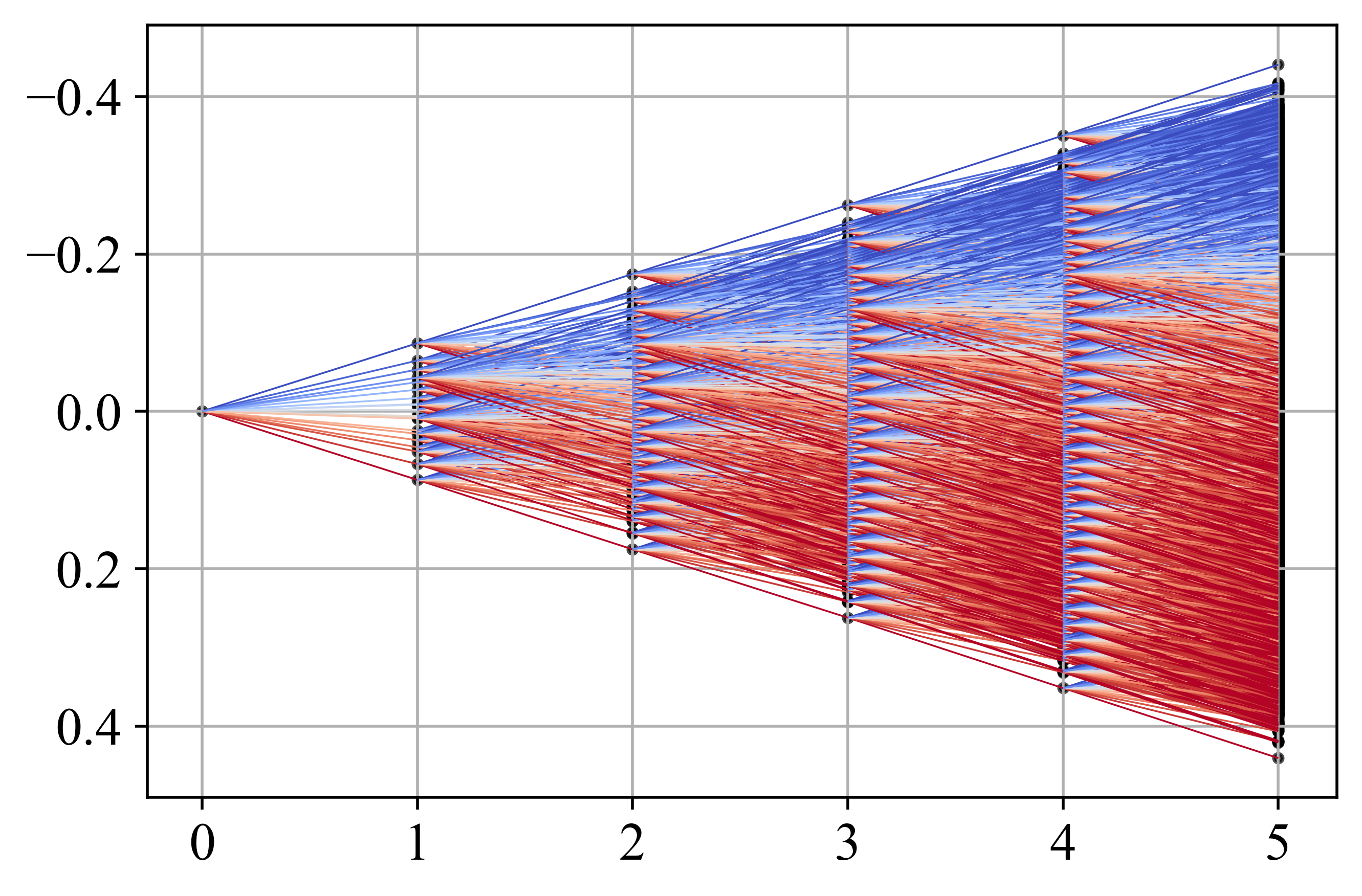}\vspace{-3pt}
\caption{\footnotesize First 6 Steps by Mixture Decoder}
\label{Figure_8b}
\end{subfigure}
\begin{subfigure}{.47\textwidth}\includegraphics[width=\linewidth]{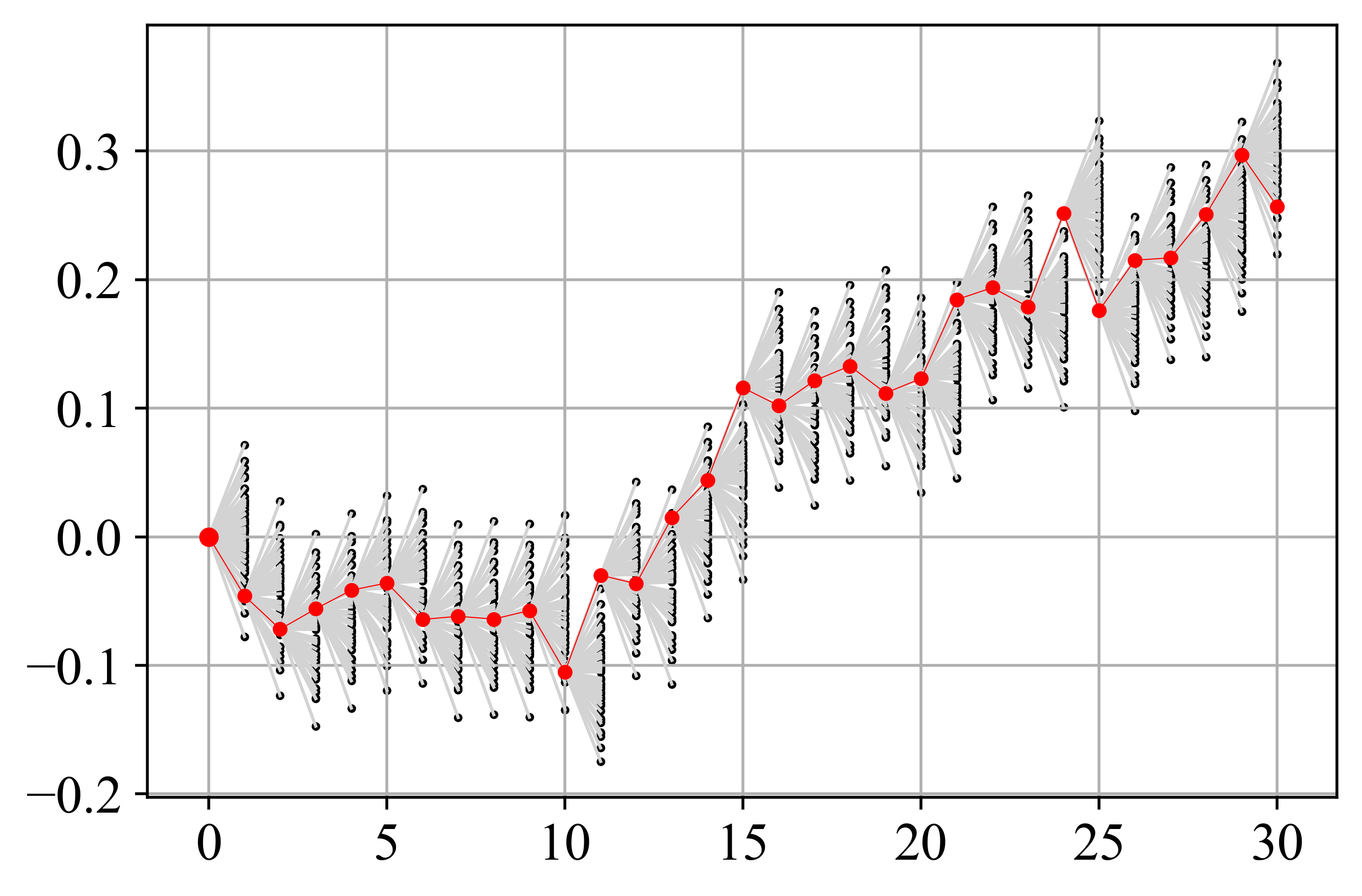}\vspace{-3pt}
\caption{\footnotesize One Walk Simulated by Mixture Decoder (30 Steps)}
\label{Figure_8c}
\end{subfigure}
\begin{subfigure}{.47\textwidth}\includegraphics[width=\linewidth]{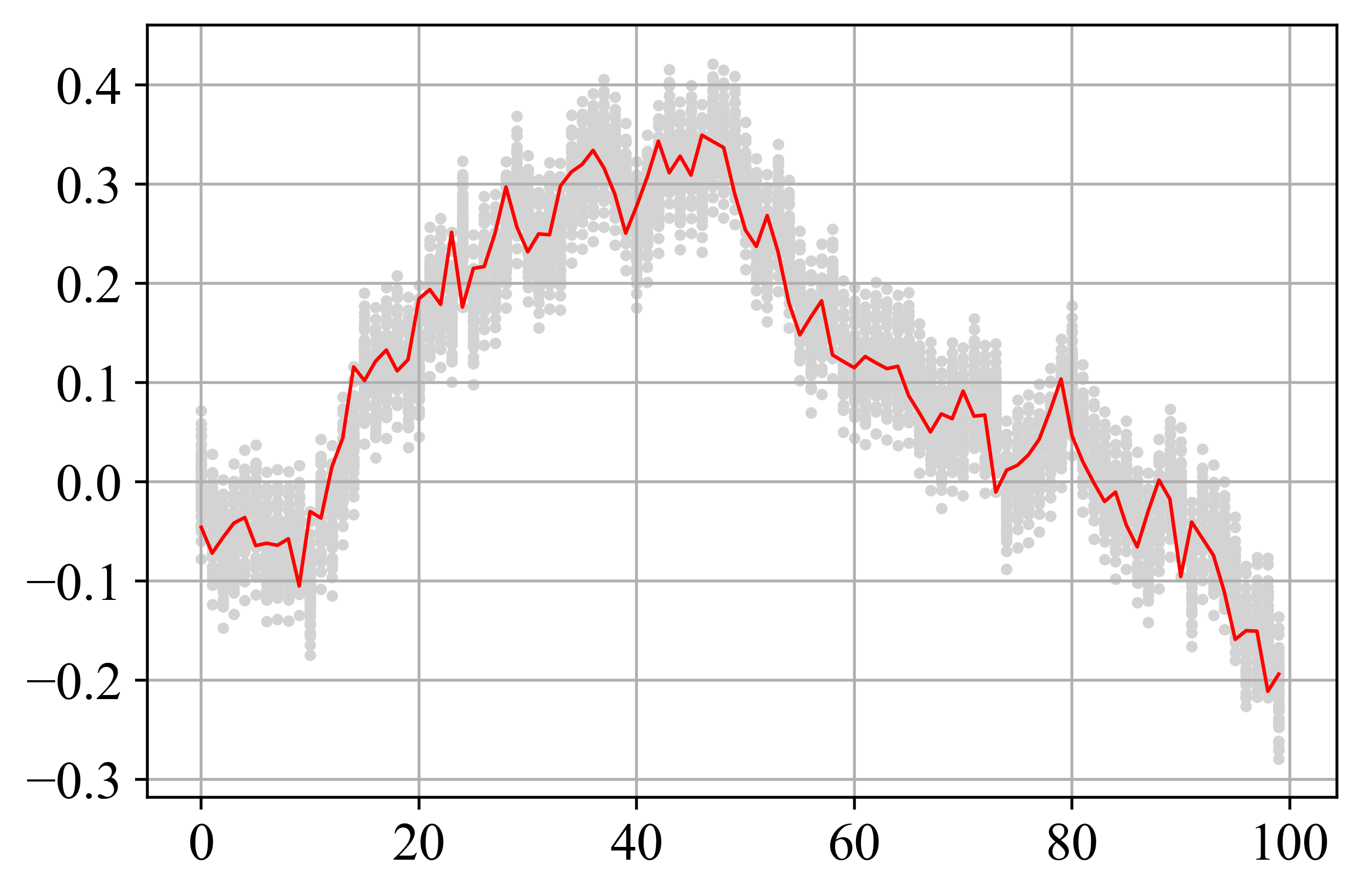}\vspace{-3pt}
\caption{\footnotesize One Walk Simulated by Mixture Decoder (100 Steps)}
\label{Figure_8d}
\end{subfigure}
\begin{subfigure}{.47\textwidth}\includegraphics[width=\linewidth]{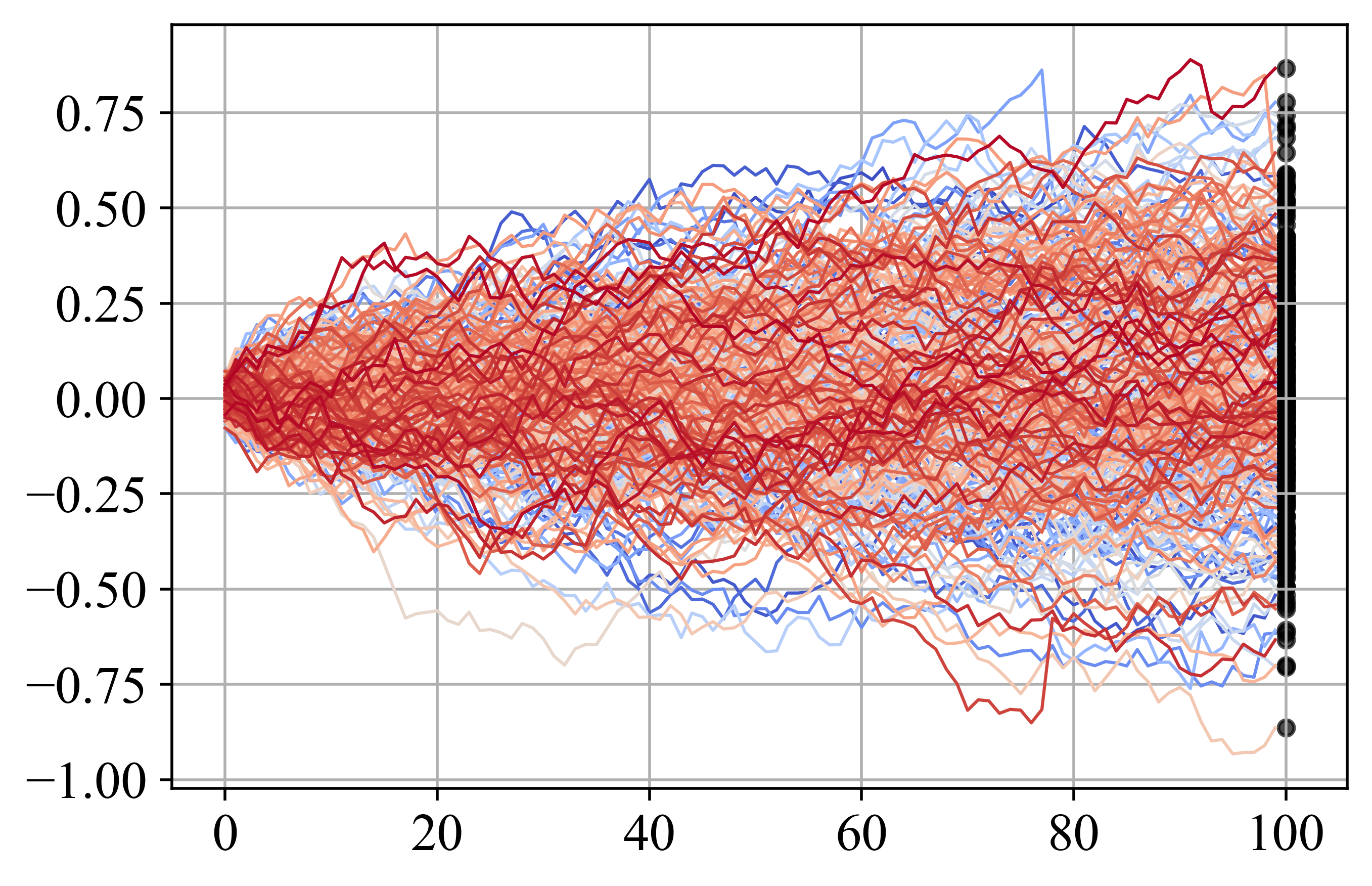}\vspace{-3pt}
\caption{\footnotesize $300$ Simulations Simulated by Mixture Decoder}
\label{Figure_8e}
\end{subfigure}
\begin{subfigure}{.47\textwidth}\includegraphics[width=\linewidth]{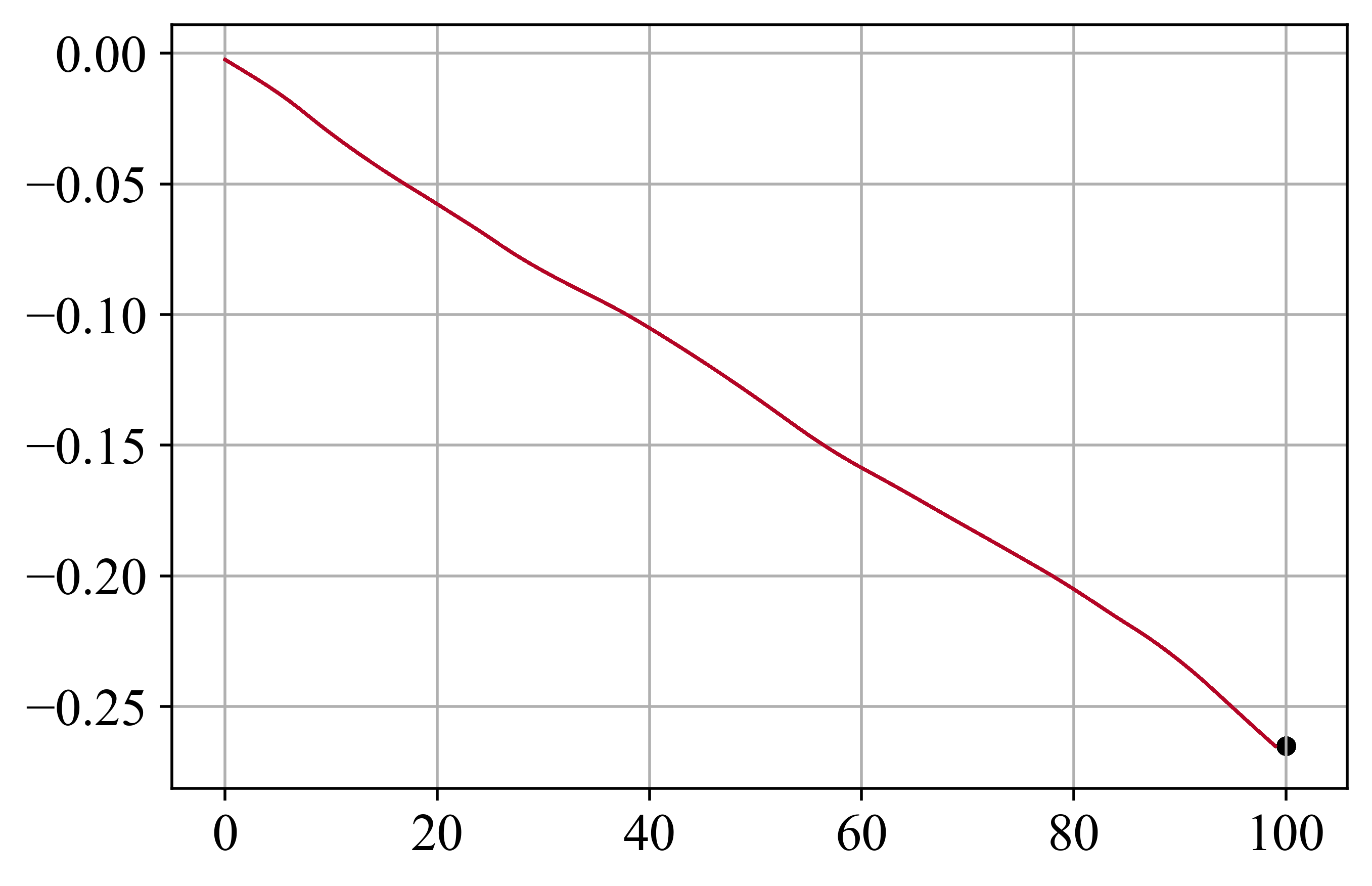}\vspace{-3pt}
\caption{\footnotesize Simulations Simulated by a Regular Deterministic Decoder }
\label{Figure_8f}
\end{subfigure}
\caption{Simulated random walks as data samples and simulations generated by a mixture decoder and a regular decoder after training. We visualize $300$ simulations shown in (a). After training the mixture decoder, we visualize all possible paths in (b). The decoder has $50$ outputs, discretizing the conditional $p(X_{t+1}|X_t)$ that is a conditional Gaussian with $X_t$ as the mean. Every output of the $50$ outputs will generate another $50$ outputs by the neural network, and so forth. (c) Visualizing the first $30$ steps of a simulation produced by the mixture decoder. At each step the neural network has $50$ outputs, and we randomly pick one of them as the next step. (d) Visualizing the $100$ steps of a random walk simulated by the mixture decoder. It can only be applied to $100$ steps as with longer steps it is difficult for the model to converge. This is because the variance can be very large at the end of $1000$ steps and it requires many samples to approximate the conditional. (e) $300$ simulations produced by the mixture decoder, which can be seen that it is very similar to (a). (f) If we just train a deterministic decoder, it can be expected that it will be just one curve.}
\end{figure}

Our first example is a simple random walk example that shows the advantage of having a multiple-output neural network to approximate the conditional. The random walk is a simple Markov process with each step being an additive Gaussian. Given $X_t$, the conditional probability of $X_{t+1}$ given $X_t$ is just a conditional Gaussian mean shift $p(X_{t+1}|X_t) = \mathcal{N}(X_{t+1}-X_t)$. We simulate $300$ walks, each with length $100$, visualized in Fig.~\ref{Figure_8a}. The reason why we do not simulate for longer lengths is that for longer length is requires more points trials/walks to fill in the blank and the approximation does not work that well for a longer length. We also divide the magnitude (y-axis) by 30 such that the activation function of the neural network can be a $\tanh$ function. 

The reason why we present this example is as follows:
\begin{enumerate}[leftmargin=*]
\item The diffusion model can be seen as a random walk walking backwards;\vspace{3pt}
\item It has very good properties. For example, the marginal at any $t$ is a Gaussian distribution with an increasing variance as $v$ increases. The region that the walks span also looks like a Gaussian. If one zooms in at any scale of this random walk, with enough points it also looks like a Gaussian. So the additivity is a Gaussian and distributions and the region of the walk are also Gaussians. It is also interesting to investigate the accumulative error in this case.\vspace{3pt}
\item It shows the advantage of approximating $p(Y|X)$ with a mixture density using a multivariate network. If $p(X_{t+1}|X_t)$ is a stochastic process not a deterministic process, and we do not know the range that the next point will land in the space. One way is to discretize the entire space and using the cross entropy to approximate it. But because such range can be very small and can be just a small neighborhood compared to the entire sample space, a more efficient and elegant way should be generating centers by ourselves and defining a mixture density with these centers to approximate the conditional of the next point. A even more elegant way is to quantify the increment because this is a random walk, and the increment is just an additive standardized Gaussian. But here we still predict the next point ($X_t$ as inputs and the probabilities of $X_{t+1}$ as outputs).\vspace{3pt}
\item This also can be seen as a discretization of the random walks. At each step of the walk, since the next step is sampled from a conditional Gaussian and there is only a small range that the point will fall into, a discrete prior is sufficient. But because the probability has to be dependent on the past sample $X_t$, the possible way to parameterize it is using this sample as the input to a neural net. The neural network is multiple-output, producing all possible outcomes in a discretized way. Combining random walks, a multiple-output mapping function, and the cost in $L_2$, we are able to approximate the conditional of $X_{t+1}$ given $X_t$, predicting the possible landing point of the next point using a discrete prior.\vspace{3pt}
\end{enumerate}

We choose the discrete prior to be $50$ discrete centers. We use the $L_2$ costs presented in Eq.~\eqref{conditional_cost1} and Eq.~\eqref{conditional_cost2}. The procedure is as follows. First sample a batch of $X_t$ randomly from all trials and all time. Then we make these sample points go through the decoder and in this example only the decoder is needed. The decoder produces $50$ outputs, each one of them corresponding to a center to define the estimation of $p(X_{t+1}|X_t)$. Having introduced in Eq.~\eqref{sample_estimate}, we need to first compute the Gaussian differences between all $50$ points of $\textbf{D}(X_t)$, the outputs of the decoder, and $X_{t+1}$, the next point as the target, and average them to get the inner product $\langle p(X_{t+1}|X_t), q(X_{t+1}|X_t)\rangle_{p(X_t)}$. Then we compute the Gaussian differences between the $50$ points of $\textbf{D}(X_t)$ themselves, for each given $50$ points, and average them to get the norm $||q(X_{t+1}|X_t)||_{p(Y)}^2$. 

\newpage

In Fig.~\ref{Figure_8a} to~\ref{Figure_8f}, we compare the simulated random walks as data samples and simulations generated by the trained mixture decoder and a regular decoder. It shows that after using our proposed cost to train a multiple-output mixture decoder, at each step there are $50$ outputs that represent the possible value of the next time step (Fig.~\ref{Figure_8b},~\ref{Figure_8c},~\ref{Figure_8d}), and we randomly sample from one of them, and simulate the random walks in $100$ steps. The simulations generated by the mixture decoder (Fig.~\ref{Figure_8e}) is very similar to the actual random walk simulations. If the neural network is deterministic, we can only get one curve in the space and it is expected (Fig.~\ref{Figure_8f}). 

\subsection{Encoder-Mixture-Decoder for Simple Datasets}
\label{section_results_mixture_data}

\begin{figure}[t]

\begin{subfigure}{.3\textwidth}\includegraphics[width=\linewidth]{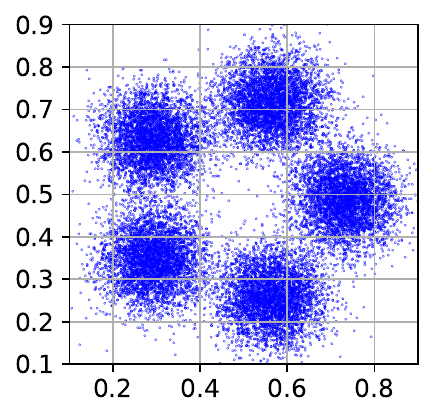}\vspace{-6pt}
\phantomcaption
\label{9a}
\end{subfigure}\hspace{-1pt}
\begin{subfigure}{.3\textwidth}\includegraphics[width=\linewidth]{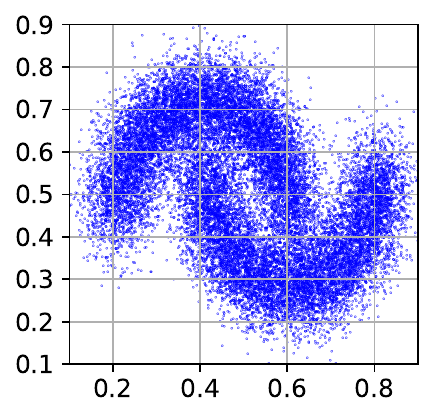}\vspace{-6pt}
\phantomcaption
\label{9b}
\end{subfigure}\hspace{-1pt}
\begin{subfigure}{.3\textwidth}\includegraphics[width=\linewidth]{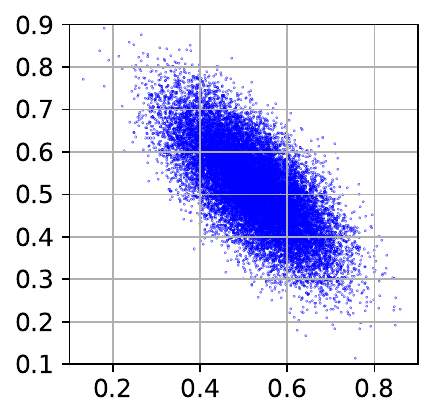}\vspace{-6pt}
\phantomcaption
\label{9c}
\end{subfigure}\vspace{-9pt}

(a) The actual data. \vspace{3pt}

\begin{subfigure}{.3\textwidth}\includegraphics[width=\linewidth]{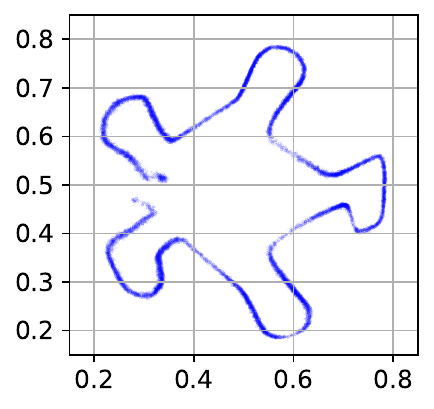}\vspace{-4pt}%
\phantomcaption
\label{9d}
\end{subfigure}\hspace{-1pt}
\begin{subfigure}{.3\textwidth}\includegraphics[width=\linewidth]{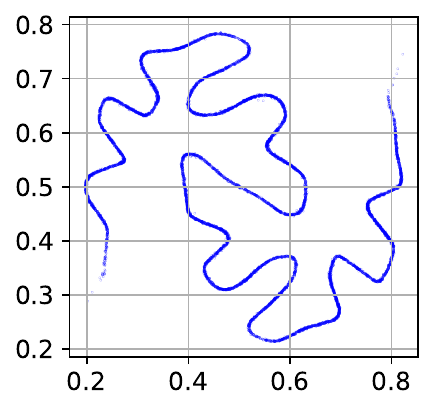}\vspace{-6pt}%
\phantomcaption
\end{subfigure}\hspace{-1pt}
\begin{subfigure}{.3\textwidth}\includegraphics[width=\linewidth]{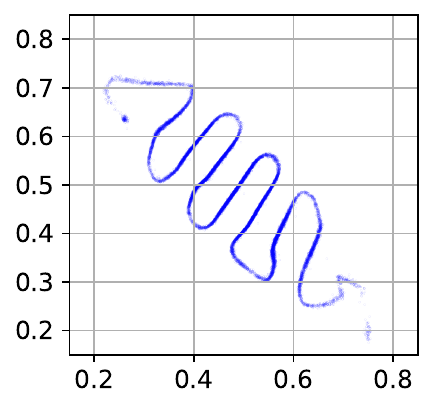}\vspace{-6pt}%
\phantomcaption
\end{subfigure}\vspace{3pt}

(b) Outputs from the regular autoencoder.\vspace{3pt}

\begin{subfigure}{.3\textwidth}\includegraphics[width=\linewidth]{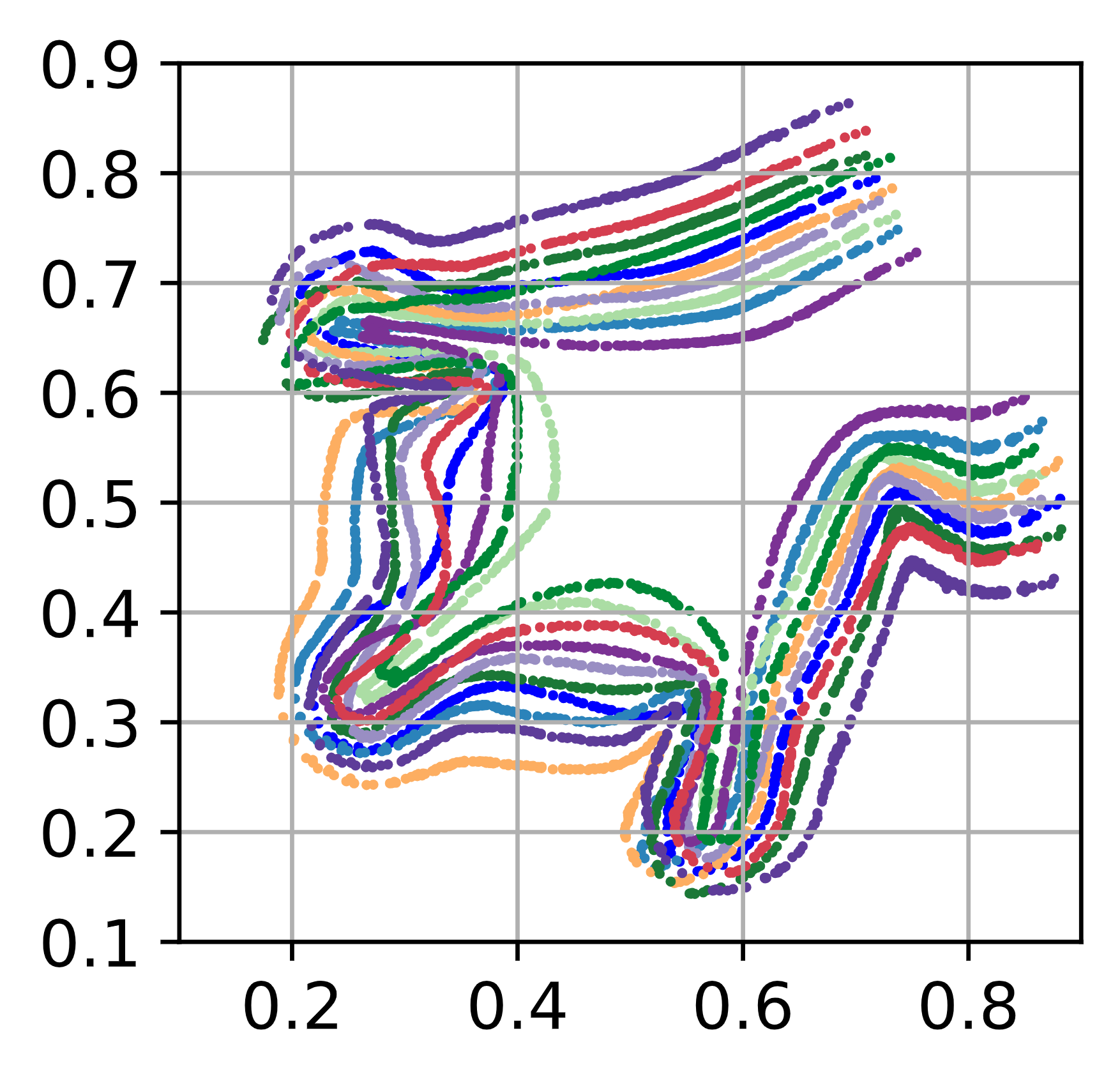}\vspace{-6pt}
\phantomcaption
\end{subfigure}\hspace{-1pt}
\begin{subfigure}{.3\textwidth}\includegraphics[width=\linewidth]{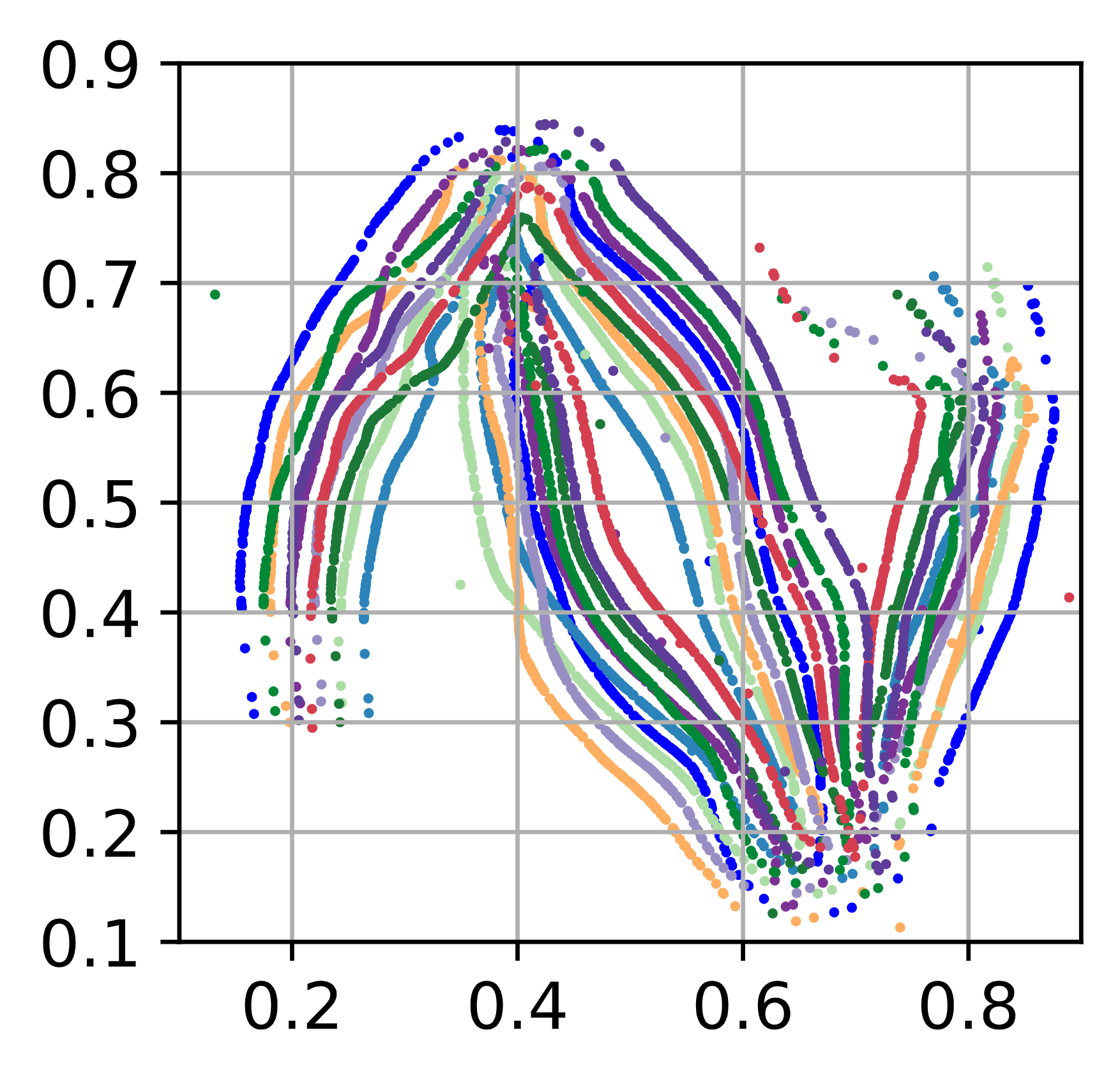}\vspace{-6pt}
\phantomcaption
\end{subfigure}\hspace{-1pt}
\begin{subfigure}{.3\textwidth}\includegraphics[width=\linewidth]{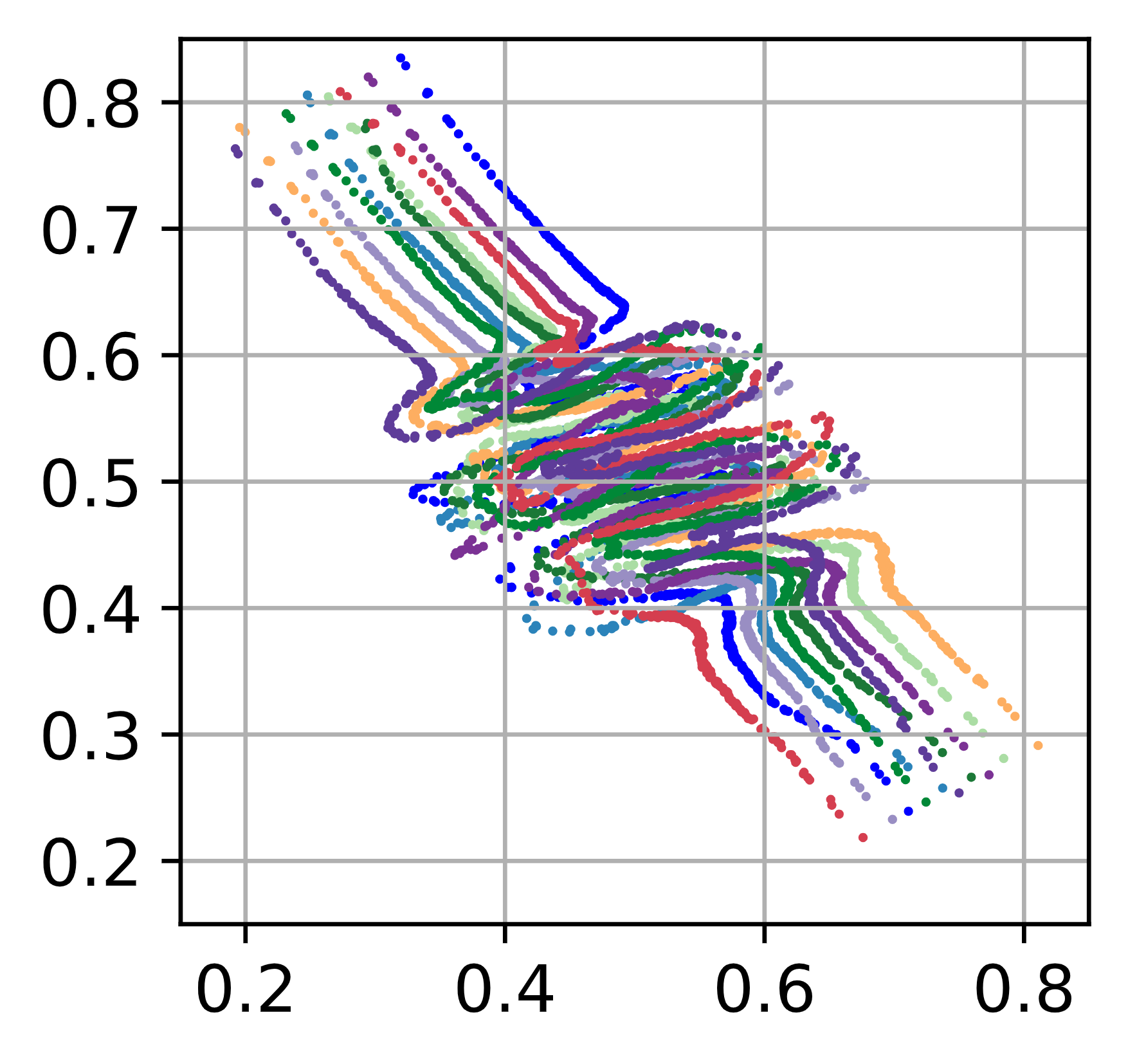}\vspace{-6pt}
\phantomcaption
\end{subfigure}\vspace{3pt}

(c) Encoder-mixture-decoder (discrete prior).\vspace{3pt}

\begin{subfigure}{.3\textwidth}\includegraphics[width=\linewidth]{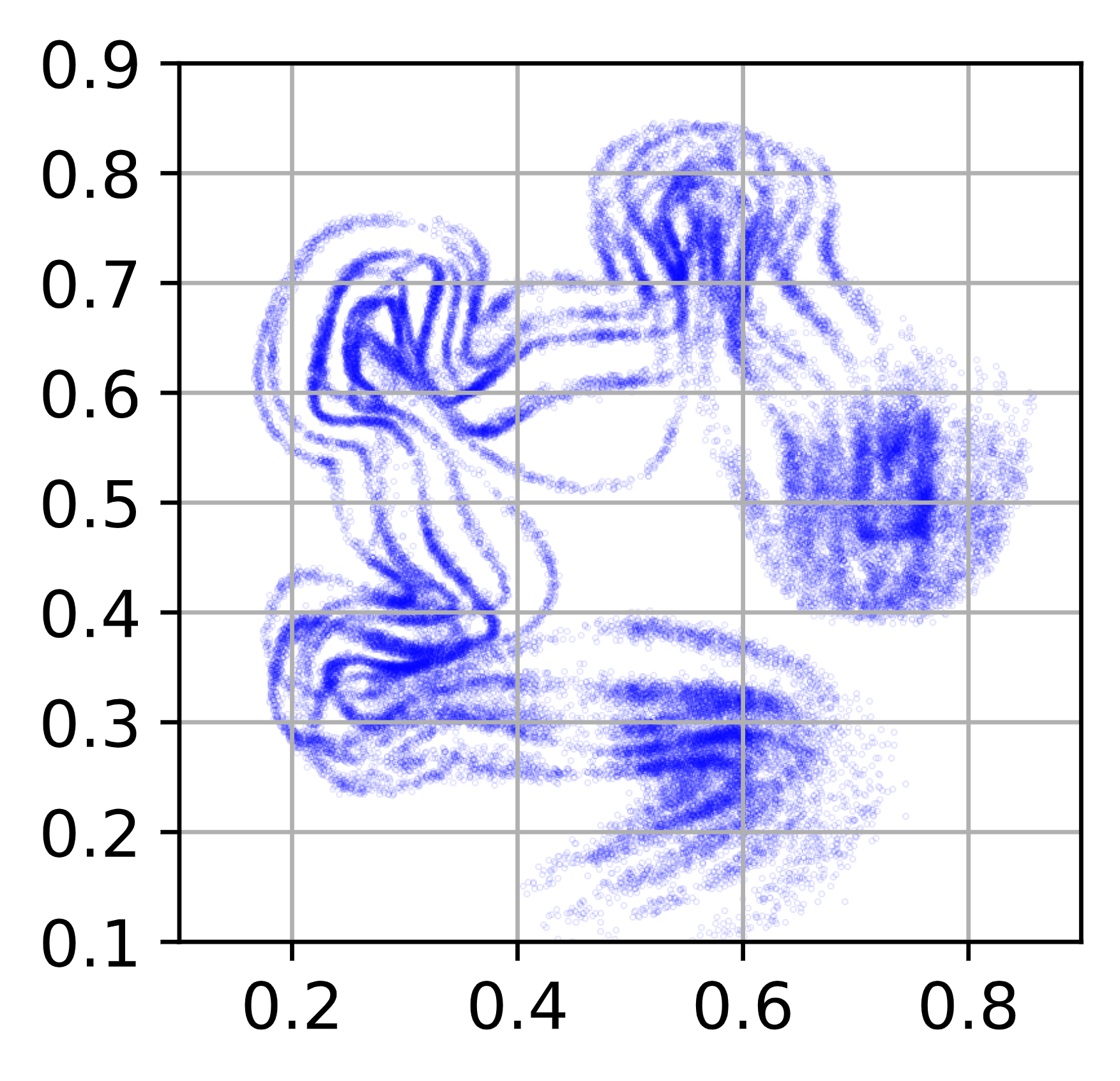}\vspace{-6pt}
\phantomcaption
\end{subfigure}\hspace{-1pt}
\begin{subfigure}{.3\textwidth}\includegraphics[width=\linewidth]{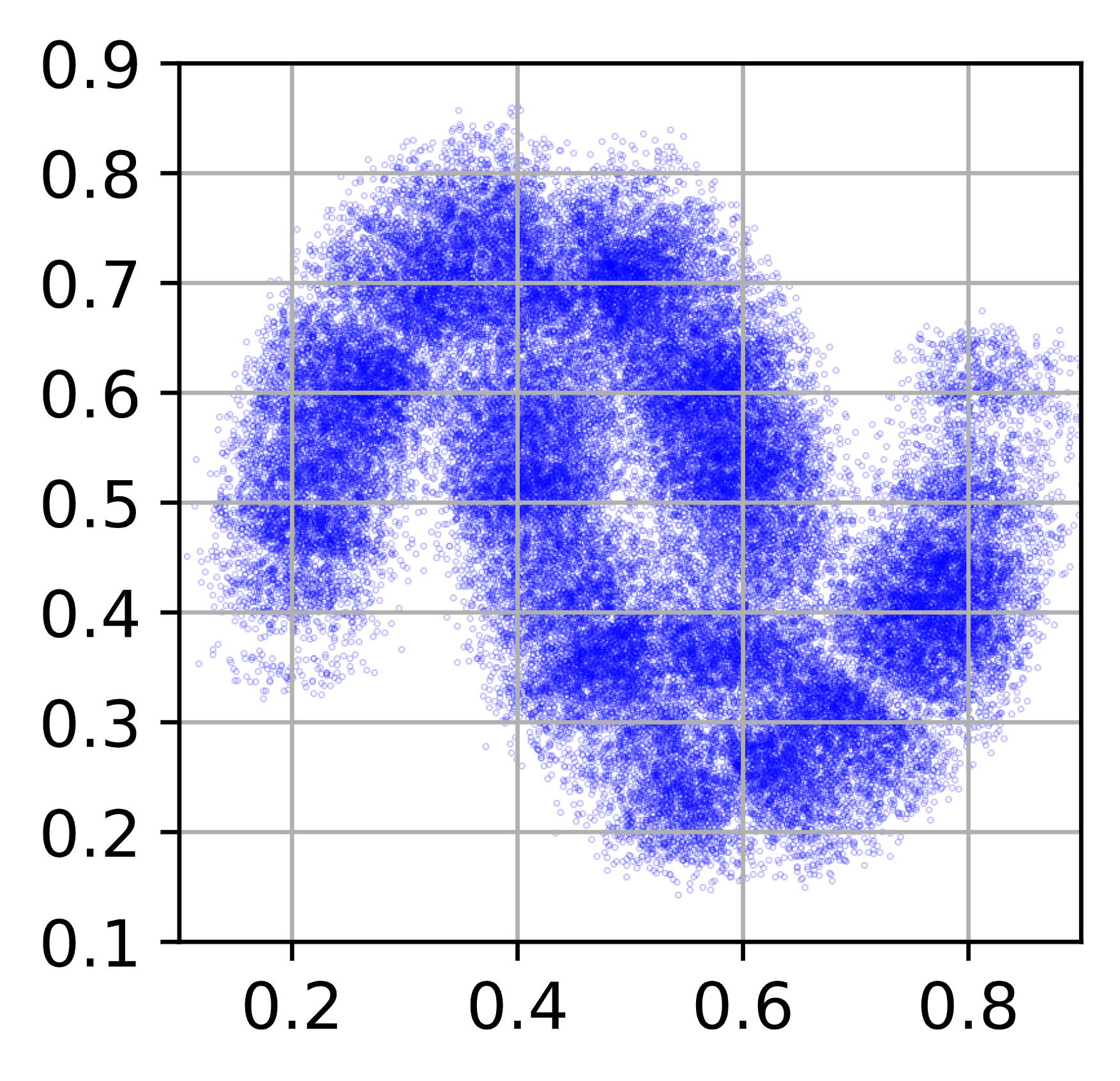}\vspace{-6pt}
\phantomcaption
\end{subfigure}\hspace{-1pt}
\begin{subfigure}{.3\textwidth}\includegraphics[width=\linewidth]{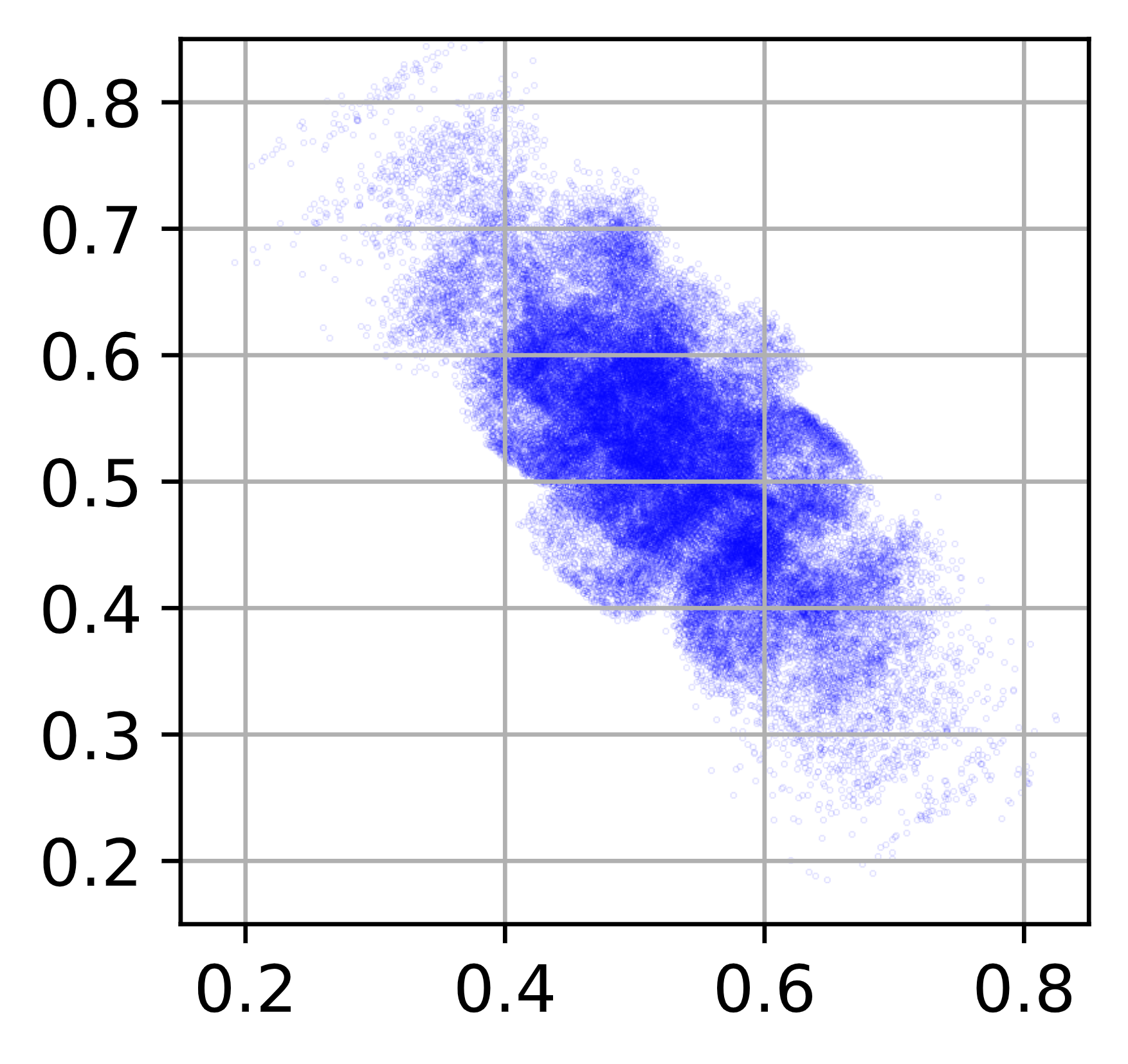}\vspace{-6pt}
\phantomcaption
\end{subfigure}\vspace{3pt}

(d) Encoder-mixture-decoder (discrete-continuous prior).\vspace{-3pt}

\caption{Results of projecting $2D$ toy datasets into $1D$ features via an encoder, and back to reconstructions via the regular decoder or the mixture decoder proposed by us. With $10$ centers for each sample, a fixed data variance $v_p$ and a fixed model variance $v_q$, the reconstructions via the mixture decoder in (c) is $10$ continuous curves in the sample space, compared to just one curve in (b). With a hybrid discrete-continuous prior, the mixture decoder can almost reconstruct the toy distributions, shown in (d).}
\label{figure9ae}
\end{figure}

We then present a side-by-side comparison results of training a regular encoder-decoder autoencoder and a encoder-mixture-decoder. We start with simple toy distributions and extend it to image datasets. We intentionally design the task to be the simplest such that the numerical quantitative analysis is possible. The task is to project $2D$ toy distributions visualized in Fig.~\ref{9a} to $1D$ features through an encoder, then back to $2D$ reconstructions through a decoder or a mixture decoder, and make the comparison. The datasets we use are a five-state Gaussian mixture, a two-moon distribution, and a simple Gaussian. Since features in $1D$ is insufficient to represent $2D$ samples if the distribution is complex, thus the reconstructions will not be exact and will have a marginal of error. This provides the possibility of investigating and comparing the solutions. 

\begin{figure}[H]
\centering
\begin{subfigure}{.23\textwidth}\includegraphics[width=\linewidth]{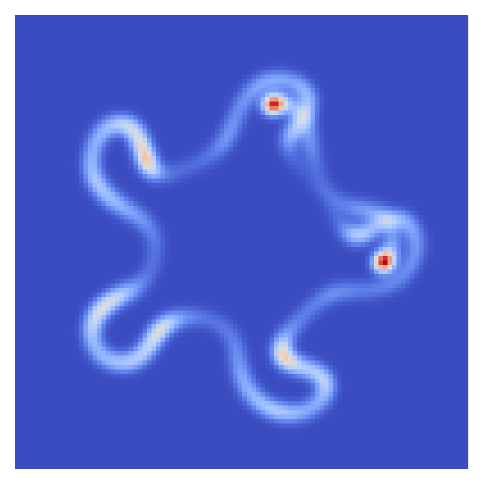}
\phantomcaption
\end{subfigure}
\begin{subfigure}{.23\textwidth}\includegraphics[width=\linewidth]{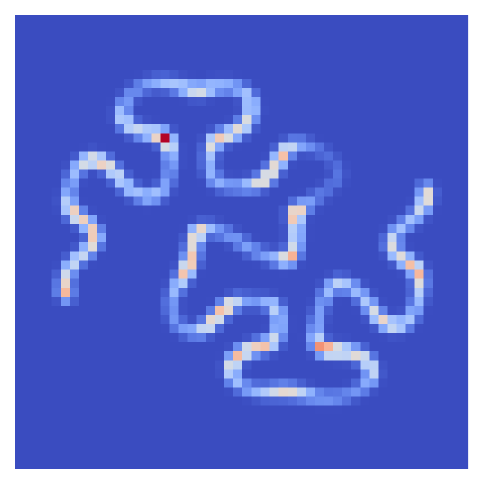}
\phantomcaption
\end{subfigure}
\begin{subfigure}{.23\textwidth}\includegraphics[width=\linewidth]
{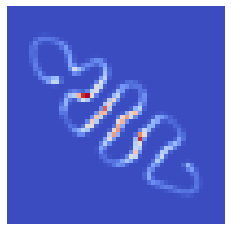}
\phantomcaption
\end{subfigure}\vspace{-5pt}
\caption{We found that even for the solution of the regular autoencoder, the solution is almost stably unique, irrelevant to the training parameters. The reason could be that the optimal solution is finding the conditional densities $p(Y|X)$ and $q(X|Y)$ that maximize an objective relevant only to the densities, regardless of the model parameterization. This can be verified by learning the densities directly through a neural network, instead of training the autoencoder. The learned solution of the densities shown in the figure matches the solution produced by the neural network in Fig.~\ref{9d}. The details of generating these heatmaps are left in Appendix.~\ref{Appendix_autoencoder}.}
\label{fig10}
\end{figure}

We first train an autoencoder, and the reconstructions are shown in Fig.~\ref{9b}. We found that the solution is almost unique, regardless of how we change the parameters and random seeds. The shape of the reconstructions looks like a continuous curve in the sample space, which is due to that the features are in $1D$. The uniqueness of the solution is because of the fact that the solution may only be related to the density, as the objective is finding the $1D$ $Y$ such that the mutual information between $X$ and $Y$ is the maximal. We find that indeed it is possible to learn this solution not from training an autoencoder, but by learning the densities $p(X|Y)$ and $q(Y|X)$ by optimizing an objective of the cross entropy with neural networks, shown in Fig.~\ref{fig10}, which may demonstrate that the solution of the autoencoder is indeed finding the optimal densities $p(X|Y)$ and $q(Y|X)$. The details of producing Fig.~\ref{fig10} are left in the Appendix~\ref{Appendix_autoencoder}. In summary, the solution when training an autoencoder to project $2D$ distributions to $1D$ and back to $2D$ is quite stable and unique, where the reconstructions look like a smooth continuous curve in the sample space. 

Now we train the proposed encoder-mixture-decoder. We first train the case of the discrete prior. We choose the output dimension $K$ to be $10$. We need to fix the data variance $v_p$ and the model variance $v_q$ to be $0.001$ as if we make them trainable to be arbitrary values, it can be expected that they will find the centers of the Gaussians or two deterministic curves of the two-moons, and all the toy example can actually be represented by deterministic center points with an additive Gaussian. Immediately we see that it is $10$ continuous curves in the reconstruction space, visualized in Fig.~\ref{9c}, instead of just one curve like the regular autoencoder. 

For the discrete-continuous case, we choose the concatenation of a $10$-state categorical distribution and a $30$-dimensional uniform distribution as $\mathbf{c}$, then we use the concatenation of $\mathbf{c}$ and features $\mathbf{Y}$ as the inputs to the mixture decoder. Visualized in Fig.~\ref{9d}, it almost reconstructs the original distribution. But also keep in mind that the mapping from samples to the reconstructions is not one-to-one, as the feature dimension is not sufficient, but the mapping retains the maximal mutual information between $X$ and $Y$, as well as matching $q(X)$ with $p(X)$. As discussed in Sec.~\ref{inspiration_from_autoencoders} and Eq.~\eqref{conditional_cost1},~\eqref{conditional_cost2}, there are two conditions for the optimal condition of the cost. First, the joint $p(X,Y) = p(X)p(Y|X)$ and $q(X,Y) = q(Y)\int q(c)\mathcal{N}(X-\textbf{D}(Y,c))dc$ has to match, which means that the marginals $p(X)$ and $q(X)$ will match, which is shown in the figure. Second, the conditional entropy of $Y$ given $X$ has to be minimized, which does not necessarily guarantee the one-to-one mapping between $X$ and $Y$, but ensures the minimal conditional entropy between $X$ and $Y$ or the maximal mutual information. The reconstructions produced by our mixture decoder clearly has a tighter bound between $p(X,Y)$ and $q(X,Y)$, thus making $p(X)$ and $q(X)$ closer (the first condition). Later on we will show quantitatively that using a mixture decoder may also potentially lift the bound and find a better conditional entropy, i.e., a better optimal solution (the second condition). 

\subsection{Encoder-Mixture-Decoder for Image Datasets}
\label{section_real_images}

We also implement the encoder-mixture-decoder for image datasets such as MNIST and CelebA. Since CelebA has $200k$ samples, we train the model on its first $40,000$ samples for simplicity, and there is no obstacle to run it for the full dataset. We still project images into $1D$ features and back to the reconstructions in the sample space as it is easier to spot the difference and do quantitative analysis. The noise is a hybrid of uniform and discrete noises. For each sample, after they are projected to $1D$ features, the features will be concatenated with random uniform noises and discrete noises sampled from categorical distributions, and we sample $30$ of them. This will produce $30$ reconstructions for each sample. Then we maximize the normalized inner product cost in $L_2$ (Eq.~\eqref{conditional_cost1},~\eqref{conditional_cost2}). We first focus on directly comparing the quality and diversity of the reconstructed samples. Later on we will perform extensive quantitative analysis in Sec.~\ref{quantitative_analysiss}.

One important question is whether using more centers with the mixture decoder, instead of using just one center with the regular decoder, will make the bound more tight and increase the value of the upper bound. The argument is because $p$ and $q$ are closer with more number of centers. However, we did not spot a significant increase of the cost value using $1$ or more centers, although we did observe an improvement of the reconstruction quality. The improvement of the optimal cost can be observed in the quantitative analysis example when we train the model on a fixed subset of $800$ samples, shown in Fig.~\ref{MNIST_NUMERICAL} and Fig.~\ref{CELEBA_NUMERICAL}.

\begin{figure}[t]
\centering
\begin{subfigure}{.8\textwidth}\includegraphics[width=\linewidth]{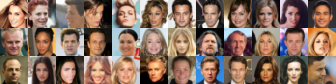}\vspace{-4pt}
\caption{Original Samples from CelebA}
\end{subfigure}
\begin{subfigure}{.8\textwidth}\includegraphics[width=\linewidth]{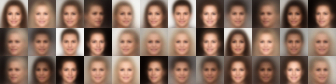}\vspace{-4pt}
\caption{Reconstructions by A Regular Autoencoder}
\end{subfigure}
\begin{subfigure}{.83\textwidth}\includegraphics[width=\linewidth]{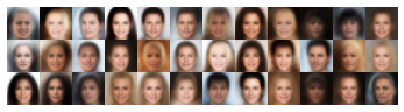}\vspace{-5pt}
\caption{Reconstructions by An Encoder-Mixture-Decoder}
\end{subfigure}
\begin{subfigure}{.8\textwidth}\includegraphics[width=\linewidth]{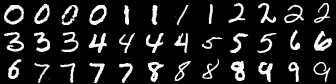}\vspace{-4pt}
\caption{Original Samples from MNIST}
\end{subfigure}
\begin{subfigure}{.8\textwidth}\includegraphics[width=\linewidth]{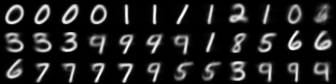}\vspace{-4pt}
\caption{Reconstructions by A Regular Autoencoder}
\end{subfigure}
\begin{subfigure}{.83\textwidth}\includegraphics[width=\linewidth]{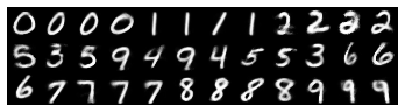}\vspace{-4pt}
\caption{Reconstructions by An Encoder-Mixture-Decoder}
\end{subfigure}\vspace{-5pt}
\caption{Side-by-side comparisons of reconstructions from the autoencoder and the encoder-mixture-decoder, for CelebA and MNIST. The feature dimension is only $1$ and the number of outputs of the mixture decoder is set to be $30$. The noise is chosen as a hybrid of continuous and discrete noise. We visualize one reconstruction for each sample. We do spot a noticeable increase in the reconstruction quality and diversity for the mixture decoder. The data and model variances $v_p$ and $v_q$ in the cost need to be handpicked, and we found that the smaller the variances, the more diversity and high-quality the reconstructions would be. But the model is trainable only within a range of variances.}
\label{fig11}
\end{figure}

Fig.~\ref{fig11} is a direct comparison of the performance of the reconstruction. We want to emphasize that the variances $v_p$ and $v_q$ do need to handpicked, and the smaller the variance value, the more diverse and higher quality the reconstruction samples would be. Each sample will produce $30$ reconstructions by the mixture decoder, and we randomly pick one of them to visualize. We do find that the reconstruction quality of the mixture decoder is better than the regular decoder. 

\begin{figure}[H]
\begin{subfigure}{.32\textwidth}\includegraphics[width=\linewidth]{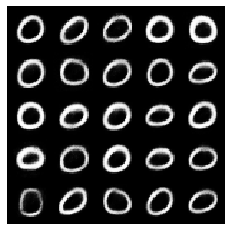}
\phantomcaption
\end{subfigure}
\begin{subfigure}{.32\textwidth}\includegraphics[width=\linewidth]{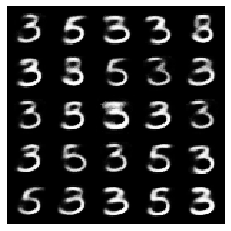}
\phantomcaption
\end{subfigure}
\begin{subfigure}{.32\textwidth}\includegraphics[width=\linewidth]{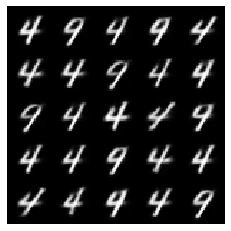}
\phantomcaption
\end{subfigure}\vspace{3pt}
\begin{subfigure}{.32\textwidth}\includegraphics[width=\linewidth]{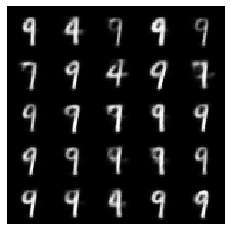}
\phantomcaption
\end{subfigure}
\begin{subfigure}{.32\textwidth}\includegraphics[width=\linewidth]{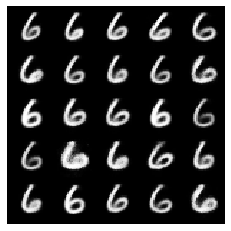}
\phantomcaption
\end{subfigure}
\begin{subfigure}{.32\textwidth}\includegraphics[width=\linewidth]{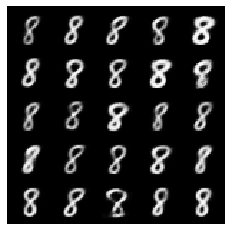}
\phantomcaption
\end{subfigure}\vspace{3pt}
\begin{subfigure}{.32\textwidth}\includegraphics[width=\linewidth]{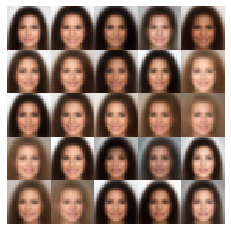}
\phantomcaption
\end{subfigure}
\begin{subfigure}{.32\textwidth}\includegraphics[width=\linewidth]{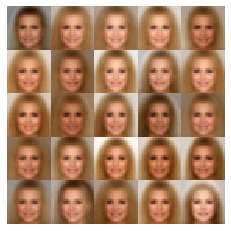}
\phantomcaption
\end{subfigure}
\begin{subfigure}{.32\textwidth}\includegraphics[width=\linewidth]{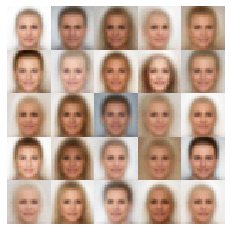}
\phantomcaption
\end{subfigure}\vspace{3pt}
\begin{subfigure}{.32\textwidth}\includegraphics[width=\linewidth]{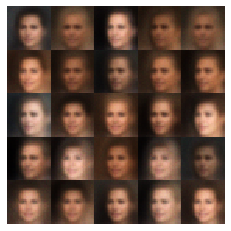}
\phantomcaption
\end{subfigure}
\begin{subfigure}{.32\textwidth}\includegraphics[width=\linewidth]{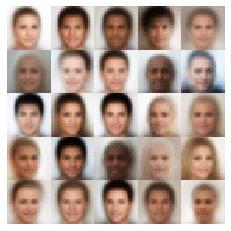}
\phantomcaption
\end{subfigure}
\begin{subfigure}{.32\textwidth}\includegraphics[width=\linewidth]{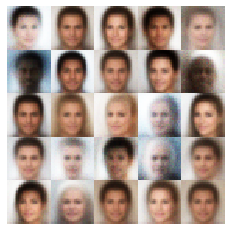}
\phantomcaption
\end{subfigure}\vspace{-7pt}
\caption{The mixture decoder will produce multiple reconstructions by sampling $\mathbf{c}$ multiple times. We visualize $25$ samples for a handful of exemplars. One training sample is mapped to multiple reconstructions by the encoder-mixture-decoder. We spot that for each sample, the bag of reconstructions are diverse. The model is less diverse when it is certain and more diverse when it is uncertain. The reconstructions are also meaningfully diverse, as the model density $q(X)$ has to match the data density $p(X)$ so the bound is tight (Eq.~\eqref{bound_conditional_entropy}).}
\label{reconstructions}
\end{figure}

\begin{figure}[t]
\centering
\begin{subfigure}{.33\textwidth}\includegraphics[width=\linewidth]{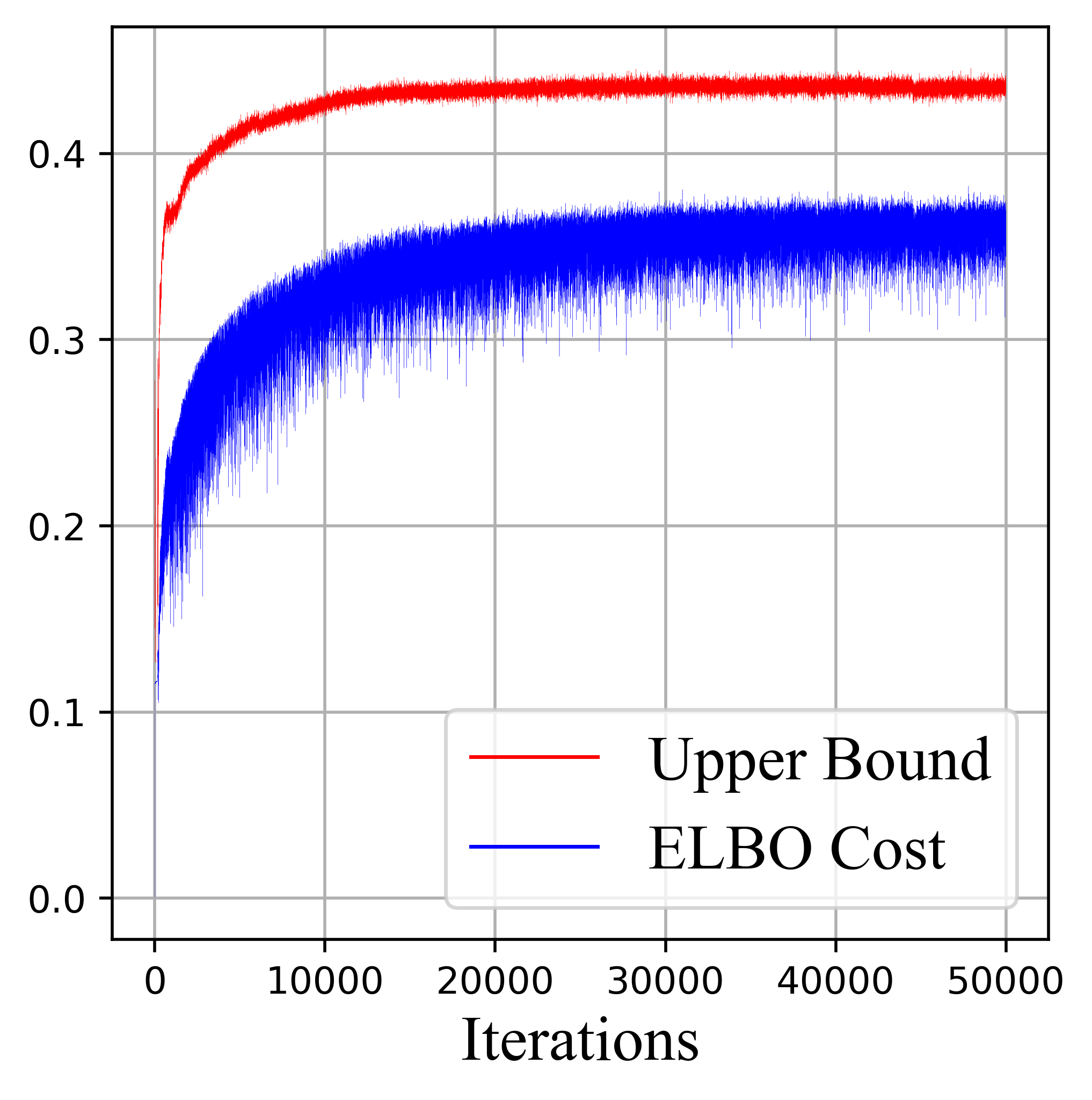}\vspace{-5pt}
\caption{\footnotesize{\textbf{\textit{1C}}: $v = 0.03$}}
\label{13a}
\end{subfigure}\hspace{-5pt}
\begin{subfigure}{.33\textwidth}\includegraphics[width=\linewidth]{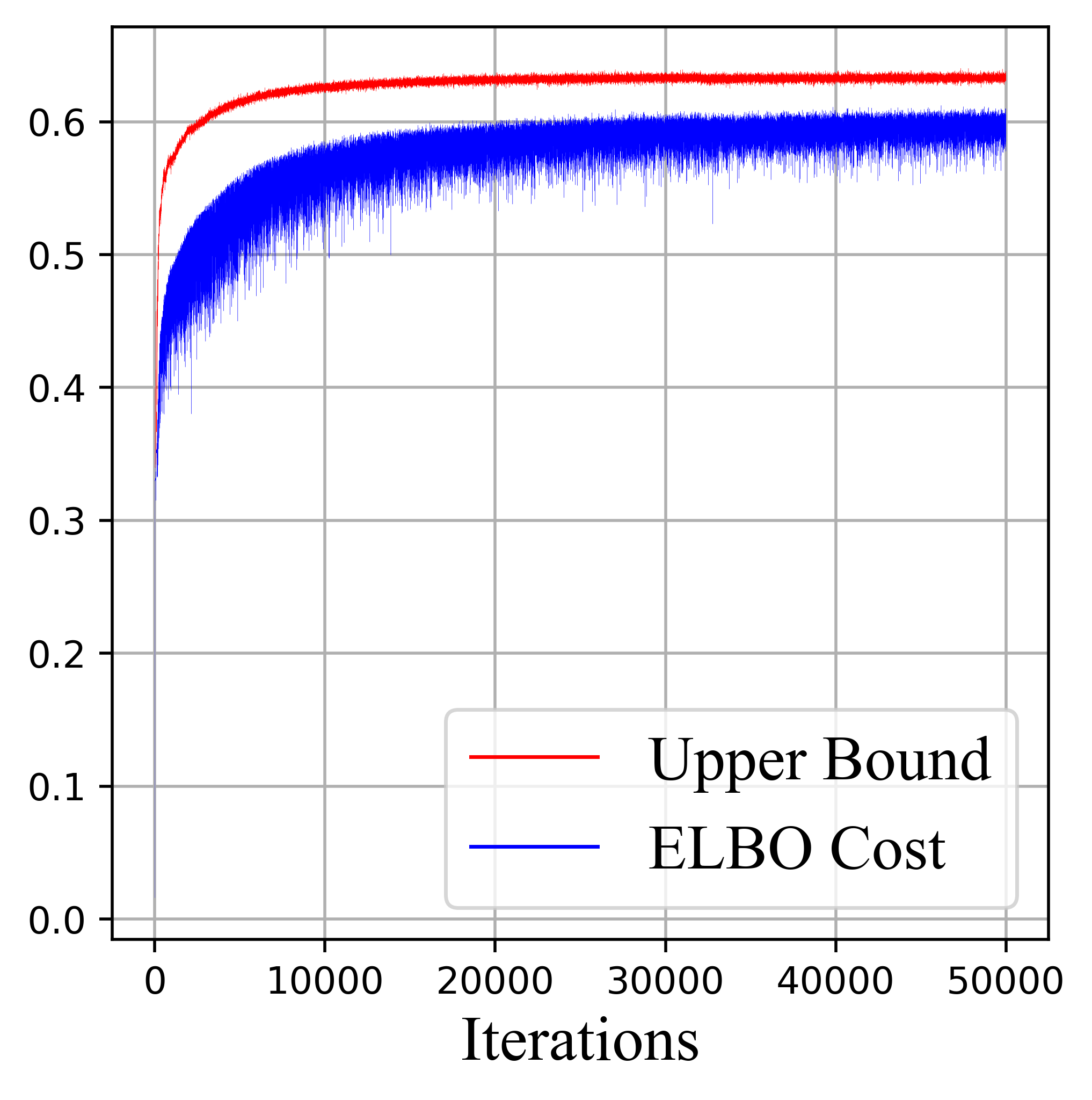}\vspace{-5pt}
\caption{\footnotesize{\textbf{\textit{1C}}: $v = 0.06$}}
\end{subfigure}\hspace{-5pt}
\begin{subfigure}{.33\textwidth}\includegraphics[width=\linewidth]{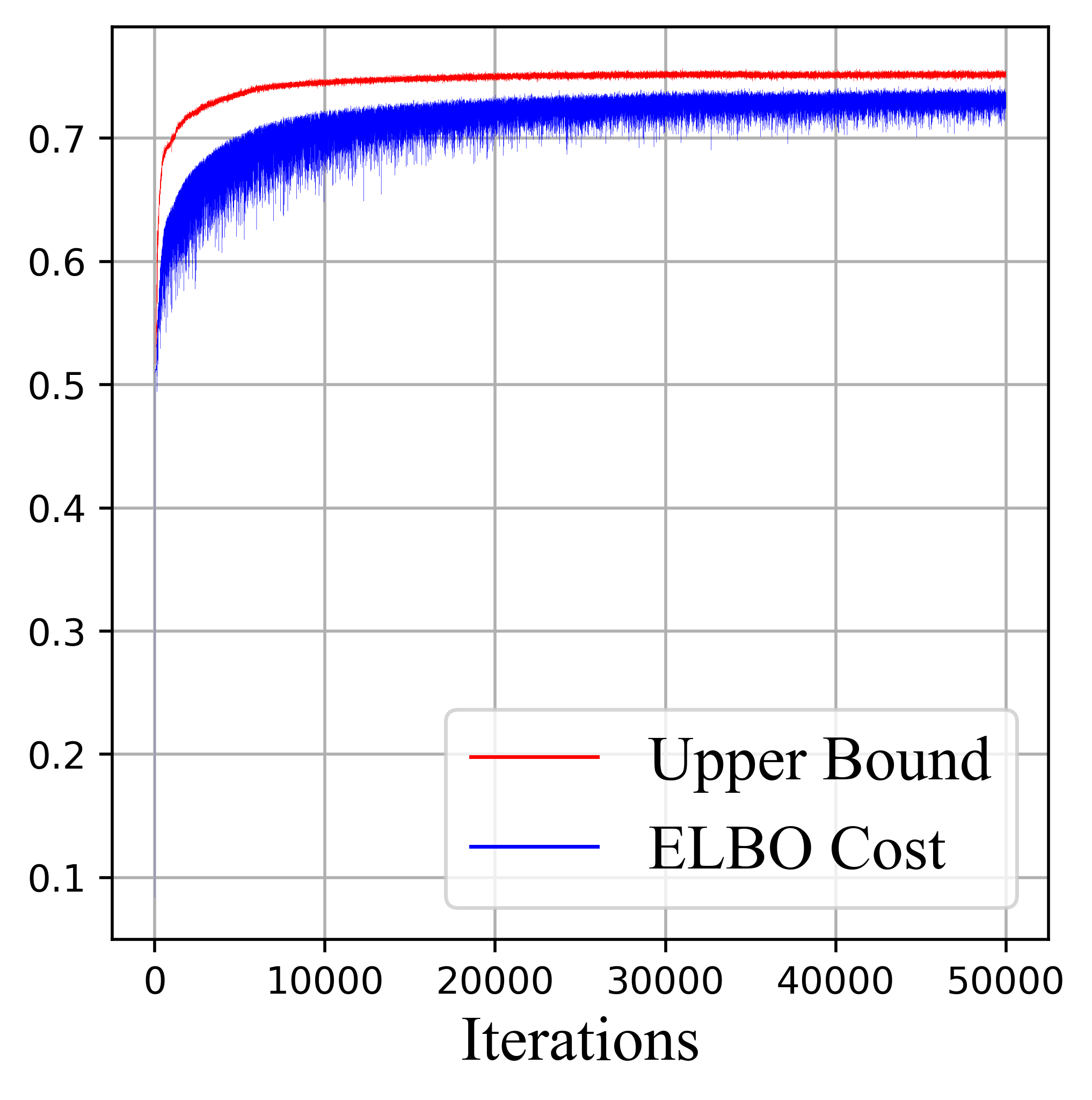}\vspace{-5pt}
\caption{\footnotesize{\textbf{\textit{1C}}: $v = 0.1$}}
\end{subfigure}
\begin{subfigure}{.33\textwidth}\includegraphics[width=\linewidth]{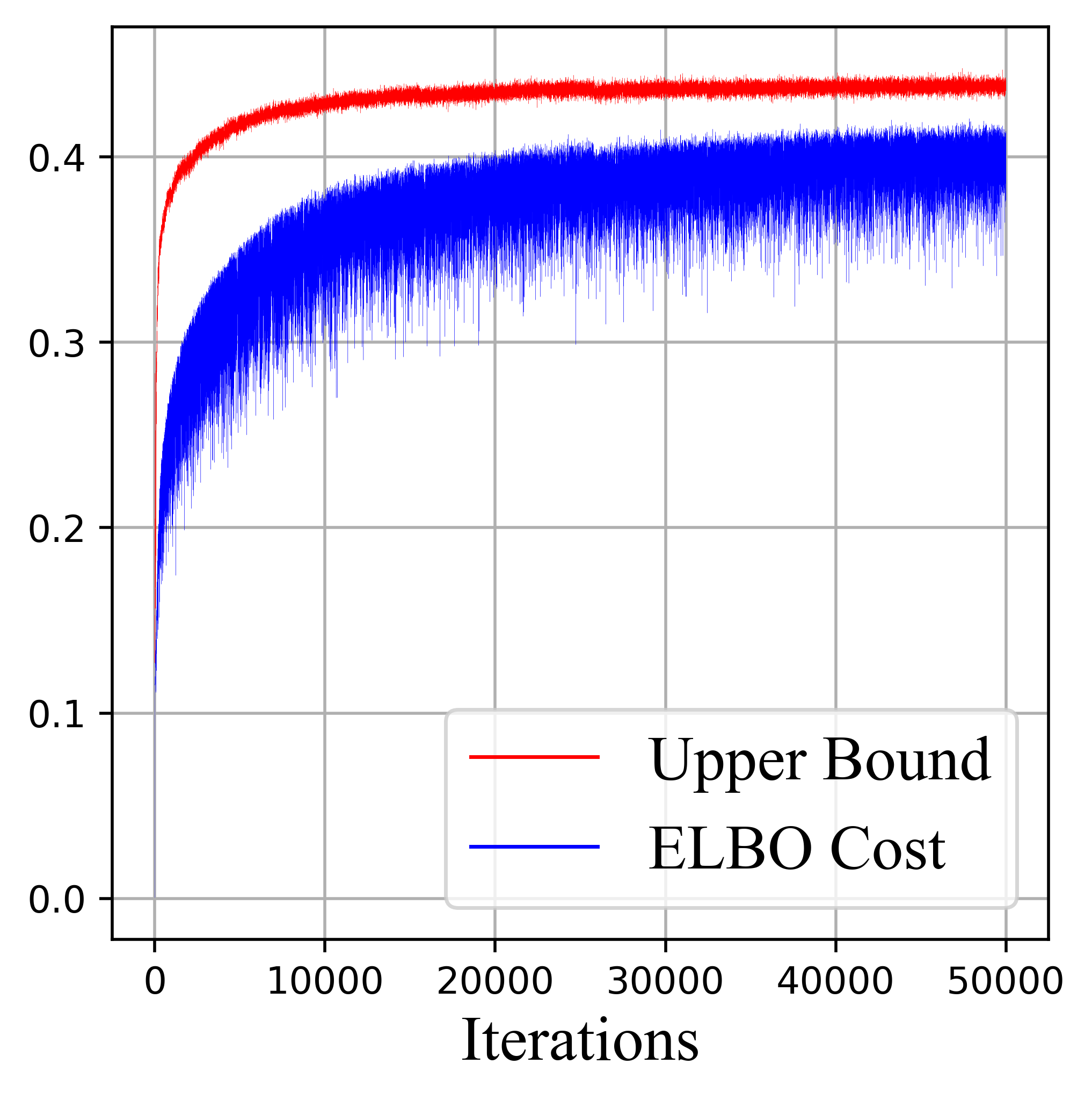}\vspace{-5pt}
\caption{\footnotesize{\textbf{\textit{30C}}: $v = 0.03$}}
\end{subfigure}\hspace{-5pt}
\begin{subfigure}{.33\textwidth}\includegraphics[width=\linewidth]{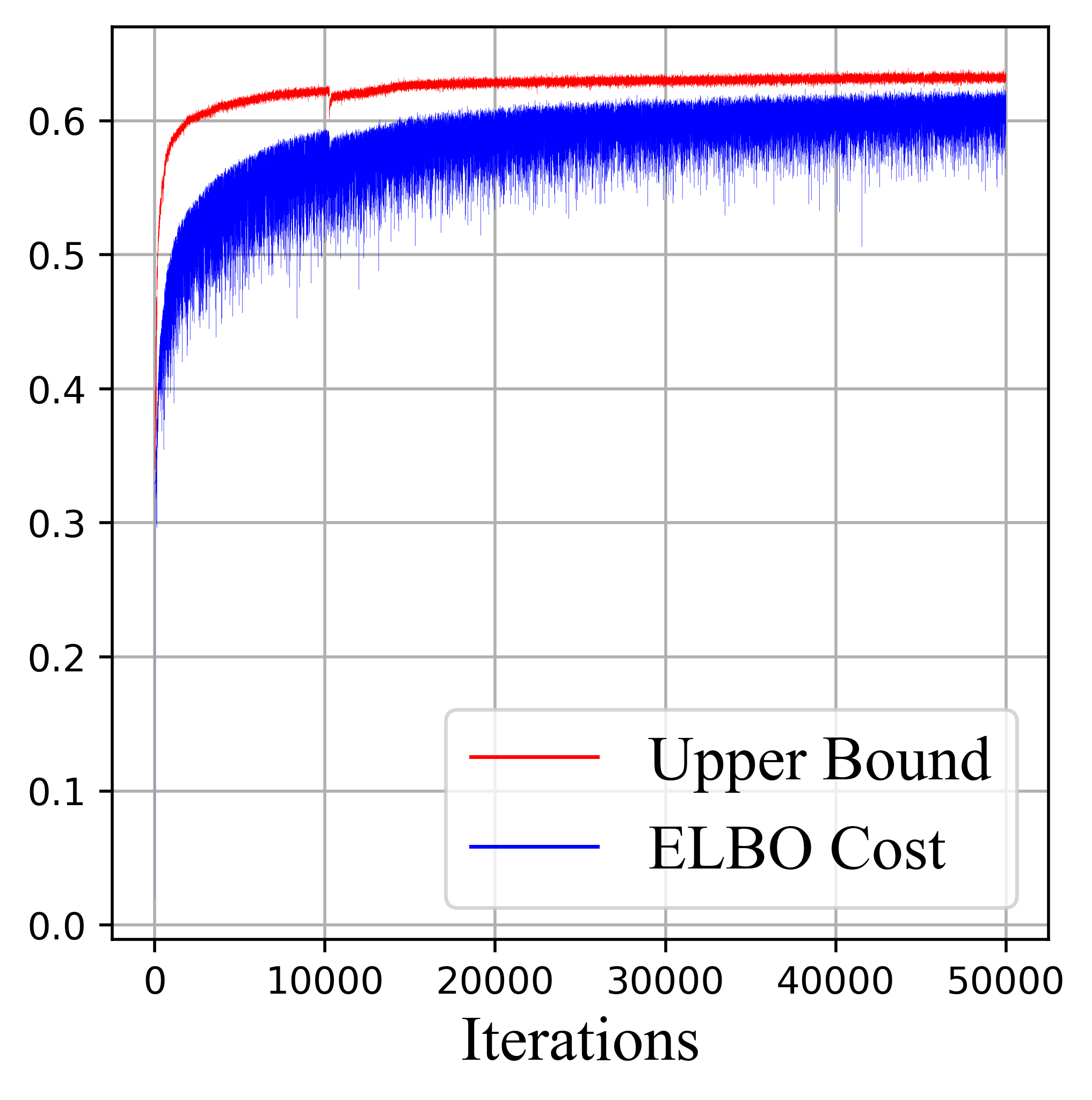}\vspace{-5pt}
\caption{\footnotesize{\textbf{\textit{30C}}: $v = 0.06$}}
\end{subfigure}\hspace{-5pt}
\begin{subfigure}{.33\textwidth}\includegraphics[width=\linewidth]{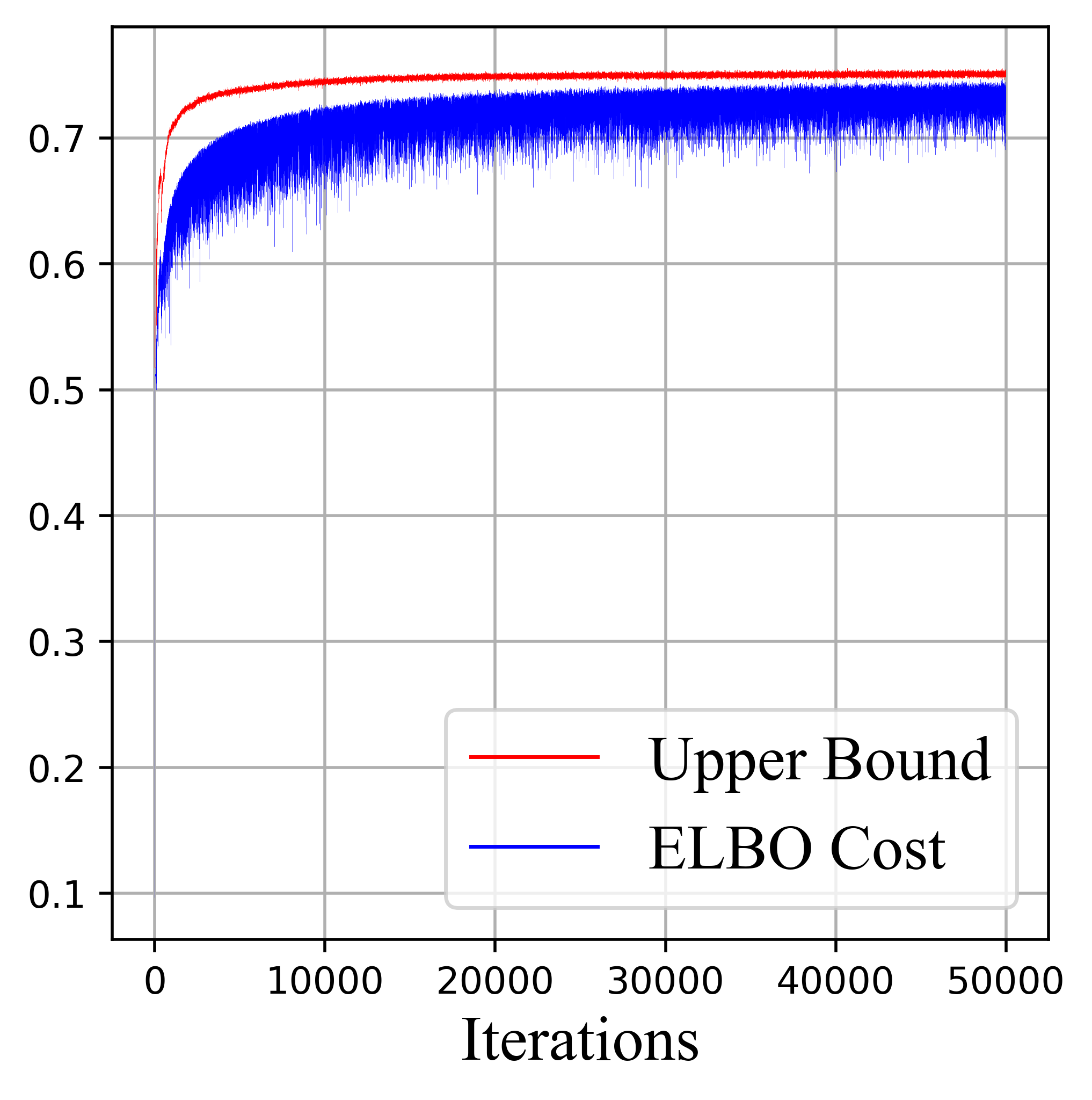}\vspace{-5pt}
\caption{\footnotesize{\textbf{\textit{30C}}: $v = 0.1$}}
\label{13f}
\end{subfigure}
\centering
\begin{subfigure}{.4\textwidth}\includegraphics[width=\linewidth]{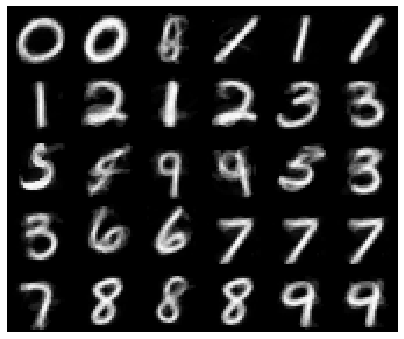}\vspace{-5pt}
\caption{\textbf{\textit{AE}}}
\label{131}
\end{subfigure}\hspace{5pt}
\begin{subfigure}{.4\textwidth}\includegraphics[width=\linewidth]{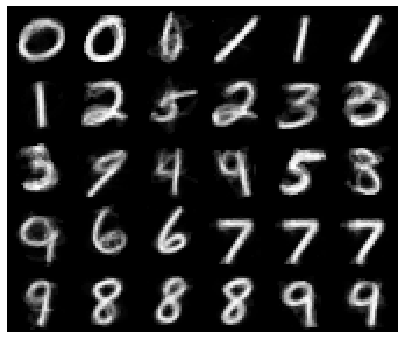}\vspace{-5pt}
\caption{\textbf{\textit{1C}}}
\label{132}
\end{subfigure}
\begin{subfigure}{.4\textwidth}\includegraphics[width=\linewidth]{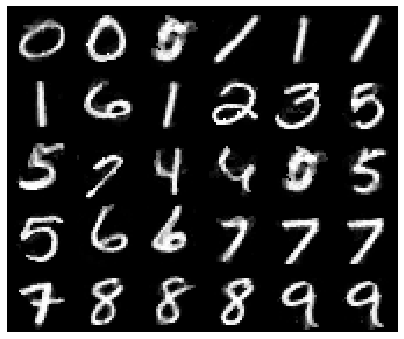}\vspace{-5pt}
\caption{\textbf{\textit{5C}}}
\label{133}
\end{subfigure}\hspace{5pt}
\begin{subfigure}{.4\textwidth}\includegraphics[width=\linewidth]{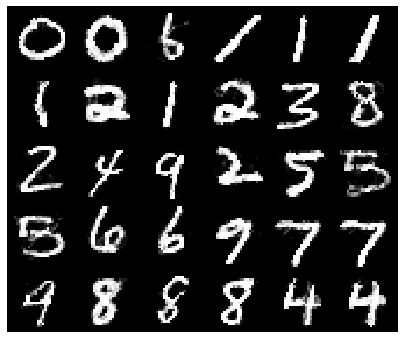}\vspace{-5pt}
\caption{\textbf{\textit{30C}}}
\label{134}
\end{subfigure}\vspace{-9pt}
\caption{Quantitative analysis for the encoder-mixture-decoder on MNIST. We investigate two impacting factors, the variances and the number of centers. (a) to (f) show that the bound is tight almost across all experiments. But it becomes harder to make the bound tight for the case of small variances. The smaller variance will improve the quality and the diversity of reconstruction samples. We found that in this example, using more centers indeed improve the tightness of the bound, and drastically improve the reconstruction quality, although the mapping relation may no longer be one-to-one.\vspace{-9pt}}
\label{MNIST_NUMERICAL}
\end{figure}

\begin{figure}[t]
\centering
\begin{subfigure}{.33\textwidth}\includegraphics[width=\linewidth]{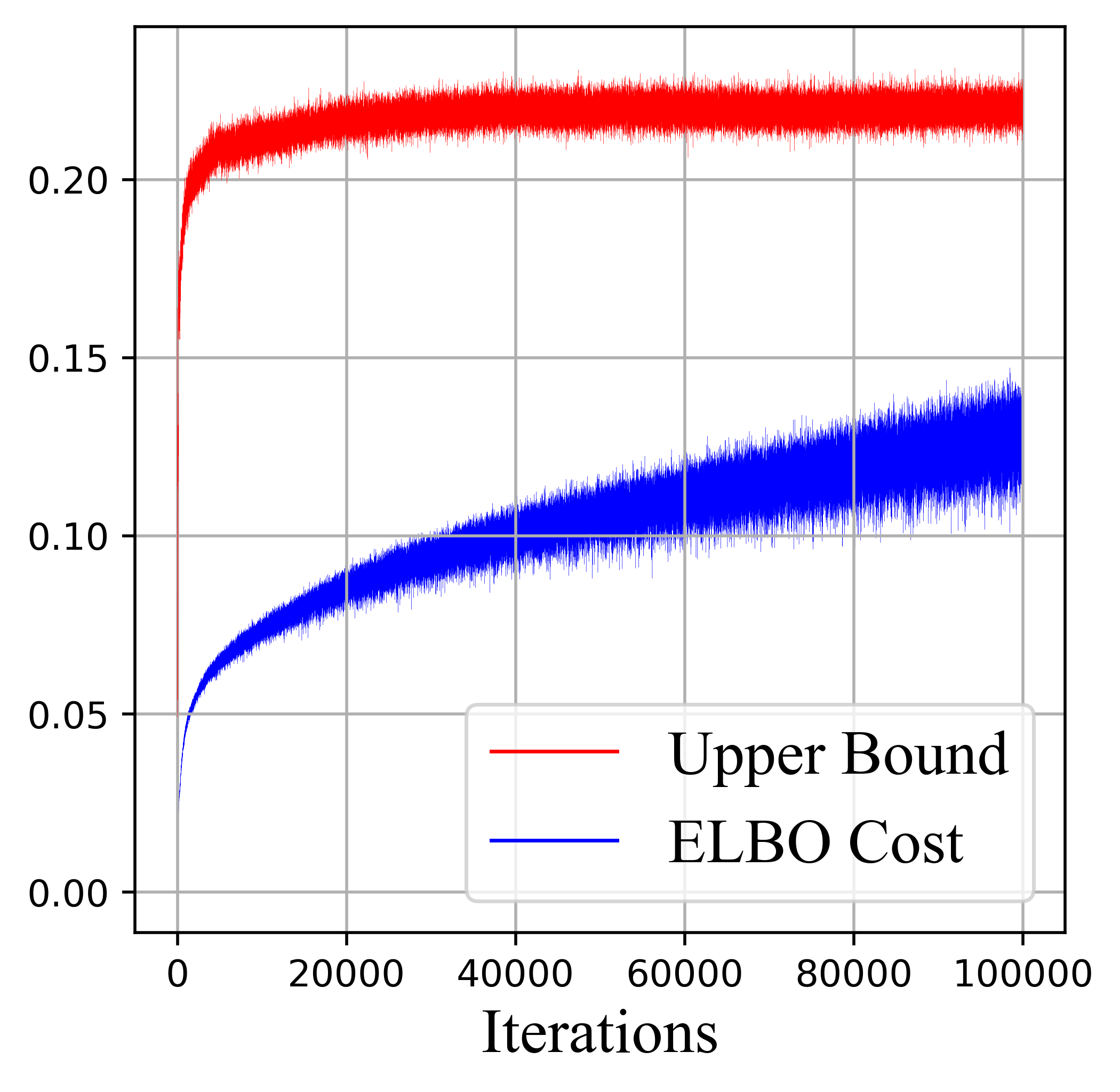}\vspace{-5pt}
\caption{\footnotesize{\textbf{\textit{1C}}: $v = 0.01$}}
\label{14a}
\end{subfigure}\hspace{-5pt}
\begin{subfigure}{.33\textwidth}\includegraphics[width=\linewidth]{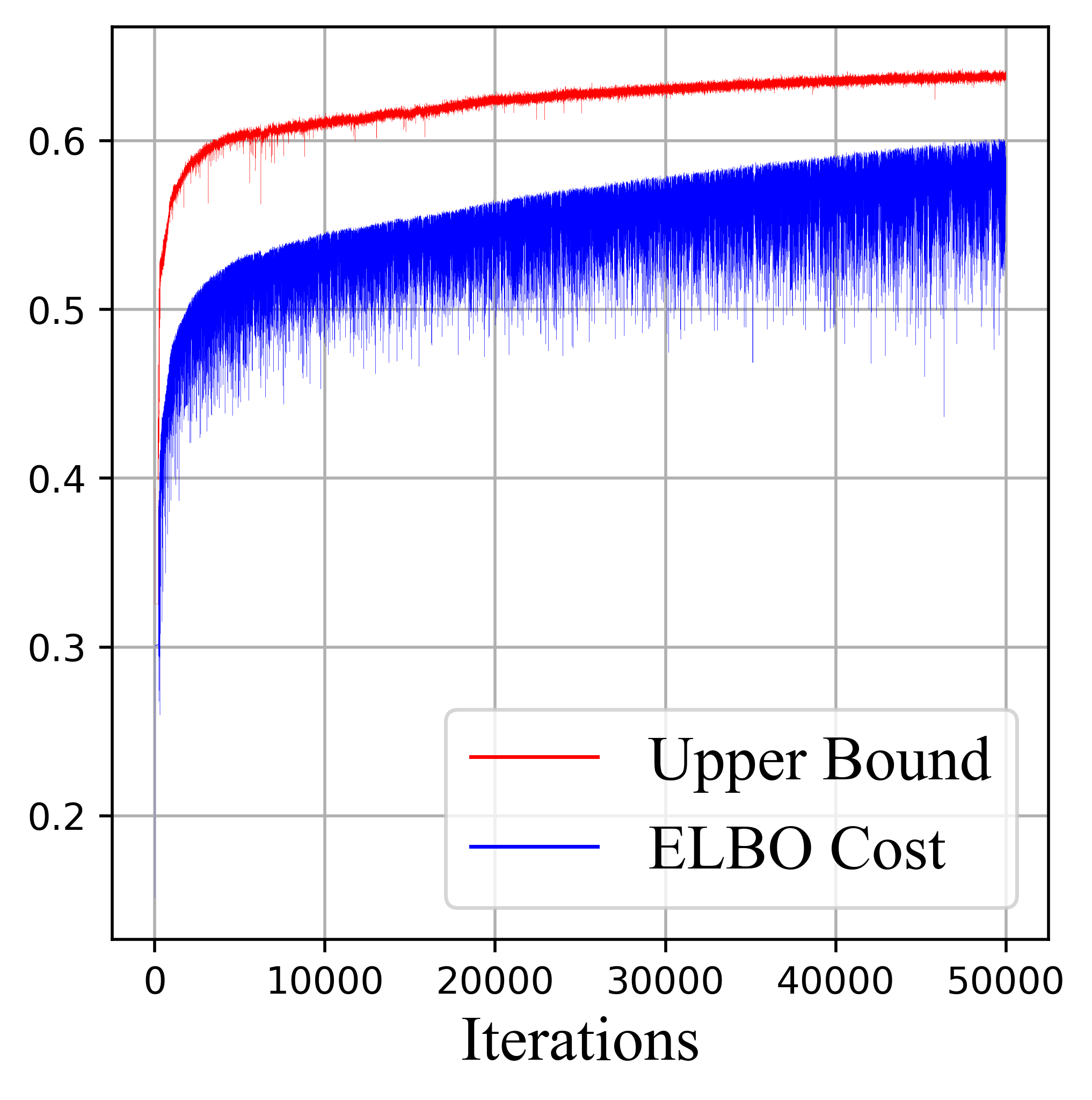}\vspace{-5pt}
\caption{\footnotesize{\textbf{\textit{1C}}: $v = 0.05$}}
\end{subfigure}\hspace{-5pt}
\begin{subfigure}{.33\textwidth}\includegraphics[width=\linewidth]{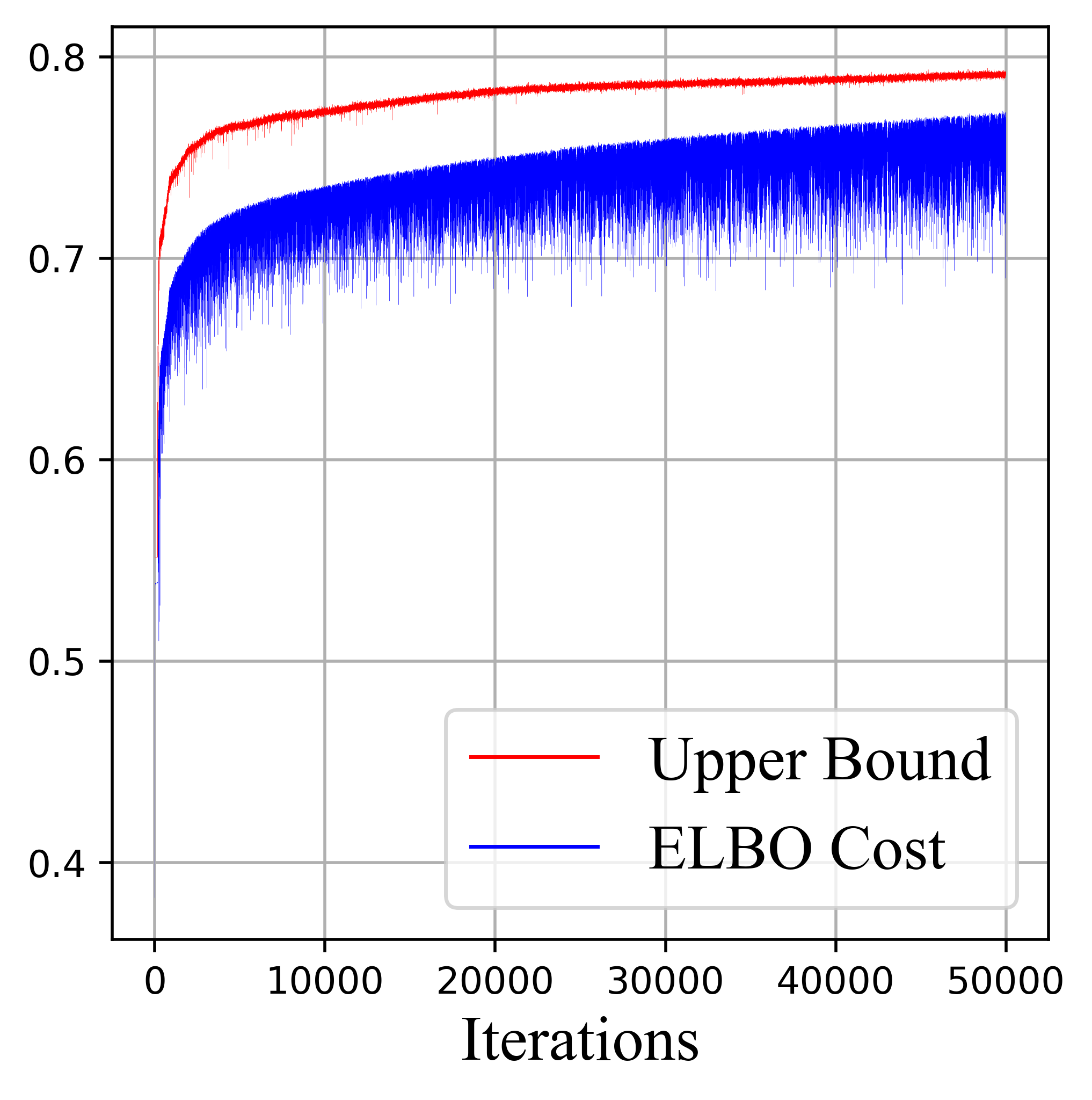}\vspace{-5pt}
\caption{\footnotesize{\textbf{\textit{1C}}: $v = 0.1$}}
\end{subfigure}
\begin{subfigure}{.33\textwidth}\includegraphics[width=\linewidth]{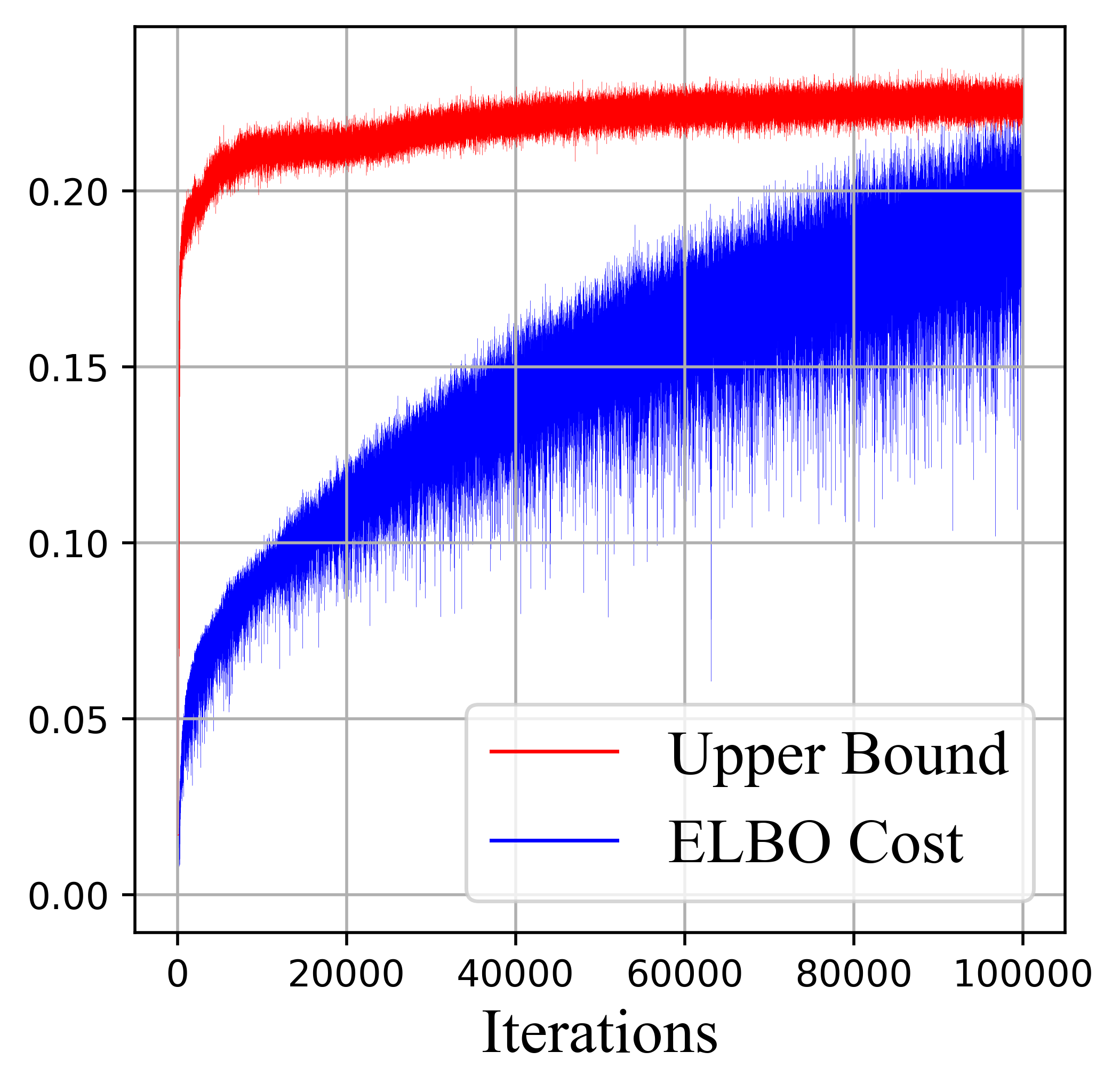}\vspace{-5pt}
\caption{\footnotesize{\textbf{\textit{30C}}: $v = 0.01$}}
\end{subfigure}\hspace{-5pt}
\begin{subfigure}{.33\textwidth}\includegraphics[width=\linewidth]{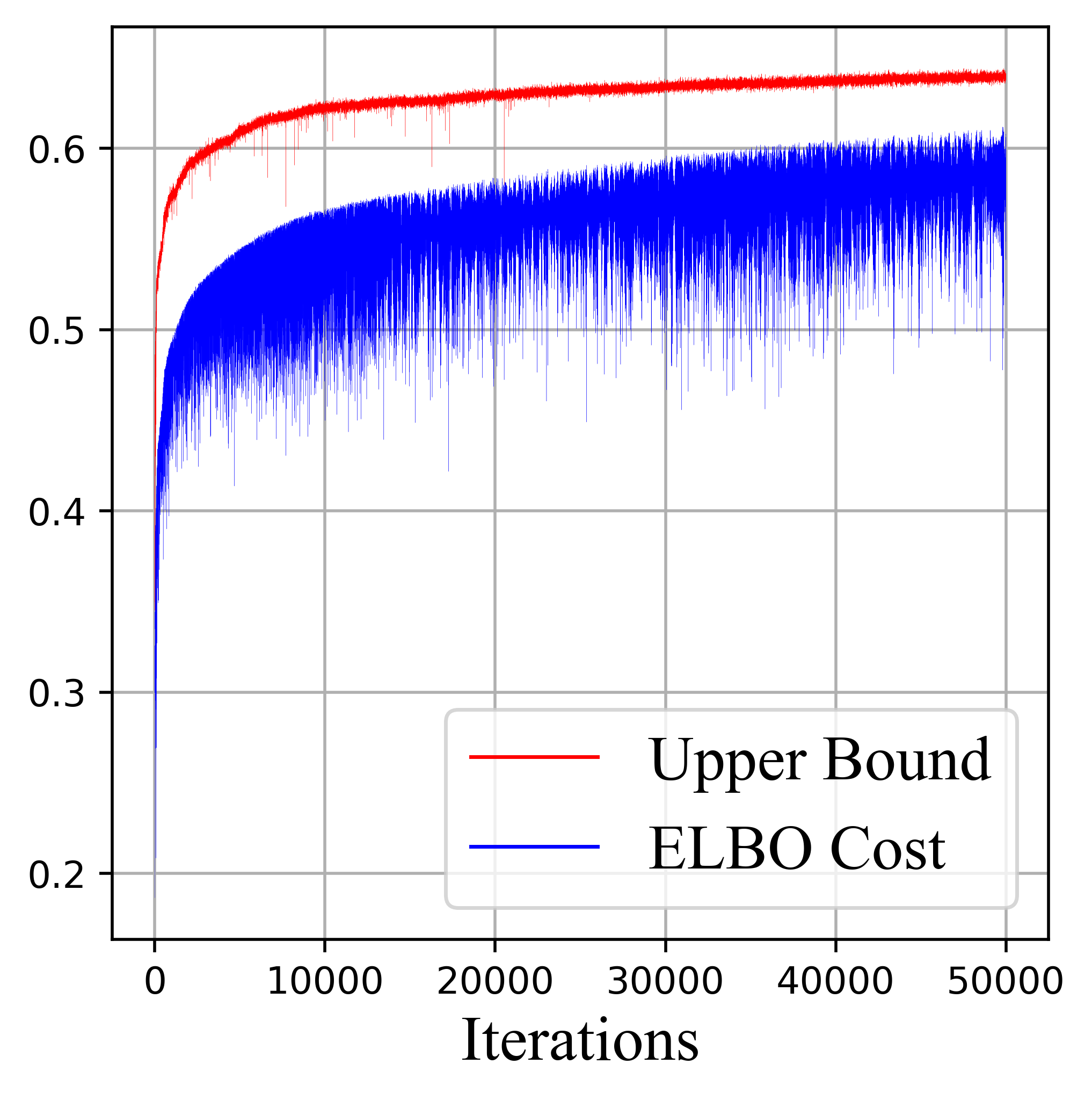}\vspace{-5pt}
\caption{\footnotesize{\textbf{\textit{30C}}: $v = 0.05$}}
\end{subfigure}\hspace{-5pt}
\begin{subfigure}{.33\textwidth}\includegraphics[width=\linewidth]{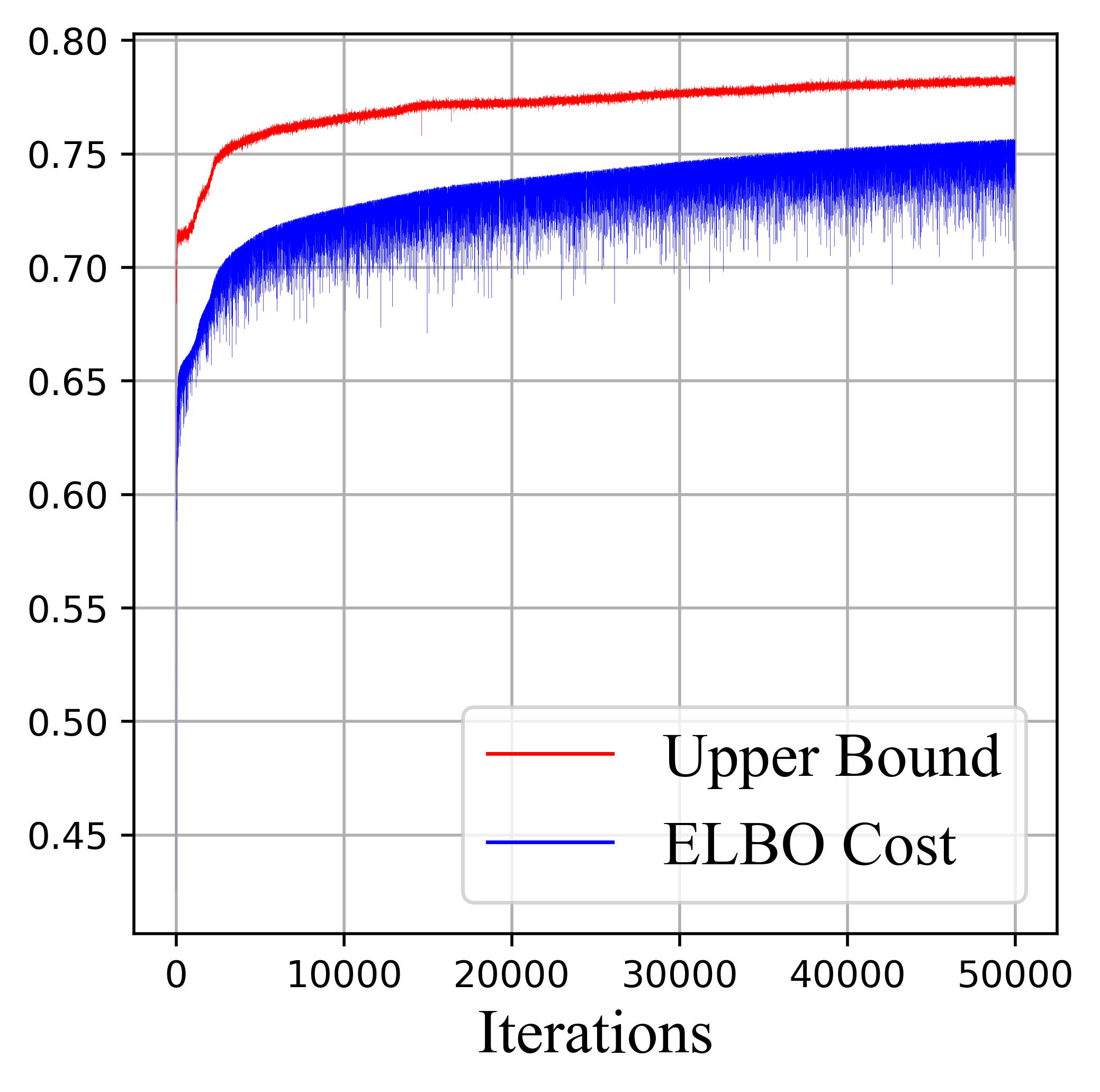}\vspace{-5pt}
\caption{\footnotesize{\textbf{\textit{30C}}: $v = 0.1$}}
\label{14f}
\end{subfigure}
\centering
\begin{subfigure}{.42\textwidth}\includegraphics[width=\linewidth]{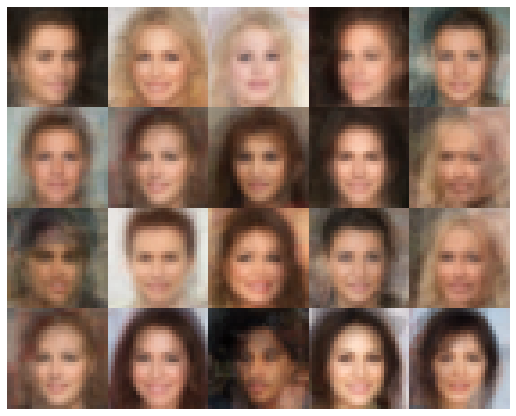}\vspace{-5pt}
\caption{\textbf{\textit{AE}}}
\label{141}
\end{subfigure}\hspace{3pt}
\begin{subfigure}{.42\textwidth}\includegraphics[width=\linewidth]{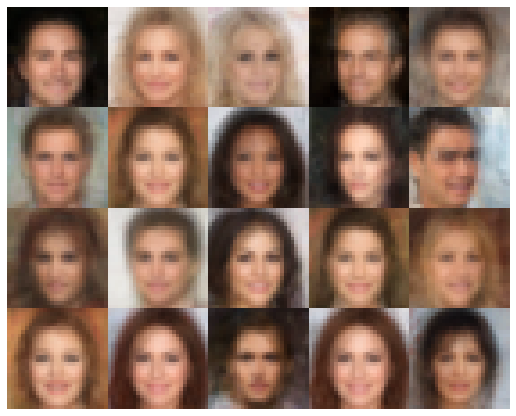}\vspace{-5pt}
\caption{\textbf{\textit{1C}}}
\label{142}
\end{subfigure}
\begin{subfigure}{.42\textwidth}\includegraphics[width=\linewidth]{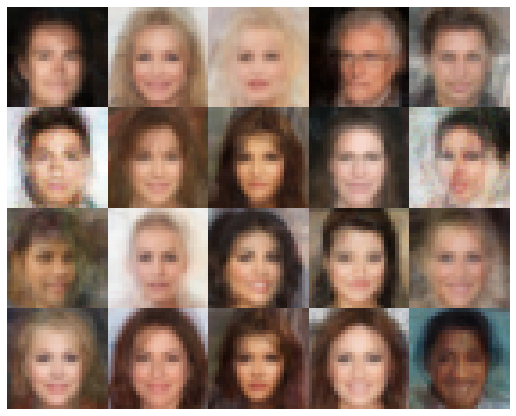}\vspace{-5pt}
\caption{\textbf{\textit{5C}}}
\label{143}
\end{subfigure}\hspace{3pt}
\begin{subfigure}{.42\textwidth}\includegraphics[width=\linewidth]{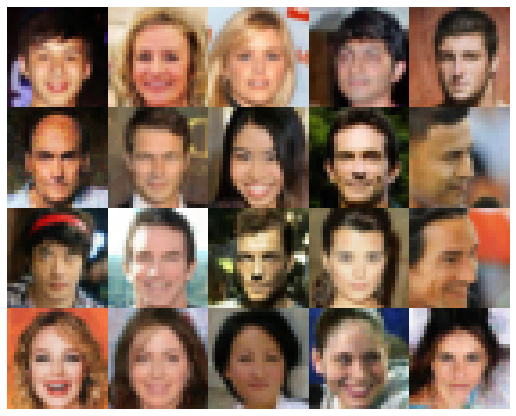}\vspace{-5pt}
\caption{\textbf{\textit{30C}}}
\label{144}
\end{subfigure}\vspace{-9pt}
\caption{Quantitative analysis for the encoder-mixture-decoder on CelebA. The results and conclusions are very similar to results on MNIST shown in Fig.~\ref{MNIST_NUMERICAL}.}
\label{CELEBA_NUMERICAL}
\end{figure}

Since the feature is only in $1D$ so the dimensionality is insufficient, and the reconstruction will not be exact and one-to-one, but one-to-many. As expected, one image sample is mapped to multiple reconstructions, shown in Fig.~\ref{reconstructions}. In our experiments, each sample is mapped to multiple reconstructions due to the use of noise $\mathbf{c}$, and we visualize $25$ reconstructions from them for a handful of samples in Fig.~\ref{reconstructions}. Indeed we spot that one sample is mapped to  diverse reconstructions by the mixture decoder. We found that when the model is more certain about a sample, the reconstructions are more close to the original sample. When the model is more uncertain about a sample, the bag of reconstructions are more diverse. An observation is that the diversity of reconstructions has semantic meanings that they do not represent just noises but also meaningful sematic features. This is because $p(X)$ and $q(X)$ has to match according to the assumption, thus even though the reconstructions are not one-to-one, their distributions still need to match the data distributions, thus they have to be semantically meaningful to make the bound tight. This explains why the reconstructions are diverse yet still still meaningful. A parallel comparison can also be made to Fig.~\ref{9c} and~\ref{9d} for toy examples. 


After comparing the reconstruction quality and demonstrating the one-to-many property of the encoder-mixture-decoder reconstructions, we move to the numerical analysis, shown in Fig.~\ref{MNIST_NUMERICAL} for MNIST and Fig.~\ref{CELEBA_NUMERICAL} for CelebA. In this example, we fix the dataset to be $800$ samples of each dataset, instead of using the full dataset as before, and we set the batch size to be $800$ and use the full subset for training at each training iteration. The reason for this is that for this $800$ samples, if we assume that the samples and features have a small Gaussian noise, the upper bound of this subset the norm $||p(X|Y)||_{p(Y)}^2$ is a fixed value that can be estimated and tracked during training. And we will see that this norm $||p(X|Y)||_{p(Y)}^2$ will always be above the cost $r_c(p, q) = \frac{\langle p(X|Y), q(X|Y)\rangle_{p(Y)}^2}{||q(X|Y)||_{p(Y)}^2}$. And we have also demonstrated the two optimal conditions that first the bound has to be tight, and second the upper bound the norm $||p(X|Y)||_{p(Y)}^2$ has to be lifted to find $Y$ for the maximal value of the upper bound. We investigate the impact of two factors, the chosen data and model variances $v$ that impact the tightness of the bound, and the number of centers for the mixture decoder that impact the reconstruction quality. 

Obtaining the actual value of the upper bound also requires detailed derivations, which we illustrated in Sec.~\ref{quantitative_analysis_for_images}. Note that as demonstrated in Sec.~\ref{SecVIB}, when we calculate the Gaussian functions, we need to ignore the constant before the exponential in the Gaussian, as in high dimensions that constant can be a small value. Also notice that we take the mean of the squared terms inside the exponential over the dimension of pixels, instead of the sum, to ensure that taking the exponential is numerically stable. These two modifications are essential for the derivations and tracking the upper bound values, the details of which can be found in Sec.~\ref{quantitative_analysis_for_images}. 

We first vary the data and model variances. The training curves of the costs ($r_c(p, q) = \frac{\langle p(X|Y), q(X|Y)\rangle_{p(Y)}^2}{||q(X|Y)||_{p(Y)}^2}$, inner products normalized by the norm of $q$) and the upper bound ($||p(X|Y)||_{p(Y)}^2$, norm of $p$) are shown in Fig.~\ref{13a} to~\ref{13f} for MNIST and Fig.~\ref{14a} to~\ref{14f} for CelebA. We found that:\vspace{3pt}
\begin{enumerate}[leftmargin=*]
\item The smaller the variance, the higher-quality the reconstructions will be. But it also becomes more difficult for the bound to be tight. The value of the cost will also decrease. The bound is relatively tight across different variances. This verifies the first part of the optimal condition, that maximizing the cost makes the bound tight and $q(X|Y)$ approach $p(X|Y)$;\vspace{3pt}
\item The upper bound is consistently above the cost, and is being maximized during training. This verifies the second part of the optimal condition, that after the bound being tight, further maximizing the cost finds the maximal value of the upper bound, the negative value of the conditional entropy of $Y$ given $X$;\vspace{3pt}
\item We do observe that in this example, using more centers makes the bound tighter and may improve the value of the bound although incrementally;\vspace{3pt}
\end{enumerate}

We then compare the quality of the reconstructions by varying the number of centers for the mixture decoder. We fix the variance to be the smallest variance such that the model is trainable, which is $0.03$ for MNIST and $0.01$ for CelebA. The results of a regular autoencoder are shown in Fig.~\ref{131} and Fig.~\ref{141}. We compare them with the results produced by the mixture decoder, shown in Fig.~\ref{132} to~\ref{134} and Fig.~\ref{142} to~\ref{144}. We indeed found in this example of using a subset of $800$ training samples, the reconstruction quality improve drastically when we vary the number of centers from $1$ to $30$. When we set the number of centers to be $30$, the reconstructed samples look exactly like the original samples, even though the feature dimension is only $1$. Notice that even though the generation quality matches the original samples, the mapping relation may not be one-to-one, that one sample is mapped to multiple reconstructions by the mixture decoder, and we only visualize one of the centers in the figures. 

\newpage

\section{Quantitative Analysis}
\label{quantitative_analysiss}

Having shown the reconstruction quality of the encoder-mixture-decoder, we further present the details and results of the quantitative numerical analysis. 

The focus is to derive the actual value of the upper bound $||p(X|Y)||_{p(Y)}^2$ using training samples, such that its value can be tracked during training to verify its tightness and the optimal value. The derived value can also be maximized to see whether the optimality of maximizing the upper bound is the same as maximizing the cost.

This is achieved by two main assumptions. First we fix the training samples to be a small subset of the training set, for example $1,000$ or $5,000$ samples. At each iteration of the training, we train the model only on this subset. So we only need to derive the empirical value of the upper bound at each iteration, which keeps the computation complexity low. The second assumption is to assume a small Gaussian ball around each sample. This makes the density, particularly the density $p(X,Y)$ tractable, which makes the upper bound $||p(X|Y)||_{p(Y)}^2$ tractable. We discuss how to construct such tractable $p$ that can be estimated empirically through samples. We also show experimentally how this bound becomes tight and maximized, and maximizing this upper bound obtains the same solution and the same optimality as maximizing the cost itself. 

\subsection{Quantitative Analysis for Simple Datasets}
\label{quantitative_analysis_simple}

For simple low-dimensional toy datasets, it is possible to derive the exact values of the upper bound $||p(X|Y)||_{p(Y)}^2$ by making the assumption that $p(X,Y)$ is a mixture of Gaussian. 

Given a series of data samples $X_1, X_2,\cdots,X_N$, apply an encoder to obtain features $Y_1,Y_2,\cdots,Y_N$ for each sample, the assumption is to assume that the joint $p(X,Y)$ is a Gaussian mixture on $X_1, X_2,\cdots,X_N$ and $Y_1,Y_2,\cdots,Y_N$ as centers, written in Eq.~\eqref{joint_gaussian}. Suppose the variance for $X$ is $v_X$ and the variance for $Y$ is $v_Y$, the assumption is to simply make $p(X,Y)$ the product of two Gaussians $\mathcal{N}(X-X_n;v_X)$ and $\mathcal{N}(Y-Y_n;v_Y)$, and take the average over all samples. Given any $X$ and $Y$, the value of the joint density $p(X,Y)$ can be computed by Eq.~\eqref{joint_gaussian}. We can also marginalize over $X$ to obtain $p(Y)$, which is a mixture with a variance $v_Y$ with $Y_1,Y_2,\cdots, Y_n$ as centers. We still assume that $q(X|Y)$ is a mixture decoder, parameterized by $q(X|Y) = \int p(c)\mathcal{N}(X-\textbf{D}(Y,c);v_q) dc$. We show how under this scenario that the inner product $\langle p(X|Y), q(X|Y)\rangle_{p(Y)}$ and the two norms $||p(X|Y)||_{p(Y)}^2$ and $||q(X|Y)||_{p(Y)}^2$ can be empirically estimated through samples. 
\begin{equation}
\begin{gathered}
p(X,Y) = \frac{1}{N}\sum_{n=1}^N\mathcal{N}(X-X_n;v_X)\mathcal{N}(Y-Y_n;v_Y),\\
p(Y) = \frac{1}{N}\sum_{n=1}^N \mathcal{N}(Y-Y_n;v_Y). 
\end{gathered}
\label{joint_gaussian}
\end{equation}

\noindent \textbf{\textit{1) The inner product $\langle p(X|Y), q(X|Y)\rangle_{p(Y)}$.}} We first check the inner product in the numerator of the cost. Using the assumption of $p(X,Y)$ and $q(X|Y)$ in the inner product gives us Eq.~\eqref{equation_simple_closed}. 

\begin{equation}
\resizebox{1\linewidth}{!}{
$\begin{aligned}
&\iint p(X,Y) q(X|Y) dXdY\\ 
& = \iint \frac{1}{N}\sum_{n=1}^N\mathcal{N}(X-X_n;v_X) \mathcal{N}(Y-Y_n;v_Y) \\ & \;\;\;\;\;\;\;\;\;\;\; \cdot \int p(c)\mathcal{N}(X-\textbf{D}(Y,c);v_q) dc \cdot dXdY\\
& = \frac{1}{N} \sum_{n=1}^N \iint p(c) \mathcal{N}(Y-Y_n;v_Y) \mathcal{N}(X_n - \textbf{D}(Y,c); v_X+v_q) dY dc\\
& \approx \frac{1}{N}\sum_{n=1}^N \int p(c) \mathcal{N}(X_n - \textbf{D}(\widehat{Y_n},c);v_X+v_q) dc \\
& \approx \frac{1}{NK} \sum_{n=1}^N\sum_{k=1}^K\mathcal{N}(X_n-\textbf{D}(\widehat{Y_n}, c_n(k));v_X+v_q).
\end{aligned}$}
\label{equation_simple_closed}
\end{equation}

\noindent We explain Eq.~\eqref{equation_simple_closed} line by line. First notice that there is the $\mathcal{N}(X-X_n;v_X)$ term in $p(X,Y)$. While in $q(X|Y)$ we also have the term $\mathcal{N}(X-\textbf{D}(\widehat{Y_n},c);v_q)$. By taking the integral of the cross product of them over $X$, by the property that the inner product of two Gaussian function has a closed form, it follows that their inner product is $\mathcal{N}(X_n - \textbf{D}(Y,c); v_X+v_q)$. 

Then we need to take the integral over $Y$ and $c$. Taking the integral over $Y$ requires the inner product between $\mathcal{N}(Y-Y_n;v_Y)$ and $\mathcal{N}(X_n - \textbf{D}(Y,c); v_X+v_q)$. However, this integral no longer has a closed form since in $\mathcal{N}(X_n - \textbf{D}(Y,c); v_X+v_q)$ as the $Y$ is inside the decoder function $\textbf{D}$. Instead, we construct $Y$ with an additive Gaussian $\widehat{Y_n} = Y_n + \sqrt{v_Y}\cdot z_n$ where $z_n$ is a series of realizations sampled from the standardized Gaussian. This constructed $\widehat{Y_n}$ will represent $\mathcal{N}(Y-Y_n;v_Y)$. Then the inner product can simply be estimated by $\mathcal{N}(X_n - \textbf{D}(\widehat{Y_n},c))$. To take the integral over $c$, we can sample a series of $c(1), c(2),\cdots,c(K)$ for each $n$, and estimate the integral empirically by averaging $\mathcal{N}(X_n-\textbf{D}(\widehat{Y_n}, c_n(k));v_X+ v_q)$. That is, we use an additive Gaussian $\widehat{Y_n} = Y_n + \sqrt{v_Y}\cdot z_n$ to estimate the integral over $Y$, and a series of priors $c_n(k)$ for each $k$ to estimate the integral over $c$. This gives us the final form in Eq.~\eqref{equation_simple_closed}.

To describe the estimation, we first use $X_n$ as inputs to the encoder to obtain $Y_n$. Each $Y_n$ is added with a Gaussian with a variance $v_Y$, combined with multiple priors $c_n(1), c_n(2),\cdots, c_n(K)$ as inputs to the decoder. Then we average the Gaussian differences $\mathcal{N}(X_n-\textbf{D}(\widehat{Y_n}, c_n(k));v_X+ v_q)$.\vspace{7pt}

\noindent \textbf{2) \textit{The norm $||q(X|Y)||_{p(Y)}^2$.}} Using the assumption of $p(Y)$ (Eq.~\eqref{joint_gaussian}) and $q(X|Y) = \int p(c)\mathcal{N}(X-\textbf{D}(Y,c);v_q) dc$ as a Gaussian mixture, we can write down Eq.~\eqref{eq_norm1}.
\begin{equation}
\resizebox{1\linewidth}{!}{
$\begin{aligned}
&\iint p(Y)q^2(X|Y) dXdY \\
&= \iint \frac{1}{N}\sum_{n=1}^N \mathcal{N}(Y-Y_n;v_Y) \cdot \big( \int p(c)\mathcal{N}(X-\textbf{D}(Y,c);v_q) dc \big)^2 dX dY \\
& \approx \frac{1}{N} \sum_{n=1}^N \big( \int p(c)\mathcal{N}(X-\textbf{D}(\widehat{Y_n},c);v_q) dc \big)^2 \\
& \approx \frac{1}{NK^2} \sum_{n=1}^N  \sum_{i=1}^K \sum_{j=1}^K \mathcal{N}(\widehat{X_n'}(i) - \widehat{X_n'}(j);2v_q). 
\end{aligned}$}
\label{eq_norm1}
\end{equation}
We also explain Eq.~\eqref{eq_norm1} in details. Since $p(Y)$ is defined by $\mathcal{N}(X-\textbf{D}(Y,c);v_q)$ and we need to compute the expectation of the squared of $q$, $q^2(X|Y)$, by integrating over $Y$. This can be achieved also by constructing a $\widehat{Y_n} = Y_n + \sqrt{v_Y}\cdot z_n$, $Y_n$ with an additive noise, and take the average over $\big( \int p(c)\mathcal{N}(X-\textbf{D}(\widehat{Y_n},c);v_q) dc \big)^2$. This squared term can be applied with the closed-form of Gaussian inner product, and estimate it with a series of $c_n(1), c_n(2),\cdots,c_n(K)$ similar to before, which gives us $\frac{1}{K^2}\sum_{k=1}^K \sum_{t=1}^K \mathcal{N}(\textbf{D}(\widehat{Y_n},c_k) - \textbf{D}(\widehat{Y_n},c_t);2v_q)$. We also simplify the notation here, for each $n$ and $k$, we denote the output of the decoder as $\widehat{X_n'}(k):=\textbf{D}(\widehat{Y_n},c_k)$, representing a reconstruction with a sample $\widehat{Y_n}$ and a noise $c_k$. Then the norm becomes the average of the Gaussian differences $\mathcal{N}(\widehat{X_n'}(i) - \widehat{X_n'}(j);2v_q)$. 

Describing the process in details, for each $X_n$, we construct a series of $\widehat{Y_n}$ that is $Y_n$ with an additive Gaussian noise. Combining $\widehat{Y_n}$ with sampled prior noise $c_n(k)$, we obtain reconstructions $\widehat{X_n'}(k)$. The norm $||q(X|Y)||_{p(Y)}^2$ is the average of Gaussian differences between two reconstructions $\widehat{X_n'}(i)$ and $\widehat{X_n'}(j)$ of the same sample $X_n$. \vspace{7pt}

\noindent \textbf{\textit{3) The norm $||p(X|Y)||_{p(Y)}^2$.}} The derivation of the norm of $p$ requires extra steps, shown in Eq.~\eqref{eq_33_norm_2}. 
\begin{equation}
\resizebox{1\linewidth}{!}{
$\begin{aligned}
&\iint p^2(X|Y) p(Y) dXdY \\ & = \iint \frac{p(X,Y)}{p(Y)}\cdot p(X,Y) dXdY \\
& = \iint \frac{\frac{1}{N}\sum_{n=1}^N\mathcal{N}(X-X_n;v_X)\mathcal{N}(Y-Y_n;v_Y)}{\frac{1}{N}\sum_{n=1}^N \mathcal{N}(Y-Y_n;v_Y)} p(X,Y) dXdY \\ 
& = \frac{1}{N} \sum_{m=1}^N \frac{\sum_{n=1}^N\mathcal{N}(\widehat{X_m}-X_n;v_X)\mathcal{N}(\widehat{Y_m}-Y_n;v_Y)}{\sum_{n=1}^N \mathcal{N}(\widehat{Y_m}-Y_n;v_Y)}.
\end{aligned}$}
\label{eq_33_norm_2}
\end{equation}
The explanation is as follows. We first write the conditional $p(X|Y)$ as $p(X|Y) = \frac{p(X,Y)}{p(Y)}$, because $p(X,Y)$ and $p(Y)$ are assumed to be Gaussian mixtures following Eq.~\eqref{joint_gaussian}. Then Eq.~\eqref{eq_33_norm_2} is an integral $\iint \frac{p^2(X,Y)}{p(Y)} dXdY$, and because of the $p(Y)$ in the denominator, this integral does not have a closed-form solution. Instead, we factorize one $p(X,Y)$ out of the square, and write it as the expectation of $\frac{p(X,Y)}{p(Y)}$, which can be written as the average of Gaussians by the assumption in Eq.~\eqref{joint_gaussian}. Then we use the same trick of constructing $\widehat{Y_n} = Y_n + \sqrt{v_Y}\cdot z_n$ and $\widehat{X_n} = X_n + \sqrt{v_X}\cdot s_n$ where $z_1,z_2,\cdots,z_N$ and $s_1,s_2,\cdots,s_N$ are two series of standardized Gaussians, representing the joint $p(X,Y)$. Then we can estimate the term empirically with $\widehat{X_n}$ and $\widehat{Y_n}$, shown in Eq.~\eqref{eq_33_norm_2}

We summarize the estimations of the here. Given $N$ samples $X_1,X_2,\cdots,X_N$, each sample $X_n$ corresponds to the output of an encoder $Y_n$. We construct $\widehat{X_n}$ and $\widehat{Y_n}$ which are $X_n$ and $Y_n$ with additive Gaussian noises, then sample a series of prior noises $c_n(k)$, combined with $\widehat{Y_n}$, to obtain outputs of the mixture decoder $\widehat{X_n'}(k):= \textbf{D}(\widehat{Y_n},c_k)$. The three terms to estimate the cost and the upper bound are as follows:
\begin{equation}
\resizebox{1\linewidth}{!}{
$\begin{aligned}
\langle p(X|Y), q(X|Y)\rangle_{p(Y)} &= \frac{1}{NK} \sum_{n=1}^N\sum_{k=1}^K\mathcal{N}(X_n-\widehat{X_n'}(k);v_X+v_q),\\
||q(X|Y)||_{p(Y)}^2 &= \frac{1}{NK^2} \sum_{n=1}^N  \sum_{i=1}^K \sum_{j=1}^K \mathcal{N}(\widehat{X_n'}(i) - \widehat{X_n'}(j);2v_q),\\
||p(X|Y)||_{p(Y)}^2 &= \frac{1}{N}\sum_{m=1}^N \frac{\sum_{n=1}^N\mathcal{N}(\widehat{X_m}-X_n;v_X)\mathcal{N}(\widehat{Y_m}-Y_n;v_Y)}{\sum_{n=1}^N \mathcal{N}(\widehat{Y_m}-Y_n;v_Y)}.
\end{aligned}$}
\label{equation_34}
\end{equation}
With these derivations, we are able to estimate the terms in the bound $\frac{\langle p(X|Y), q(X|Y)\rangle_{p(Y)}^2}{||q(X|Y)||_{p(Y)}^2} \leq ||p(X|Y)||_{p(Y)}^2$ from samples and obtain their exact values. This can be used to verify the theory, that an autoencoder or an encoder-mixture-decoder is enforcing $q(X|Y)$ to approximate $p(X|Y)$. This can also be used to investigate the impact of the number of centers, the claim that by changing $q(X|Y)$ from a simple Gaussian to a mixture and increasing the centers, it can increase the representation capability of $q(X|Y)$, and make it closer to $p(X|Y)$, such that the bound is tighter and we can find a better optimality. Thirdly, we can also directly maximize $||p(X|Y)||_{p(Y)}^2$ as it can be empirically estimated from samples independent of $q$, and compare the solution of $p(X|Y)$ with the solution from training an encoder-decoder with both $p$ and $q$. Note that the estimation form for $||p(X|Y)||_{p(Y)}^2$ may only apply to small datasets. 

It is also possible to compare the solution with finding the $p(X|Y)$ that maximizes the mutual information between $X$ and $Y$, with Shannon's mutual information or R\'enyi's mutual information bounds. Suppose $p(X, Y)$ follows the assumption, then Shannon's mutual information and R\'enyi's mutual information can directly be estimated by
\begin{equation}
\resizebox{1\linewidth}{!}{
$\begin{aligned}
& \iint p(X,Y) \log \frac{p(X,Y)}{p(X)p(Y)} dXdY \\
& = \frac{1}{N}\log  \left( N\cdot \sum_{m=1}^N \frac{\sum_{n=1}^N\mathcal{N}(\widehat{X_m}-X_n;v_X)\mathcal{N}(\widehat{Y_m}-Y_n;v_Y)}{\sum_{n=1}^N \mathcal{N}(\widehat{X_m}-X_n;v_X) \sum_{n=1}^N \mathcal{N}(\widehat{Y_m}-Y_n;v_Y)} \right), \\ \vspace{-35pt} \\
&\iint \frac{p^2(X,Y)}{p(X)p(Y)} dXdY \\
&  = N\cdot \sum_{m=1}^N \frac{\sum_{n=1}^N\mathcal{N}(\widehat{X_m}-X_n;v_X)\mathcal{N}(\widehat{Y_m}-Y_n;v_Y)}{\sum_{n=1}^N \mathcal{N}(\widehat{X_m}-X_n;v_X) \sum_{n=1}^N \mathcal{N}(\widehat{Y_m}-Y_n;v_Y)}. 
\end{aligned}$}
\label{EQ_35}
\end{equation}
The term $\iint p(X,Y) \log \frac{p(X,Y)}{p(X)p(Y)} dXdY$ is an estimation for Shannon's mutual information. The term $\iint \frac{p^2(X,Y)}{p(X)p(Y)} dXdY$ is an estimation for R\'enyi's mutual information. R\'enyi's mutual information in $L_2$ (the integral over $\frac{p^2(X,Y)}{p(X)p(Y)}$) and the conditional entropy (the integral over $\frac{p^2(X,Y)}{p(Y)}$) differ from one $p(X)$ term in the denominator. The reason why optimizing the encoder-mixture-decoder estimates and maximizes the conditional entropy, not R\'enyi's mutual information is because the decoder $q(X|Y)$ is a conditional, not a joint density. We will show that minimizing teh conditional entropy in $L_2$ and maximizing R\'enyi's mutual information, which differ in a $p(X)$ term will reach different optimal solutions (Fig.~\ref{solution_comparison_17}).

For toy distributions with sufficient number of samples, estimating these two terms and maximizing them is possible, not requiring Mutual Information Neural Estimators (MINE) although possible: 
\begin{equation}
\resizebox{1\linewidth}{!}{
$\begin{gathered}
\max_{f_\theta} \;\; \mathbb{E}_{\mathbf{X}, \mathbf{Y} \sim p(X,Y)}\left [f_\theta(\mathbf{X}, \mathbf{Y}) \right] - \log \,\mathbb{E}_{\mathbf{X}, \mathbf{Y} \sim p(X)p(Y)}\left [e^{f_\theta(\mathbf{X}, \mathbf{Y})} \right],\\
\max_{f_\theta} \;\; \frac{\left(\mathbb{E}_{\mathbf{X}, \mathbf{Y} \sim p(X,Y)} \left [f_\theta (\mathbf{X}, \mathbf{Y}) \right] \right )^2}{ \mathbb{E}_{\mathbf{X}, \mathbf{Y} \sim p(X)p(Y)} \left [f_\theta^2 (\mathbf{X}, \mathbf{Y}) \right]}. 
\end{gathered}$}
\label{eq_36}
\end{equation}
The first maximization in Eq.~\eqref{eq_36} is for Shannon's mutual information, while the second maximization is for R\'enyi's mutual information, justified by the Donsker–Varadhan inequality and the Cauchy-Schwarz inequality:
\begin{equation}
\resizebox{.95\linewidth}{!}{
$\begin{aligned}
&\left \langle f_\theta(X,Y) \sqrt{p(X)}\sqrt{p(Y)}, \frac{p(X,Y)}{\sqrt{p(X)}\sqrt{p(Y)}} \right \rangle^2  \\
& \;\;\;\;\;\;\;\;\;\;\;\;\;\;\; \leq \iint f_\theta^2(X,Y) p(X)p(Y) dXdY \cdot \iint \frac{p^2(X,Y)}{p(X)p(Y)} dXdY.
\end{aligned}$}
\end{equation}
One can use the optimizations in Eq.~\eqref{eq_36} to estimate the mutual information value, and find the encoder that maximizes the mutual information between $X$ and $Y$. But we find that for simple toy distributions, Eq.~\eqref{EQ_35} of estimating the density and mutual information directly from samples is sufficient and produces the same result as Eq.~\eqref{EQ_35}.

\subsection{Results}
\label{quantitative_analysis_simple_results}

Here we show the results for the constructed toy datasets. The following experiments and results are shown: (1) Tracking the cost ($\frac{\langle p(X|Y), q(X|Y)\rangle_{p(Y)}^2}{||q(X|Y)||_{p(Y)}^2}$ in Eq.~\eqref{equation_34}) and the bound (the norm $||p(X|Y)||_{p(Y)}^2$ in Eq.~\eqref{equation_34}) for the encoder-mixture-decoder, we show that the bound indeed holds, and the bound becomes tighter when the number of centers used in the mixture decoder increases; (2) For these simple datasets when the upper bound can be estimated through Eq.~\eqref{equation_34} from samples, we found that the learned solution and the maximal value of the bound by maximizing the bound directly match the results of maximizing the cost, particularly when multiple centers are used in the mixture decoder; (3) Comparing the optimal solutions by visualizing the features when maximizing the conditional entropy in $L_2$ and Shannon's mutual information, as well as R\'enyi's mutual information, we can show that the solution when maximizing the conditional entropy in $L_2$ is different from the other two, with a sharper and more clear boundary.\vspace{9pt}

\begin{figure}[h]
\centering
\begin{subfigure}{.32\textwidth}\includegraphics[width=\linewidth]{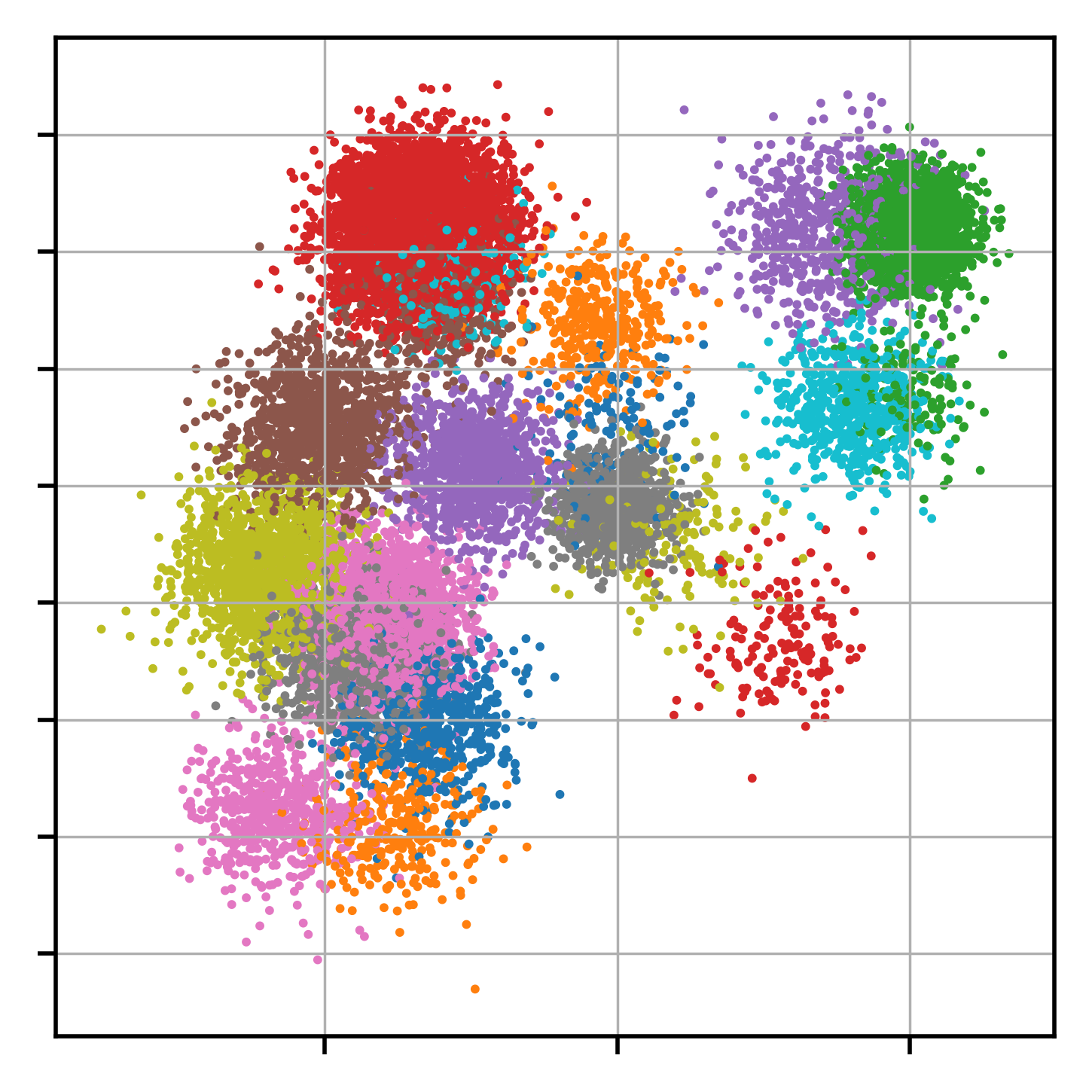}\vspace{-5pt}
\caption*{\textbf{MIX1}}
\end{subfigure}
\begin{subfigure}{.32\textwidth}\includegraphics[width=\linewidth]{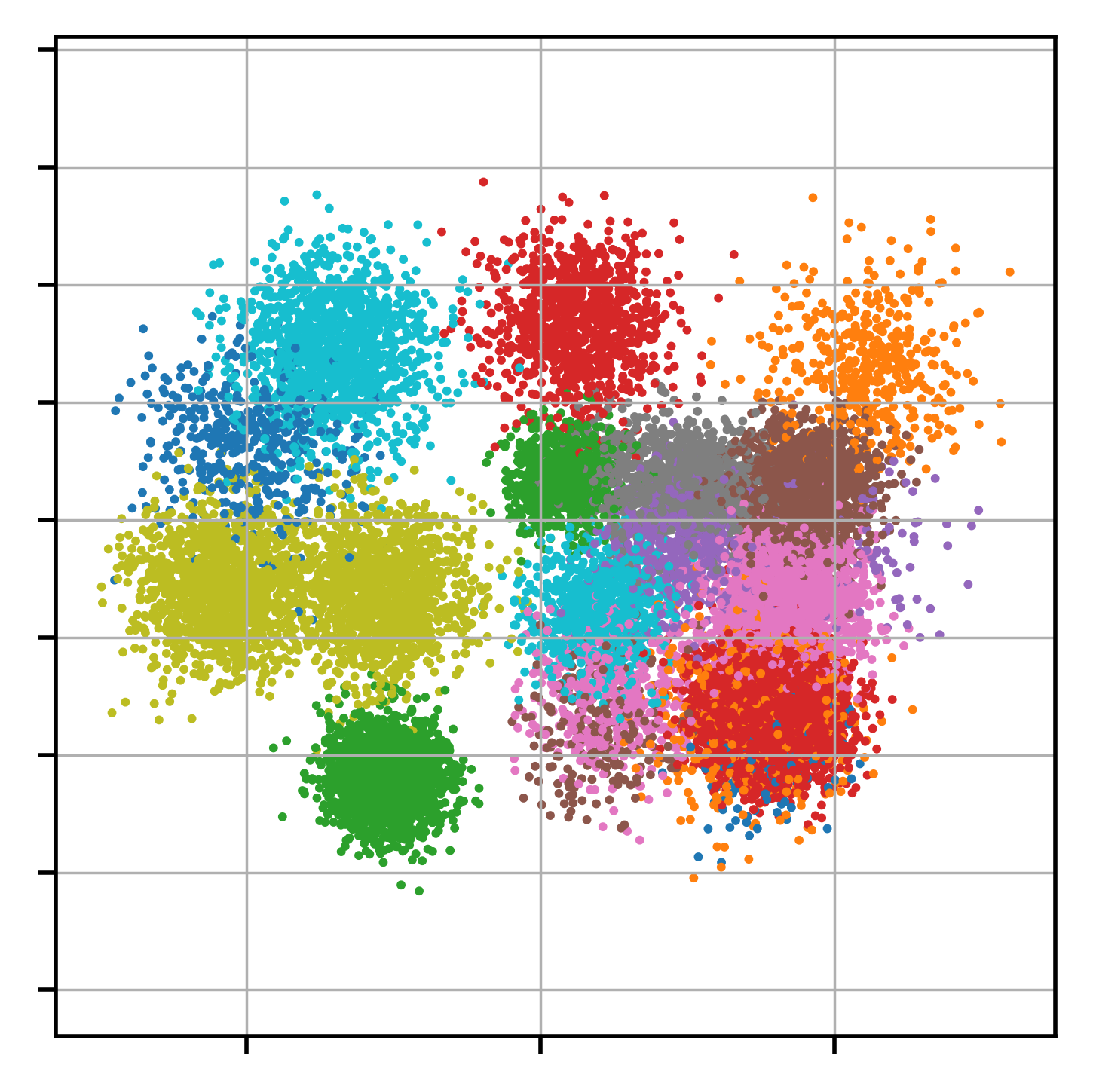}\vspace{-5pt}
\caption*{\textbf{MIX2}}
\end{subfigure}
\begin{subfigure}{.32\textwidth}\includegraphics[width=\linewidth]{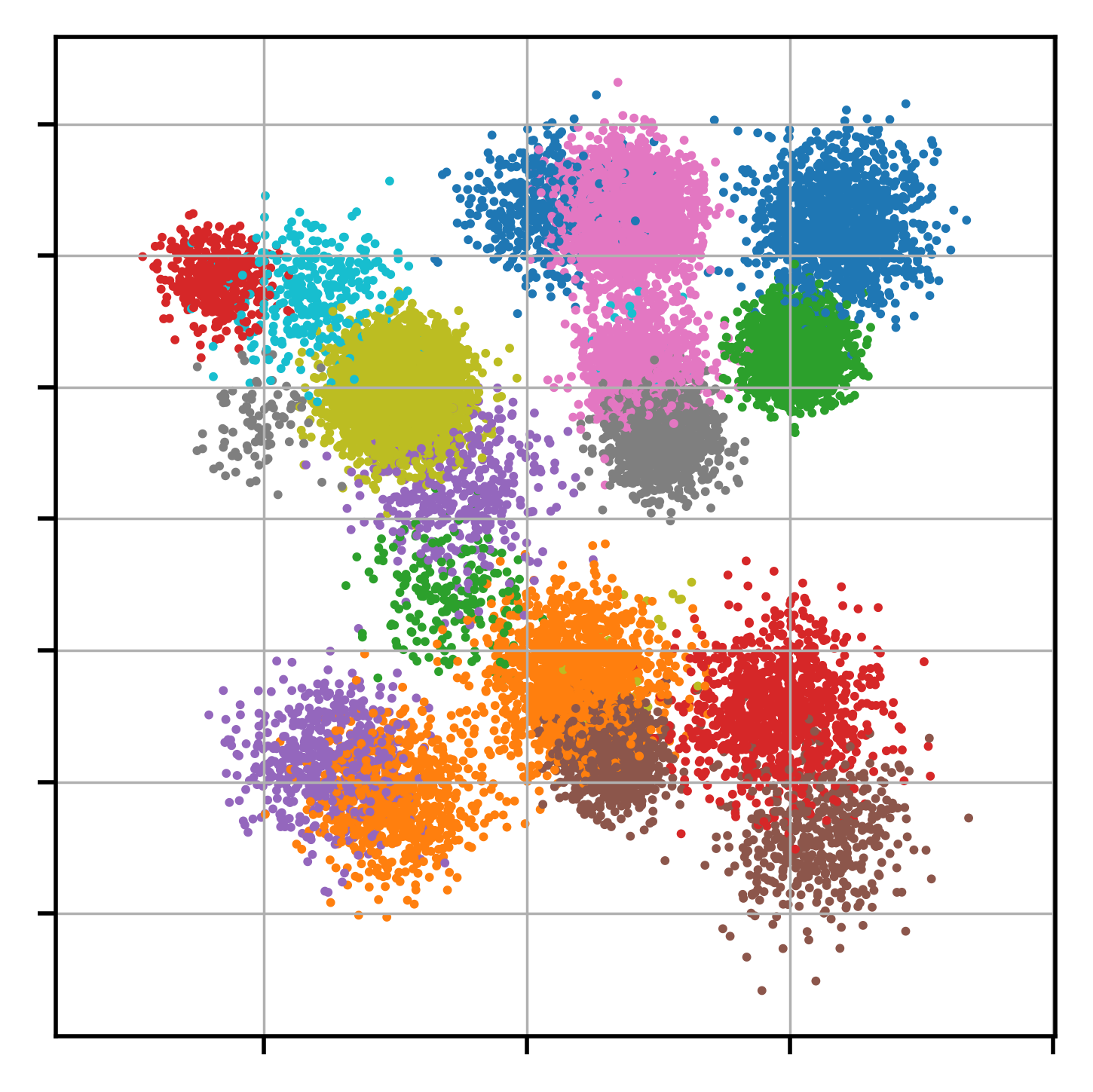}\vspace{-5pt}
\caption*{\textbf{MIX3}}
\end{subfigure}\vspace{-7pt}
\caption{Additional datasets for the quantitative experiments.}
\label{figure_dataset_mix}
\end{figure}

\noindent \textbf{{Dataset.}} Including the three toy datasets we used in Fig.~\ref{9a}, a five-center Mixture of Gaussians ({MOG}), a two-moon distribution ({TM}) and a Gaussian distribution ({GAUSS}), we also include three Gaussian mixture distributions as in Fig.~\ref{figure_dataset_mix}, {MIX1}, {MIX2} and {MIX3}. Each of them will have $20$ Gaussian components with a diagonalized covariance matrix with the variance value randomly sampled ranging from $0.2$ to $0.8$. 

\vspace{-5pt}`

The encoder still maps the $2D$ toy datasets into $1D$ features, and the decoder maps the datasets back to $2D$, similar to Sec.~\ref{section_results_mixture_data}. Since the feature dimension is insufficient, the reconstructions will not be exact, which provides the opportunity for quantitative analysis and feature visualization. When the dataset is in $2D$, the optimal solutions with different centers and baselines may be different but not significantly, so we also construct a dataset sampled from a $5D$ uniform distribution, which truly shows the difference in the solution, but may cause difficulties in visualizing the features. The $2D$ datasets all have $2,000$ samples. The $5D$ uniform dataset contains $10,000$ samples.\vspace{9pt}

\noindent \textbf{{Baselines.}} We also implement multiple baselines, including the Mutual Information 
Neural Estimators (MINE) for Shannon's and R\'enyi's mutual information, indicated in Eq.~\eqref{eq_36} using a neural network and sample estimators. We estimate and maximize the mutual information values simultaneously during training, where the mutual information estimator estimates the value, and the encoder maximizes this value through gradient ascent.

As the objective function involves the dependence measures, we also include the kernel dependence estimators, the Kernel Independent Component Analysis (KICA)~\cite{bach2002kernel} and Hilbert-Schmidt Independence Criterion (HSIC)~\cite{gretton2005measuring}, as evaluations for the dependence between the data and the learned features. However, these kernel measures are more difficult to use as objective functions to maximize.\vspace{9pt}

\begin{figure}[t]
\begin{subfigure}{.33\textwidth}\includegraphics[width=\linewidth]{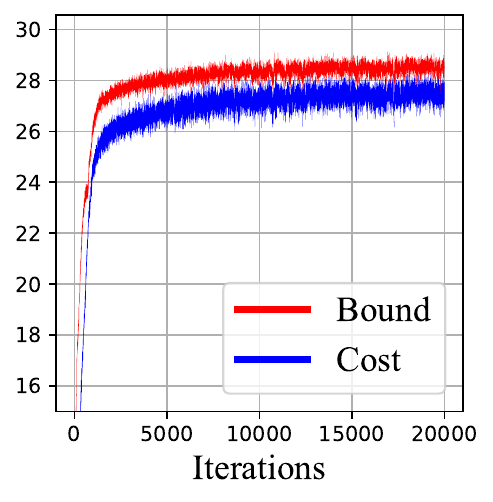}\vspace{-5pt}
\caption*{\textbf{MIX1}: \textbf{\textit{1C}}}
\end{subfigure}\hspace{-5pt}
\begin{subfigure}{.33\textwidth}\includegraphics[width=\linewidth]{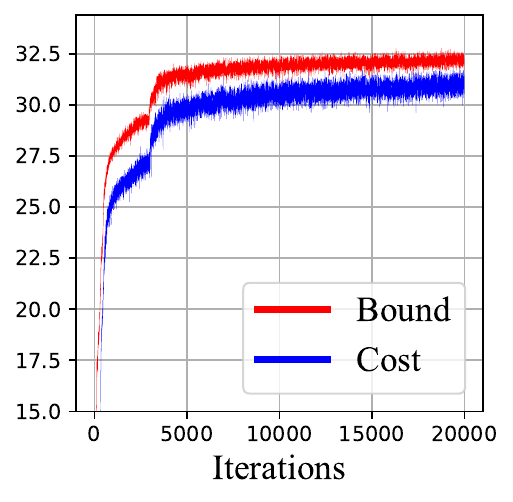}\vspace{-5pt}
\caption*{\textbf{MIX3}: \textbf{\textit{1C}}}
\end{subfigure}\hspace{-5pt}
\begin{subfigure}{.33\textwidth}\includegraphics[width=\linewidth]{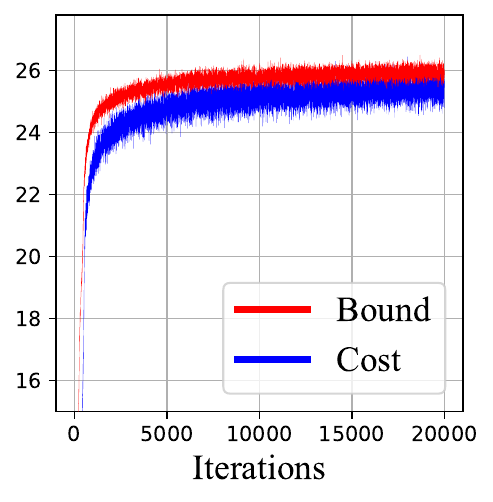}\vspace{-5pt}
\caption*{\textbf{MOG}: \textbf{\textit{1C}}}
\end{subfigure}\hspace{-5pt}
\begin{subfigure}{.33\textwidth}\includegraphics[width=\linewidth]{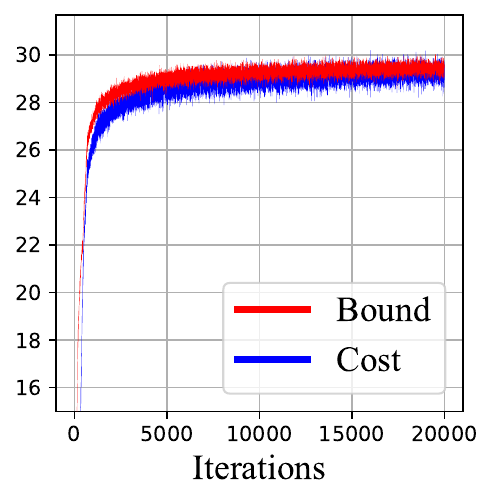}\vspace{-5pt}
\caption*{\textbf{MIX1}: \textbf{\textit{30C}}}
\end{subfigure}\hspace{-5pt}
\begin{subfigure}{.33\textwidth}\includegraphics[width=\linewidth]{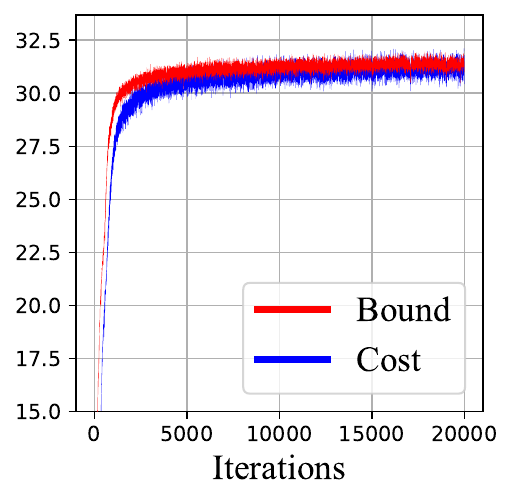}\vspace{-5pt}
\caption*{\textbf{MIX3}: \textbf{\textit{30C}}}
\end{subfigure}\hspace{-5pt}
\begin{subfigure}{.33\textwidth}\includegraphics[width=\linewidth]{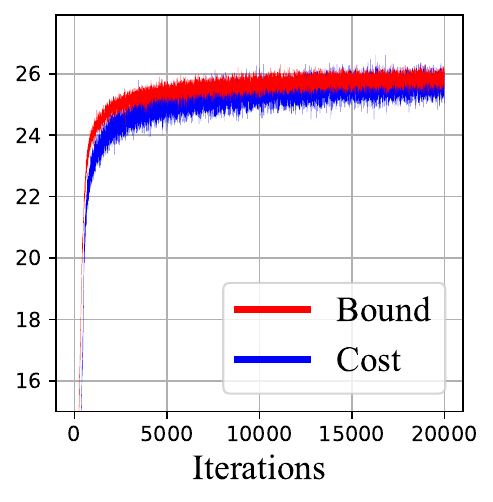}\vspace{-5pt}
\caption*{\textbf{MOG}: \textbf{\textit{30C}}}
\end{subfigure}\vspace{10pt}
\begin{subfigure}{.33\textwidth}\includegraphics[width=\linewidth]{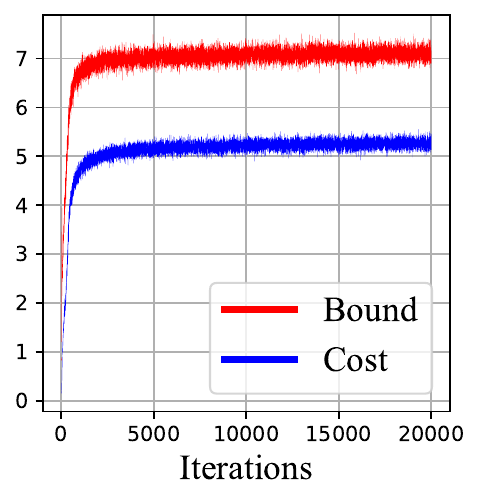}\vspace{-5pt}
\caption*{\textbf{5D Uniform}: \textbf{\textit{1C}}}
\label{15g}
\end{subfigure}\hspace{-5pt}
\begin{subfigure}{.33\textwidth}\includegraphics[width=\linewidth]{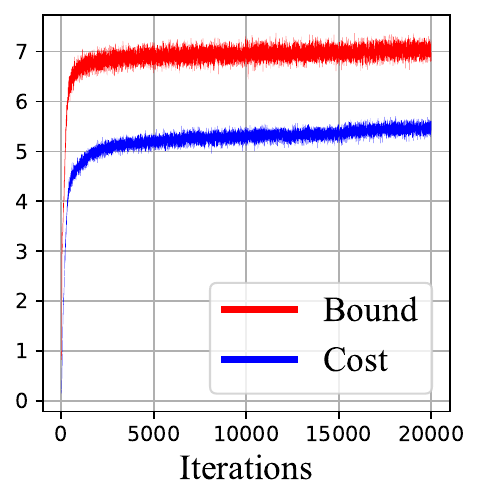}\vspace{-5pt}
\caption*{\textbf{5D Uniform}: \textbf{\textit{3C}}}
\end{subfigure}\hspace{-5pt}
\begin{subfigure}{.33\textwidth}\includegraphics[width=\linewidth]{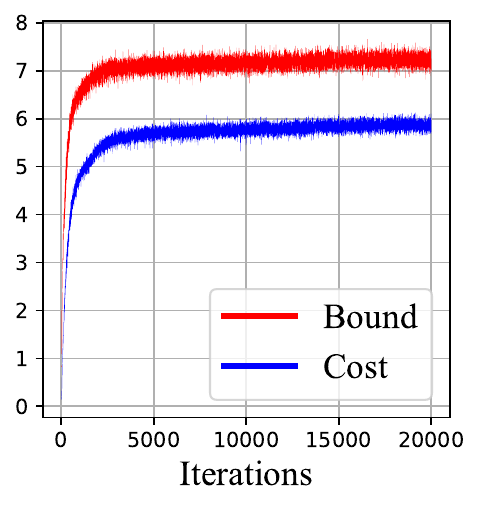}\vspace{-5pt}
\caption*{\textbf{5D Uniform}: \textbf{\textit{5C}}}
\end{subfigure}\hspace{-5pt}
\begin{subfigure}{.33\textwidth}\includegraphics[width=\linewidth]{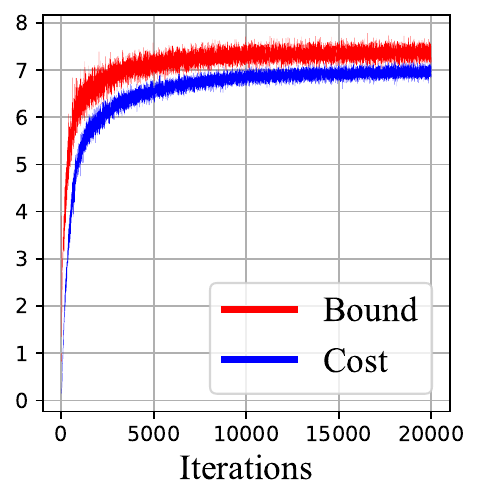}\vspace{-5pt}
\caption*{\textbf{5D Uniform}: \textbf{\textit{30C}}}
\end{subfigure}\hspace{-5pt}
\begin{subfigure}{.33\textwidth}\includegraphics[width=\linewidth]{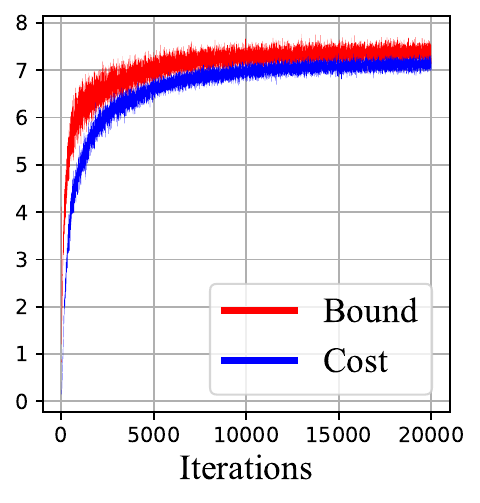}\vspace{-5pt}
\caption*{\textbf{5D Uniform}: \textbf{\textit{100C}}}
\label{15j}
\end{subfigure}
\caption{Visualizing the learning curves for datasets \textbf{MIX1}, \textbf{MIX2}, \textbf{MIX3} and \textbf{5D Uniform} when varying the number of centers. For $2D$ datasets, the bound becomes more tight from $1$ center (\textbf{\textit{1C}}) to $30$ centers (\textbf{\textit{30C}}). This is more evident on the $5D$ uniform dataset. We maximize the cost and track the bound values during training, following Eq.~\eqref{equation_34}.}
\label{learning_curve_distributions}
\end{figure}

\noindent \textbf{\textit{Table~\ref{tab:model_accuracy}: Increasing the number of centers makes the bound tighter and increases the optimal bound value.}} We estimate and maximize the cost, and track the bound value during training, following Eq.~\eqref{equation_34}, with results shown in Table~\ref{tab:model_accuracy} and visualized in Fig.~\ref{learning_curve_distributions}. 

We vary the number of centers for the mixture decoder to be $1$, $3$, $5$, $30$. Apart from the values obtained from maximizing the cost, we also record the values obtained from maximizing the bound with Eq.~\eqref{equation_34}, which requires only an encoder without the mixture decoder, denoted as \textbf{Max Bound} in the table, used as the ground truth value that maximizing the cost should reach. The purpose is to show that with the number of centers increasing, the bound will become tighter, and the value of the upper bound will increase. And eventually maximizing the cost and maximizing the bound (\textbf{Max Bound}) will reach the same maximal value.

\begin{table}[t]
\setlength{\tabcolsep}{4pt}
\centering
\caption{Results for maximizing the cost when varying the number of centers to be $1$, $3$, $5$, $30$, compared with the maximal value when maximizing the bound directly (\textbf{Max Bound}), following Eq.~\eqref{equation_34}. We found that increasing centers indeed makes the bound more tight and increases the maximal bound value. And maximizing the cost (such as \textbf{\textit{5C}} and \textbf{\textit{30C}}) matches the optimal values from maximizing the bound (\textbf{Max Bound}).}
\resizebox{1\linewidth}{!}{
\footnotesize
\label{tab:model_accuracy}
\begin{tabular}{r S S S S S S S S S S S}
    \toprule[2pt] \vspace{3pt}
    \textbf{} & \textbf{MOG} & \textbf{TM} & \textbf{GAUSS} & \textbf{MIX1} & \textbf{MIX2} & \textbf{MIX3} & \textbf{5D-U} \\ 
    \midrule[1pt] \\ \vspace{-14pt} \\
    \textbf{Cost (\textit{1C})} & 25.1 & 21.99 & 39.82 & 27.94 & 27.2 & 29.74 & 5.27 \vspace{3pt}\\
    \textbf{Bound (\textit{1C})} & 25.67 & 22.61 & 40.23 & 28.94 & 28.0 & 30.81 & 7.10 \\ \vspace{-7pt} \\
    \midrule[0.1pt] \\ \vspace{-14pt} \\
    \textbf{Cost (\textit{3C})} & 25.4 & 22.86 & 39.81 & 28.7 & 27.82 & 31.21 & 5.46 \vspace{3pt}\\
    \textbf{Bound (\textit{5C})} & 25.76 & 23.23 & 40.21 & 29.12 & 28.22 & 31.61 & 7.01\\ \vspace{-7pt} \\
    \midrule[0.1pt] \\ \vspace{-14pt} \\
    \textbf{Cost (\textit{5C})} & 25.57 & 23.58 & 39.85 & 28.83 & 27.75 & 31.09 & 5.86 \vspace{3pt}\\
    \textbf{Bound (\textit{5C})} & 25.9 & 23.85 & 40.16 & 29.13 & 28.08 & 31.49 & 7.20 \\ \vspace{-7pt} \\
    \midrule[0.1pt] \\ \vspace{-14pt} \\
    \textbf{Cost (\textit{30C})} & 25.53 & 23.57 & 39.88 & 28.75 & 27.96 & 31.02 & 6.97 \vspace{3pt}\\
    \textbf{Bound (\textit{30C})} & 25.78 & 23.88 & 40.25 & 29.07 & 28.22 & 31.3 & 7.35 \\ \vspace{-7pt} \\
    \midrule[0.1pt] \\ \vspace{-13pt} \\
   \textbf{Max Bound} & 25.80 & 23.95 & 41.26 & 29.28 & 27.46 & 31.62 & 7.35\\ \vspace{-6pt} \\
    \bottomrule[2pt]
\end{tabular}}\vspace{-3pt}
\end{table}

We break down the table as the following. First looking at the scores with $1$ center ($\textbf{\textit{1C}}$), the cost is constantly below the bound across all datasets, which is more evident on the three mixture datasets ($\textbf{{MIX1, MIX2, MIX3}}$) and the $5D$ uniform dataset (\textbf{5D-U}). The regular autoencoder 
not using the mixture decoder also corresponds to the case with $1$ center. 

Second when the number of centers increases from $1$ to $3$, the bound becomes more tight for the most datasets, with a tiny increase in the bound value for $2D$ datasets and a more visible increase in the bound value for the $5D$ uniform dataset. 

Further increasing the number of centers, from $\textbf{5D-U}$, it can be seen that the bound gets tighter with a visible increase in the value of the bound. 

Next, if we compare the bound value from maximizing the cost (\textbf{Bound}) and maximizing the bound directly as an objective (\textbf{Max Bound}), the values are extremely close. This comparison is possible only for toy distributions when the bound can be estimated from samples and used as the objective function (Eq.~\eqref{equation_34}).

Thus from the table we can conclude that for simple $2D$ datasets, increasing the number of centers from $\textbf{\textit{1C}}$ to $\textbf{\textit{3C}}$ makes the bound tighter and increases the bound value. For the $5D$ uniform datasets, the tightness and the value of the bound increases visibly for $1$, $3$, $5$ and $30$ centers. The optimal value of the maximizing the cost and maximizing the bound (\textbf{Max Bound}) matches, only when multiple centers are used in the mixture decoder. So using multiple centers in a mixture decoder, corresponding to the encoder-mixture-decoder, indeed has an advantage to using a regular decoder with only one center, corresponding to the regular autoencoder. The regular autoencoder is the case of the mixture decoder with only one center.\vspace{9pt}

\noindent \textit{\textbf{Fig.~\ref{learning_curve_distributions}: Learning curves show that the bound becomes tighter when increasing the number of centers.}} We further visualize the learning curves for obtaining the values in Table~\ref{tab:model_accuracy}, shown in Fig.~\ref{learning_curve_distributions}. We pick \textbf{MIX1}, \textbf{MIX2}, \textbf{MIX3} and \textbf{5D Uniform} to visualize. 

The first row of the figure is when using $1$ center in the decoder and maximizing the cost. A visible gap can be seen between the cost and the bound. The second row of the figure is when we increase the number of centers to $30$, and the gap is visibly smaller than the $1$-center case. This is more evident on the $5D$ uniform example when we visualize the learning curves for the cost and the bound for $1$, $5$, $5$, $30$, $100$ centers. It can be seen clearly that the bound becomes tighter when the number of centers increases. The value of the bound follows Table~\ref{tab:model_accuracy} and will increase from $7.1$ to $7.35$ with the increasing number of centers.\vspace{9pt}

\noindent \textit{\textbf{Table~\ref{table2}: Maximizing the upper bound directly, comparing with the baselines.}} The learning curves in Fig.~\ref{learning_curve_distributions} are the visualization for the scenario of maximizing the cost in Table~\ref{tab:model_accuracy}. We further provide more results for the scenario of maximizing the upper bound directly, which is shown in Table~\ref{table2}. 

As having been introduced in~\eqref{equation_34} and~\eqref{EQ_35}, for simple datasets, the conditional entropy in $L_2$ (our objective function), as well as Shannon's mutual information and R\'enyi's information can be estimated directly through samples, thus can be used directly as the objective function to maximize, to compare the differences in the optimal solutions when using the variational cost of these quantities as the objective. We reiterate the procedure here. Given a series of samples $X_1,X_2,\cdots,X_N$, we apply an encoder to obtain $Y_1,Y_2,\cdots,Y_N$. Then we apply the additive noise on each sample $\widehat{X_n} = X_n+ \sqrt{v_X} \cdot z_n$ and $\widehat{Y_n} = Y_n + \sqrt{v_Y} \cdot s_n$, with $z_n$ and $s_n$ sampled from standardized Gaussians. We first estimate the densities by\vspace{-5pt}
\begin{equation}
\resizebox{1\linewidth}{!}{
$\begin{gathered}
p(X,Y) = \frac{1}{N}\sum_{n=1}^N\mathcal{N}(X-X_n;v_X)\cdot\mathcal{N}(Y-Y_n;v_Y), \vspace{-5pt}\\
p(X) = \frac{1}{N}\sum_{n=1}^N \mathcal{N}(X-X_n;v_X), \;\; p(Y) = \frac{1}{N}\sum_{n=1}^N \mathcal{N}(Y-Y_n;v_Y).
\end{gathered}$}
\label{mixtures_form_equation}
\end{equation}

\noindent Then the conditional entropy in $L_2$, Shannon's and R\'enyi's mutual information can be estimated by
\begin{equation}
\resizebox{1\linewidth}{!}{
$\begin{aligned}
&||p(X|Y)||_{p(Y)}^2 = \frac{1}{N}\sum_{m=1}^N \frac{\sum_{n=1}^N\mathcal{N}(\widehat{X_m}-X_n;v_X)\mathcal{N}(\widehat{Y_m}-Y_n;v_Y)}{\sum_{n=1}^N \mathcal{N}(\widehat{Y_m}-Y_n;v_Y)},\\ \vspace{-35pt} \\ 
& \iint p(X,Y) \log \frac{p(X,Y)}{p(X)p(Y)} dXdY \\
& = \frac{1}{N}\log  \left( N\cdot \sum_{m=1}^N \frac{\sum_{n=1}^N\mathcal{N}(\widehat{X_m}-X_n;v_X)\mathcal{N}(\widehat{Y_m}-Y_n;v_Y)}{\sum_{n=1}^N \mathcal{N}(\widehat{X_m}-X_n;v_X) \sum_{n=1}^N \mathcal{N}(\widehat{Y_m}-Y_n;v_Y)} \right), \\ \vspace{-35pt} \\
&\iint \frac{p^2(X,Y)}{p(X)p(Y)} dXdY \\
&  = N\cdot \sum_{m=1}^N \frac{\sum_{n=1}^N\mathcal{N}(\widehat{X_m}-X_n;v_X)\mathcal{N}(\widehat{Y_m}-Y_n;v_Y)}{\sum_{n=1}^N \mathcal{N}(\widehat{X_m}-X_n;v_X) \sum_{n=1}^N \mathcal{N}(\widehat{Y_m}-Y_n;v_Y)}. 
\end{aligned}$}
\label{sample_estimator_equation}
\end{equation}

\noindent The procedure follows that the densities $p(X)$, $p(Y), p(X,Y)$ are assumed to be the average of Gaussian components in the forms of mixtures (Eq.~\eqref{mixtures_form_equation}). For each of the quantities $||p(X|Y)||_{p(Y)}^2$, $\iint p(X,Y) \log \frac{p(X,Y)}{p(X)p(Y)} dXdY$ and $\iint \frac{p^2(X,Y)}{p(X)p(Y)} dXdY$, we first leave one joint term $p(X,Y)$ and substitute all other forms with the mixture forms~\eqref{mixtures_form_equation}. Then the integral over $p(X,Y)$ can be estimated with empirical samples $\widehat{X_n}$, $\widehat{Y_n}$ with additive Gaussian noises because the joint of them will follow the $p(X,Y)$ assumed to be a mixture density. We optimize the encoder, the mapping from $X$ to $Y$, to maximize these quantities. The scores of them in Table~\ref{table2} are denoted as \textbf{S-MI} (Shannon's MI, $\iint p(X,Y) \log \frac{p(X,Y)}{p(X)p(Y)} dXdY$), \textbf{R-MI} (R\'enyi's MI, $\iint \frac{p^2(X,Y)}{p(X)p(Y)} dXdY$) and \textbf{R-CondEntr} (the conditional entropy in $L_2$, the norm $||p(X|Y)||_{p(Y)}^2$).

In addition to the maximization of these upper bounds, we also implemented MINE for Shannon's mutual information (\textbf{MINE-S}) and R\'enyi's mutual information (\textbf{MINE-R}) following the procedure introduced in Eq.~\eqref{eq_36}. We need to initiate a mutual information estimator $f_\theta(X,Y)$ that estimates the mutual information by maximizing Eq.~\eqref{eq_36}, then updates the encoder to further maximize this estimation, which can be done simultaneously at each iteration with gradient ascent. The inputs to MINE has to be the samples and features with additive Gaussian noises $\widehat{X_n}$, $\widehat{Y_n}$ such that their joint follows the assumption of $p(X,Y)$ as a Gaussian mixture. We also implement the regular autoencoder (\textbf{AE}) that minimizes the mean-squared error to learn features $Y$. 

After training the encoder with the objectives \textbf{S-MI}, \textbf{R-MI}, \textbf{R-CondEntr}, \textbf{MINE-S}, \textbf{MINE-R} and \textbf{AE}, we further use the estimators \textbf{S-MI}, \textbf{R-MI}, and \textbf{R-CondEntr}, with additional kernel dependence estimators \textbf{KICA} and \textbf{HSIC}, to evaluate the dependence between samples $X$ and learned features $Y$.\vspace{5pt}

\begin{table}[t]
\centering
\caption{The optimal values of Shannon's mutual information (\textbf{S-MI}), R\'enyi's mutual information (\textbf{R-MI}), the conditional entropy in $L_2$ which our approach optimizes (\textbf{R-CondEntr}), cross evaluated by the three scores of them after training. Indeed they reach different solutions, and the maximal score is obtained only when using the score as an objective, marked in red. Using neural mutual information estimator (\textbf{MINE-S} and \textbf{MINE-R}) and the regular autoencoder (\textbf{AE}) produces a similar result. The estimations made by MINE are marked in blue, followed by their empirical estimations (Eq.~\eqref{sample_estimator_equation}). In addition, we also use the kernel dependence estimators (\textbf{KICA} and \textbf{HSIC}) for evaluation.}
\resizebox{.95\linewidth}{!}{%
\begin{tabular}{c}
  \subcaptionbox{\large Mix1}{
    \begin{tabular}{c c c c c c}
      \toprule[2pt]
      \diagbox{Cost}{Eval} & \textbf{S-MI} & \textbf{R-MI} & \textbf{R-CondEntr} & \textbf{KICA} & \textbf{HSIC} \\
      \midrule[1pt]
      \textbf{S-MI}   & \textcolor{red}{\textbf{1.83}} & 7.81 & 28.03 & 25.33 & 24.42  \\
      \textbf{R-MI} & 1.79 & \textcolor{red}{\textbf{8.17}} & 26.24 & 27.87 & 25.99
 \\
      \textbf{R-CondEntr} & 1.76 & 7.12 & \textcolor{red}{\textbf{29.19}} & 22.51 & 21.76 \\
        \textbf{MINE-S} & \textbf{\textcolor{blue}{1.81}} / 1.83 & 7.84 & 27.81 & 25.09 & 23.74 \\
        \textbf{MINE-R} & 1.81 & \textbf{\textcolor{blue}{6.50}} / 7.7 & 27.65 & 23.94 & 23.78 \\
      \textbf{AE} & 1.71 & 7.18 & 24.54 & 17.48 & 18.05 \\
    \bottomrule[2pt]
    \end{tabular}
  } \vspace{7pt} \\[7pt]
  \subcaptionbox{\large Mix2}{
    \begin{tabular}{c c c c c c}
      \toprule[2pt]
      \diagbox{Cost}{Eval} & \textbf{S-MI} & \textbf{R-MI} & \textbf{R-CondEntr} & \textbf{KICA} & \textbf{HSIC} \\
      \midrule[1pt]
      \textbf{S-MI} & \textcolor{red}{\textbf{1.81}} & 7.8 & 25.98 & 22.38 & 22.06 \\
      \textbf{R-MI} & 1.74 & \textcolor{red}{\textbf{8.08}} & 23.72 & 25.2 & 24.21  \\
      \textbf{R-CondEntr} & 1.74 & 7.01 & \textcolor{red}{\textbf{27.74}} & 22.87 & 22.11 \\
      \textbf{MINE-S} & \textbf{\textcolor{blue}{1.8}} / 1.82 & 7.84 & 26.41 & 23.82 & 23.08 \\
      \textbf{MINE-R} & 1.8 & \textbf{\textcolor{blue}{6.65}} / 7.72 & 25.78 & 22.39 & 22.24 \\
    \textbf{AE} & 1.76 & 7.78 & 24.59 & 19.85 & 20.44 \\
      \bottomrule[2pt]
    \end{tabular} 
  } \vspace{7pt} \\[7pt]
    \subcaptionbox{\large Mix3}{
      \begin{tabular}{c c c c c c}
        \toprule[2pt]
      \diagbox{Cost}{Eval} & \textbf{S-MI} & \textbf{R-MI} & \textbf{R-CondEntr} & \textbf{KICA} & \textbf{HSIC} \\
        \midrule[1pt]
        \textbf{S-MI} & \textcolor{red}{\textbf{1.91}} & 8.22 & 30.47 & 26.8 & 23.6 \\
        \textbf{R-MI} & 1.86 & \textcolor{red}{\textbf{8.43}} & 28.93 & 27.0 & 24.39 \\
        \textbf{R-CondEntr} & 1.86 & 7.8 & \textcolor{red}{\textbf{31.62}} & 26.18 & 23.42 \\
        \textbf{MINE-S} & \textcolor{blue}{\textbf{1.86}} / 1.89 & 8.13 & 30.07 & 25.04 & 23.33 \\
        \textbf{MINE-R} & 1.88 & \textcolor{blue}{\textbf{6.81}} / 8.07 & 30.08 & 23.27 & 22.12 \\
        \textbf{AE} & 1.86 & 7.92 & 28.12 & 21.98 & 20.79 \\
        \bottomrule[2pt]
      \end{tabular}
    }
\end{tabular}%
}\vspace{-5pt}
\label{table2}
\end{table}

\begin{figure*}[t]
\centering
\begin{subfigure}{.11\textwidth}\includegraphics[width=\linewidth]{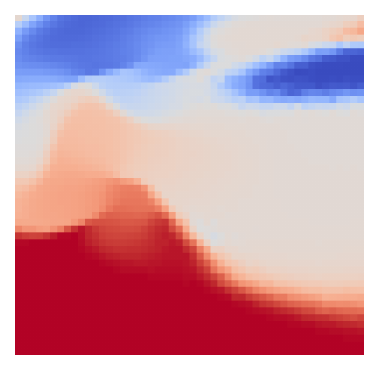}
\caption*{\textbf{MIX1} \textbf{\textit{1C}}}
\end{subfigure}
\begin{subfigure}{.11\textwidth}\includegraphics[width=\linewidth]{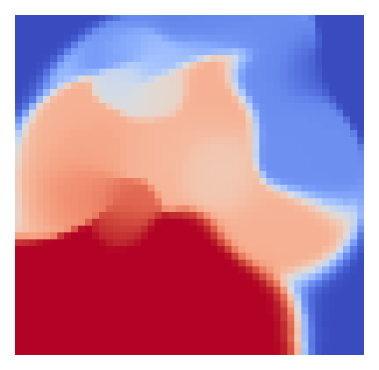}
\caption*{\textbf{MIX1} \textbf{\textit{3C}}}
\end{subfigure}
\begin{subfigure}{.11\textwidth}\includegraphics[width=\linewidth]{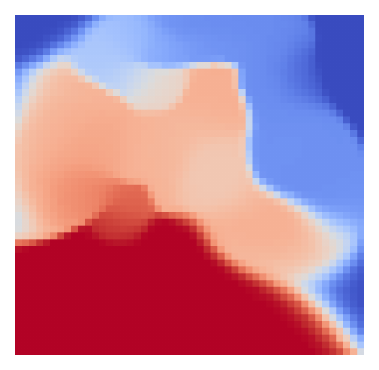}
\caption*{\textbf{MIX1} \textbf{\textit{5C}}}
\end{subfigure}
\begin{subfigure}{.11\textwidth}\includegraphics[width=\linewidth]{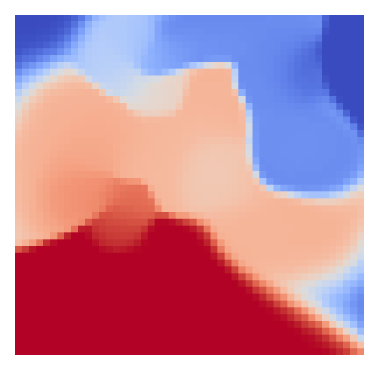}
\caption*{\textbf{MIX1} \textbf{\textit{MAX}}}
\end{subfigure}
\begin{subfigure}{.11\textwidth}\includegraphics[width=\linewidth]{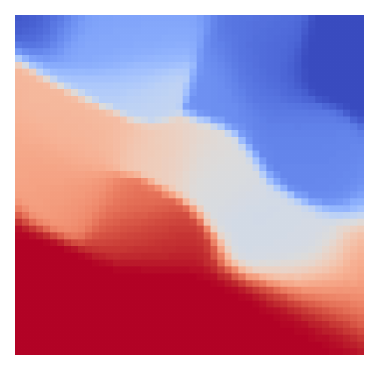}
\caption*{\textbf{MIX1} \textbf{\textit{S-MI}}}
\end{subfigure}
\begin{subfigure}{.11\textwidth}\includegraphics[width=\linewidth]{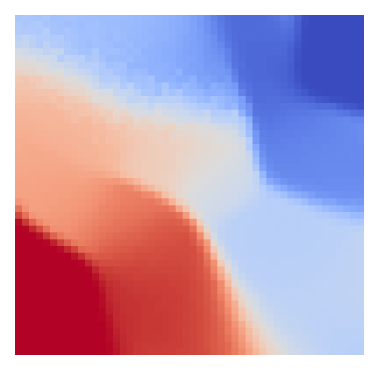}
\caption*{\textbf{MIX1} \textbf{\textit{R-MI}}}
\end{subfigure}
\begin{subfigure}{.11\textwidth}\includegraphics[width=\linewidth]{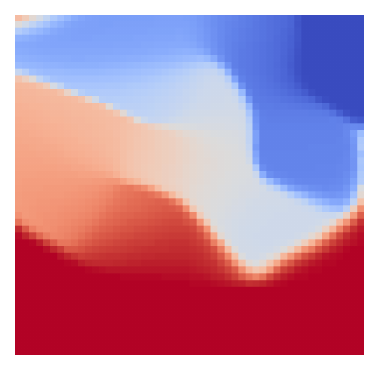}
\caption*{\textbf{MIX1} \textbf{\textit{MINE-S}}}
\end{subfigure}
\begin{subfigure}{.11\textwidth}\includegraphics[width=\linewidth]{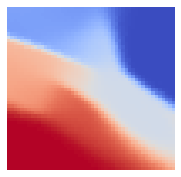}
\caption*{\textbf{MIX1} \textbf{\textit{AE}}}
\end{subfigure}\vspace{5pt}

\begin{subfigure}{.11\textwidth}\includegraphics[width=\linewidth]{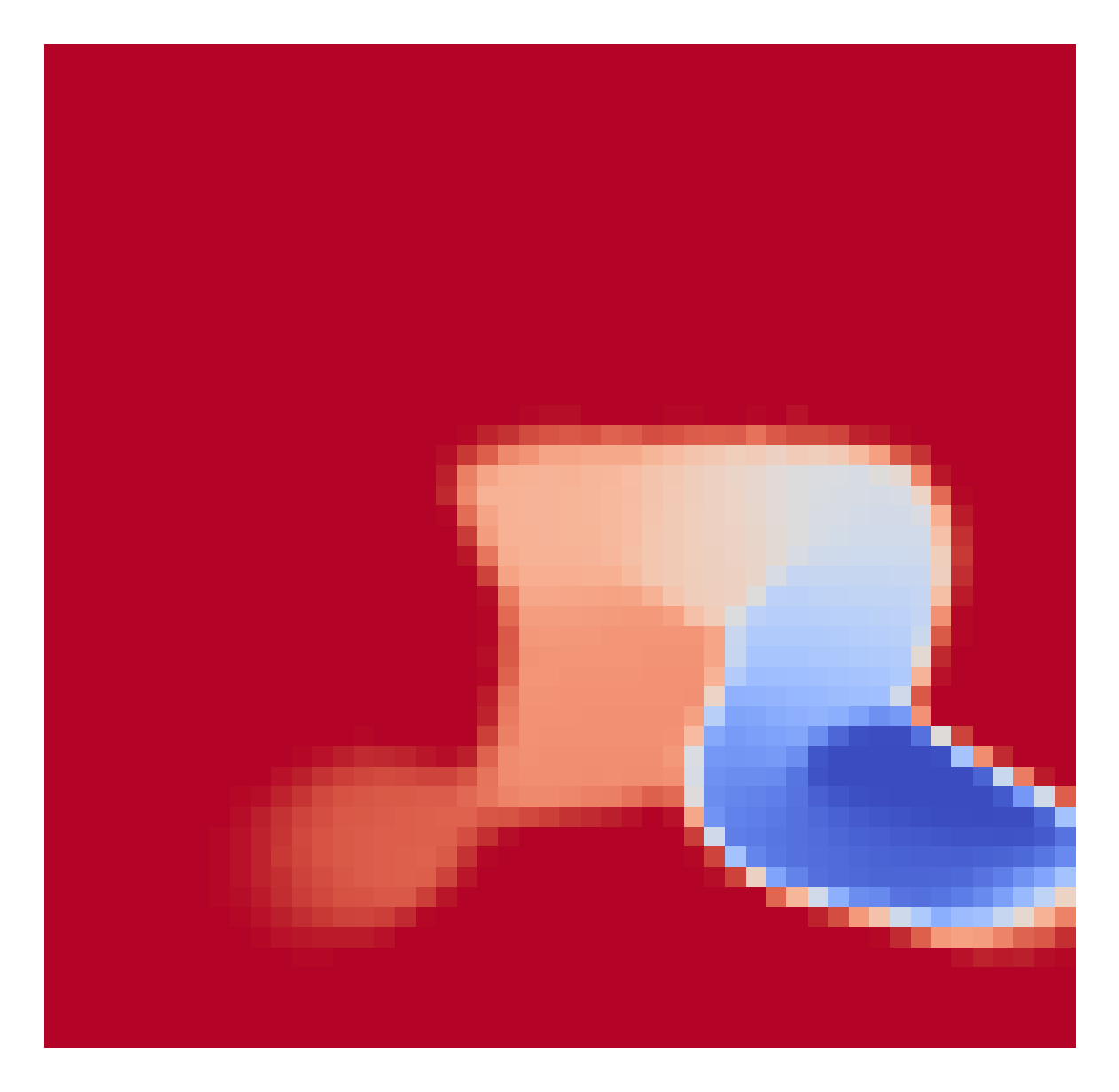}
\caption*{\textbf{MIX2} \textbf{\textit{1C}}}
\end{subfigure}
\begin{subfigure}{.11\textwidth}\includegraphics[width=\linewidth]{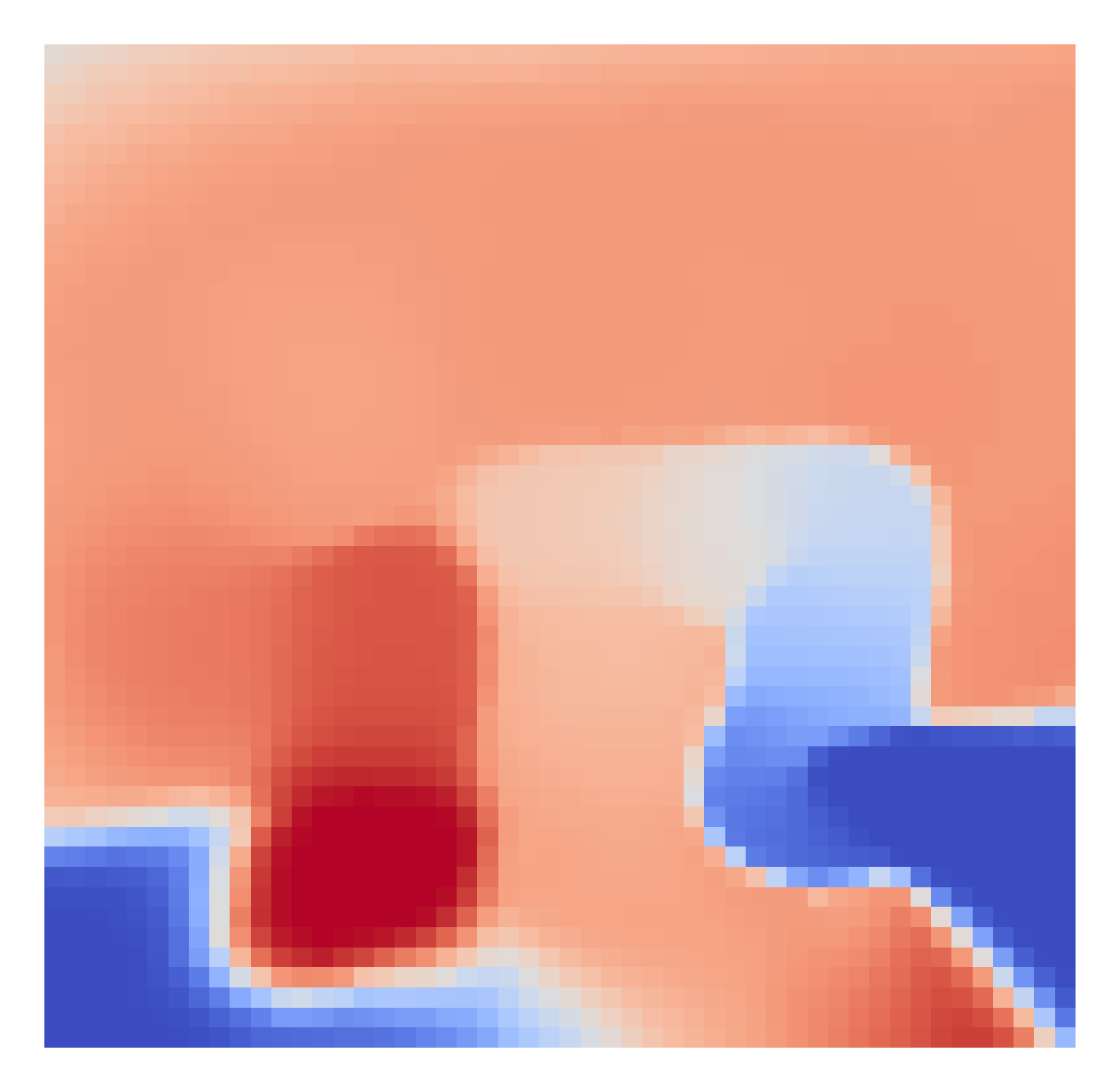}
\caption*{\textbf{MIX2} \textbf{\textit{3C}}}
\end{subfigure}
\begin{subfigure}{.11\textwidth}\includegraphics[width=\linewidth]{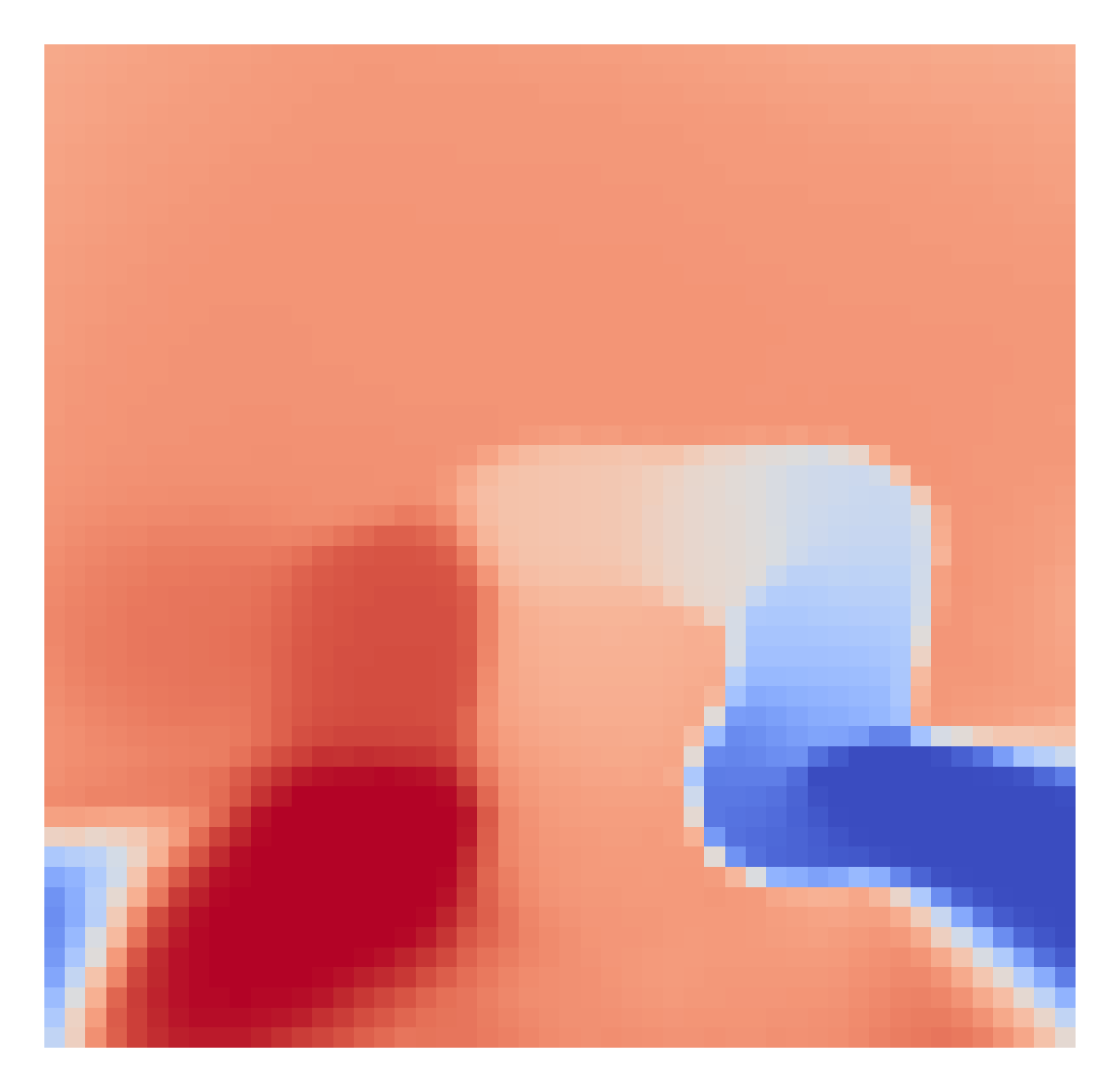}
\caption*{\textbf{MIX2} \textbf{\textit{5C}}}
\end{subfigure}
\begin{subfigure}{.11\textwidth}\includegraphics[width=\linewidth]{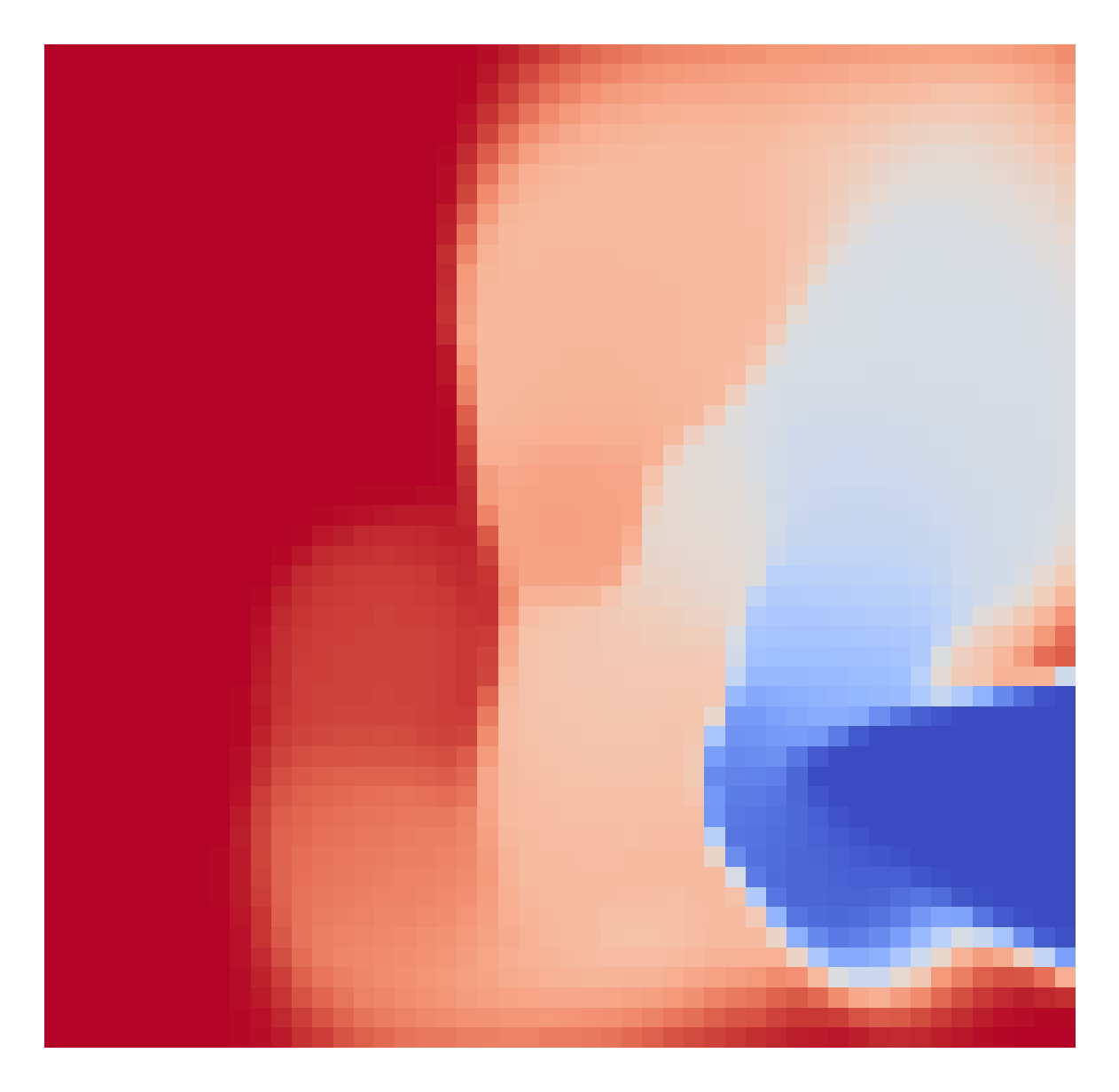}
\caption*{\textbf{MIX2} \textbf{\textit{MAX}}}
\end{subfigure}
\begin{subfigure}{.11\textwidth}\includegraphics[width=\linewidth]{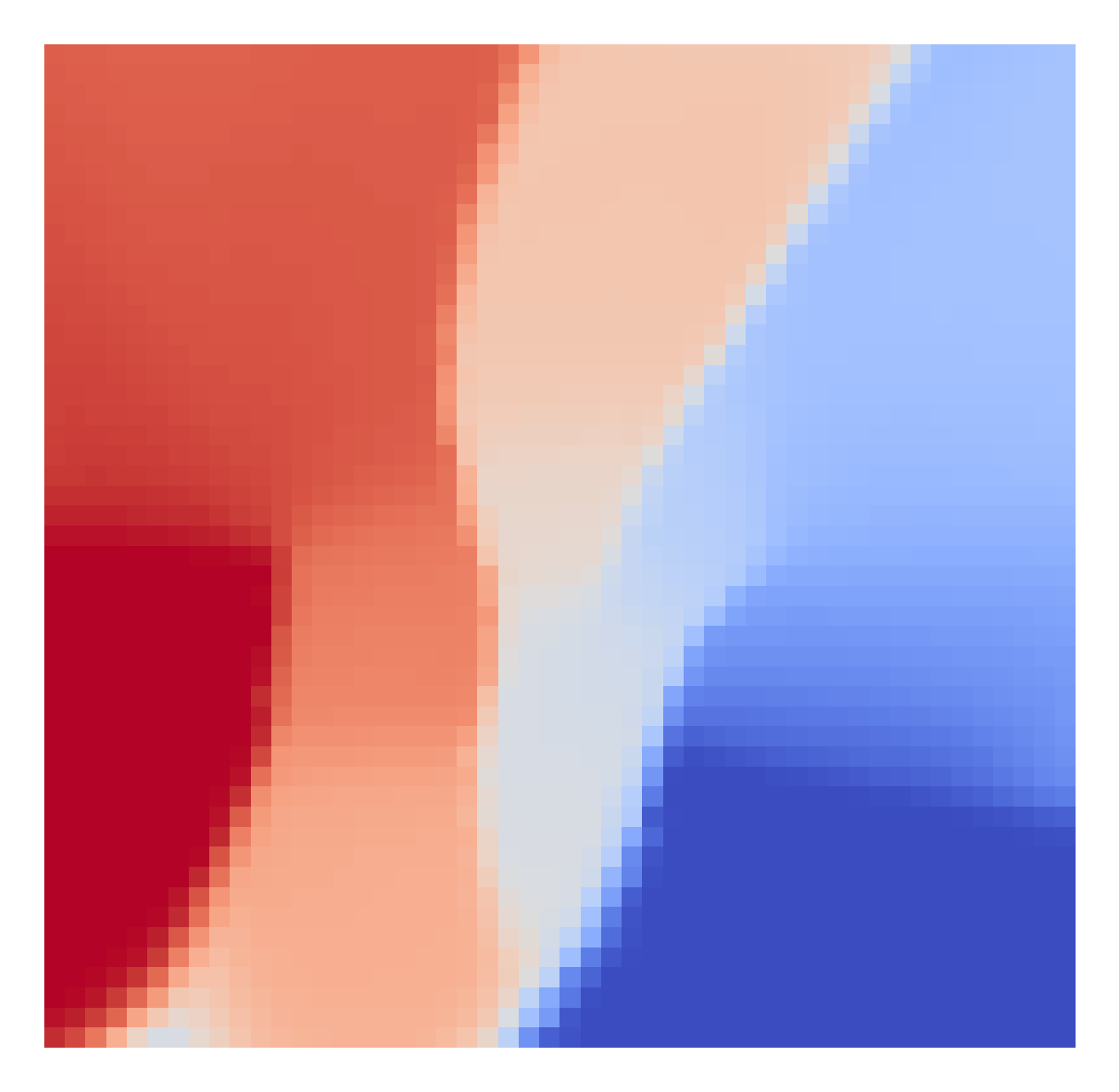}
\caption*{\textbf{MIX2} \textbf{\textit{S-MI}}}
\end{subfigure}
\begin{subfigure}{.11\textwidth}\includegraphics[width=\linewidth]{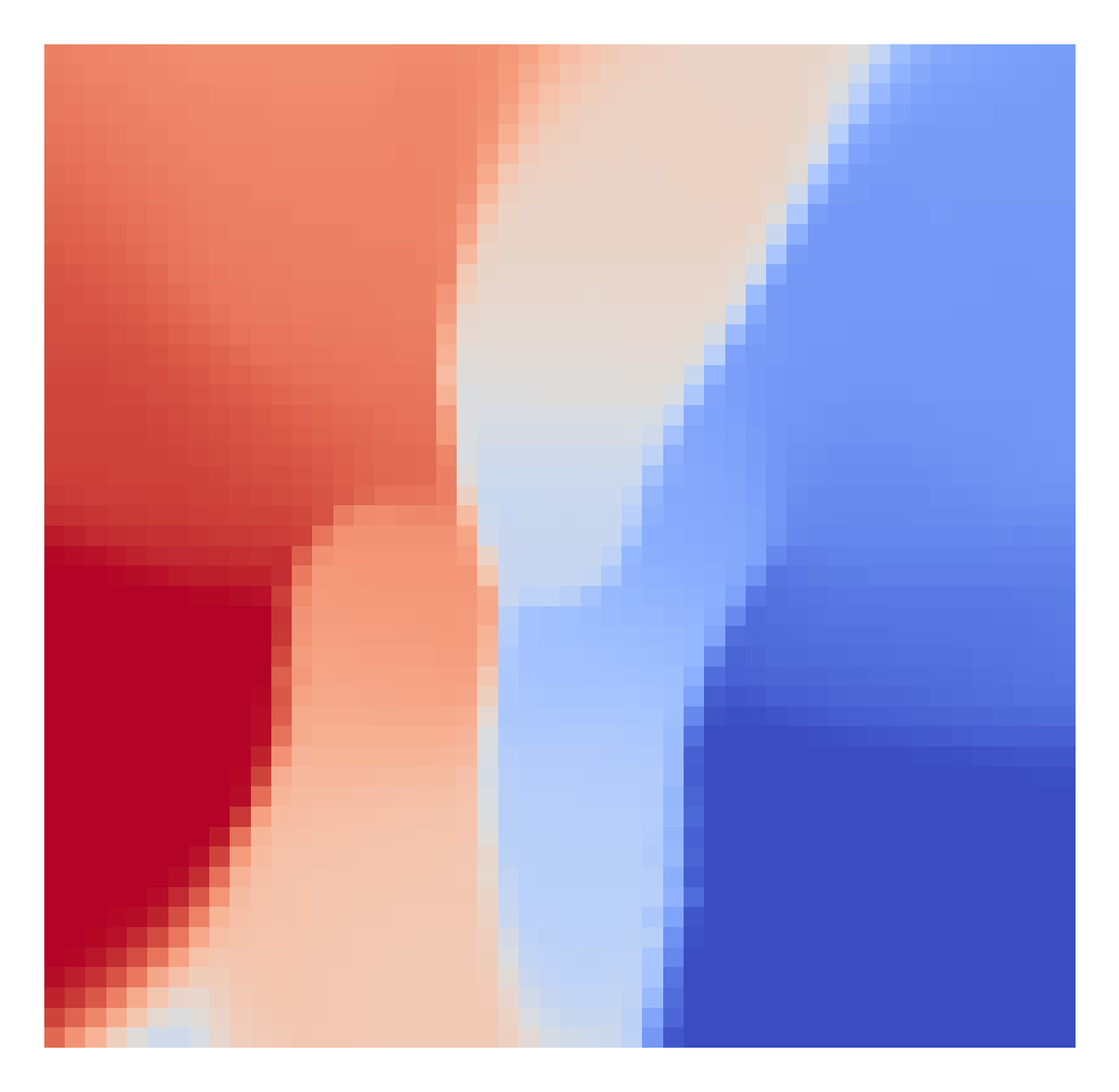}
\caption*{\textbf{MIX2} \textbf{\textit{R-MI}}}
\end{subfigure}
\begin{subfigure}{.11\textwidth}\includegraphics[width=\linewidth]{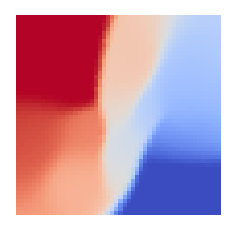}
\caption*{\textbf{MIX2} \textbf{\textit{MINE-S}}}
\end{subfigure}
\begin{subfigure}{.11\textwidth}\includegraphics[width=\linewidth]{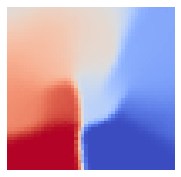}
\caption*{\textbf{MIX2} \textbf{\textit{AE}}}
\end{subfigure}\vspace{5pt}

\begin{subfigure}{.11\textwidth}\includegraphics[width=\linewidth]{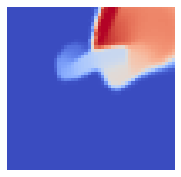}
\caption*{\textbf{MIX3} \textbf{\textit{1C}}}
\end{subfigure}
\begin{subfigure}{.11\textwidth}\includegraphics[width=\linewidth]{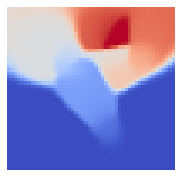}
\caption*{\textbf{MIX3} \textbf{\textit{3C}}}
\end{subfigure}
\begin{subfigure}{.11\textwidth}\includegraphics[width=\linewidth]{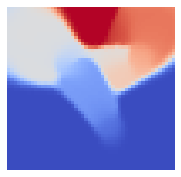}
\caption*{\textbf{MIX3} \textbf{\textit{5C}}}
\end{subfigure}
\begin{subfigure}{.11\textwidth}\includegraphics[width=\linewidth]{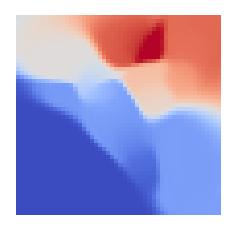}
\caption*{\textbf{MIX3} \textbf{\textit{MAX}}}
\end{subfigure}
\begin{subfigure}{.11\textwidth}\includegraphics[width=\linewidth]{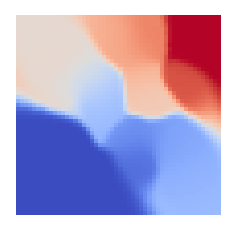}
\caption*{\textbf{MIX3} \textbf{\textit{S-MI}}}
\end{subfigure}
\begin{subfigure}{.11\textwidth}\includegraphics[width=\linewidth]{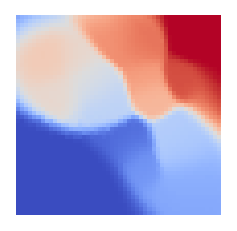}
\caption*{\textbf{MIX3} \textbf{\textit{R-MI}}}
\end{subfigure}
\begin{subfigure}{.11\textwidth}\includegraphics[width=\linewidth]{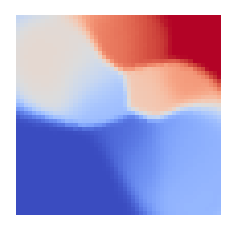}
\caption*{\textbf{MIX3} \textbf{\textit{MINE-S}}}
\end{subfigure}
\begin{subfigure}{.11\textwidth}\includegraphics[width=\linewidth]{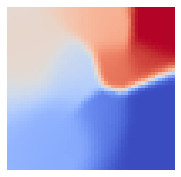}
\caption*{\textbf{MIX3} \textbf{\textit{AE}}}
\end{subfigure}

\caption{Since we purposely conduct the experiments with the encoder projecting $2D$ data into $1D$ features, it is possible to visualize the features after training as heatmaps by using interpolated grid points as the encoder inputs. We visualize the features for training the encoder-mixture-decoder by maximizing the cost for $1$, $3$ and $5$ centers (\textit{\textbf{1C}}, \textit{\textbf{3C}} and \textit{\textbf{5C}}), training the encoder alone by maximizing directly the estimators of the conditional entropy in $L_2$ (\textit{\textbf{MAX}}), Shannon's and R\'enyi's mutual information (\textit{\textbf{S-MI}} and \textit{\textbf{R-MI}}), as well as using MINE (\textit{\textbf{MINE-S}}) and a regular autoencoder (\textit{\textbf{AE}}). We obtain the following conclusions. Using only $1$ center in the decoder is in sufficient to reach the same optimality as using multiple centers. Maximizing the cost and maximizing the bound (\textbf{\textit{MAX}}) reaches the same solution, validating Table~\ref{tab:model_accuracy}. The features learned by maximizing three different bounds, \textbf{\textit{MAX}}, \textbf{\textit{S-MI}} and \textbf{\textit{R-MI}} also have different solutions, validating Table~\ref{table2}. The solutions for maximizing the conditional entropy in $L_2$ may appear sharper and more detailed, but it is difficult to conclude which features are better. To make a fair comparison, the distributions also need to be fairly complex such that the solution is unique without having multiple solutions.}
\label{solution_comparison_17}
\end{figure*}

Table~\ref{table2} shows the results of this comparison. From the table we obtain the following conclusions. First the three methods of maximizing the upper bound \textbf{S-MI}, \textbf{R-MI}, and \textbf{R-CondEntr} all reach a different solution. If we cross-evaluate the three methods with the three scores of them after training, the approach with a maximal score for each is the one that uses that score as the objective to optimize, labeled in red. This implies that the three objectives do arrive at three different optimality condition, with the maximum of each evaluator can only be found when using that score as the objective function to maximize. 

Second, the neural estimator of mutual information (\textbf{MINE-S} and \textbf{MINE-R}) performs similarly to maximizing the mutual information values directly (\textbf{S-MI} and \textbf{R-MI}). This shows the consistency of the estimators. There are two ways to compute the scores for the neural estimator cases \textbf{MINE-S} and \textbf{MINE-R}. Since the cost of these estimators are the variational bound of the mutual information, their optimal costs are the estimator to the mutual information, which is marked in blue in the table. After training, we can also apply the estimators from \textbf{S-MI} and \textbf{R-MI} to estimate the mutual information values, which are also presented in the table. 

In summary, maximizing three objectives, Shannon's and R\'enyi's mutual information, as well as the conditional entropy in $L_2$, reach different solutions. Using a neural estimator with MINE produces similar results to maximizing the objective. We can further illustrate this by visualizing the optimal learned features in Fig.~\ref{solution_comparison_17}.\vspace{7pt}

\noindent \textit{\textbf{Fig.~\ref{solution_comparison_17}: Comparing the solutions of optimal features.}} Our final result is to further demonstrate that using only $1$ center in the decoder only generates a suboptimal solution, and maximizing the cost for the encoder-mixture-decoder matches the solution when maximizing the upper bound, the conditional entropy in $L_2$. We also show a comparison with maximizing Shannon's and R\'enyi's mutual information, MINE and a regular autoencoder. 

Since we set up the experiments as projecting from $2D$ data to $1D$ features, it is possible to visualize the features in the $2D$ space after training, by using the interpolated grid points as inputs to the encoder. Then the features can be visualized as heatmaps. We visualize in Fig.~\ref{solution_comparison_17} results for two datasets, \textbf{MIX1} and \textbf{MIX2}.

We first visualize the features when training an encoder-mixture-decoder, changing the number of centers to be $1$, $3$, and $5$. It can be seen that the solutions for $1$ center (\textbf{MIX1 \textit{1C}} and \textbf{MIX2 \textit{1C}}) differs from $3$ and $5$ centers (\textit{\textbf{3C}} and \textit{\textbf{5C}}), where $\textbf{\textit{1C}}$ appears more blurry. This is consistent with the scores shown in Table~\ref{tab:model_accuracy}, where the bound is not tight at $1$ center and becomes more tight with an increased bound value at $3$ or $5$ centers. 

Then we visualize the features obtained from maximizing the upper bound directly, the results we have shown in Table~\ref{tab:model_accuracy} (\textbf{Max Bound}) and Table~\ref{table2} (\textbf{R-CondEntr}), which are shown as $\textbf{MIX1 \textit{MAX}}$ and $\textbf{MIX2 \textit{MAX}}$ in Fig.~\ref{solution_comparison_17}. It can be seen that they almost match the solutions from the encoder-mixture-decoder when maximizing the cost, when using $3$ or $5$ centers. This further validates the results in Table~\ref{tab:model_accuracy} where we show that eventually the maximal values of the bound found by maximizing the cost will match the values from maximizing the bound. Here we further show that their solutions in terms of the learned features are consistent. 

Next we also visualize the features learned from maximizing Shannon's mutual information and R\'enyi's mutual information, including both the approach of maximizing the scores directly (Eq.~\eqref{sample_estimator_equation}) and using MINE (\textit{\textbf{MINE-S}} for Shannon's mutual information) with a neural estimator apart from the encoder. We also visualize the features learned from optimizing a regular autoencoder by minimizing the mean-squared error. It shows that the solutions of maximizing Shannon's and R\'enyi's mutual information (\textbf{\textit{S-MI}} and \textbf{\textit{R-MI}}) differs from maximizing the conditional entropy in $L_2$ (\textbf{\textit{MAX}}) and optimizing the encoder-mixture-decoder. The solution from optimizing the regular autoencoder (\textbf{\textit{AE}}) is also different. We found that in general, the solutions from maximizing the conditional entropy in $L_2$ produces a sharper and more detailed boundary while the baselines appear less sharp and more generalized. 

In short, we cannot conclude that the features learned by maximizing the conditional entropy in $L_2$ is better than maximizing the various forms of mutual information, but they do reach different solutions. Also, if we maximize the cost with an encoder and a mixture decoder, the solution reaches the maximal value from maximizing the bound only if multiple centers are used. 

Notice that in order to produce this result, the $2D$ distribution needs to sufficiently complex, or there could be multiple optimal solutions for the optimization, making the comparison more difficult. This is why the results are shown on datasets \textbf{MIX1} and \textbf{MIX2}. 

\subsection{Explaining the Difference in the Assumption}

The assumption $p(X,Y)$ we made in this section is slightly different from the assumption over in a regular autoencoder. If we put the assumptions side by side, it becomes
\begin{equation}
\resizebox{.85\linewidth}{!}{
$\begin{aligned}
p(X,Y) &= \frac{1}{N}\sum_{n=1}^N\mathcal{N}(X-X_n;v_X)\cdot\mathcal{N}(Y-Y_n;v_Y), \\
p(X,Y) & = p(X) p(Y|X) \\ &= \frac{1}{N} \sum_{n=1}^N \mathcal{N}(X-X_n;v_X) \cdot \mathcal{N}(Y-\textbf{E}(X); v_p), 
\end{aligned}$}
\label{two_form_of_assumptions}
\end{equation}
The first form is the assumption we made in this section, that the joint of the inputs $X_n$ and the outputs of the encoder $Y_n$ follows a Gaussian mixture distribution; the second form is the regular assumption of an autoencoder, that the encoder parameterized a conditional Gaussian distribution $p(Y|X) = \mathcal{N}(Y-\textbf{E}(X);v_p)$. The difference is in the second Gaussian. The first form assumes that the Gaussian over $Y$ has $Y_n$ as a centers; in the second form, the centers are $\textbf{E}(X)$ with $X$ the variable for the data. Both assume that $p(X)$ is a mixture with centers on data samples. The difference is only on the second term. 

This change in the assumption is needed for the quantitative analysis. In Eq.~\eqref{equation_simple_closed}, when we assume that the decoder follows $q(X|Y) = \mathcal{N}(X-\textbf{D}(Y);v_q)$ and compute the inner product $\iint p(X,Y) q(X|Y) dXdY$, we use the property that the inner product between two Gaussians have a closed form:\vspace{-3pt}
\begin{equation}
\resizebox{1\linewidth}{!}{
$\begin{aligned}
&\iint p(X,Y) q(X|Y) dXdY\\ 
& = \iint \frac{1}{N}\sum_{n=1}^N\mathcal{N}(X-X_n;v_X) \mathcal{N}(Y-Y_n;v_Y) \\ & \;\;\;\;\;\;\;\;\;\;\; \cdot \int p(c)\mathcal{N}(X-\textbf{D}(Y,c);v_q) dc \cdot dXdY\\
& = \frac{1}{N} \sum_{n=1}^N \iint p(c) \mathcal{N}(Y-Y_n;v_Y) \mathcal{N}(X_n - \textbf{D}(Y,c); v_X+v_q) dY dc,
\end{aligned}$}
\end{equation}
which relies on putting $\mathcal{N}(X-X_n;v_X)$ and $\mathcal{N}(X-X_n;v_X)$ and $\mathcal{N}(X-\textbf{D}(Y,c);v_q)$ together and the closed-form solution when integrating over $X$, as $\int\mathcal{N}(X-X_n;v_X)\cdot \mathcal{N}(X-\textbf{D}(Y,c);v_q) dX = \mathcal{N}(X_n - \textbf{D}(Y,c); v_X+v_q)$. This is because the term $\mathcal{N}(Y-Y_n;v_Y)$ only involves $Y$ and the outputs $Y_n$, not $X$. 

But if we choose the second form in Eq.~\eqref{two_form_of_assumptions}, with the form that has $\textbf{E}(X)$ in the Gaussian of $Y$ instead of $X_n$, the inner product becomes\vspace{-3pt}
\begin{equation}
\resizebox{.75\linewidth}{!}{
$\begin{aligned}
&\iint p(X,Y) q(X|Y) dXdY\\ 
& = \iint \frac{1}{N}\sum_{n=1}^N\mathcal{N}(X-X_n;v_X) \mathcal{N}(Y-\textbf{E}(X);v_Y) \\ & \;\;\;\;\;\;\;\;\;\;\; \cdot \int p(c)\mathcal{N}(X-\textbf{D}(Y,c);v_q) dc \cdot dXdY.
\end{aligned}$}
\end{equation}
The closed form can no longer be applied since the three Gaussian terms all contain $X$. Furthermore, the variable $X$ is inside the encoder function $\textbf{E}(X)$ in the second Gaussian function $\mathcal{N}(Y-\textbf{E}(X);v_Y)$ due to the assumption. The inner product of these three Gaussian terms, one involving a function on $\textbf{E}(X)$ no longer has a closed form that can be applied. This makes the computation of this inner product less convenient. Thus for this reason, we choose the first form of the assumption about $p(X,Y)$ in Eq.~\eqref{two_form_of_assumptions}. 

\subsection{Quantitative Analysis for High-Dimensional Real Datasets}

We also have conducted the quantitative experiments on the real image datasets including MNIST and CelebA in Sec.~\ref{section_real_images}. However, we have mentioned that the quantitative procedure requires extra steps compared to the low-dimensional datasets with the exact value of the cost and the bound being computable. In image datasets, we can no longer compute the exact but only the relative values for the bound and the cost. We present the reasons as follows. 

Notice that the density of a Gaussian function, assuming that the the data $X$ is $L$-dimensional, can be written as follows:
\begin{equation}
\resizebox{.9\linewidth}{!}{
$\begin{aligned}
\mathcal{N}(X-X_n;v_X) &= \frac{1}{({2\pi v_X\cdot d_X})^{\frac{d_X}{2}}} \exp \left( -\frac{||X-X_n||_2^2}{2 v_X\cdot d_X}\right).
\end{aligned}$}
\label{gaussian_function}
\end{equation}
To ensure that the value after the exponential does not vanish, we take the mean over the $d_X$ dimensions instead of the sum, where the norm $||X-X_n||_2^2$ is divided by $d_X$. The issue with $X$ in high dimensions is that the constant $\frac{1}{{2\pi v_X\cdot d_X}}$ in front of the exponential will be arbitrarily small. When we optimize the cost functions for the real image data, we ignore this constant and only consider the exponential term, as introduced. In quantitative analysis however, we need to deal with this issue and investigate what the cost approaches when ignoring this exponential term. 

As introduced before, the Cauchy-Schwarz inequality is used to define the bound:
\begin{equation}
\resizebox{.85\linewidth}{!}{
$\begin{aligned}
\frac{ \big ( \iint p(X,Y)q(X|Y) dX dY  \big )^2 }{\iint q^2(X|Y) p(Y) dX dY } \leq \iint p^2(X|Y) p(Y) dX dY.
\end{aligned}$}
\label{inequality_restate}
\end{equation}
We show that the constant in the Gaussian function can be canceled on both sides. As shown before, suppose the inputs to the encoder-mixture-decoder are $X_1, X_2, \cdots, X_N$, for each $X_n$, the outputs of the mixture decoder have $K$ samples $\widehat{X_n'}(1), \widehat{X_n'}(2), \cdots, \widehat{X_n'}(K)$. Here we also use the $\widehat{X_n}$ as samples $X_n$ added with noise, and $\widehat{X_n'}(k)$ as the $k$-th output of the mixture decoder with $\widehat{Y_n}$, the features $Y_n$ with additive noises as inputs. Given each sample ${X_n}$ and a series of outputs $\widehat{X_n'}(k)$ from the mixture decoder, we have shown the inner product can be written as $\langle p(X|Y), q(X|Y)\rangle_{p(Y)} = \frac{1}{NK} \sum_{n=1}^N\sum_{k=1}^K\mathcal{N}(X_n-\widehat{X_n'}(k);v_X+v_q)$, the average of the Gaussian differences between an input sample $X_n$ and a series of decoder outputs $\widehat{X_n'}(1), \widehat{X_n'}(2),\cdots,\widehat{X_n'}(K)$ for this sample. As we will show, we have to assume $v_q = v_X$ for the derivation, thus the inner product becomes $\langle p(X|Y), q(X|Y)\rangle_{p(Y)} = \frac{1}{NK} \sum_{n=1}^N\sum_{k=1}^K\mathcal{N}(X_n-\widehat{X_n'}(k);2v_X)$. Similarly, the norm in the denominator can be written as $||q(X|Y)||_{p(Y)}^2 = \frac{1}{NK^2} \sum_{n=1}^N  \sum_{i=1}^K \sum_{j=1}^K \mathcal{N}(\widehat{X_n'}(i) - \widehat{X_n'}(j);2v_X)$, which is the pairwise Gaussian differences between all decoder outputs from one sample, averaged across all samples. 

Then the left side of the inequality~\eqref{inequality_restate}, the cost, can be rewritten as
\begin{equation}
\resizebox{.9\linewidth}{!}{
$\begin{aligned}
\mathbf{cost} & = \frac{ \big ( \iint p(X,Y)q(X|Y) dX dY  \big )^2 }{\iint q^2(X|Y) p(Y) dX dY } \\
& = \frac{\left(  \frac{1}{NK} \sum_{n=1}^N\sum_{k=1}^K\mathcal{N}(X_n-\widehat{X_n'}(k);2v_X) \right) ^2}{\frac{1}{NK^2} \sum_{n=1}^N  \sum_{i=1}^K \sum_{j=1}^K \mathcal{N}(\widehat{X_n'}(i) - \widehat{X_n'}(j);2v_X)} \\
& = \frac{1}{N} \cdot \frac{1}{({4\pi v_X\cdot d_X})^{\frac{d_X}{2}}} \cdot \frac{ \left( \sum_{n,k} \exp\left( - \frac{||X_n - \widehat{X_n'}(k)||_2^2}{4v_X\cdot d_X}\right) \right)^2 }{\sum_{n,k,q} \exp \left(  - \frac{||\widehat{X_n'}(k) - \widehat{X_n'}(q)||_2^2}{4v_X\cdot d_X} \right) },
\end{aligned}$}
\label{definition_cost}
\end{equation}

\noindent which shows that because the numerator is the square of the inner product, if we assume $v_q = v_X$ to be the same in the assumption, there will be one constant term $\frac{1}{({4\pi v_X\cdot d_X})^{\frac{d_X}{2}}}$ left in the cost, the left side of the inequality~\eqref{inequality_restate}. 

Now if we check the bound, the right side of the inequality, which is irrelevant to $q(X,Y)$ or $v_q$ but only the probability $p(X,Y)$, it follows that
\begin{equation}
\resizebox{.95\linewidth}{!}{
$\begin{aligned}
\mathbf{bound} &= \iint p^2(X|Y) p(Y) dX dY \\
& = \iint p(X,Y) \cdot \frac{p(X,Y)}{p(Y)} dXdY \\
& = \iint \frac{1}{N}\sum_{n=1}^N\mathcal{N}(X-X_n;v_X)\cdot\mathcal{N}(Y-Y_n;v_Y)  \\
&\;\;\;\;\;\;\;\;\; \cdot \frac{ \frac{1}{N}\sum_{n=1}^N\mathcal{N}(X-X_n;v_X)\cdot\mathcal{N}(Y-Y_n;v_Y)}{\frac{1}{N}\sum_{n=1}^N\mathcal{N}(Y-Y_n;v_Y)} dXdY. 
\end{aligned}$}
\label{bound_term}
\end{equation}
The difference of this derivation with the derivation for simple datasets shown in Eq.~\eqref{eq_33_norm_2} is that instead of using using $\widehat{X_n}$ and $\widehat{Y_n}$, the samples $X_n$ and features $Y_n$ with additive noises, to directly estimate the double integral over $p(X,Y)$, here we still use the mixture density form of $p(X,Y)$ first. Then notice that there are two Gaussian terms in the double integral, one from $p(X,Y)$, one from $\frac{p(X,Y)}{p(Y)}$, by the special assumption we made for $p(X,Y)$, which means that we can use the closed form of the inner products between Gaussians to further simplify the term. We can not do the same to Gaussians over $Y$, because there are three terms in total involving $Y$, in the numerator and the denominator, so we still need to use $\widehat{Y_n}$ for the empirical estimations as we have shown before. This turns the term Eq.~\eqref{bound_term} into 
\begin{equation}
\resizebox{1\linewidth}{!}{
$\begin{aligned}
\mathbf{bound} &= \int  \frac{1}{N} \sum_{m=1}^N \mathcal{N}(Y-Y_m;v_Y) \\
&\;\;\;\;\;\cdot \frac{\sum_{n=1}^N \mathcal{N}(X_m-X_n;2v_X)\cdot\mathcal{N}(Y-Y_n;v_Y)}{\sum_{n=1}^N \mathcal{N}(Y-Y_n;v_Y)} dY \\
 \approx &\frac{1}{N} \sum_{m=1}^N \frac{\sum_{n=1}^N \mathcal{N}(X_m-X_n;2v_X)\cdot\mathcal{N}(Y-Y_n;v_Y)}{\sum_{n=1}^N \mathcal{N}(\widehat{Y_m}-Y_n;v_Y)} \\
 = &\frac{1}{N} \cdot \frac{1}{({4\pi v_X\cdot d_X})^{\frac{d_X}{2}}} \cdot \sum_m \frac{\sum_n \exp(-\frac{||X_m-X_n||_2^2}{4v_X\cdot d_X}) \cdot \mathcal{N}({\widehat{Y_m} - Y_n};v_Y) }{\sum_n \mathcal{N}(\widehat{Y_m} - Y_n;v_Y)}.
\end{aligned}$}
\label{definition_bound}
\end{equation}
This shows that from the Gaussian differences over $X_m$ and $X_n$, after applying the closed form to the inner product between $\mathcal{N}(X-X_m;v_X)$ and $\mathcal{N}(X-X_n;v_X)$, one constant term $\frac{1}{({4\pi v_X\cdot d_X})^{\frac{d_X}{2}}}$ can also be factorized out of the Gaussian term. Also, since both the numerator and the denominator contain the Gaussians over $Y$, we also only need to use the exponential terms for them while the constant terms can be ignored. 


Now comparing the definition of the bound~\eqref{definition_bound} and the definition of the cost~\eqref{definition_cost}, both have a constant $\frac{1}{({4\pi v_X\cdot d_X})^{\frac{d_X}{2}}}$. Therefore, this constant is canceled on both sides, and the tightness of the bound is irrelevant to this constant. Let us redefine
\begin{equation}
\resizebox{.95\linewidth}{!}{
$\begin{aligned}
\mathbf{cost}_{new} & = \frac{1}{N} \cdot \frac{ \left( \sum_{n,k} \exp\left( - \frac{||X_n - \widehat{X_n'}(k)||_2^2}{4\pi v_X\cdot d_X}\right) \right)^2 }{\sum_{n,k,q} \exp \left(  - \frac{||\widehat{X_n'}(k) - \widehat{X_n'}(q)||_2^2}{4\pi v_X\cdot d_X} \right) },\\
\mathbf{bound}_{new} & = \frac{1}{N} \cdot \sum_m \frac{\sum_n \exp(-\frac{||X_m-X_n||_2^2}{4\pi v_X\cdot d_X}) \cdot \mathcal{N}({\widehat{Y_m} - Y_n};v) }{\sum_n \mathcal{N}(\widehat{Y_m} - Y_n;v)},\\
& \;\;\;\;\;\;\;\;\;\;\;\;\;\mathbf{cost}_{new} \leq \mathbf{bound}_{new}.
\end{aligned}$}
\label{bound_term}
\end{equation}
Since in $\mathbf{bound}_{new}$, the constant term in the Gaussian $\mathcal{N}(\widehat{Y_m} - Y_n;v)$ can also be canceled, and the the Gaussian terms can be substituted by $\exp(-\frac{||\widehat{Y_m} - Y_n||_2^2}{4\pi v_Y\cdot d_Y})$. All the computations involve the exponential terms, irrelevant to the constant terms for the Gaussian. We also found that the two new terms $\mathbf{bound}_{new}$ and $\mathbf{cost}_{new}$ are bounded from $0$ to $1$. Thus this quantitative analysis can be extended to high dimensions, to investigate the tightness of the bound and the optimal value of the bound, as performed in Sec~\ref{section_real_images}.
\newpage

\appendices 

\section{Algorithms}

\noindent \textbf{Algorithm Details.} To further clarify the pipeline of our proposed algorithms, we provide a detailed descriptions of them in the following, including Algorithm~\ref{alg1} for the MDNs with nuclear norms (Sec.~\ref{nuclear_norm_section}), Algorithm~\ref{alg2} for the MDNs with the inner product and norms of densities (Sec.~\ref{inner_product_norm}), Algorithm~\ref{alg3} for the encoder-mixturer-decoder (Sec~\ref{encoder_mixture_decoder}), and Algorithm~\ref{alg4} for the quantitative analysis we have accomplished on simple and real datasets (Sec.~\ref{quantitative_analysiss}).

\begin{algorithm}[H]
\small
\caption{MDNs w/ Nuclear Norms}
\begin{algorithmic}[1]\vspace{3pt}
\Statex \textbf{At each iteration do:}
\State Sample $c_1,c_2,\cdots,c_K$ from a noise distribution;
\State Pass the noise samples through a decoder network to produce $K$ generated samples $X_1',X_2',\cdots,X_K'$;
\State Sample $N$ data samples $X_1,X_2,\cdots,X_N$;
\State Pick a variance $v$, compute the Gaussian cross Gram matrix between them; 
\State Compute the SVD of the cross Gram matrix. Maximize the sum of its singular values. 
\end{algorithmic}
\label{alg1}
\end{algorithm}

\begin{algorithm}[H]
\small
\caption{MDNs w/ Inner Products and Norms}
\begin{algorithmic}[1]\vspace{3pt}
\Statex \textbf{At each iteration do:}
\State Sample $c_1,c_2,\cdots,c_K$ from a noise distribution;
\State Pass the noise samples through a decoder network to produce $K$ generated samples $X_1',X_2',\cdots,X_K'$;
\State Compute the Gaussian differences between every pair of the generated samples $\mathcal{N}(X_i'-X_j';2v)$;
\State Sample $N$ data samples $X_1,X_2,\cdots,X_N$;
\State Compute the Gaussian differences between every generated sample and the data sample $\mathcal{N}(X_i-X_j';2v)$. 
\State Construct the inner product $\langle p, q\rangle = \frac{1}{NK}\sum_{i,j}\mathcal{N}(X_i - X_j';2v)$ and the norm $||q||_2^2 = \frac{1}{N^2}\sum_{i,j}\mathcal{N}(X_i'-X_j';2v)$. 
\State Maximize the cost $\frac{\langle p, q\rangle^2}{||q||_2^2}$. 
\end{algorithmic}
\label{alg2}
\end{algorithm}\vspace{-5pt}

\begin{algorithm}[H]
\small
\caption{Encoder-Mixture-Decoder}
\begin{algorithmic}[1]\vspace{3pt}
\Statex \textbf{At each iteration do:}
\State Sample $X_1,X_2,\cdots,X_N$ from the data distribution;
\State Pass $X_1,X_2,\cdots,X_N$ through an encoder and obtain features $Y_1,Y_2,\cdots,Y_N$;
\State Add additive Gaussian noise to $Y$ with $\widehat{Y_n} = Y_n + \sqrt{v_Y}\cdot z_n$, obtain $\widehat{Y_1}, \widehat{Y_2},\cdots,\widehat{Y_N}$;
\State For each feature $\widehat{Y_n}$, sample a series of $K$ random noises $c_n(1),c_n(2),\cdots,c_n(K)$; Concatenate each $Y_n$ with $c_n(k)$ as inputs to the mixture decoder, so for each $Y_n$, the mixture decoder will produce a series of samples $X_n'(1), X_n'(2),\cdots,X_n'(K)$ for various $c_n(k)$. Another approach is to have the network produce multiple centers, instead of setting $c_n(k)$ in the input, but we found the former to work better. 
\State Compute the Gaussian differences between each sample and its reconstructions $\mathcal{N}(X_n-\widehat{X_n}(k);2v)$. Average them over $k$ and $n$ to estimate the inner product $\langle p(X|Y), q(X|Y) \rangle_{p(Y)}$. 
\State Compute the Gaussian differences between all reconstructions $\mathcal{N}(\widehat{X_n}(i) - \widehat{X_n}(j);2v)$ for this sample $X_n$. Average them over $i$, $j$, $n$ to estimate the norm $||q(X|Y)||_{p(Y)}^2$.
\State Maximize the cost $\frac{\left( \langle p(X|Y), q(X|Y) \rangle_{p(Y)} \right)^2}{||q(X|Y)||_{p(Y)}^2}$. 
\end{algorithmic}
\label{alg3}
\end{algorithm}

\begin{algorithm}[H]
\small
\caption{Quantitative Analysis - Maximizing the Upper Bound Directly for the Encoder}
\begin{algorithmic}[1]\vspace{3pt}
\Statex \textbf{At each iteration do:}
\State Sample $X_1,X_2,\cdots,X_N$ from the data distribution;
\State Pass $X_1,X_2,\cdots,X_N$ through an encoder and obtain features $Y_1,Y_2,\cdots,Y_N$;
\State Add additive Gaussian noises to each $X_n$ and $Y_n$ with $\widehat{X_n} = X_n+ \sqrt{v_X} \cdot z_n$ and $\widehat{Y_n} = Y_n + \sqrt{v_Y} \cdot s_n$. 
\State Compute the conditional entropy in $L_2$ $||p(X|Y)||_{p(Y)}^2 = \frac{1}{N}\sum_{m=1}^N \frac{\sum_{n=1}^N\mathcal{N}(\widehat{X_m}-X_n;v_X)\mathcal{N}(\widehat{Y_m}-Y_n;v_Y)}{\sum_{n=1}^N \mathcal{N}(\widehat{Y_m}-Y_n;v_Y)}$, or other quantities such as Shannon's or R\'enyi's mutual information (Eq.~\eqref{sample_estimator_equation}) to maximize, by optimizing the encoder network. 
\end{algorithmic}
\label{alg4}
\end{algorithm}

All the Gaussian differences in the algorithms can be substituted by the exponential term, without the normalizing constant, to ensure the numerical stability, unless the weights and variances are also parameterized to be trainable, following Appendix.~\ref{no_one_cares}. The mean over the data dimension is also preferred than the summation inside the exponential.\vspace{5pt}

\noindent\textbf{Baselines.} In Sec.~\ref{quantitative_analysis_simple}, we have extensively described the procedure of applying MINE. After passing samples $X_1,X_2,\cdots,X_N$ through an encoder and produce features $Y_1,Y_2,\cdots,Y_N$, we add Gaussian noises to them and construct $\widehat{X_n} = X_n+ \sqrt{v_X} \cdot z_n$ and $\widehat{Y_n} = Y_n + \sqrt{v_Y} \cdot s_n$. Such $\widehat{X_n}$, $\widehat{Y_n}$ follow the joint distribution $p(X,Y) = \frac{1}{N}\sum_{n=1}^N\mathcal{N}(X-X_n;v_X)\cdot\mathcal{N}(Y-Y_n;v_Y)$. We then shuffle them, such that the joint after shuffling follows the product of the marginals $p(X)p(Y)$. Then we use the $\widehat{X_n}$, $\widehat{Y_n}$ and the shuffled samples to optimize the costs Eq.~\eqref{eq_36} with a neural network $f_\theta(X,Y)$. Mind that for Shannon's mutual information (MINE-S), the neural network does not have a final activation, but ends with a linear projection. For R\'enyi's mutual information (MINE-R), we found that it is the most stable when final activation is chosen to be $\mathbf{sigmoid}(X)+0.1$, a sigmoid function with a small added constant. 

The implementations of KICA and HSIC for evaluations are as follows. First, individual Gram matrices $\mathbf{R}_{X}$ and $\mathbf{R}_{Y}$ are estimated for given Gaussian kernel $\mathcal{N}(X_i-X_j;2v)$ with $v$ the hyperparameters for the variance. Then we construct the normalization matrix $\mathbf{N}_{i,j}$ and normalize the two Gram matrices as $\widehat{\mathbf{R}_{X}} = \mathbf{N}\mathbf{R}_{X}\mathbf{N}$ and $\widehat{\mathbf{R}_{Y}} = \mathbf{N}\mathbf{R}_{Y}\mathbf{N}$. For KICA-KGV, matrices $\mathbf{A}$ and $\mathbf{B}$ are constructed:
\begin{equation}
\resizebox{1\linewidth}{!}{
$\begin{gathered}
\mathbf{A} = \begin{bmatrix}\mathbf{A}_1 & 0\\ 0 & \mathbf{A}_2 \end{bmatrix}, \quad \mathbf{A}_1 = \widehat{\mathbf{R}_{X}}\,\widehat{\mathbf{R}_{Y}}, \quad \mathbf{A}_2 = \widehat{\mathbf{R}_{Y}}\,\widehat{\mathbf{R}_{X}}, \\
\mathbf{B} = \begin{bmatrix}\mathbf{B}_1 & 0\\ 0 & \mathbf{B}_2 \end{bmatrix}, \quad \mathbf{B}_1 = (\widehat{\mathbf{R}_{X}}+\epsilon \mathbf{I}) (\widehat{\mathbf{R}_{X}}+\epsilon \mathbf{I}), \quad \mathbf{B}_2 = (\widehat{\mathbf{R}_{Y}}+\epsilon \mathbf{I}) (\widehat{\mathbf{R}_{Y}}+\epsilon \mathbf{I}),
\end{gathered}$}
\end{equation}
\noindent Then, solve the generalized eigenvalue problem for KICA $\mathbf{A} \mathbf{v}_i = \sigma_i \mathbf{B} \mathbf{v}_i$, where $i=1, \cdots, 2N$. This generalized eigenproblem generates $2N$ eigenvalues that are symmetric over the real line. Only $N$ positive eigenvalues of them are used to compute the measure, obtaining KICA's Kernel Generalized Variance (KGV) measure.
For HSIC, we construct matrix $\mathbf{C}$:
\begin{equation}
\mathbf{C} = \mathbf{B}_1^{-\frac{1}{2}} \mathbf{A}_1 \mathbf{B_2}^{-\frac{1}{2}},
\end{equation}
\noindent and solve the eigenvalue problem $\mathbf{C}\mathbf{v}_i = \sigma_i \mathbf{v}_i$, where $i=1,\cdots, N$. Compute $T_{HSIC} = Trace(\mathbf{C})$. HSIC's measure is named the Normalized Cross-Covariance Operator (NOCCO). Hyperparameters are set as kernel size $v=0.001$ and regularization constant $\epsilon=1$.

\section{Training the Variances and Weights}
\label{no_one_cares}

It is possible to train the variances and the weights of the learned mixture density, not just keeping them as fixed constants such as $v=0.001$ or equal weights. We can assign each sample $X_n$ with its own variance and weight in MDNs, or assign for each sample, each output of the multiple outputs of the mixture decoder with its own variance and weight in the encoder-mixture-decoder. In fact, this is required to obtain the quantitative results we show for simple toy datasets, to match the exact value of the cost with the maximal value of the upper bound, the conditional entropy in $L_2$. But the trainable variances and weights could be unnecessary for the real image datasets as in high dimensions, the impact of the variances and weights could be trivial. 

For MDNs, when pass the random noises $c_1,c_2,\cdots,c_K$ through the decoder and produce the generated samples $X_1',X_2',\cdots,X_K'$, not only we make the encoder produce the mean values of the Gaussian components $X_k':= {m}(c_k)$, we also produce the the variances ${v}(c_k)$ and the weights ${w}(c_k)$ for this noise $c_k$ and its reconstructions $X_k'$. Suppose the data samples have a fixed variance $v$, as we have assumed before. The density is defined by
\begin{equation}
\resizebox{1\linewidth}{!}{
$\begin{aligned}
q(X) &= \int p(c)w(c)\mathcal{N}(X-m(c);v(c)) dc \\
&\approx \frac{1}{K}\sum_{k=1}^K w(c_k) \mathcal{N}(X-m(c_k);v(c_k)), \\
&\hspace{-30pt}\int q^2(X) dX = \frac{1}{K^2}\sum_{i=1}^K  \sum_{j=1}^K w(c_i)w(c_j)\mathcal{N}(m(c_i)-m(c_j);v(c_i)+v(c_j)), \\
&\hspace{-30pt}\int p(X)q(X) dX = \frac{1}{NK}\sum_{i=1}^N \sum_{j=1}^K w(c_j) \mathcal{N}(X_i-m(c_j);v(c_j)).
\end{aligned}$}
\end{equation}
Then we maximize the cost $\frac{\left( \int p(X) q(X)dX \right)^2}{\int q^2(X) dX}$ by optimizing the mean $m(c)$, the variance $v(c)$, the weights $w(c)$ by gradient ascent. Suppose the $X$ is in dimension $d_X$, then the means and the variances can be vectors or the covariance matrix also in dimension $d_X$, instead of being just a scalar constant for each noise sample. In this case, each Gaussian difference is rewritten as 
\begin{equation}
\resizebox{1\linewidth}{!}{
$\begin{aligned}
& \mathcal{N}(X-m(c_k);v(c_k))  = \prod_{l=1}^{d_X} \mathcal{N}(X^{(l)} - m^{(l)}(c_k); v^{(l)}(c_k))\\
& =\frac{1}{(2\pi \cdot \prod_{l=1}^{d_X} v^{(l)}(c_k)\cdot d_X)^{\frac{d_X}2{}}} \exp \left(\frac{1}{d_X}\sum_{l=1}^{d_X}-\frac{(X^{(l)} - m^{(l)}(c_k))^2}{2\cdot v^{(l)}(c_k)} \right). 
\end{aligned}$}
\label{equation_difference1}
\end{equation}
We use $X^{(l)}$, $m^{(l)}$, $v^{(l)}$ for the $l$-th of the $d_X$ dimensions for the data, mean values and variances. The joint density is modeled by the product of densities over $l$. This becomes a sum over the data dimension and a product of the variances in the normalizing constant. We change the sum to the mean in the exponential to ensure the numerical stability, which also adds a $d_X$ term to the normalizing constant. The tricky part is the product term $\prod_{l=1}^{d_X} v^{(l)}$, which can vanish in high dimensions. Thus this applies to simple datasets, but in real image datasets it is more suitable to keep the variance a constant such as $v=0.001$. 

The norm term in this parameterization of a vector of weights and variances has the form 
\begin{equation}
\resizebox{0.85\linewidth}{!}{
$\begin{aligned}
& w(c_i)w(c_j)\mathcal{N}(m(c_i)-m(c_j);v(c_i)+v(c_j))  \\
& = w(c_i)w(c_j) \prod_{l=1}^{d_X} \mathcal{N}(m^{(l)}(c_i) - m^{(l)}(c_j); v^{(l)}(c_i) + v^{(l)}(c_j)) \\
& = w(c_i)w(c_j) \cdot \frac{1}{\left(2\pi \cdot \prod_{l=1}^{d_X} (v^{(l)}(c_i) + v^{(l)}(c_j))\cdot d_X\right)^{\frac{d_X}2{}}}\\
& \;\;\;\;\;\;\;\;\;\;\;\;\; \cdot \exp \left(\frac{1}{d_X}\sum_{l=1}^{d_X}-\frac{(m^{(l)}(c_i) - m^{(l)}(c_j))^2}{2\cdot (v^{(l)}(c_i) + v^{(l)}(c_j))} \right),
\end{aligned}$}
\label{equation_difference2}
\end{equation}
which requires the summation of the variances from a sample pair, similar to the $L_2$ differences of a sample pair, as well as the product of two weights $w(c_i)$ and $w(c_j)$. We have introduced before an approach in Sec.~\ref{saving_computing_memory} that uses the matrix operation to save the computational memory when computing the $L_2$ differences of the mean values, which can also applied to the variances and the weights. The advantage is that in this case, the density $q(X)$ can potentially approximate the exact $p(X)$ and produce the density value for simple toy datasets.\vspace{5pt}

It is similar for the encoder-mixture-decoder. If we make the variances and weights trainable for the mixture decoder $q(X|Y)$, it can potentially match the exact $p(X|Y)$ instead of having a bias due to the fixed variances and equal weights. Similarly, the mixture decoder will produce not just a series of means, but also the variances and weights. The norm and the inner product will follow
\begin{equation}
\resizebox{1\linewidth}{!}{
$\begin{aligned}
&q(X|Y)  = \int p(c)w(Y,c)\mathcal{N}(X- m(Y,c);v(Y,c)) dc,\\
&\iint p(X,Y) q(X|Y) dXdY  \\ 
&= \frac{1}{NK} \sum_{n,k} w(Y_n,c_n(k)) \mathcal{N}\Big(X_n - m(Y_n, c_n(k)); v_X + v(Y_n, c_n(k))\Big),\\
&\iint q^2(X|Y)  p(Y) dXdY \\
&= \frac{1}{N K^2} \sum_{n,i,j}  w(Y_n,c_n(i)) w(Y_n,c_n(j)) \\
& \;\;\;\;\;\;\;\;\cdot \mathcal{N}\Big( m(Y_n, c_n(i)) - m(Y_n, c_n(j)); v(Y_n, c_n(i)) + v(Y_n, c_n(j)) \Big). 
\end{aligned}$}
\end{equation}
The computation of the the Gaussian terms is the same as in Eq.~\eqref{equation_difference1} and Eq.~\eqref{equation_difference2}, thus we no longer reiterate the derivation again here. The only difference is that now the means, variances and weights no longer depends on $c_n$ alone, but $c_n$ and $Y_n$ both. The assumption of the data variance $v_X$ is still needed to ensure the numerical stability and for quantitative analysis. 

In fact, the results we have shown in Sec.~\eqref{quantitative_analysis_simple} and Sec~\eqref{quantitative_analysis_simple_results} are based on the trainable variances and weights, such that the $q(X|Y)$ can match $p(X|Y)$ exactly and we obtain the exact match of the cost and the bound for the quantification. 

\section{Theoretical Details for the Nuclear Norm Decomposition}

One question to investigate is what the optimal solution for maximizing the nuclear norm, the sum of singular values of the Gaussian cross Gram matrix will reach, and why it can be used to learn a generative model that matches the model density $q$ with $p$. In this section, we provide a more detailed illustration of it. More details of derivations can also be found in the paper~\cite{lll}.\vspace{7pt}

\noindent \textbf{The simplified explanation.} A simplified justification that maximizing the singular values of the Gaussian cross Gram matrix $\mathbf{K}$ (Eq.~\eqref{gram_matrices}), where each element of $\mathbf{K}$ is a Gaussian difference between a sample $X_i$ and a generated sample $X_j'$ mapped from the random noise $c_j$, the output of a decoder. If in the extreme case that the data samples and the generated samples match one-by-one, with $X_n= X_n'$, then the cross-correlation matrix will become an auto-correlation matrix, thus Hermitian. In this case, the sum of its singular values is the sum of its eigenvalues, which is also the matrix trace. The diagonal elements of a Hermitian Gaussian Gram matrix are all constants $\mathcal{N}(X_n-X_n) = \mathcal{N}(0)$, so the trace is $N\cdot \mathcal{N}(0)$. We found that this trace, a constant $N\cdot \mathcal{N}(0)$, is the maximal value that this cost, the sum of singular values can reach. Although this optimal solution that $X_n=X_n'$ match exactly might not be the unique solution such that the cost is the maximal, the nuclear norm of $\mathbf{K}$ will be smaller than this constant value when $p(X)$ and $q(X)$ are far apart.\vspace{5pt}

\noindent \textbf{The detailed analysis.} We provide the detailed analysis that in the optimal condition of maximizing the nuclear norm, the sum of singular values of a Gaussian cross Gram matrix, it is equivalent to finding an orthonormal decomposition of a kernel function $\mathcal{K}(X,X')$ with basis functions orthonormal to the probability distributions $p(X)$ and $q(X)$. We illustrate this statement in details with the following lemma.\vspace{5pt}

\begin{lemma}
With a continuous kernel function $\mathcal{K}(X,X')$ and density functions $p(X)$, $q(X)$, we propose two decompositions based on Mercer's theorem: decomposing $\sqrt{p(X)}\mathcal{K}(X,X')\sqrt{q(X')}$ with orthonormal bases w.r.t. Lebesgue measure $\mu$ (Eq.~\eqref{decomposition1}); and decomposing $\mathcal{K}(X,X')$ with bases orthonormal w.r.t. probability measures $p$, $q$ (Eq.~\eqref{decomposition2}). The two decompositions share the same singular values. Their discrete equivalents as the Hermitian matrix $\mathbf{K}_{XX'}$ and discrete densities are shown in Eq.~\eqref{discrete1} and Eq.~\eqref{discrete2}.\vspace{-7pt}
\begin{equation}
\resizebox{.8\linewidth}{!}{
$\begin{gathered}
    \sqrt{p(X)}\mathcal{K}(X,X')\sqrt{q(X')} = \sum_{k=1}^K \sigma_k {\phi_k}(X){\psi_k}(X'),\vspace{-3pt}\\ \vspace{-3pt}
\int {\phi_i}{\phi_j} dX = \int {\psi_i}{\psi_j} dX' = \begin{cases} 1, \;i = j& \\ 0, \; i\neq j \end{cases}\hspace{-9pt}. 
\end{gathered}$}
\label{decomposition1}
\end{equation}\vspace{-5pt}
\begin{equation}
\resizebox{.8\linewidth}{!}{
$\begin{gathered}
\mathcal{K}(X,X') = \sum_{k=1}^K \sigma_k \widehat{\phi_k}(X) \widehat{\psi_k}(X'),\\
\int \widehat{\phi_i} \widehat{\phi_j} p(X) dX = \int \widehat{\psi_i} \widehat{\psi_j} q(X') dX' = \begin{cases} 1, \;i = j& \\ 0, \; i\neq j \end{cases}\hspace{-7pt}. 
\end{gathered}$}
\label{decomposition2}
\end{equation}
\begin{equation}
\resizebox{1\linewidth}{!}{
$\begin{gathered}
{diag}(\sqrt{\mathbf{P}_X}) \, \mathbf{K}_{X,X'} \, diag(\sqrt{\mathbf{Q}_{X
'}}) = \mathbf{U}\mathbf{S}\mathbf{V},\;\;
\mathbf{U}\mathbf{U}^\intercal  = \mathbf{I}, \;\; \mathbf{V}\mathbf{V}^\intercal = \mathbf{I}. 
\end{gathered}$}
\label{discrete1}
\end{equation}\vspace{-10pt}
\begin{equation}
\resizebox{1\linewidth}{!}{
$\begin{gathered}
\mathbf{K}_{X,X'} = \mathbf{U}\mathbf{S}\mathbf{V}, \, \mathbf{U} diag(\mathbf{P}_X)\, \mathbf{U}^\intercal  = \mathbf{I}, \, \mathbf{V} diag(\mathbf{Q}_X) \mathbf{V}^\intercal = \mathbf{I}. 
\end{gathered}$}
\label{discrete2}
\end{equation}
\noindent Since the matrix $\mathbf{K}_{XX'}$ is Hermitian, we can decompose it with $\mathbf{K}_{XX'} = \mathbf{Q}_{\mathbf{N}}\mathbf{\Lambda}_{\mathbf{N}}\mathbf{Q}_{\mathbf{N}}$. Define $\mathbf{A}:= diag(\sqrt{\mathbf{P}_X}) \mathbf{Q}_{\mathbf{N}} \mathbf{\Lambda}_{\mathbf{N}}^{\frac{1}{2}}$ and $\mathbf{B} :=  diag(\sqrt{\mathbf{Q}_{X}}) \mathbf{Q}_{\mathbf{N}} \mathbf{\Lambda}_{\mathbf{N}}^{\frac{1}{2}}$, applying the inequality of the nuclear norm:
\begin{equation}
\resizebox{.8\linewidth}{!}{
$\begin{aligned}
& \hspace{-6pt} ||\mathbf{A}\mathbf{B}^\intercal||_*  \leq \sqrt{||\mathbf{A}\mathbf{A}^\intercal||_*} \cdot \sqrt{||\mathbf{B}\mathbf{B}^\intercal||_*}, \\
||\mathbf{A}\mathbf{A}^\intercal||_* &=  ||{diag}(\sqrt{\mathbf{P}_X}) \mathbf{K}_{XX'} diag(\sqrt{\mathbf{P}_{X}})||_* \\
&= Trace({diag}(\sqrt{\mathbf{P}_X}) \mathbf{K}_{XX'} diag(\sqrt{\mathbf{P}_{X}})) \\
&= \mathcal{N}(0) = ||\mathbf{B}\mathbf{B}^\intercal||_*. 
\end{aligned}$}
\label{bound_AB}
\end{equation}
\noindent That is, the nuclear norm of the defined matrix ${diag}(\sqrt{\mathbf{P}_X}) \mathbf{K}_{XX'} diag(\sqrt{\mathbf{Q}_{X}})$ is upper bounded by the constant $\mathcal{N}(0)$. The bound is tight when $\mathbf{A} = \mathbf{B}$  for positive eigenvalues of $\mathbf{K}_{XX'}$, i.e., when $diag(\sqrt{\mathbf{P}_X}) \mathbf{Q}_{\mathbf{N}} =  diag(\sqrt{\mathbf{Q}_{X}}) \mathbf{Q}_{\mathbf{N}}$.
\label{property8}
\end{lemma}\vspace{5pt}

\noindent Lemma~\ref{property8} shows the decomposition of a continuous function $\sqrt{p(X)}\mathcal{K}(X,X')\sqrt{q(X')}$ using Mercer's theorem (Eq.~\eqref{decomposition1}). Suppose $\phi_i$ and $\psi_i$ are the bases of this function, by applying variational trick $\widehat{\phi_i} = \phi_i / \sqrt{p(X)}$ and $\widehat{\psi_i} = \psi_i / \sqrt{q(X)}$, we obtain basis functions $\widehat{\phi_i}$ and $\widehat{\psi_i}$, orthonormal to the probability measures $p(X)$ and $q(X)$, which decompose the kernel $\mathcal{K}(X,X')$ (Eq.~\eqref{decomposition2}).

In summary, the variational trick transforms the decomposition of $\sqrt{p(X)}\mathcal{K}(X,X')\sqrt{q(X')}$ (Eq.~\eqref{decomposition1}) into decomposing $\mathcal{K}(X,X')$ (Eq.~\eqref{decomposition2}), by changing the measures from the Lebesgue measure to probability measures. This decomposition of $\mathcal{K}(X,X')$ with bases orthonormal to probability measures is the SVD of the Gaussian cross-correlation matrix $\mathbf{K}$ in the cost function we defined in Eq.~\eqref{gram_matrices} of the main section. We summarize this lemma in the following:\vspace{1pt}
\begin{enumerate}[leftmargin=*]
\item Given samples $X_i$ following $p(X)$ and generated samples $X_j'$ following $q(X)$, if we perform the singular value decomposition of the Gaussian cross Gram matrix between them, it is equivalent to decomposing the continuous kernel $\mathcal{K}(X,X')$ with basis functions orthonormal to the probability measures $p(X)$ and $q(X)$, further equivalent to decomposing the continuous function $\sqrt{p(X)}\mathcal{K}(X,X')\sqrt{q(X')}$ with basis functions orthonormal to the Lebesgue measure. The equivalence is proved by a variational trick.\vspace{5pt}
\item Then we can discuss the optimal condition of maximizing the cost, the optimal $q(X)$ such that the sum of singular values is the maximal.\vspace{3pt}

By computing the SVD to the kernel function $\mathcal{K}(X,X')$ and applying the inequality of the nuclear norm in Eq.~\eqref{bound_AB}, we show that the optimal condition is $diag(\sqrt{\mathbf{P}_X}) \mathbf{Q}_{\mathbf{N}} =  diag(\sqrt{\mathbf{Q}_{X}}) \mathbf{Q}_{\mathbf{N}}$, when the half power of the densities, projected to the bases of the Gaussian functions. Here we apply the trick of using the discrete equivalents of $p(X)$, $q(X)$, $\mathcal{K}(X,X')$ for the derivation, denoted as the vector $P_X$, $Q_X$ and the Hermitian matrix $\mathbf{K}_{X,{X'}}$, which can be extended to the continuous case simply.\vspace{7pt}
\end{enumerate}

\noindent \textbf{The decomposition to the density ratio $\frac{p(X)}{q(X)}$ is not convenient, but the decomposition to the continuous function $\sqrt{p(X)}\mathcal{N}(X,X')\sqrt{q(X')}$ is possible.} The inspiration also comes from the following. A conventional $f$-divergence~\cite{jordan1, jordan2} is a functional on the density ratio $\frac{p(X)}{q(X)}$. Suppose the densities are discrete. This ratio is a vector that does not have a standard convenient orthonormal decomposition. But if we define a quantity like ${diag}(\sqrt{\mathbf{P}_X}) \mathbf{K}_{XX'} diag(\sqrt{\mathbf{Q}_{X}})$, then the decomposition becomes possible. If we pick $\mathbf{K}_{X,X'}$ to be an identity matrix, correspondingly the identify function $\mathcal{K}(X,X') = \mathbbm{1}\{X=X'\}$, then the matrix to decompose becomes ${diag}(\sqrt{\mathbf{P}_X}) \;diag(\sqrt{\mathbf{Q}_{X}})$, a diagonal matrix with elements $\sqrt{P_X} \sqrt{Q_X}$. Summing its singular values gives us $\int \sqrt{p(X)}\cdot \sqrt{q(X)}\; dX$, a form of Hellinger distance. One can spot the drawback that for continuous $p$ and $q$, this decomposition will generate an infinite number of singular values at each point in the sample domain when $\sqrt{p(X)}\sqrt{q(X)}$ is positive. So a smoother like a Gaussian function as $\mathcal{K}(X,X')$ is a must such that we can still measure the distance between $\sqrt{p(X)}$ and $\sqrt{q(X)}$ but with a finite number of singular values.\vspace{7pt}

\noindent \textbf{Another way of decomposing $\mathbf{K}_{X,{X'}}$, instead of using the SVD.} It is also possible to discuss the optimal singular functions when $q = p$. Not only can we can apply eigendecomposition $\mathbf{K}_{XX'} = \mathbf{Q}_{\mathbf{N}}\mathbf{\Lambda}_{\mathbf{N}}\mathbf{Q}_{\mathbf{N}}$, but we can also decompose a Gaussian function using the closed-form inner product $\mathcal{N}(X-X';2v) = \int \mathcal{N}(X-X'';v)\mathcal{N}(X'-X'';v) dX''$. So $\mathbf{K}_{XX'}$ can also be decomposed as $\mathbf{K}_{XX'} = \mathbf{K}_{XX''}\mathbf{K}_{XX''}^\intercal$, with $\mathbf{K}_{XX''}$ having half of the variance of $\mathbf{K}_{XX'}$. Denote $\mathbf{C} = diag(\sqrt{\mathbf{P}_X}) \mathbf{K}_{XX''}$, then
\begin{equation}
\begin{aligned}
{diag}(\sqrt{\mathbf{P}_X}) \mathbf{K}_{XX'} diag(\sqrt{\mathbf{P}_{X}}) = \mathbf{C}\mathbf{C}^\intercal,
\end{aligned}
\label{CCT}
\end{equation}
implying that the eigenvector of ${diag}(\sqrt{\mathbf{P}_X}) \mathbf{K}_{XX'} diag(\sqrt{\mathbf{P}_{X}})$ must match the left singular vector of $\mathbf{C}$. One can further see that $\mathbf{C}$ represents $\sqrt{p(X)}\mathcal{N}(X-X'';v)$, the square root of a joint density $p(X,X'') = p(X) \mathcal{N}(X-X'';2v)$ up to a constant, representing the data density $p(X)$ passed through a Gaussian conditional.\vspace{7pt}

\noindent \textbf{Approximating the identity function with Gaussian residuals.} The decomposition can also be said to decompose the identity function with Gaussian residuals. Since the matrix $\mathbf{K}$ in the cost function is a Gaussian cross Gram matrix, each of its element is a Gaussian difference $\mathcal{N}(X_i-X_j';v_X+v_q) = \int \mathcal{N}(X-X_i;v_X) \mathcal{N}(X-X_j';v_q) dX$, which can be written as an inner product between a Gaussian residual $\mathcal{N}(X-X_i;v_X)$ and a residual $\mathcal{N}(X-X_j';v_q)$. By the closed form, their inner product has a variance $v_X+v_q$. 

Suppose the data samples are $X_1,X_2,\cdots,X_N$, and the generated samples are $X_1',X_2',\cdots,X_K'$. Suppose each data sample will define a Gaussian residual $p_i = \mathcal{N}(X-X_i;v_X)$, with each generated sample defining a residual $q_j = \mathcal{N}(X-X_j';v_q)$. Let us write them as vectors, $\mathbf{f} = [p_1,p_2,\cdots,p_N]^\intercal$ and $\mathbf{g} = [q_1,q_2,\cdots,q_K]^\intercal$, then the Gaussian cross Gram matrix can be written as $\mathbf{K} = \int \mathbf{f} \;\mathbf{g}^\intercal dX$, the inner product between two vectors, integrated over $X$. We perform the SVD with $\mathbf{K} = \mathbf{U}\mathbf{S}\mathbf{V}$. 

The trick we apply is to put an identity function $\mathbbm{1}(X=X')$ in the inner product, such that $\mathbf{K} = \int \mathbf{f}(X)\mathbf{g}^\intercal(X) dX = \iint p(X)\mathbbm{1}(X=X') \mathbf{g}^\intercal (X') dXdX'$, which does not change the computation for the results, but change the integral over $X$ into a double integral over two variables $X$ and $X'$. Now we apply the following variational procedure:
\begin{equation}
\resizebox{.89\linewidth}{!}{
$\begin{gathered}
\mathbf{K} = \iint \mathbf{f}(X)\mathbbm{1}(X=X') \mathbf{g}^\intercal (X') dXdX' = \mathbf{U}\mathbf{S}\mathbf{V}, \vspace{7pt}\\
\widehat{\mathbf{f}}(X) := \mathbf{U}^\intercal\mathbf{f}(X),\;\;\widehat{\mathbf{g}}(X) = \mathbf{V}^\intercal\mathbf{g}(X), \vspace{7pt}\\ 
 \iint \widehat{\mathbf{f}}(X) \mathbbm{1}(X=X') \widehat{\mathbf{g}}(X') dXdX' = \mathbf{S} =  \begin{bmatrix}
    \sigma_{1} &  \\
    & \ddots  \\
    & & \sigma_{N}
\end{bmatrix}. 
\end{gathered}$}
\end{equation}
It shows that if we apply the variational trick, constructing $\widehat{\mathbf{f}}(X)$ and $\widehat{\mathbf{g}}(X)$, the new double integral $\iint \widehat{\mathbf{f}}(X) \mathbbm{1}(X=X') \widehat{\mathbf{g}}(X') dXdX'$ will become a diagonal matrix of singular values. Our cost function is maximizing the sum of singular values. Looking at this doubler integral more closely, we can show that the vectors of functions $\widehat{\mathbf{f}}(X)$, $\widehat{\mathbf{g}}(X)$ are essentially decomposing and approximating the identity function $\mathbbm{1}(X=X')$, if we maximize the sum of singular values:
\begin{equation}
\begin{gathered}
\mathbbm{1}(X=X') \approx \sum_{k=1}^K \sigma_k \widehat{{\mathbf{f}}_k}(X) \widehat{{\mathbf{g}}_k}(X'),\vspace{-4pt}
\end{gathered}
\label{quantity}
\end{equation}
which can be described as using $\widehat{\mathbf{f}}$ and $\widehat{\mathbf{g}}$, the vectors of Gaussian residuals with affine transformations $\mathbf{U}$ and $\mathbf{V}$, to approximate and decompose the identity function $\mathbbm{1}\{X=X'\}$, i.e., using Gaussian residuals to come up with the best approximator of the identity $\mathbbm{1}\{X=X'\}$. 

\section{Other Theoretical Results}

We present two more additional results here. First, in the main paper, we have shown that the $L_2$ norm of a mixture density has a closed form solution. We further show in Property~\ref{property_other_moments} that any moment of the density function has a closed form if it is a mixture. The second result is that in the mixture decoder, we construct a topology where each data sample $X_n$ is paired with the noise variable $\mathbf{c}$ as the input to the mixture decoder $\textbf{D}$ to generate a series of outputs by sampling the noise $\mathbf{c}$ different, here we propose an extension in Property~\ref{property_6_iteration_process} that uses an iterative process to apply this procedure multiple times. 

\begin{property}
(Other moments of the mixture density.) Suppose $p(X)$ is a mixture density. For simplicity of the notations, we ignore the normalization constant and focus on the exponential part of the Gaussians with a fixed constant variance of $1$. Let us also assume that $X$ is in 1D for simplicity of the derivations, then it follows that $p(X) = \frac{1}{N}\sum_{n=1}^N\exp\left(-(X-X_n)^2\right)$. The third power of $p$ will have the form 
\begin{equation}
\resizebox{1\linewidth}{!}{
$\begin{aligned}
{p^3}(X) &= \frac{1}{N^3} \sum_{n,p,q} \exp (-(X-X_n)^2 - (X-X_p)^2 - (X -X_q)^2).
\end{aligned}$}
\end{equation}
\noindent The term inside the exponential can be written as
\begin{equation}
\begin{aligned}
&-(X-X_n)^2 - (X-X_p)^2 - (X -X_q)^2 \\
& = -3X^2 + 2X(X_n+X_p+X_q) - X_n^2 - X_p^2 - X_q^2\\ 
& = -3(X-\frac{1}{3}(X_n+X_p+X_q))^2 \\
&\;\;\;\;\;\;\;\;\;\;\;\;\;\;+ \frac{1}{3}(X_n+X_p+X_q)^2 - X_n^2 - X_p^2 - X_q^2,
\end{aligned}
\end{equation}
which is a quadratic difference term $(X-\frac{1}{3}(X_n+X_p+X_q))^2$ with a mean value of $\frac{1}{3}(X_n+X_p+X_q)$. Therefore, the integral of ${p^3}(X)$ will also has a closed form with 
\begin{equation}
\resizebox{.85\linewidth}{!}{
$\begin{aligned}
\int p^3(X) dX &= \frac{1}{3} \cdot \frac{1}{N^3} \sum_{n,p,q}\exp\Big(\frac{1}{3}\big(-(X_n-X_p)^2 \\ 
&\;\;\;\;\;\;\;\;\;\;\;\;\; - (X_p-X_q)^2  - (X_n-X_q)^2\big)\Big).\vspace{5pt}
\end{aligned}$}
\end{equation}
The same analysis generalizes to the case with multiple dimensions and the normalizing constant. The same analysis also generalizes to every order $p$, not just the third power, by induction. Therefore, the $L_p$ norm of a mixture of density has a closed form, and it may also be reasonable to use a series of norms of $p(X)$, with various polynomial powers, to define its entropy. 
\label{property_other_moments}
\end{property}

The second supplemented proposition is the definition of a recursive procedure for the mixture decoder. Our inspiration is as follows. If we start with a sample $X_1$, concatenated $X_1$ with random noises $\mathbf{c}$, then sampling various $\mathbf{c}$ will produce a series of generated samples through the mixture decoder. Now if we pick one sample $X_2$ from these generated samples, concatenated again with randomly sampled $\mathbf{c}$ and feed the samples through a decoder, it can again generate a series of samples $X_3$. In this way, the mixture decoder can be applied recursively. 
\begin{prop}
(Making the mixture decoder recursive.) Start with one sample $X_1$, we define recursively the conditional probabilities $p(X_2|X_1), p(X_3|X_2), \cdots$ with 
\begin{equation}
\begin{gathered}
p(X_2|X_1) = \int p(c) \mathcal{N}(X_2 - \textbf{D}(X_1,c);v)  dc,\\
p(X_3|X_2) = \int p(c) \mathcal{N}(X_3 - \textbf{D}(X_2,c);v)  dc,\\
p(X_4|X_3) = \int p(c) \mathcal{N}(X_4 - \textbf{D}(X_3,c);v)  dc,\\
\cdots
\end{gathered}
\end{equation}
Following this procedure, one can produce a series of variables $X_1, X_2, X_3,\cdots, X_S$, where one $X_1$ corresponds to multiple $X_2$, one $X_2$ corresponds to multiple $X_3$, and so forth. It has the following property:\vspace{7pt}

\noindent 1) Assuming that $p(X_1)$ is a mixture, then $p(X_2) = \int p(X_1)p(X_2|X_1) dX_1$, integrated over $X_1$, will also be a mixture. Thus $p(X_t)$ at any given $t$ will be a mixture. So its norm has a closed form.\vspace{9pt}

\noindent 2) Since the conditionals $p(X_2|X_1)$ and $p(X_3|X_2)$ are both mixture densities, the conditional $p(X_3|X_1)$ will also be a mixture density, with centers defined on all possible generated samples that can be reached from $X_1$. Thus the norm such as $\iint p^2(X_3|X_1)p(X_1)dX_1dX_3$ also has a closed form solution, which can also be generalized to any $X_t$.\vspace{9pt}

\noindent 3) Compared with random walks where each step $p(X_{t+1}|X_{t})$ is a standardized white Gaussian, here each step is a nonlinear function on $X_t$ parameterized by a neural network. But the analysis on the random walks also applies here.
\label{property_6_iteration_process}
\end{prop}

\begin{figure}[H]
\centering
\includegraphics[width=.8\linewidth]{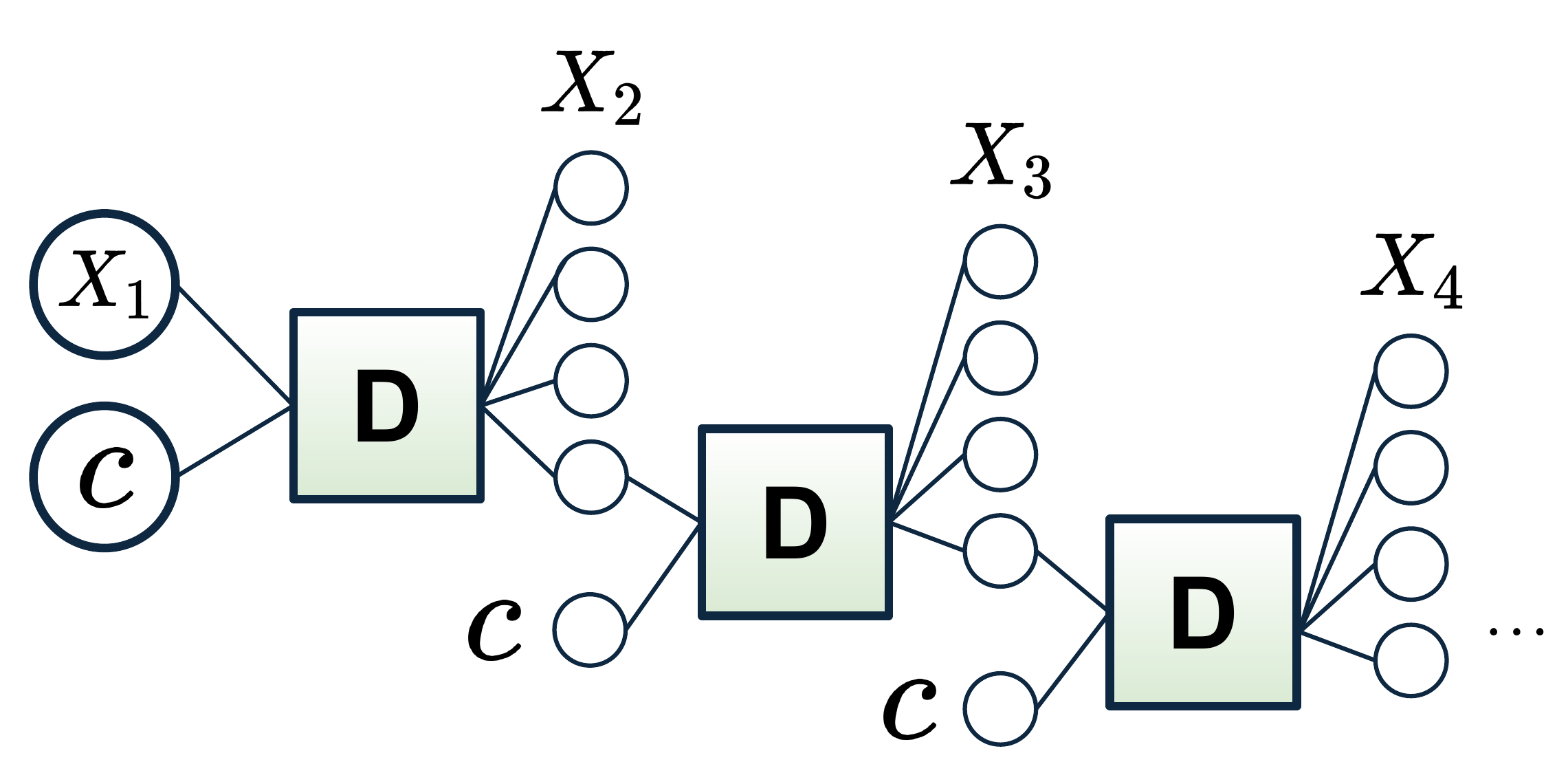}
\caption{The illustration of the proposed recursive procedure of a mixture decoder network. One sample $X_1$ is mapped to multiple $X_2$; one $X_2$ is mapped to multiple $X_3$, and so forth, by applying the decoder and the noise variable.}
\label{proposed_diagram_2}
\end{figure}

We also provide a diagram of the proposed process in Fig.~\ref{proposed_diagram_2}. One sample $X_1$ is concatenated with the noise variable $\mathbf{c}$ to produce multiple generated samples, from which we pick one $X_2$, concatenated with $\mathbf{c}$ to further generate multiple samples, and repeat the process recursively. This can be viewed as an extension of the mixture decoder to a recursive procedure, which can also be seen as a nonlinear extension of a random walk where each step $p(X_2|X_1)$ is no longer a simple Gaussian variance, but a complex nonlinear mapping. 

\section{Other Experimental Results}

\subsection{Validating the Optimal Solutions of an Autoencoder}
\label{Appendix_autoencoder}

In Sec.~\ref{section_results_mixture_data}, we discussed that when training an autoencoder for projecting the $2D$ toy data samples into $1D$ features, the solution is almost unique, regardless of the parameters and models used for the training.

Sec.~\ref{section_results_mixture_data} shows the results for toy datasets such as two moons and Gaussians, we further apply the same procedure to the three more complex mixture datasets shown in Sec.~\ref{quantitative_analysiss} and Fig.~\ref{figure_dataset_mix}. This also shows that the heatmaps generated from optimizing the densities matches training an end-to-end autoencoder. We explain the procedure of obtaining these heatmaps below. 

Although in autoencoders, we did not specifically set the variance for the features, which means that we always use the deterministic features as direct inputs to the decoder network. But if we look at the reconstructions from the decoder (Fig.~\ref{9b} and Fig.~\ref{figure_dataset_mix}), they look like a continuous curve on the $2D$ space, similar to a manifold. Thus, we investigate whether a implicit assumption of the Gaussian exist for the features, that even we did not specifically use a additive noise for the features, there is still a small Gaussian variance associated with the features, such that the reconstructions form a continuous curve like a manifold.

\begin{figure}[h]
\centering
\begin{subfigure}{.32\textwidth}\includegraphics[width=\linewidth]{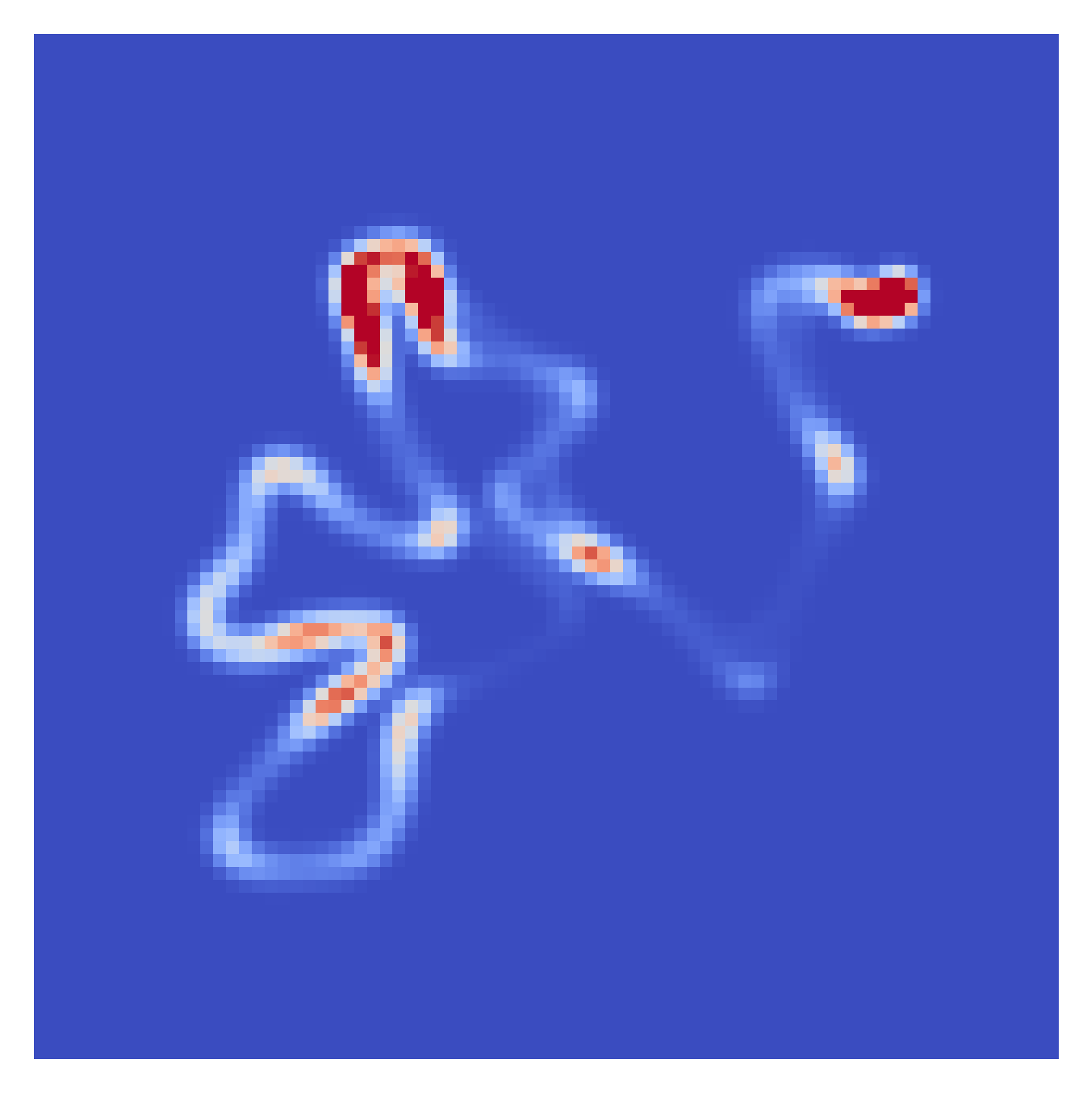}\vspace{-5pt}
\caption{\textbf{}}
\end{subfigure}
\begin{subfigure}{.32\textwidth}\includegraphics[width=\linewidth]{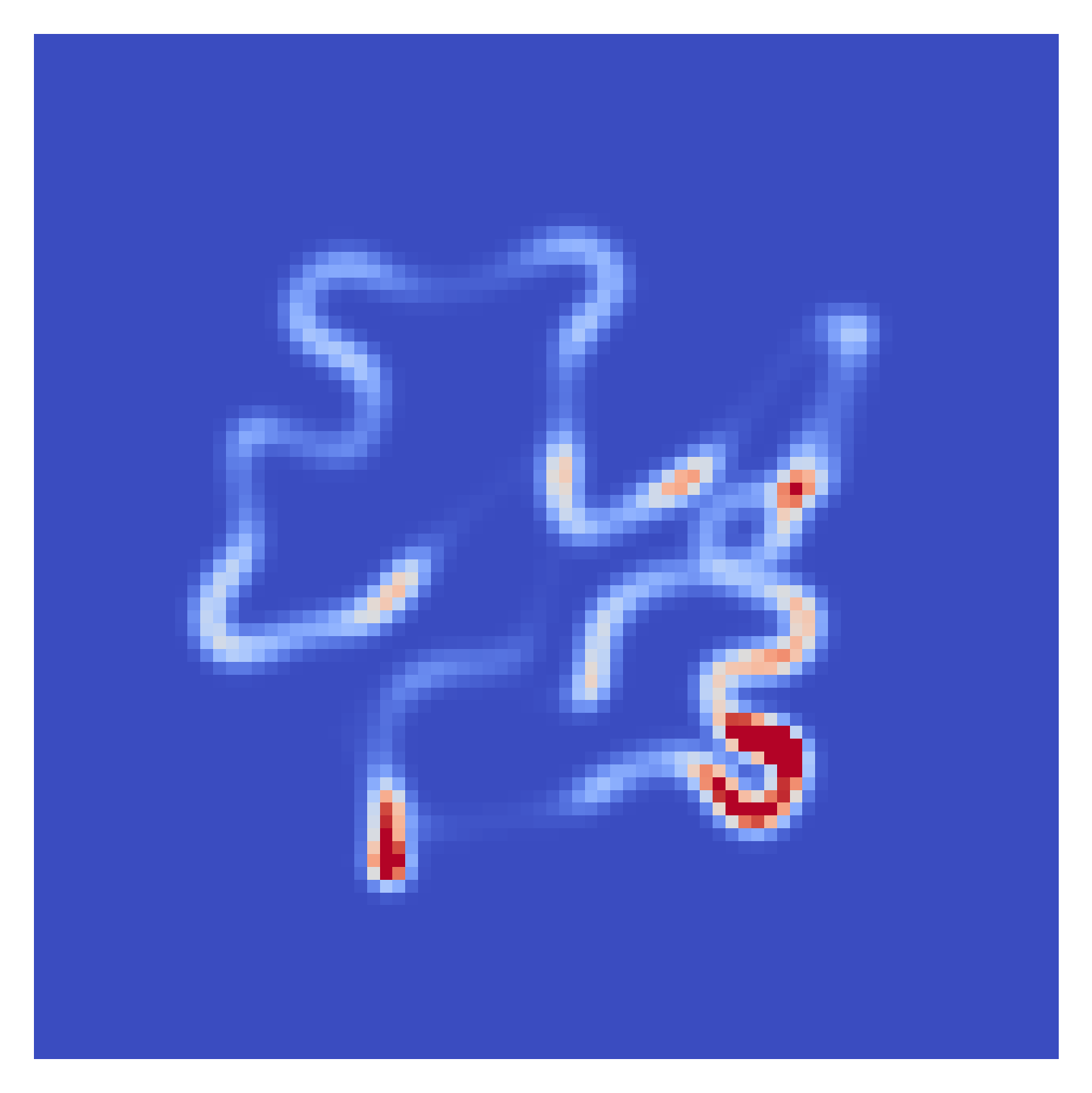}\vspace{-5pt}
\caption{\textbf{}}
\end{subfigure}
\begin{subfigure}{.32\textwidth}\includegraphics[width=\linewidth]{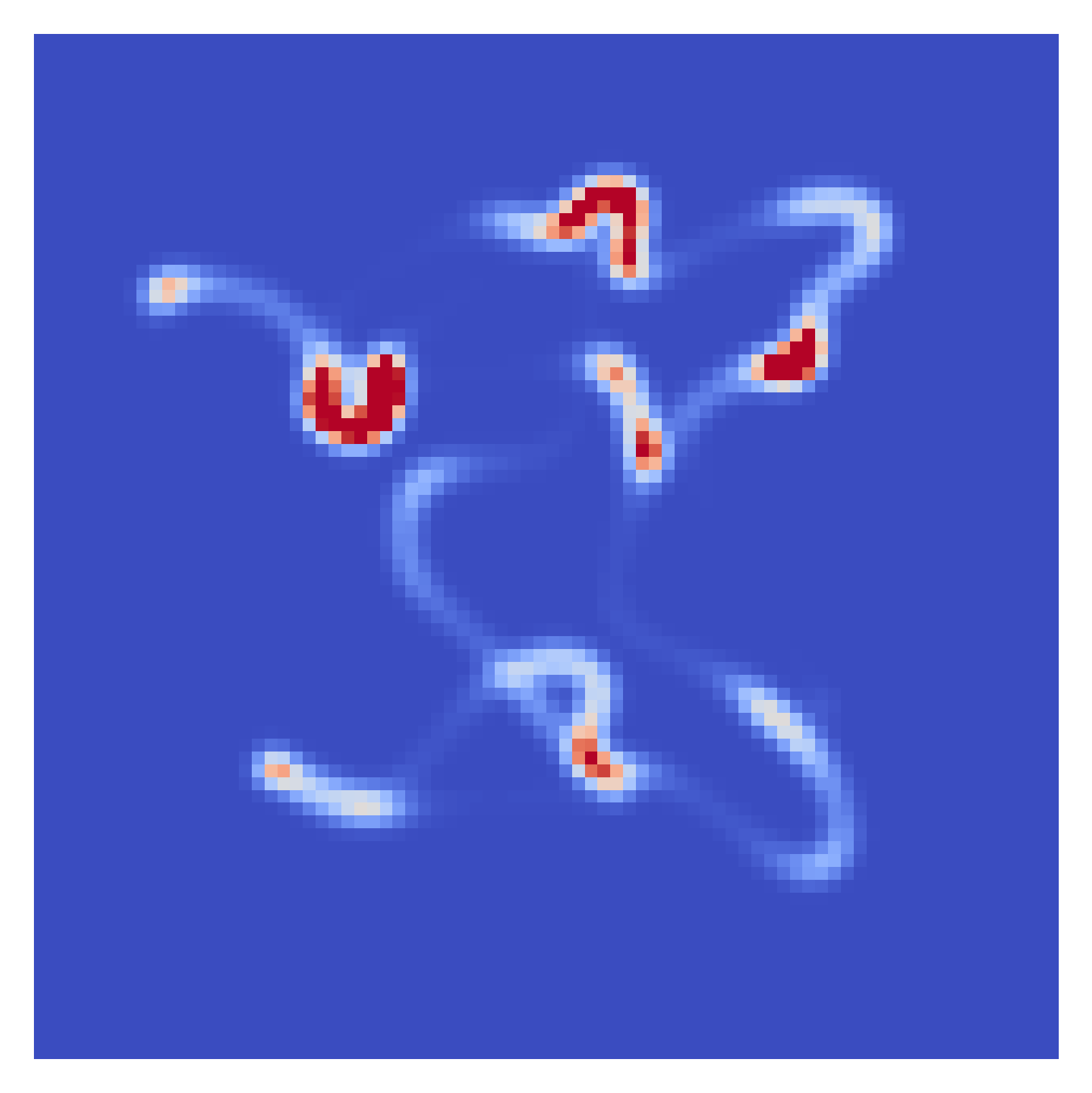}\vspace{-5pt}
\caption{\textbf{}}
\end{subfigure}
\begin{subfigure}{.32\textwidth}\includegraphics[width=\linewidth]{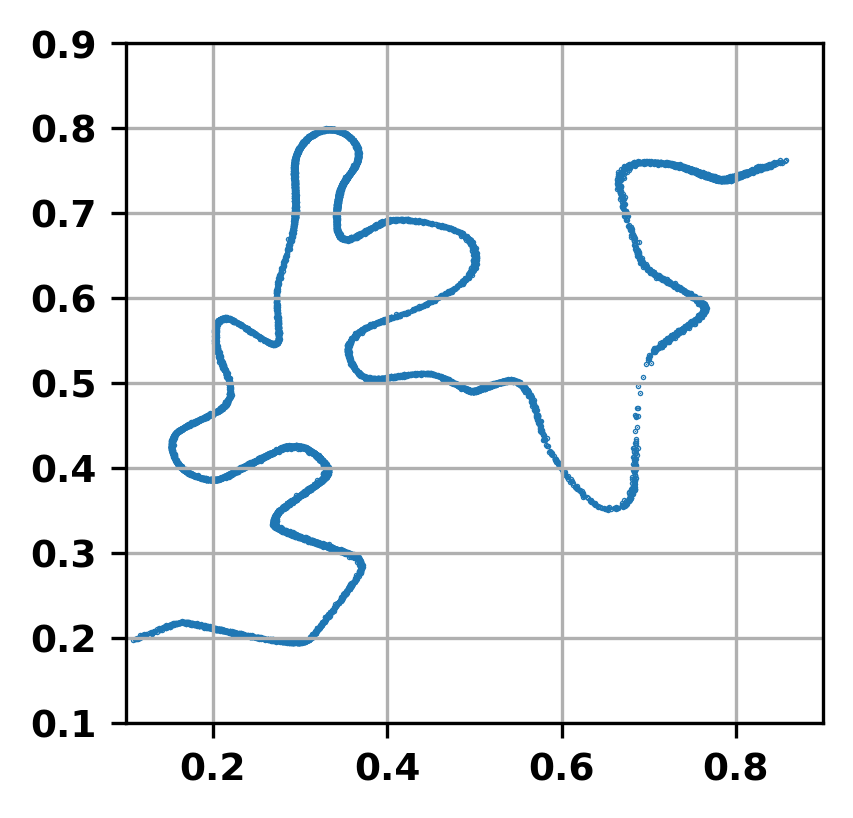}\vspace{-5pt}
\caption{\textbf{}}
\end{subfigure}
\begin{subfigure}{.32\textwidth}\includegraphics[width=\linewidth]{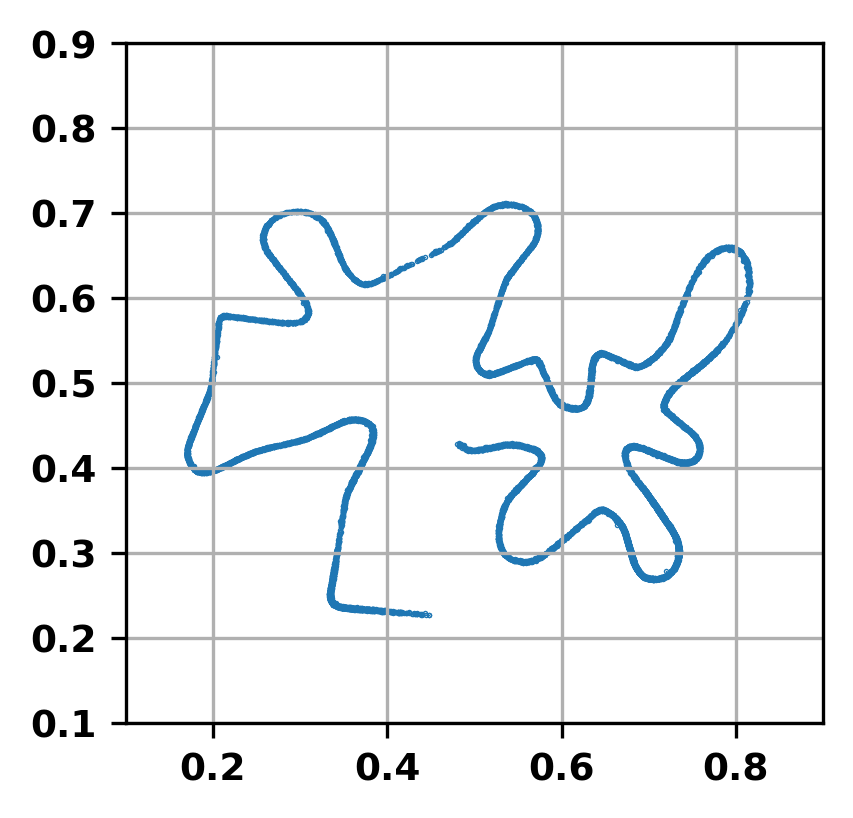}\vspace{-5pt}
\caption{\textbf{}}
\end{subfigure}
\begin{subfigure}{.32\textwidth}\includegraphics[width=\linewidth]{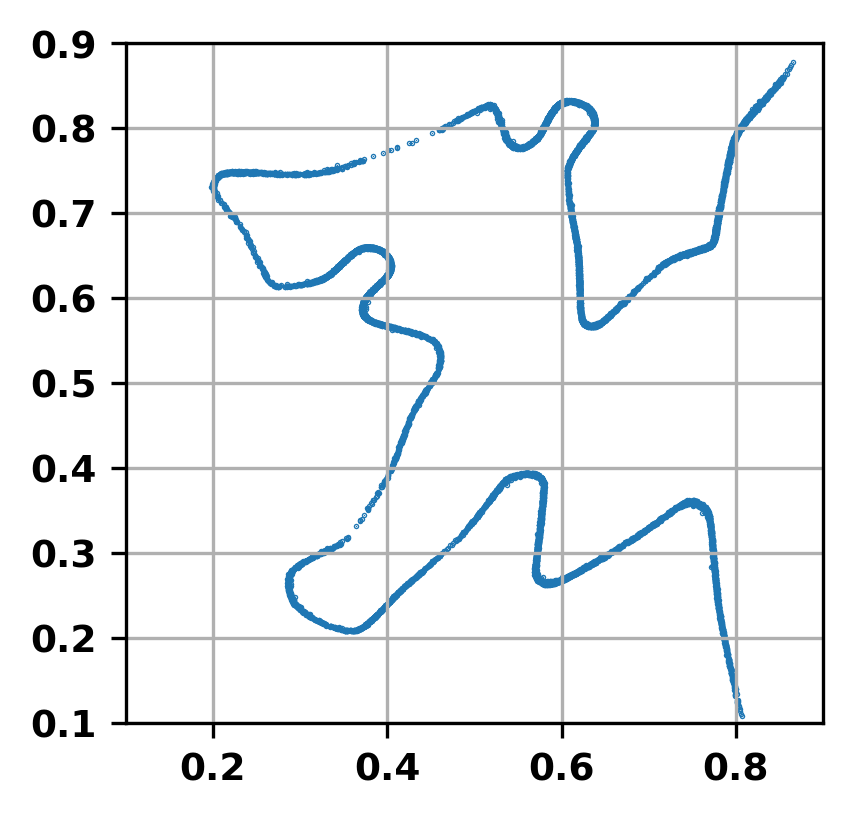}\vspace{-5pt}
\caption{}
\end{subfigure}
\caption{In addition to the datasets we show for Fig.~\ref{9b}, we also show the comparison of density heatmaps obtained from optimizing directly the objective of an autoencoder from the principle of optimizing densities (figures (a) to (c)), and the reconstructions of training an end-to-end autoencoder that minimizes the mean-squared error. The purpose is to validate that the autoencoder objective indeed can be modeled by finding the optimal conditional densities. This also shows that an implicit Gaussian variance may exist in the features, even when we use deterministic features as inputs to the decoder for optimization.}
\end{figure}

As we have discussed, the objective of an autoencoder is maximizing $\iint p(X,Y)\log q(X|Y) dXdY$, when $p(Y|X)$ and $q(X|Y)$ are parameterized as $p(Y|X) = \mathcal{N}(Y-\textbf{E}(X))$ and $q(X|Y) = \mathcal{N}(X-\textbf{D}(Y))$. We want to show that the almost unique solution we found is exactly the solution to this maximization of the objective. Our test is to show that from this principle of maximizing over an objective of densities, we can obtain the same solution as optimizing an autoencoder that minimizes the mean-squared error. 

We find one implementation that can achieve this goal, to generate the consistent results with an autoencoder but from the principle of maximizing the objective involving only the probability densities. Since for toy datasets the data are in $2D$, we can directly estimate $p(X)$ with interpolated grids. We normalize the data such that the data range falls between $0$ and $1$. Then we estimate the density $p(X)$ with grid points interpolated in the sample space of $[0, 1]\times [0, 1]$. We use $50$ points for each axis, therefore the data density is discretized into a $50\times 50$ matrix. 

Then we still use an encoder and a decoder to parameterize the two conditional densities $p(Y|X)$ and $q(X|Y)$. We want to parameterize the encoder with $p(Y|X) = \mathcal{N}(Y-\textbf{E}(X))$ and the decoder with the $\log$, $\log q(X|Y)$, with the mean-squared error $||X-\textbf{D}(Y)||_2^2$. The procedure of the parameterization is as follows. 

For $p(Y|X)$, we still parameterize it as a Gaussian mixture. Since $p(Y|X)$ is a function of $Y$ conditioned on $X$, we set the inputs to the network a diagonal matrix of size $2500\times 2500$, with all diagonal entries of the matrix being $1$ and all other entries being $0$, representing $X$. This input can be viewed as a batch of samples with a batch size of $2500$, where each sample in the batch is an one-hot vector with one positive element in the $n$-th entry. Each of the $2500$ one-hot vectors represents a point of the $50\times 50$ grid points in the interpolated sample space. The output of this encoder is a $1D$ scalar for each sample of this batch of one-hot vectors, thus the output is a $2500\times 1$ vector. The network has an activation function of $\tanh$, thus the outputs range from $-1$ to $1$. 

We then use $3000$ point interpolated from $-1.1$ to $1.1$ to represent $Y$, generating a $3000\times 1$ vector. To compute $p(Y|X) = \mathcal{N}(Y-\textbf{E}(X);v)$, we compute the Gaussian difference between each point of the output of the encoder, a $2500\times 1$ vector representing $50 \times 50$ discretized grid points for $X$, and each point of the $3,000$ interpolated points for $Y$. This makes $p(Y|X)$ a $2500 \times 3000$ matrix. We found that this approach of directly constructing the $p(Y|X)$ from interpolated points can indeed generate heatmaps similar to the solution when training an end-to-end autoencoder. The results are most similar when the variance $v$ is set to be $0.00025$. From this procedure we parameterized directly the density $p(Y|X)$ as a discretized $2500 \times 3000$ matrix, where $2500$ refers to the $50\times 50$ grid points for $X$ and $3000$ refers to the $3000$ interpolated points from $-1.1$ to $1.1$ that represent $Y$. 

For the decoder $q(X|Y)$ we apply a similar procedure. Now we initiate a diagonal matrix of size $3000\times 3000$, each sample of this batch of $3000$ samples is an one-hot vector representing one of the $3000$ interpolated points. The output of the decoder is $2D$, representing $X$. We need to compute the mean-squared distance $||X-\textbf{D}(Y)||_2^2$ for $\log q(X|Y)$. We reuse the interpolated points we use for $X$, the $2500 \times 1$ vector representing $50\times 50$ points in the interpolated space of $[0, 1]\times [0,1]$. We compute the pairwise $L_2$ distances between all elements from the $2500 \times 1$ vector and the $3000\times 2$ matrix from the outputs of the decoder, producing a $2500\times 3000$ matrix, representing $\log q(X|Y)$. We do not need to use the Gaussian functions in this case since we want to obtain $\log q(X|Y)$. 

Now we have the discretized $p(X)$, $p(Y|X)$ and $\log q(X|Y)$ from the principle of directly estimating and parameterizing the densities. Denote three matrices as $\mathbf{P}_X$, $\mathbf{P}_{Y|X}$ and the matrix of mean-squared error $\mathbf{M}$. We directly minimize the objective $\frac{1}{2500\cdot 3000}\sum_{i=1}^{2500}\sum_{j=1}^{3000}\left( \left(diag(\mathbf{P}_X)\mathbf{P}_{Y|X}\right) \odot \mathbf{M} \right)_{i,j}$. Here the multiplication of $diag(\mathbf{P}_X\mathbf{P}_{Y|X})$ represents the joint $p(X,Y)$. Then we multiply this matrix with the matrix of $L_2$ distance $\mathbf{M}$. When we take the mean over all elements of the final matrix, it is equivalent to the double integral over $X$ and $Y$. Minimizing the quantities based on the discretized matrices of the densities, we find the solution obtained from this approach matches closely with training an end-to-end autoencoder. Here we can also directly visualize the densities as heatmaps, as shown in Fig.~\ref{fig10}, different from training an autoencoder when we visualize the outputs of the decoder that looks like a manifold in the $2D$ space (Fig.~\ref{9b}). 

\subsection{Additional results for MDNs}

Here we also include additional experimental results for the MDNs in addition to what we presented in Sec.~\ref{experiments_mdns}. These experiments are presented as follows.\vspace{5pt}

\noindent \textbf{{Interpolations of noises for generated samples.}} A test conventionally done for an MDN or a general generative model is sampling two noise samples, interpolating between the two noise samples as inputs, and visualizing the outputs for these interpolated samples. 

\begin{figure}[h]
  \centering
\includegraphics[width=.62\linewidth]{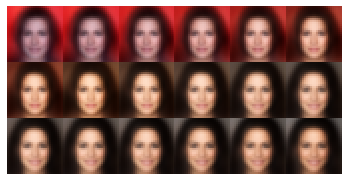}\vspace{-3pt}
\caption{Samples generated from interpolations of two randomly sampled noises.}
\label{interpolated_samples}
\end{figure}

Fig.~\ref{interpolated_samples} shows the interpolations of the samples. The top-left and the top-right are two samples generated from two randomly sampled noises. Those in between are generated from the interpolations of the two noises. The transition is smooth, meaning that the model indeed finds a smooth continuous mapping from a continuous noise distribution to generated images. 

\noindent \textbf{{Difference in the range of variances when the model is trainable.}} We have compared three costs for training an MDN, using the cross entropy cost based on the Kullback-Libeler divergence, the normalized inner product based on the Cauchy-Schwarz inequality, and the sum of singular values for the Gaussian cross Gram matrix. All these methods require setting the variance of the Gaussian functions in the cost functions. We do find that the three costs have different ranges of Gaussian variances that make the model trainable. 

\begin{figure}[h]
  \centering
\begin{subfigure}{.75\textwidth}\includegraphics[width=\linewidth]{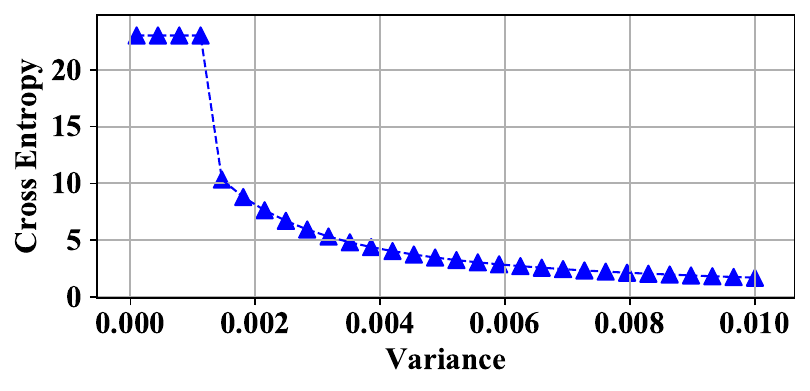}\vspace{-3pt}
\caption{The cross entropy cost from the Kullback-Libeler divergence.}
\label{21a}
\end{subfigure}\vspace{8pt}
\begin{subfigure}{.75\textwidth}\includegraphics[width=\linewidth]{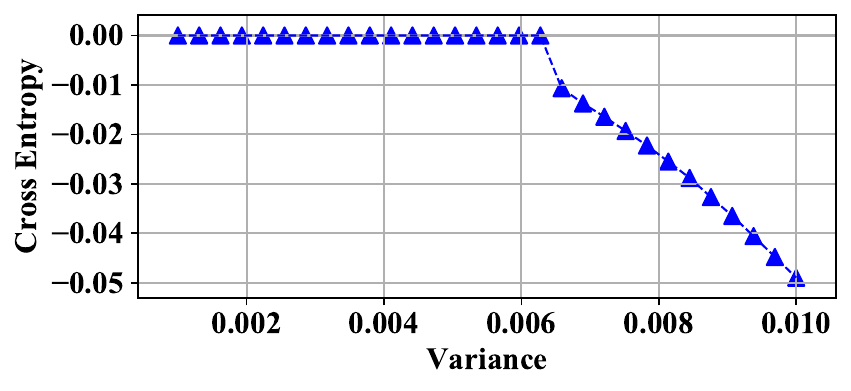}\vspace{-3pt}  
\caption{The $L_2$ normalized inner product cost from the Cauchy-Schwarz inequality.}
\label{21b}
\end{subfigure}\vspace{4pt}
\begin{subfigure}{.75\textwidth}\includegraphics[width=\linewidth]{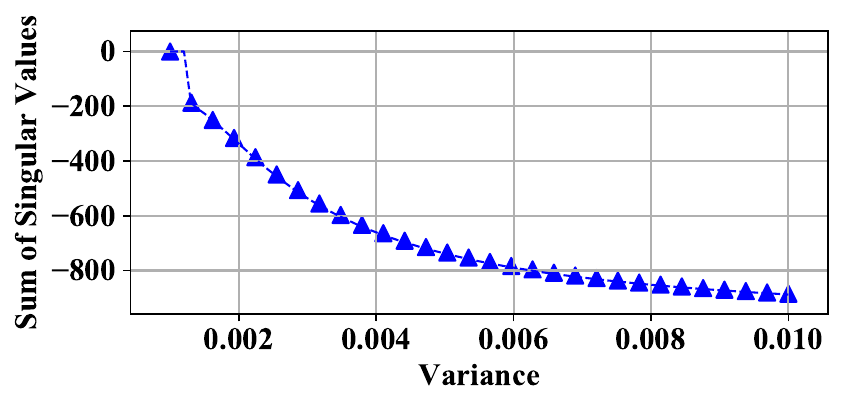}\vspace{-3pt}
\caption{The nuclear norm, the sum of singular values from the SVD.}
\label{21c}
\end{subfigure}\vspace{-1pt}
\caption{The impact of the variances used in the objective functions over the cost functions. When the variance is too small, the model is not trainable. When the variance is too large, the model is trainable but the performance will decay. In Fig.~\ref{diversity_comparison}, we will further compare the generation quality. In general, the smaller the variance, the more diverse the generated samples will be. The generation quality and diversity will decay when increasing the variance.}
\label{figure_renyi_shannon}
\end{figure}

We compare the values and present them in Fig.~\ref{21a} to Fig.~\ref{21c}. The three figures show how the three respective costs change to the variances we set for the Gaussian differences. We find that the nuclear norm cost (Fig.~\ref{21c}) has the widest range that the model can be trained, followed by the $L_2$ cost (Fig.~\ref{21b}), and finally the cross entropy from Kullback-Libeler divergence (Fig.~\ref{21c}). The figures has $30$ interpolated points from variance $0.001$ to $0.1$.\vspace{5pt}

\noindent \textbf{{Impact of the variances over the generation quality.}} In general, we find that smaller variances will increase the generative diversity and quality. If the variance is too large, for the regular cross entropy cost, the samples to go to the mean of the dataset; for the other two related to the $L_2$ scenario with normalizations, the samples will look similar when decomposing a matrix of $L_2$ distances without the exponential function, when the Gaussian function is not used. 

\begin{figure}[h]
\centering
\begin{subfigure}{.3\textwidth}\includegraphics[width=\linewidth]{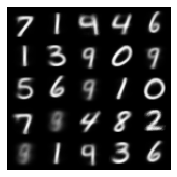}
\caption{\textit{var = 0.0015}}
\label{22a}
\end{subfigure}
\begin{subfigure}{.3\textwidth}\includegraphics[width=\linewidth]{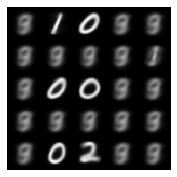}
\caption{\textit{var = 0.0025}}
\end{subfigure}
\begin{subfigure}{.3\textwidth}\includegraphics[width=\linewidth]{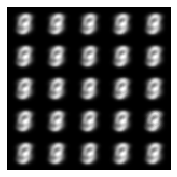}
\caption{\textit{var = 0.01}}
\label{22c}
\end{subfigure}\vspace{3pt}

\begin{subfigure}{.3\textwidth}\includegraphics[width=\linewidth]{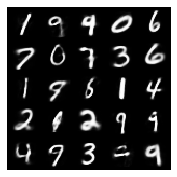}
\caption{\textit{var = 0.007}}
\label{22d}
\end{subfigure}
\begin{subfigure}{.3\textwidth}\includegraphics[width=\linewidth]{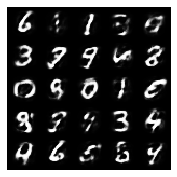}
\caption{\textit{var = 0.1}}
\end{subfigure}
\begin{subfigure}{.3\textwidth}\includegraphics[width=\linewidth]{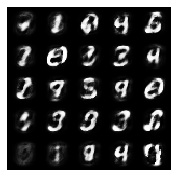}
\caption{\textit{var = 1}}
\label{22f}
\end{subfigure}\vspace{3pt}

\begin{subfigure}{.285\textwidth}\includegraphics[width=\linewidth]{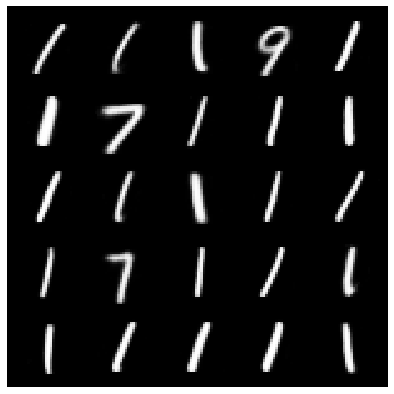}
\caption{\textit{var = 0.001}}
\label{22g}
\end{subfigure}\hspace{3pt}
\begin{subfigure}{.285\textwidth}\includegraphics[width=\linewidth]{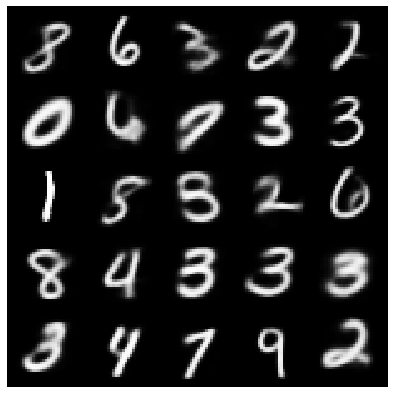}
\caption{\textit{var = 1}}
\end{subfigure}\hspace{3pt}
\begin{subfigure}{.285\textwidth}\includegraphics[width=\linewidth]{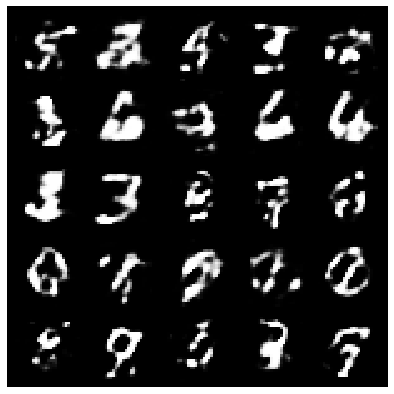}
\caption{\textit{var = 10}}
\label{22i}
\end{subfigure}
\caption{Generated samples when changing the variance in the Gaussian function. (a) to (c) correspond to the regular cross entropy cost; (d) to (f) correspond to the $L_2$ normalized inner product cost; (g) to (i) correspond to the nuclear norm cost. We find that when the variance is too large, for the $L_2$ case and the nuclear case, the results become similar to decomposing the minimizing the sum of singular values of the cross matrix of $L_2$ distances, shown in Fig.~\ref{figure_23}.}
\label{diversity_comparison}
\end{figure}

In Fig.~\ref{diversity_comparison}, the first three figures~\ref{22a} to~\ref{22c} correspond to the generated samples from the regular cross entropy cost. The generated samples go to the mean values when the variance is too large. 

Figures~\ref{22d} to~\ref{22f} correspond to the outputs for the $L_2$ normalized inner product cost. Figures~\ref{22g} to~\ref{22i} correspond to the maximization of the nuclear norm cost. For these two costs that use normalizations, when the variance of the Gaussian function becomes too large, the Gaussian function will be reduced to the $L_2$ distance, thus the generated samples will look like the samples if we substitute the Gaussian functions by the $L_2$ distances and decompose directly the cross Gram matrix of $L_2$ distance. We also find that when the variance is too small for the nuclear norm case, the diversity of the generated samples will become worse and generate repeated samples of digit $1$. 

\begin{figure}[h]
\centering
\includegraphics[width=.3\linewidth]{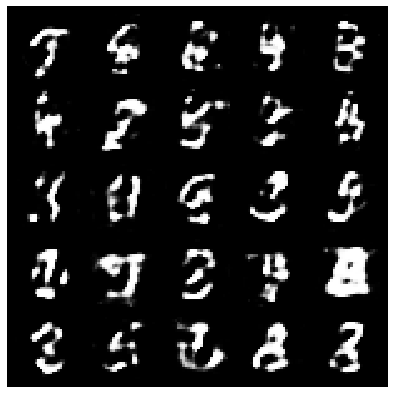}
\caption{The generated samples if we maximize the singular values for $-\mathbf{M}$, the negative of the the matrix of $L_2$ distances without the exponential. Note that the maximization is stable only when we perform a normalization by dividing all singular values by the largest singular value, such that all singular values are bounded by 1. We find that the generated results are similar to the results when the variance is too large in the nuclear norm cost case (Fig.~\ref{22i}). The generated samples are diverse but does not have semantic meanings. This shows that the Gaussian functions in the cost is necessary and cannot be substituted by the $L_2$ distances only.}
\label{figure_23}
\end{figure}

The Gaussian cross Gram matrix is computed by $\mathbf{M} \coloneqq
\frac{1}{d_X}\begin{bmatrix}
||X_1-X_1'||_2^2 & \cdots & ||X_1-X_N'||_2^2 \\
\vdots      & \ddots & \vdots  \\
||X_N-X_1'||_2^2 & \cdots & ||X_N-X_N'||_2^2  
\end{bmatrix}, \;\mathbf{K} \coloneqq \exp(-\frac{1}{2v_X}\mathbf{M})$, and we maximize the singular values from decomposing $\mathbf{K}$. Now we directly decompose $-\mathbf{M}$, the matrix of $L_2$ distances without the exponential. Suppose the singular values are $\sigma_1, \sigma_2,\cdots,\sigma_{N}$. We find that directly maximizing the sum of these singular values as a cost  will produce an unbounded value. Instead, we find it stable to first perform an normalization by dividing all singular values by the first largest singular value, and then maximize the sum $\max_{\sigma_1,\sigma_2,\cdots,\sigma_N} \sum_{n=1}^N \frac{\sigma_n}{\sigma_1}$. After normalization, the first value in the series will be set to $1$, and all other values will be smaller than $1$. We find that maximizing this normalized sum is stable, with generated samples shown in Fig.~\ref{figure_23}. This normalization step is needed for the $L_2$ distance matrix only. In the Gaussian cross Gram matrix, the singular values are already bounded.

Comparing the nuclear norm cost of the Gaussian Gram matrix when the variance is large (Fig.~\ref{22i}) and the generation result for this matrix of $L_2$ distances (Fig.~\ref{figure_23}), we can show that their generation results are consistent and highly similar, but the generation samples do not have semantic meanings. This proves that the Gaussianity in the cost is essential to learn generated samples that are semantically meaningful and resemble digits, even though two distributions do not match exactly. This highlights the importance of Gaussuianity. 

\newpage \;
\newpage

\bibliography{reference}



\end{document}